\documentclass[runningheads]{llncs}

\PassOptionsToPackage{numbers, compress, sort}{natbib}

\usepackage{eccv}

\usepackage{eccvabbrv}
\usepackage[font={footnotesize}]{caption}

\usepackage{graphicx}
\usepackage{booktabs}

\usepackage[accsupp]{axessibility}  % Improves PDF readability for those with disabilities.

\usepackage{hyperref}

\usepackage{orcidlink}

\usepackage{algorithm}
\usepackage{algpseudocode}

\newcommand{\Def}[0]{\mathrel{\mathop:}=}

\usepackage{pifont}
\newcommand{\cmark}{\textcolor{green}{\ding{51}}}
\newcommand{\xmark}{\textcolor{red}{\ding{55}}}

\usepackage{color, colortbl}
\definecolor{Gray}{gray}{0.93}
\definecolor{Orange}{rgb}{1,0.5,0}
\definecolor{Green}{rgb}{0,0.80,0}
\definecolor{Blue}{rgb}{0,0,0.92}
\definecolor{Red}{rgb}{0.90,0,0}
\definecolor{DGray}{gray}{0.83}
\definecolor{LightCyan}{rgb}{0.88,1,1}

\definecolor{bluegray}{rgb}{0.4, 0.6, 0.8}
\definecolor{ceruleanblue}{rgb}{0.16, 0.32, 0.75}

\newcommand{\MU}{{\text{MU}}}
\newcommand{\Retrain}{{\text{Retrain}}}
\newcommand{\Salun}{{\text{SalUn}}}
\newcommand{\MUSparse}{{\text{$\ell_1$-sparse}}}
\newcommand{\FT}{{\text{FT}}}
\newcommand{\RL}{{\text{RL}}}
\newcommand{\EUk}{{\text{EU-$k$}}}
\newcommand{\CFk}{{\text{CF-$k$}}}
\newcommand{\BE}{{\text{BE}}}
\newcommand{\BS}{{\text{BS}}}
\newcommand{\Scrub}{{\text{SCRUB}}}
\newcommand{\thetau}{{\btheta_\mathrm{u}}}
\newcommand{\thetao}{{\btheta_\mathrm{o}}}

\newcommand{\Puwr}{{\text{$P_\text{u}^{(\text{w/r})}$}}}
\newcommand{\Puw}{{\text{$P_\text{u}^{(\text{w})}$}}}
\newcommand{\Pur}{{\text{$P_\text{u}^{(\text{r})}$}}}
\newcommand{\Pn}{{\text{$P_\text{n}$}}}

\newcommand{\ourset}{worst-case forget set}
\newcommand{\Ourset}{Worst-case forget set}
\newcommand{\OurSet}{Worst-Case Forget Set}
\newcommand{\randomset}{random forget set}

\newcommand{\RandomSet}{Random Forget Set}

\usepackage{wrapfig}

\usepackage{multirow,mathtools}
\usepackage{pifont}
\usepackage{color, colortbl}

\usepackage{blindtext}
\usepackage{lipsum}

\usepackage{multirow}
\usepackage{makecell}
\usepackage{graphicx}
\usepackage{listings}

\usepackage{bbm}

\usepackage [english]{babel}
\usepackage [autostyle, english = american]{csquotes}

\usepackage{pifont}
\usepackage{url}
\usepackage[most]{tcolorbox}

\usepackage{lipsum}
\usepackage{xcolor}
\usepackage{wrapfig}
\usepackage{booktabs}

\usepackage{bbold}
\usepackage{amsmath,amsfonts,bm}

\def\eqref#1{(\ref{#1})}
\def\1{\bm{1}}
\newcommand{\train}{\mathcal{D}}
\newcommand{\Dr}{\mathcal{D_{\mathrm{r}}}}
\newcommand{\Df}{\mathcal{D_{\mathrm{f}}}}

\DeclareMathAlphabet{\mathsfit}{\encodingdefault}{\sfdefault}{m}{sl}
\SetMathAlphabet{\mathsfit}{bold}{\encodingdefault}{\sfdefault}{bx}{n}

\def\gS{{\mathcal{S}}}

\DeclareMathOperator*{\argmin}{arg\,min}

\DeclareMathOperator*{\st}{\text{subject to}}

\newcommand{\btheta}{{\boldsymbol{\theta}}}

\newcommand{\bx}{\mathbf{x}}
\newcommand{\bw}{\mathbf{w}}

\newcommand{\bz}{\mathbf{z}}

\hypersetup{
 colorlinks=true,
 citecolor=ceruleanblue,
 linkcolor=ceruleanblue,
 urlcolor=black}

\begin{document}

% ---------------------------------------------------------------
% TODO REVIEW: Replace with your title
\title{
%Forgettable or Unforgettable? Who Is the Nonconformist in Machine Unlearning?
% Unlearning Under Fire: Identifying Worst-Case Scenarios in Data Influence Erasure
Challenging Forgets: Unveiling the Worst-Case Forget Sets in Machine Unlearning
} 

% TODO REVIEW: If the paper title is too long for the running head, you can set
% an abbreviated paper title here. If not, comment out.
\titlerunning{Unveiling the Worst-Case Forget Sets in Machine Unlearning}

% TODO FINAL: Replace with your author list. 
% Include the authors' OCRID for the camera-ready version, if at all possible.
\author{Chongyu Fan\inst{*,1}, Jiancheng Liu\inst{*,1}\orcidlink{0009-0000-3650-3097}, Alfred Hero\inst{2}\orcidlink{0000-0002-2531-9670}, Sijia Liu\inst{1,3}\orcidlink{0000-0003-2817-6991}}

% TODO FINAL: Replace with an abbreviated list of authors.
\authorrunning{Chongyu Fan$^{*}$, Jiancheng Liu$^{*}$, Alfred Hero, Sijia Liu}
% First names are abbreviated in the running head.
% If there are more than two authors, 'et al.' is used.

% TODO FINAL: Replace with your institution list.
\institute{OPTML@CSE, Michigan State University \and EECS, University of Michigan, Ann Arbor \and 
MIT-IBM Watson AI Lab, IBM Research
\\ $^{*}$Equal contribution
\\ \email{\{fanchon2,liujia45,liusiji5\}@msu.edu, hero@eecs.umich.edu}
}

\maketitle

\setcounter{footnote}{0}

\everydisplay{\small}

\begin{abstract} 
The trustworthy machine learning (ML) community is increasingly recognizing the crucial need for models capable of selectively `unlearning' data points after training. This leads to the problem of \textit{machine unlearning} ({\MU}), aiming to eliminate the influence of chosen data points on model performance, while still maintaining the model's utility post-unlearning. Despite various {\MU} methods for data influence erasure, evaluations have largely focused on \textit{random} data forgetting, ignoring the vital inquiry into which subset should be chosen to truly gauge the authenticity of unlearning performance. To tackle this issue, we introduce a new evaluative angle for {\MU} from an adversarial viewpoint. We propose identifying the data subset that presents the most significant challenge for influence erasure, \textit{i.e.}, pinpointing the \textit{worst-case} forget set. Utilizing a bi-level optimization principle, we amplify unlearning challenges at the upper optimization level to emulate worst-case scenarios, while simultaneously engaging in standard training and unlearning at the lower level, achieving a balance between data influence erasure and model utility. Our proposal offers a worst-case evaluation of {\MU}'s resilience and effectiveness. Through extensive experiments across different datasets (including CIFAR-10, 100, CelebA, Tiny ImageNet, and ImageNet) and models (including both image classifiers and generative models), we expose critical pros and cons in existing (approximate) unlearning strategies. Our results illuminate the complex challenges of {\MU} in practice, guiding the future development of more accurate and robust unlearning algorithms.
The code is available at \url{https://github.com/OPTML-Group/Unlearn-WorstCase}.
% \keywords{Machine unlearning \and Bi-level optimization \and Data selection}
\end{abstract}

\section{Introduction}
\label{sec: intro}

%%% Motivation on MU
In this work, we study the problem of machine unlearning ({\MU}) \cite{cao2015towards,bourtoule2021machine,nguyen2022survey,liu2024rethinking}, which aims to erase unwanted data influences (\textit{e.g.}, specific data points, classes, or knowledge concepts) from a machine learning (ML) model, while preserving the utility of the model post-unlearning (termed `unlearned model') for information not targeted by the unlearning process.
The concept of {\MU} was initially developed to meet data protection regulations, \textit{e.g.}, the `right to be forgotten' \cite{hoofnagle2019european,rosen2011right}.
Given its ability to evaluate data's impact on model performance, the application of {\MU} has expanded to address a variety of trustworthy ML challenges.
These include defending against ML security threats \cite{liu2022backdoor,jia2023model}, removing data biases for enhanced model fairness \cite{oesterling2023fair,sattigeri2022fair,chen2024fast}, protecting copyright and privacy \cite{zhang2024unlearncanvas,achille2023ai,eldan2023whos}, and mitigating sociotechnical harms by, \textit{e.g.}, erasing generative models' propensity for producing toxic, discriminatory, or otherwise undesirable outputs \cite{gandikota2023erasing,zhang2023forget, fan2023salun}.

%%% evaluation challenge -> problem of interest
With the growing significance and popularity of {\MU}, a wide array of unlearning methods has been devised across various domains, such as image classification \cite{warnecke2021machine, izzo2021approximate, golatkar2020eternal, thudi2022unrolling, jia2023model,  fan2023salun}, text-to-image generation \cite{gandikota2023erasing,zhang2023forget,kumari2023ablating,gandikota2024unified,heng2023selective, fan2023salun,zhang2024unlearncanvas}, federated learning \cite{wang2022federated, liu2022right, wu2022federated, che2023fast}, graph neural networks \cite{chen2022graph, chien2022certified, cheng2023gnndelete}, and large language modeling \cite{eldan2023whos,wu2023depn,yu2023unlearning,zhang2023composing,yao2023large}. 
In this work, we delve into {\MU} in vision tasks, primarily concentrating on image classification but also extending our investigation to text-to-image generation.
For a detailed review of existing {\MU} methods, we refer readers to Secs.\,\ref{sec: related works} and \ref{sec: preliminary}.

Our study concentrates on improving the reliability of {\MU}  evaluation, considering the diversity of unlearning tasks and methodologies.
Our motivation stems from the limitations in current {\MU}  evaluation methods, which heavily rely on artificially constructed \textit{random data forgetting} scenarios \cite{kurmanji2023towards,jia2023model,fan2023salun,cotogni2023duck}, 
particularly noticeable in {\MU} for image classification.
However, the observation in \cite{fan2023salun} and our own investigations (in Sec.\,\ref{sec: preliminary}) indicate that the effectiveness of unlearning methods can significantly vary with the selection of the forget set (\textit{i.e.}, the specific data points designated for forgetting), resulting in substantial performance variance. This unlearning variability based on forget set choices leads us to reconsider the possibility of exploring a \textit{worst-case forget set} selection scenario. 
Such a scenario would ideally represent the most challenging conditions for an unlearning method's performance, reducing unlearning variance and facilitating a more reliable assessment.
This leads us to the following research question:
\begin{center}
    \textit{(Q)} \textit{How can we pinpoint the worst-case forget set for {\MU} to accurately assess its unlearning efficacy under such challenging conditions?}
\end{center}

\begin{wrapfigure}{r}{0.6\textwidth}
\centering
\vspace{-8mm}
\includegraphics[width=0.6\textwidth]{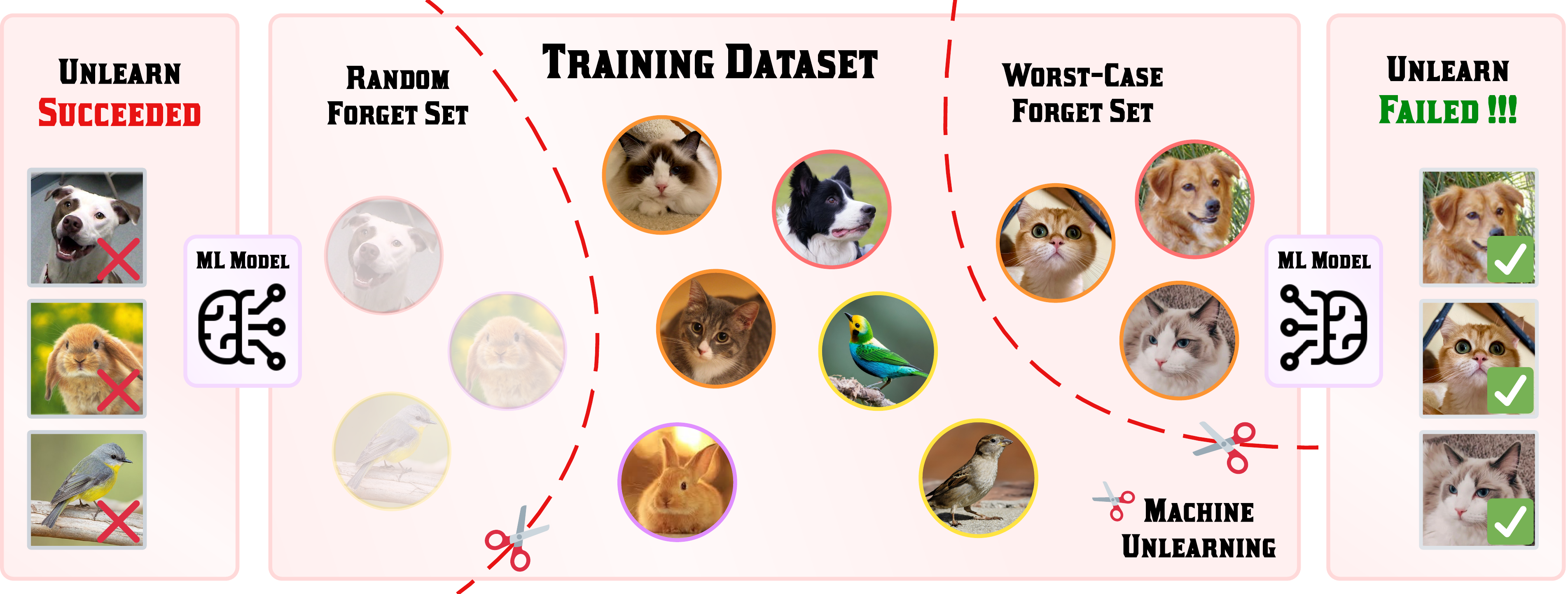}
\vspace{-6.5mm}
\caption{Overview of unlearning under our proposal ({\ourset}) vs. {\randomset}. The data influence is difficult to remove under {\ourset} vs. {\randomset}.
}
\vspace{-7mm}
\label{fig: teaser}
\end{wrapfigure}
Tackling \textit{(Q)} can also be viewed as creating an adversarial evaluation to challenge the effectiveness of {\MU}.  Yet, different from existing adversarial metrics such as membership inference attacks (MIAs) \cite{shokri2017membership,carlini2022membership} 
or adversarial prompts in {\MU} for image generation  \cite{zhang2023generate}, 
the worst-case forget set method assesses {\MU} from a \textit{data selection} perspective, and is compatible with other metrics for evaluating unlearning effectiveness and utility.
We will show that the method of selecting a worst-case forget set can be readily extended to different unlearning scenarios, including class-wise forgetting  (aiming to remove the impact of an entire image class) and prompt-wise forgetting in text-to-image generation (aimed at avoiding generation conditioned on certain text prompts). \textbf{Fig.\,\ref{fig: teaser}}\,\footnote{Thanks to Naughty, Fries, Crescent, Catcat, Wula, and other cuties for their appearance in Fig.\,\ref{fig: teaser}.}
shows an overview of the {\ourset} vs. {\randomset}. 

\noindent \textbf{Contributions.} We summarize our {contributions} below. 

\noindent $\bullet$ We are the first, to our knowledge, to highlight the necessity of identifying the worst-case forget set for {\MU}, %introduce its concept, 
and develop a solid formulation and optimization foundation through the lens of bi-level optimization (BLO).

\noindent $\bullet$ We introduce an effective algorithmic framework for identifying the worst-case forget set, offering two distinct benefits:  optimization efficiency, which reduces the complexity of BLO through the sign-based gradient unrolling method, and optimization generality, making it easily adaptable to worst-case evaluations in both class-wise and prompt-wise forgetting scenarios.

\noindent $\bullet$ We assess the empirical effectiveness of worst-case forget set-based MU evaluation, showcasing the strength of our approach and elucidating the rationale behind the chosen forget sets in terms of coreset selection and class-discriminative ability. Additionally, we explore the applicability of the worst-case forget set, extending from image classification to text-to-image generation.

\section{Related Works}
\label{sec: related works}
\noindent\textbf{Machine unlearning ({\MU}).}
{\MU} involves modifying ML models to eliminate the influence of specific data points, classes, or even broader knowledge-level concepts \cite{ginart2019making,neel2021descent,ullah2021machine,sekhari2021remember,cao2015towards}.
A widely recognized \textit{exact} unlearning strategy refers to \textit{retraining the model from scratch} (termed \textit{\Retrain}), executed by omitting the data points designated for forgetting from the training set \cite{thudi2022necessity,thudi2022unrolling}.
While {\Retrain} offers a solid guarantee of the data influence erasure \cite{dwork2006our, graves2021amnesiac}, it demands significant training costs, rendering it practically challenging for deep models.
To overcome the efficiency challenges of {\MU}, many research efforts have shifted focus towards creating \textit{approximate} unlearning methods \cite{warnecke2021machine, izzo2021approximate, golatkar2020eternal, thudi2022unrolling, jia2023model,  fan2023salun,ginart2019making,guo2019certified,neel2021descent,ullah2021machine,sekhari2021remember}.
%, despite the absence of theoretical guarantees for unlearning.
Some representative methods will be reviewed in Sec.\,\ref{sec: preliminary}.
\iffalse 
Representative methods include fine-tuning the pre-trained model on the retain set to induce catastrophic forgetting of the forget set  \cite{warnecke2021machine}, performing gradient ascent on the forget set to counterbalance standard training processes \cite{thudi2022unrolling}, 
relabeling the forget set followed by fine-tuning to unlearn its accurate information \cite{golatkar2020eternal}, reducing the forget set's impact through the influence function method \cite{izzo2021approximate}, and 
identifying a specific subset of model parameters crucial for unlearning, leading to localization-informed unlearning tactics \cite{jia2023model,fan2023salun}. 
\fi 
\iffalse
Alongside efforts to boost the efficiency of {\MU}, various studies \cite{ginart2019making,guo2019certified,neel2021descent,ullah2021machine,sekhari2021remember} have investigated probabilistic approaches leveraging differential privacy (DP) principles. These methods provide theoretical unlearning guarantees under specific DP assumptions.
However, they fall short in effectively guarding unlearning against MIAs (membership inference attacks) \cite{dwork2006our, graves2021amnesiac}, and do not match the computational efficiency of approximate unlearning methods. 
In this work, our study centers on approximate unlearning methods; see a summary presented in Sec.\,\ref{sec: preliminary}.
\fi

In the above literature, the applications of  {\MU} have mainly focused on image classification tasks, targeting either class-wise forgetting, which seeks to erase the impact of an entire image class, or random data forgetting, aimed at removing randomly chosen data points from the training set.
However, the scope and use cases of {\MU} have significantly broadened recently.
For instance, within the field of text-to-image generation using diffusion models (DMs), several studies \cite{gandikota2023erasing,zhang2023forget,kumari2023ablating,gandikota2024unified,heng2023selective, fan2023salun,zhang2024unlearncanvas} have applied the concept of {\MU} to mitigate the harmful effects of inappropriate or sensitive prompts on image generation, aiming to enhance the safety of DMs.
In addition, the significance of {\MU}  in enhancing the trustworthiness of data-models has been recognized across other non-vision domains, including graph neural networks \cite{chen2022graph, chien2022certified, cheng2023gnndelete}, federated learning \cite{wang2022federated, liu2022right, wu2022federated, che2023fast}, and the rapidly evolving field of large language models (LLMs) \cite{eldan2023whos,wu2023depn,yu2023unlearning,zhang2023composing,yao2023large}. 
In this study, we focus on vision-related tasks.

\noindent \textbf{Evaluation of {\MU}.}
Assessing the effectiveness of {\MU} presents its own set of challenges \cite{jia2023model,thudi2022necessity,zhang2023generate,gandikota2023erasing}. 
\iffalse 
Ideally, one would compare the performance of an approximate unlearning method against {\Retrain}. However, while the performance discrepancy with {\Retrain} may be theoretically intriguing, it is often impractical for real-world applications. Therefore, empirical evaluation metrics have been developed for {\MU}. 
The empirical metrics employed are also varied 
%as depending solely on a single performance metric may offer only a partial perspective of the performance  
\cite{jia2023model,thudi2022necessity,zhang2023generate,gandikota2023erasing}.
\fi 
Generally speaking, \textit{unlearning effectiveness} and post-unlearning \textit{model utility} are the two primary factors considered for assessing the performance of {\MU}. 
When applying {\MU} to classification tasks,
effectiveness-oriented metrics include {unlearning accuracy}, which relates to the model's performance accuracy after unlearning on the forget set \cite{golatkar2020eternal}, and {MIAs} to determine if a data point in the forget set was part of the model's training set post-unlearning \cite{song2021systematic}.
Utility-oriented metrics include {remaining accuracy}, which evaluates the performance of the updated model post-unlearning on the retain set \cite{becker2022evaluating}, and testing accuracy, assessing the updated model's generalization capability.
When applying {\MU} to generation tasks, accuracy-based metrics are also employed through the use of a post-generation classifier applied to the generated content \cite{zhang2024unlearncanvas,zhang2023generate},
while quality metrics of generations are employed to assess utility \cite{gandikota2023erasing}. 
However, a notable limitation of the above evaluation metrics
%, especially when measuring the unlearning effectiveness, 
is their significant dependency on the specific unlearning tasks at hand.
For example, the task of random data forgetting in image classification may result in considerable variance in the measurements of unlearning effectiveness  \cite{fan2023salun}, 
attributed to the randomness in selecting the forget set.

\noindent  \textbf{Data selection for deep learning.}
Data selection methods, such as dataset pruning \cite{schioppa2022scaling,paul2021deep,yang2022dataset,pruthi2020estimating,zhang2024selectivity} and coreset selection \cite{huggins2016coresets,xia2022moderate,borsos2020coresets,kim2023coreset,sener2017active}, serve as valuable strategies for enhancing the efficiency of model training \cite{coleman2019selection}. The aim of data selection is to identify the most representative training samples or eliminate the least influential ones, remaining the model's performance unaffected after training on the chosen data points.
In addition to efficiency, data selection is also employed to enhance model security by detecting and filtering out poisoned data points \cite{zeng2022sift}, as well as to increase fairness by eliminating biased data points \cite{sattigeri2022fair}.
In this work, the challenge of determining the worst-case forget set resonates with data selection strategies but applies to the novel realm of machine unlearning for the first time. 
Methodology-wise, the idea of bi-level optimization (BLO), previously applied in data selection contexts \cite{borsos2020coresets,yang2022dataset}, will also be used for addressing our focused problem. Nonetheless, we will adapt BLO  for the unlearning context and devise computationally efficient optimization strategies.

\section{Preliminaries, Motivation, and Problem Statement}
\label{sec: preliminary}

\noindent 
\textbf{Objective and setup of MU.} 
The objective of {\MU} is to negate the impact of a specific subset of training data points on a (pre-trained) model, while preserving its utility for data not subject to unlearning.
For a concrete setup of {\MU}, consider the training dataset $\train = \{\bz_i\}_{i=1}^N$, consisting of $N$ data samples.
Each sample $\bz_i$ includes a feature vector $\bx_i$ and a possible label $\textbf{y}_i$ for supervised learning. 
Let $\Df \subseteq \train$ represent the subset of data targeted for unlearning, with its complement, $\Dr = \train \setminus \Df$, being the dataset to retain. We refer to $\Df$ as the \textit{forget set} and $\Dr$ as the \textit{retain set}, respectively. 
Prior to unlearning, we have access to an initial model, denoted by $\thetao$, which could be trained on the full dataset $\train$ using methods like empirical risk minimization (ERM).

Given the above setup, 
{\Retrain}, an exact yet expensive unlearning approach, entails retraining the model $\thetao$ from scratch, exclusively utilizing the retain set $\Dr$.
It is typically regarded as the gold standard in {\MU}  \cite{thudi2022unrolling,jia2023model}.
However, due to the prolonged training time and the high cost, {\Retrain} is often impractical.
Consequently, approximate unlearning methods have emerged as efficient alternatives. Their objective is to efficiently create an \textit{unlearned model}, denoted as $\thetau$, leveraging prior knowledge of $\thetao$ and the forget set $\Df$ and/or the retain set $\Dr$.
Following the conceptual framework of {\MU} in \cite{liu2024rethinking}, the optimization problem to obtain $\thetau$ can be expressed as
\begin{align}
    \thetau = \argmin_{\btheta} ~ \ell_\MU(\btheta) \Def
    \ell_\mathrm{r} (\btheta; \Dr ) + \lambda  \ell_\mathrm{f}(\btheta; \Df),
    \label{eq: unlearn_objective}
\end{align}%
where $\ell_\mathrm{f}$ and $\ell_\mathrm{r}$ represent the forget loss and the retain loss, respectively, with $\lambda \geq 0$ acting as a regularization parameter. For instance, fine-tuning using the retain set {$\Dr$} equates to setting $\lambda = 0$, aimed to impose catastrophic forgetting of over $\Df$ after model fine-tuning.
Note that the specifics of $\ell_\mathrm{f}$, $\ell_\mathrm{r}$, and $\lambda $ can differ across various
 {\MU} methodologies.

\begin{wraptable}{r}{0.5\textwidth}
\vspace*{-12.3mm}
\caption{
Overview of examined {\MU} methods highlighting differences in relabeling-based forget loss, necessity of random re-initialization, partial model updates during unlearning, and the retain-forget regularization parameter $\lambda$ within   \eqref{eq: unlearn_objective}.
}
% \vspace{-1.3mm}
\centering
\resizebox{0.5\textwidth}{!}{
\renewcommand\tabcolsep{5pt}
\begin{tabular}{c|cccc}
\toprule[1pt]
\midrule
\multirow{1}{*}{\textbf{Method}} &
\begin{tabular}[c]{@{}c@{}}\textbf{Relabeling} \end{tabular} &
\begin{tabular}[c]{@{}c@{}}\textbf{Random} \\ \textbf{re-initialization} \end{tabular} &
\begin{tabular}[c]{@{}c@{}}\textbf{Partial model} \\ \textbf{update} \end{tabular} & 
\begin{tabular}[c]{@{}c@{}}$\mathbf{\lambda = 0}$\end{tabular} \\
% \begin{tabular}[c]{@{}c@{}}\textbf{Forget Loss ($\ell_f$)}\end{tabular}\\
\midrule
{\Retrain} &
\xmark &
\cmark &
\xmark &
%\xmark
\cmark
\\
\rowcolor{Gray}
\midrule
\FT \cite{warnecke2021machine} &
\xmark &
\xmark & 
\xmark &
%\xmark
\cmark
\\
\EUk \cite{goel2022towards} &
\xmark &
\cmark & 
\cmark &
%\xmark
\cmark
\\
\rowcolor{Gray}
\CFk \cite{goel2022towards} &
\xmark &
\xmark & 
\cmark &
%\xmark
\cmark
\\
\Scrub \cite{kurmanji2023towards} &
\xmark &
\xmark & 
\xmark &
%\cmark
\xmark
% KL divergence
% ($KL(P[\thetao(\bx_i)||\thetau(\bx_i)])$)
\\
\rowcolor{Gray}
\MUSparse \cite{jia2023model} &
\xmark &
\xmark& 
\xmark &
%\cmark
\xmark
% $\ell_1$-norm
% ($\|\thetau\|_1$)
\\
\RL \cite{golatkar2020eternal} &
\cmark &
\xmark & 
\xmark &
%\cmark
\xmark
% Random relabeling
% ($c_i \neq y_i$)
\\
\rowcolor{Gray}
\BE \cite{chen2023boundary} &
\cmark &
\cmark & 
\xmark &
%\cmark
\xmark
% Shadow class relabeling
% ($c_i = y_{shadow} \notin \textbf{y}$)
\\ 
\BS \cite{chen2023boundary} &
\cmark &
\xmark & 
\xmark &
%\cmark
\xmark
% Neighbor class relabeling
% ($c_i = f(\thetao; \text{FGSM}(\bx_i))$) 
\\
\rowcolor{Gray}
\Salun \cite{fan2023salun} &
\cmark &
\xmark & 
\cmark &
%\cmark
\xmark
% Random relabeling
% ($c_i \neq y_i$)
\\
\midrule
\bottomrule[1pt]
\end{tabular}
}
\vspace*{-7mm}
\label{tab: mu_method_summary}
\end{wraptable}
\noindent \textbf{Reviewing representative {\MU} methods.}
Assisted by \eqref{eq: unlearn_objective}, we provide an overview of $9$
existing (approximate) unlearning methods examined in this study; see \textbf{Table\,\ref{tab: mu_method_summary}} for a summary.
These methods can be roughly categorized into two main groups based on the choice of the forget loss $\ell_\mathrm{f}$: \textit{relabeling-free} and \textit{relabeling-based}.
 The latter, relabeling-based methods, assign an \textit{altered} label, distinct from the true label, to the data point targeted for forgetting. Consequently, minimizing $\ell_\mathrm{f}$ compels the unlearned model to discard the accurate label of the points to be forgotten.
These methods include random labeling (\textbf\RL) \cite{golatkar2020eternal},
boundary expanding (\textbf\BE) \cite{chen2023boundary},
boundary shrinking (\textbf\BS) \cite{chen2023boundary},
and saliency unlearning (\textbf\Salun) \cite{fan2023salun}.
In contrast, relabeling-free methods
utilize fine-tuning on the retain set $\Dr$ to induce catastrophic forgetting or apply gradient ascent on the forget set $\Df$  to achieve the forgetting objective.
These methods include 
 {f}ine-{t}uning (\textbf\FT) \cite{warnecke2021machine},
 \underline{e}xact \underline{u}nlearning restricted to the last \underline{$k$} layers (\textbf\EUk) \cite{goel2022towards},
\underline{c}atastrophically \underline{f}orgetting the last \underline{$k$} layers (\textbf\CFk) \cite{goel2022towards},
 \underline{sc}alable \underline{r}emembering and \underline{u}nlearning un\underline{b}ound (\textbf\Scrub) \cite{kurmanji2023towards} 
and \textbf{\MUSparse} MU \cite{jia2023model}.

 In addition to relabeling differences, the aforementioned {\MU} methods also vary in several other aspects. For instance, {\RL} can apply exclusively to the forget loss, corresponding to $\lambda  \to + \infty$ in \eqref{eq: unlearn_objective}. Methods including {\Salun}, {\EUk}, and {\CFk} target only a subset of model parameters, not the entire model $\btheta$ during the optimization process. Furthermore, methods like {\BE} and {\EUk} necessitate re-initializing the pre-trained model state.
{Table\,\ref{tab: mu_method_summary}} summarizes the configurations for the examined unlearning methods in this study.

\noindent \textbf{Challenge in {\MU} evaluation: Sensitivity to forget set selection.}
When assessing the effectiveness of {\MU}, a typical approach is \textit{random data forgetting}, which measures the unlearning ability when eliminating the influence of \textit{randomly} selected data points from the training set. 
However, evaluations based on the random selection of both data points and their quantity for forgetting can lead to  \textit{high variance} in the performance of a specific unlearning method, complicating fair comparisons across different {\MU} methods. Most importantly, a randomly selected forget set \textit{cannot} reveal the \textit{worst-case} unlearning performance, raising concerns about the reliability of an {\MU} method. 

\begin{figure}[htb]
\vspace*{-5mm}
\resizebox{\textwidth}{!}{
    \includegraphics[width=0.21\textwidth]{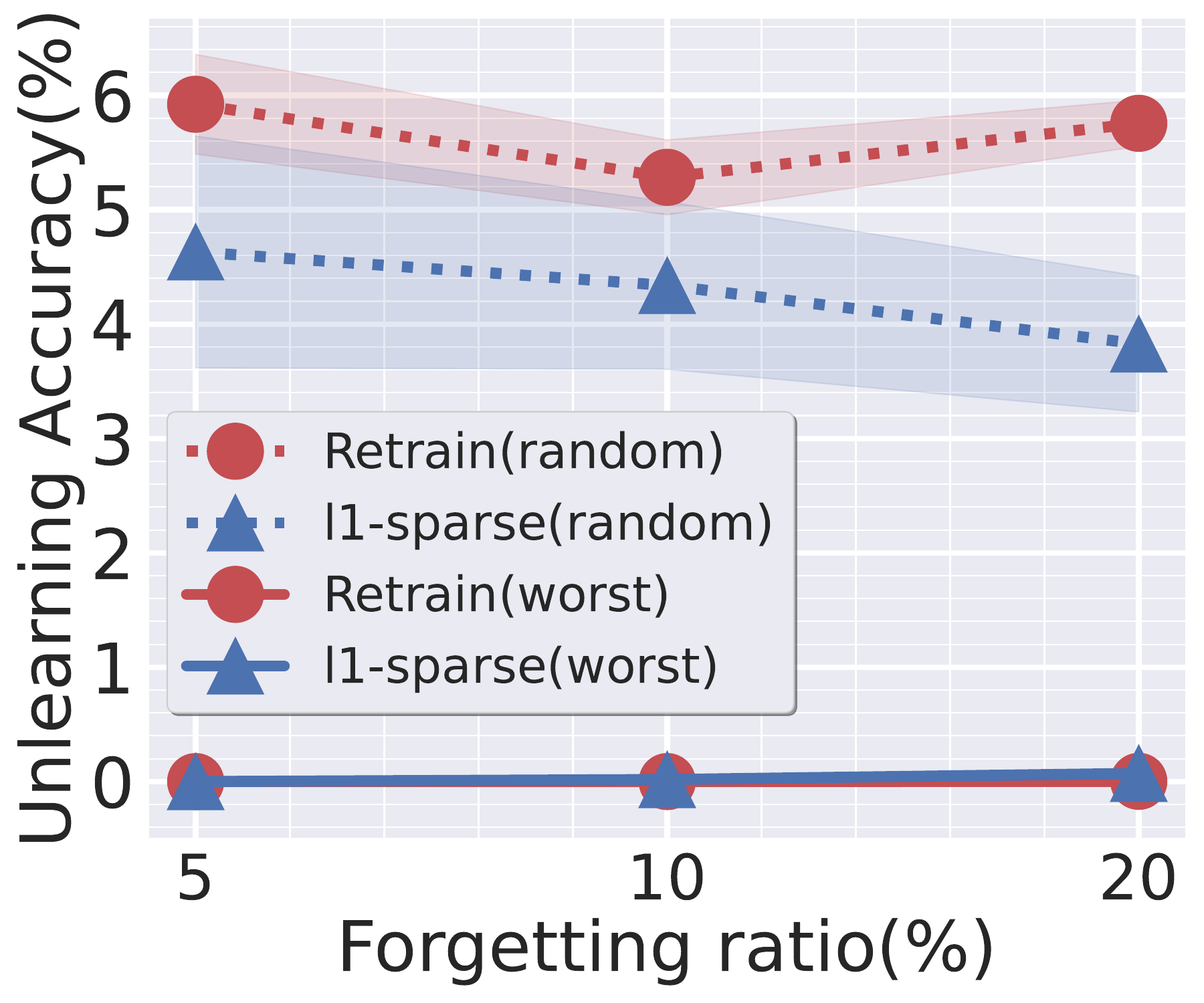}
    \includegraphics[width=0.21\textwidth]{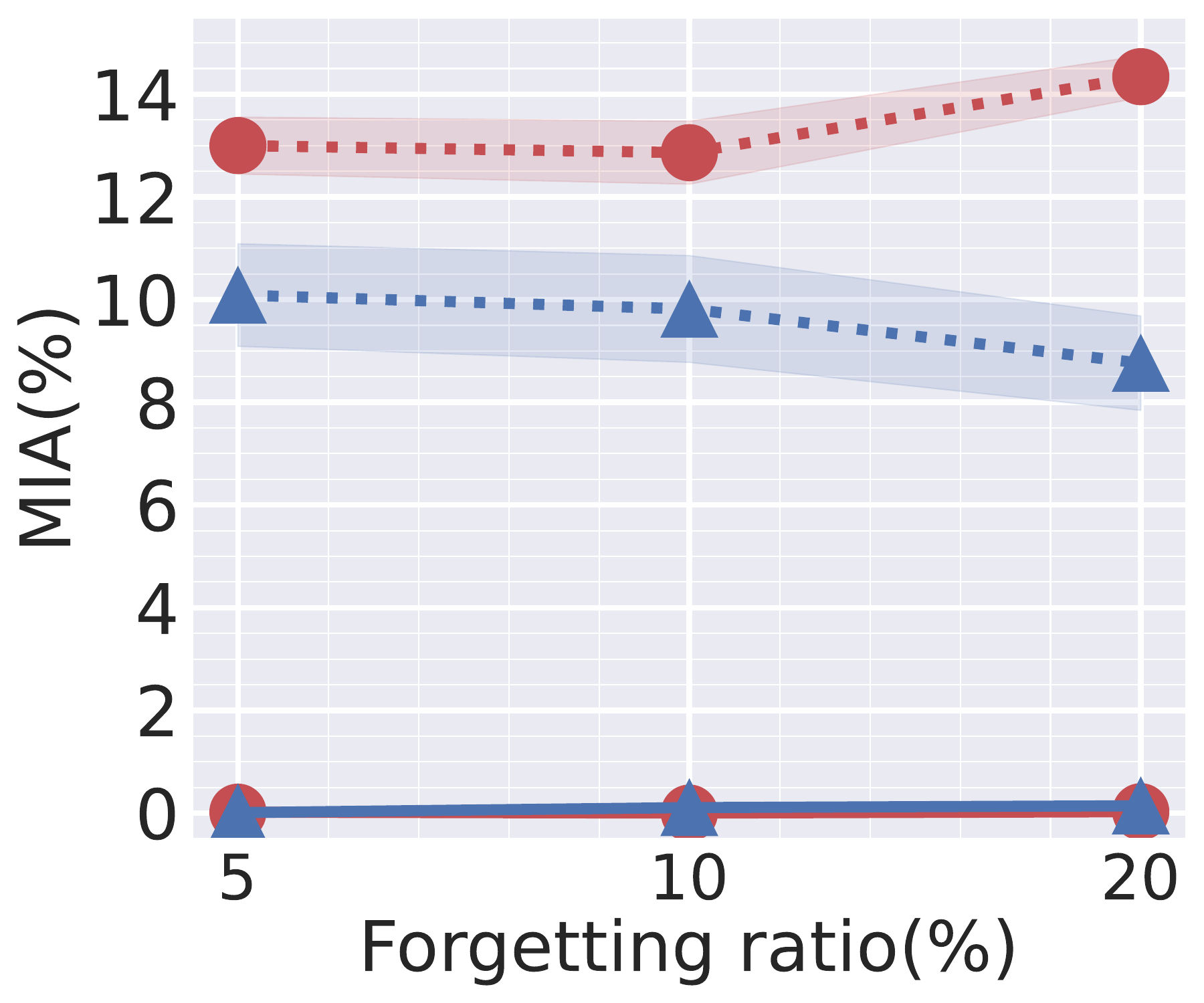}
    \includegraphics[width=0.21\textwidth]{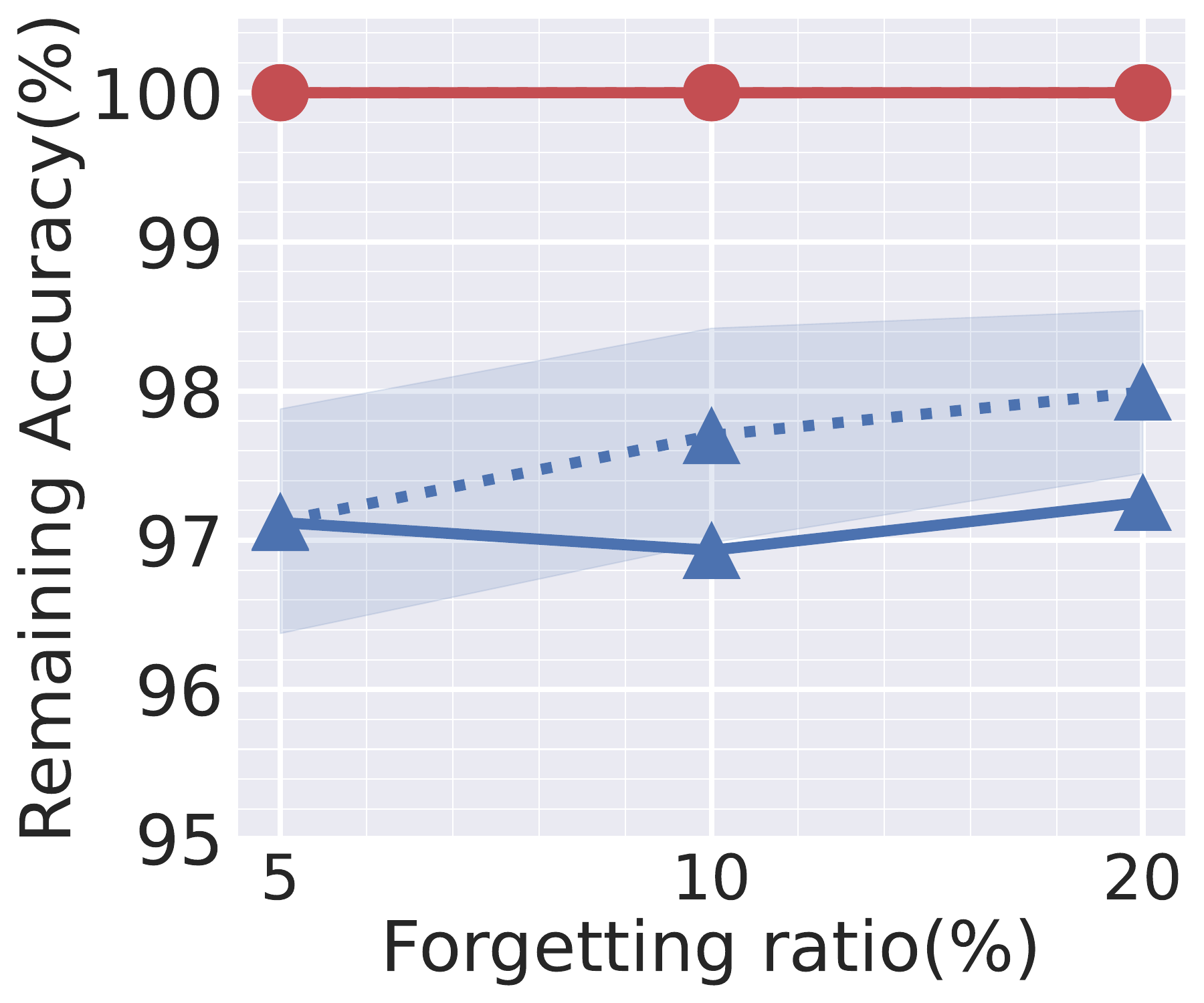}
    \includegraphics[width=0.21\textwidth]{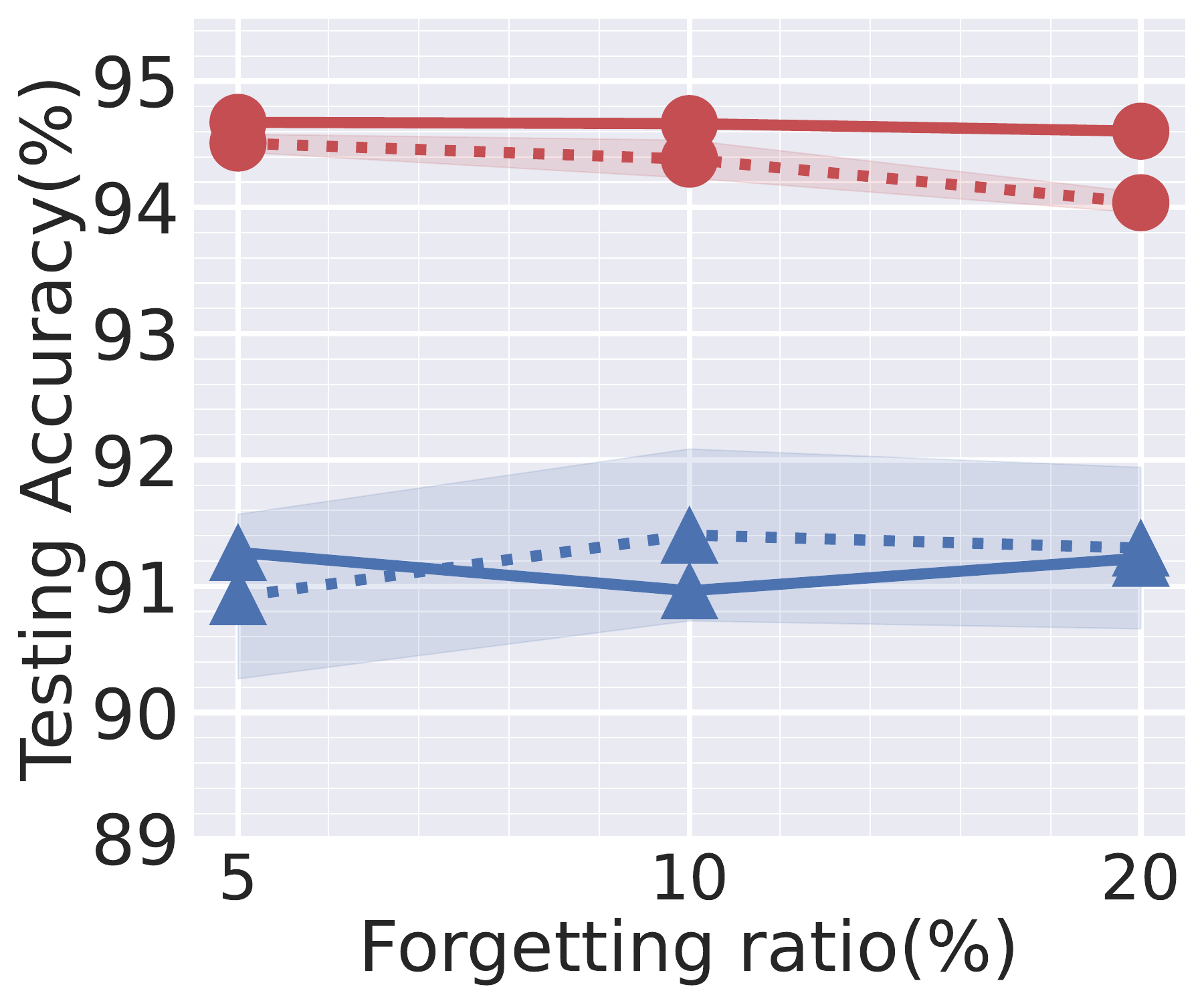}
}
\definecolor{Fig_variance_retrain}{RGB}{76,114,176}
\definecolor{Fig_variance_sparse}{RGB}{196,78,82}
\vspace*{-6.5mm}
\caption{Performance of   \textcolor{Fig_variance_retrain}{\Retrain} and \textcolor{Fig_variance_sparse}{\MUSparse} unlearning under random and worst-case forgetting scenarios at different forgetting data ratios on (CIFAR-10, ResNet-18).
Variance over 10 random selections is indicated by the shaded areas of the dashed lines.
}
\vspace*{-5mm}
\label{fig: variance}
\end{figure}

\textbf{Fig.\,\ref{fig: variance}} provides a  motivating example showcasing the sensitivity of {\MU} evaluation to the (random) forget set selection. 
We examine two unlearning approaches: {\Retrain} (exact unlearning) and  {\MUSparse} (a state-of-the-art approximate unlearning method), 
and explore two evaluation scenarios: traditional random data forgetting and our proposed worst-case forget set evaluation, which will be elaborated in Sec.\,\ref{sec: method}.
For a given unlearning method and forget set selection, we assess the unlearning effectiveness and model utility of the unlearned model $\thetau$ following the evaluation metrics used in \cite{jia2023model}.  The effectiveness metrics include unlearning accuracy (\textbf{UA}), calculated as {$\text{\textit{1--}}$\textit{the accuracy of $\thetau$}} on the forget set $\Df$, and \textbf{MIA} (membership inference attack) on $\Df$, which determines whether a data point in $\Df$ was {unused} in training $\thetau$. The utility metrics include remaining accuracy (\textbf{RA}), measured by the accuracy of $\thetau$ on the retain set $\Dr$, and testing accuracy (\textbf{TA}), the accuracy of $\thetau$ on the test set. 

Two main observations can be drawn from Fig.\,\ref{fig: variance}. First, both exact unlearning (\Retrain) and approximate unlearning ({\MUSparse}) with {\randomset}s result in \textit{increased variance} in UA and MIA, compared to the performance with identified worst-case forget sets (Sec.\,\ref{sec: method}).
Second, 
%when evaluated using unlearning effectiveness metrics, 
our approach effectively highlights unlearning challenges, demonstrated by the lowest UA and MIA at various forgetting ratios (a lower UA or MIA corresponds to a higher unlearning difficulty). Importantly, these identified worst-case forget sets do \textit{not} compromise model utility, as shown by the RA and TA of  {\Retrain}.

\noindent \textbf{Problem of interest: Identification of the worst-case forget set.}
As inspired by Table\,\ref{tab: mu_method_summary}, evaluating {\MU} through random data forgetting can lead to a high performance variance and provides limited insight into the worst-case performance of {\MU}.
To tackle these challenges, we propose to devise a systematic strategy to identify the data subset that presents the most significant challenge for influence erasure in {\MU}, while preserving the utility of the unlearned model. We define this identified subset as the \textbf{worst-case forget set}. Approaching from an adversarial perspective, our interest also lies in identifying the forget set that could diminish the unlearning performance, unveiling the worst-case unlearning scenario. In the next section, we will address the problem of identifying the worst-case forget set via a BLO-based data selection framework.

\section{Identifying the Worst-case Forget Set through BLO}
\label{sec: method}

 \noindent \textbf{A BLO view on the worst-case forget set identification for {\MU}.}
BLO (bi-level optimization) offers a hierarchical learning framework, featuring two tiers of optimization tasks, \textit{i.e.}, the upper and lower levels. In this structure, the objective and variables of the upper-level problem are contingent upon the solution of the lower-level problem.
In the context of identifying the worst-case forget set, we optimize the selection of a forget set at the \textit{upper} level to maximize the difficulty of unlearning. Concurrently, the \textit{lower} level is dedicated to generating the unlearned model, 
aiming to meet the unlearning objectives without compromising the utility on non-forgetting data points.

We introduce an optimization variable $\bw \in \{0, 1\}^{N}$, recalling that $N$ represents the total number of training data points. Here $w_i = 1$ signifies that the $i$-th training data point is included in the forget set, \textit{i.e.}, $\Df = \{\mathbf z_i | w_i = 1\}$.
Thus, our objective is to optimize the data selection scheme $\bw$, such that the associated $\Df$ can characterize the worst-case performance of an unlearned model, \textit{i.e.}, challenging the model $\thetau$ in \eqref{eq: unlearn_objective} post the unlearning of the designated forget set.

%The aforementioned learning objective aligns closely with the BLO principle. 
We first form the \textit{lower-level} optimization problem to determine the unlearned model $\thetau$ based on the forget set defined by $\bw$.
%We  first form the upper-level optimization problem to find the optimal data selection scheme $\bw$ that worsens the performance of the unlearned model.
%Simultaneously, a coupled lower-level {\MU} problem exists, tasked with determining the unlearned model $\thetau$ based on the forget set defined by $\bw$.
By integrating $\bw$ into \eqref{eq: unlearn_objective}, the unlearning problem in \textit{lower-level optimization} can be cast as
\begin{align}
   \displaystyle  \thetau(\mathbf w) = \argmin_{\btheta} ~ \ell_\MU(\btheta;  \bw) \Def  \sum_{
    \mathbf z_i \in \train}  [  w_i \ell_\mathrm{f}(\btheta; \mathbf z_i) + {(1 - w_i) \ell_\mathrm{r}(\btheta; \mathbf z_i )} ],
    \label{eq: lower_level_MU_loss}
\end{align}%
where $\thetau(\mathbf w)$ signifies the resulting  unlearned model that is a function of  $\bw$, and
the loss terms $  \sum_{
    \mathbf z_i \in \train} [w_i \ell_f(\btheta; \mathbf z_i)]$ and $ \sum_{
    \mathbf z_i \in \train} [{(1 - w_i) \ell_\mathrm{r}(\btheta; \mathbf z_i )} ]$ correspond to the forget loss and the retain loss  in \eqref{eq: unlearn_objective} on the forget set $\Df$ and the retain set $\Dr$, respectively. 
   Unless specified otherwise, we specify the unlearning objective \eqref{eq: lower_level_MU_loss} through the {\FT}-based unlearning strategy, with $\lambda  = 1$ and $\ell_\mathrm{f} = -\ell_\mathrm{r}$ in \eqref{eq: unlearn_objective}. Here, both loss functions are given by the training loss  $\ell$ (\textit{e.g.}, the cross-entropy loss for image classification) over $\btheta$, with the forget loss $\ell_\mathrm{f} = - \ell$ designed to counteract the training, thereby enforcing the unlearning.

With the unlearned model $\thetau(\mathbf w)$ defined as a function of the data selection scheme $\bw$, we proceed to outline the BLO framework by incorporating an \textit{upper-level} optimization. This is designed to optimize $\mathbf w$ for the worst-case unlearning performance, yielding the \textit{overall BLO problem}:
% 
% \vspace{-5.5mm}
% \begin{small}
\begin{align}
    \begin{array}{ll}
    \displaystyle \min_{\bw \in \gS} & 
     \underbrace{
   \displaystyle   \sum_{ \mathbf z_i \in  \train} [w_i \ell ( \thetau(\mathbf w); \mathbf z_i ) ] + \gamma \|\bw\|_2^2
     }_\text{Upper-level objective $\Def f (\mathbf w, \thetau(\bw))$} ; ~~
    % \vspace*{1mm}\\
    % \text{s.t.} &
    \st ~  %& ~
    \underbrace{
   \displaystyle  \thetau(\mathbf w) = \argmin_{\btheta} ~ \ell_\MU(\btheta;   \bw),
    }_\text{Lower-level optimization}
    \end{array}
    \label{eq: bilevel_MU}
\end{align}%
% \end{small}
% 
% \noindent
where $\bw$ is the upper-level optimization variable 
subject to the data selection constraint set $\gS$,  \textit{e.g.}, $\gS = \{\bw|\bw \in \{0, 1\}^N, \mathbf{1}^\top\bw = m \}$ with $m$ being the forget set size,   the lower-level objective function $\ell_\MU$ has been defined in \eqref{eq: lower_level_MU_loss}, and $\ell$ denotes the training loss. In addition, minimizing $ \sum_{ \mathbf z_i \in  \train} [w_i \ell ( \thetau(\mathbf w); \mathbf z_i ) ]$ renders the worst-case scenario of the unlearned model $\thetau(\mathbf w)$ (derived from the lower-level optimization), \textit{i.e.}, making it ineffective at erasing the influence of the forget set (corresponding to $\{w_i = 1\}$) on model performance. 
Furthermore, we introduce an $\ell_2$ regularization term  with the regularization parameter $\gamma \geq 0$ in the upper-level objective function. This has dual purposes: it encourages sparsity in the data selection scheme $\bw$ (when relaxed to continuous variables) and enhances the stability of  BLO  by including a strongly convex regularizer.

\noindent \textbf{A scalable BLO solver for worst-case forget set identification.}
%Next, we delve into the `implicit gradient (IG)' challenge inherent in solving the BLO problem  \eqref{eq: bilevel_MU}, and develop an efficient first-order BLO solver to circumvent this computational difficulty.
In \eqref{eq: bilevel_MU}, 
%given that the unlearned model $\thetau(\bw)$ is a function of $\bw$, 
addressing the upper-level optimization presents a significant complexity, as illustrated in the following gradient descent framework. Consider the gradient of the upper-level objective function of \eqref{eq: bilevel_MU},  $f(\mathbf w, \thetau(\mathbf w))$:
% 
% \vspace{-4.5mm}
% \begin{small}
\begin{align}
    \frac{d  f(\mathbf w, \thetau(\mathbf w)) }{d \mathbf w}
    = \nabla_{\mathbf w}f(\mathbf w, \thetau(\mathbf w)) +
    % \underbrace{
    \frac{d \thetau(\mathbf w)^\top}{d \mathbf w}
    % }_\text{IG}
    \nabla_{\btheta}f(\mathbf w, \btheta) |_{\btheta = \thetau(\mathbf w)},
    \label{eq: upper_gradient}
\end{align}%
% \end{small}
% 
% \vspace{-3mm}\noindent
where $\frac{d \cdot}{ d \mathbf w}$ denotes the \textit{full derivative} with respect to (\textit{w.r.t.}) $\mathbf w$, while $\nabla_{\bw} f$ and $\nabla_{\btheta} f$ represent the \textit{partial derivatives} of the bi-variate function $f$ \textit{w.r.t.}  $\bw$ and $\btheta$, respectively.
In \eqref{eq: upper_gradient}, the vector-wise full derivative $\frac{d \thetau(\mathbf w)^\top}{d \mathbf w}$
% (where $\top$ indicates transpose)
is commonly known as implicit gradient (IG) \cite{zhang2023introduction} since $\thetau(\mathbf w)$ is an implicit function of $\bw$, determined by lower-level optimization of  \eqref{eq: bilevel_MU}. 
Considering the difficulty of obtaining the closed-form $\thetau(\mathbf w)$,  the computation of IG introduces high complexity.

In the optimization literature,  two primary methods are used to derive the IG:
(1) The influence function (IF) approach \cite{zhang2023introduction,krantz2002implicit}, which leverages the stationarity condition of the lower-level objective function that  $\thetau(\mathbf w)$ satisfies; And (2) the gradient unrolling (GU) approach \cite{zhang2023introduction,shaban2019truncated}, which utilizes an unrolled version of a specific lower-level optimizer as an intermediate step, linking the solution of the lower-level problem to the upper-level optimization process.
However, the IF approach necessitates computing the \textit{inverse-Hessian} gradient product \cite{zhang2023introduction,krantz2002implicit} for the lower-level loss \textit{w.r.t.} $\btheta$. Consequently, it encounters scalability issues given the fact that $\btheta$ represents the parameters of a neural network.  
Therefore, we adopt GU to solve the BLO problem \eqref{eq: bilevel_MU}, as illustrated below. 

The GU strategy mainly contains two steps: 
\textbf{(S1)}
Identifying a particular lower-level optimizer to approximate the solution $\thetau(\mathbf w)$ through a finite sequence of unrolled optimization steps; And \textbf{(S2)} employing auto-differentiation to calculate the IG by tracing the solution path unfolded in (S1).
Consequently, the computation complexity of GU is dependent on the choice of the lower-level optimizer in (S1). 
In our study, we propose employing the \textit{sign}-based stochastic gradient descent (signSGD) \cite{bernstein2018signsgd} as the lower-level optimizer in (S1). As we will demonstrate, the adoption of signSGD greatly simplifies the computation of the IG.
Specifically, we derive an approximate solution of the lower-level problem by implementing a $K$-step unrolling using signSGD. This  yields
% 
% \vspace{-5.5mm}
% \begin{small}
\begin{align}
      \thetau(\mathbf w) =  \btheta_K; \quad 
  \btheta_j =   \btheta_{j-1} - \beta \cdot \mathrm{sign}(\nabla_{\btheta} 
  \ell_\mathrm{MU} (\btheta_{j-1};  \mathbf w)
  ), ~  j = 1, 2, \ldots, K,
     \label{eq: sign_SGD_unrolling}
\end{align}%
% \end{small}%
where $j$ represents the lower-level iteration index, $\mathrm{sign}(\cdot)$ is the element-wise sign operation, $\beta > 0$ specifies the learning rate, and $\btheta_{0}$ is a random initialization. Given  signSGD  \eqref{eq: sign_SGD_unrolling}, the computation of IG can be simplified to 
% 
% \vspace{-3mm}
% \begin{small}
\begin{align}
  \mathrm{IG} =  \frac{d \thetau (\mathbf w)^\top}{d \mathbf w} = \frac{d \btheta_{j-1}^\top}{d \mathbf w} - \beta \frac{ d \mathrm{sign}(\nabla_{\btheta} 
  \ell_\mathrm{MU} (\btheta_{j-1}; \mathbf w)
  )^\top}{d \mathbf w} = \ldots = \frac{d \btheta_{0}^\top}{d \mathbf w} = \mathbf 0,
  \label{eq: signIG}
\end{align}%
% \end{small}%
where we used the facts that $ \frac{d \mathrm{sign}(\mathbf x)^\top}{d \mathbf x} = \mathbf 0$ which holds almost surely and $\btheta_{0}$ is a random initialization. 
We highlight that this IG simplification is induced by the sign operation. If we replace signSGD with the vanilla SGD, then  the intermediate second-order derivatives, such as $\frac{d \nabla_{\btheta} 
  \ell_\mathrm{MU} (\btheta_{j-1};  \mathbf w)}{d \mathbf w}$,  would not be omitted.

By utilizing signSGD, we could effectively address the computational challenge of calculating the IG, enabling us to solve the problem  \eqref{eq: bilevel_MU}  using solely first-order information. For example, 
substituting \eqref{eq: signIG} into \eqref{eq: upper_gradient}, the upper-level gradient \textit{w.r.t.} $\mathbf w$ reduces to the first-order partial derivative $\nabla_{\mathbf w}f(\mathbf w, \thetau(\mathbf w))$. Subsequently, 
we can solve the BLO problem \eqref{eq: bilevel_MU} for identifying the worst-case forget set through an \textit{alternating optimization} strategy, formed by projected gradient descent (PGD) for the upper-level optimization and signSGD for the lower-level optimization. We summarize it below:
% 
% \vspace{-4mm}
% \begin{small}
\begin{align}
 &   \text{Upper-level PGD:} ~ \mathbf w_{i} = \mathrm{Proj}_{\mathbf w \in \gS} \left ( \mathbf w_{i-1} - \alpha \nabla_{\mathbf w}f(\bw, \thetau(\mathbf w_{i-1})) |_{\mathbf w = \mathbf w_{i-1}} \right ), \label{eq: U-PGD}\\
  &  \text{Lower-level signSGD:} ~  \thetau(\mathbf w_{i-1}) =  \btheta_K, 
  ~ \text{given by \eqref{eq: sign_SGD_unrolling} at $\mathbf w = \mathbf w_{i-1}$}, \label{eq: L-signSGD}
\end{align}%
% \end{small}%
where $i$ represents the step in the upper-level optimization process, $\mathbf w_0$ is an initial data selection scheme (\textit{e.g.}, a random binary vector in classification tasks),
and $\mathrm{Proj}_{\mathbf w \in \gS} (\mathbf a)$ indicates the projection of a constant $\mathbf a$ onto the constraint set $\gS$. This projection operation is defined as solving the auxiliary minimization problem $\mathrm{Proj}_{\mathbf w \in \gS} (\mathbf a) = \argmin_{\mathbf w \in \gS} \| \mathbf w - \mathbf a \|_2^2$. 
For ease of optimization, we relax the binary constraint $\gS$ into its continuous counterpart,  with $\mathbf w \in [\mathbf 0, \mathbf 1] $ and $\mathbf 1^\top \bw = m$. 
This facilitates us to obtain a closed-form solution for this projection problem, as shown in  Appendix\,\ref{app: closed_form_derivations}. \nocite{boyd2004convex}
Additionally, a relaxed version of $\mathbf w$ offers a continuous forget score, enabling us to identify not only the worst-case forget set (identified by selecting $\mathbf w$ with the top $m$ largest magnitudes) but also the set of data points that are easiest to unlearn  (identified by selecting $\mathbf w$ with the top $m$ smallest magnitudes).
In practice, our alternating PGD  and signSGD method  \eqref{eq: U-PGD}-\eqref{eq: L-signSGD}  demonstrates good convergence properties. Typically, the upper-level optimization converges within {20} iterations, while for the lower-level problem, setting {$K = 10$} epochs is found to be sufficient.

\noindent \textbf{Extending to class-wise  or prompt-wise forgetting.}
The previously proposed BLO problem \eqref{eq: bilevel_MU} was conceived for identifying worst-case forget set in the context of data-wise forgetting. However,
our approach can be easily extended to other {\MU} scenarios, such as \textit{class-wise forgetting}
\cite{jia2023model,golatkar2020eternal,graves2021amnesiac},
 and \textit{prompt-wise forgetting}
% (\textit{i.e.}, preventing a generative model from producing content conditioned on an undesirable text prompt) 
\cite{fan2023salun,gandikota2023erasing}.
For class-wise forgetting, the data selection variables $\bw$ in \eqref{eq: lower_level_MU_loss}-\eqref{eq: bilevel_MU} can be reinterpreted as class selection variables. Here, $w_i = 1$ indicates the selection of the $i$th class for targeted worst-case unlearning. In the context of prompt-wise forgetting, we interpret $\bw$ as prompt selection variables. Here a prompt refers to a text condition used for text-to-image generation, known as a `concept' within {\MU}  for generative models  \cite{gandikota2023erasing}. See Appendix\,\ref{app: data_prompt_closed_form_derivations} for details.

\section{Experiments}
\label{sec: exp}
\subsection{Experiment Setups}
\label{sec: exp_setup}

\noindent 
\textbf{Unlearning tasks and setups.} 
For \textit{data-wise} forgetting, our primary experiments are conducted on the CIFAR-10 dataset \cite{krizhevsky2009learning} and the ResNet-18 model \cite{he2016deep}. We further extend our evaluation to include CIFAR-100 \cite{krizhevsky2009learning}, CelebA \cite{liu2015deep}, and Tiny ImageNet datasets, alongside VGG \cite{simonyan2014very} and ResNet-50 \cite{he2016deep}, detailed in {Appendix\,\ref{app: data_experiment_setup}}.
For \textit{class-wise} forgetting, we focus on the ImageNet \cite{deng2009imagenet} dataset with the ResNet-18 model.  For \textit{prompt-wise} forgetting, we consider the unlearning task of preventing the latent diffusion model \cite{rombach2022high} from generating artistic painting styles alongside image objects within the UnlearnCanvas dataset \cite{zhang2024unlearncanvas}.

\noindent 
\textbf{Unlearning methods and evaluation metrics.} 
To validate the efficacy of {\MU} evaluation on the worst-case forget set, we examine the exact unlearning method {\Retrain} and $9$ approximate unlearning methods as described in Table\,\ref{tab: mu_method_summary}.
During the evaluation, we mainly adopt $4$ performance metrics as introduced in Sec.\,\ref{sec: preliminary}: 
\textit{UA} and \textit{MIA} are used to measure the unlearning effectiveness, \textit{RA} and \textit{TA} are used to assess the model utility post unlearning.

\noindent 
\textbf{BLO implementation.} 
In the upper-level optimization, we set the regularization parameter $\gamma$ in \eqref{eq: bilevel_MU} to $10^{-4}$, and applied PGD by \eqref{eq: U-PGD} with $20$ iterations at the learning rate $\alpha=10^{-3}$.
In the lower-level optimization, SignSGD is performed with 10 epochs. 
% \CF{This optimization process for worst-case forget set selection can be performed offline. Once the worst-case set is identified, the resulting unlearning challenge is method- and model-agnostic.} \JC{[delete.]} 
% See Appendix\,\ref{app: data_experiment_setup} for more implementation details.

\subsection{Experiment Results}
\label{sec: exp_results}

 \begin{table}[htb]
\vspace*{-8mm}
\caption{Performance of exact unlearning ({\Retrain}) under {\randomset} and {\ourset} at different forgetting data ratios on CIFAR-10 using ResNet-18. The result format is given by $a_{\pm b}$, with mean $a$ and standard deviation $b$  over $10$ independent trials. The performance difference is provided in Diff, represents the worst-case performance is \textcolor{Green}{lower$\blacktriangledown$}, \textcolor{Blue}{equal$-$}, or \textcolor{Red}{higher$\blacktriangle$} than random-case performance.
}
\label{tab: main_retrain_ratio}
\begin{center}
\vspace*{-6mm}
\resizebox{0.95\textwidth}{!}{

\renewcommand\tabcolsep{5pt}
\begin{tabular}{c|ccc|ccc|ccc|ccc}
\toprule[1pt]
\midrule
\multirow{2}{*}{\textbf{Metrics}} & \multicolumn{3}{c|}{\textbf{1\%-Data Forgetting}} & \multicolumn{3}{c|}{\textbf{5\%-Data Forgetting}} & \multicolumn{3}{c|}{\textbf{10\%-Data Forgetting}} & \multicolumn{3}{c}{\textbf{20\%-Data Forgetting}} \\
 & \multicolumn{1}{c|}{Random} & \multicolumn{1}{c|}{Worst-case} & Diff& \multicolumn{1}{c|}{Random} & \multicolumn{1}{c|}{Worst-case}& Diff & \multicolumn{1}{c|}{Random} & \multicolumn{1}{c|}{Worst-case} & Diff& \multicolumn{1}{c|}{Random} & \multicolumn{1}{c|}{Worst-case} & Diff\\
\midrule
UA & $5.85_{\pm 0.69}$ & $0.00_{\pm 0.00}$ & \textcolor{Green} {$ 5.85 \blacktriangledown$} & $5.92_{\pm 0.44}$ & $0.00_{\pm 0.00}$ & \textcolor{Green} {$ 5.92 \blacktriangledown$} & $5.28_{\pm 0.33}$ & $0.00_{\pm 0.00}$ & \textcolor{Green} {$ 5.28 \blacktriangledown$} & $5.76_{\pm 0.20}$ & $0.00_{\pm 0.00}$ & \textcolor{Green} {$ 5.76 \blacktriangledown$}\\
\rowcolor{Gray}
MIA & $12.89_{\pm 1.27}$ & $0.00_{\pm 0.00}$ & \textcolor{Green} {$ 12.89 \blacktriangledown$} & $13.00_{\pm 0.55}$ & $0.02_{\pm 0.02}$ & \textcolor{Green} {$ 12.98 \blacktriangledown$} & $12.86_{\pm 0.61}$ & $0.00_{\pm 0.00}$ & \textcolor{Green} {$ 12.86 \blacktriangledown$} & $14.34_{\pm 0.40}$ & $0.03_{\pm 0.01}$ & \textcolor{Green} {$ 14.31 \blacktriangledown$}\\
RA & $99.96_{\pm 0.00}$ & $99.95_{\pm 0.02}$ & \textcolor{Green} {$ 0.01 \blacktriangledown$} & $100.00_{\pm 0.00}$ & $100.00_{\pm 0.00}$ & \textcolor{Blue}{$ 0.00 - $} & $100.00_{\pm 0.00}$ & $100.00_{\pm 0.00}$ & \textcolor{Blue}{$ 0.00 - $} & $100.00_{\pm 0.00}$ & $100.00_{\pm 0.00}$ & \textcolor{Blue}{$ 0.00 - $}\\
\rowcolor{Gray}
TA & $93.17_{\pm 0.15}$ & $93.45_{\pm 0.17}$ & \textcolor{Red} {$ 0.28 \blacktriangle$} & $94.51_{\pm 0.07}$ & $94.67_{\pm 0.08}$ & \textcolor{Red} {$ 0.16 \blacktriangle$} & $94.38_{\pm 0.15}$ & $94.66_{\pm 0.09}$ & \textcolor{Red} {$ 0.28 \blacktriangle$} & $94.04_{\pm 0.08}$ & $94.60_{\pm 0.08}$ & \textcolor{Red} {$ 0.56 \blacktriangle$}\\
\midrule
\bottomrule[1pt]
\end{tabular}
}

\end{center}
\vspace*{-8mm}
\end{table}

\noindent \textbf{Validating the worst-case forget set via {\Retrain}.}
We begin by justifying the worst-case unlearning performance of the chosen forget set through the exact unlearning method, {\Retrain}.
% \CF{Lower UA and MIA indicate that it is easier to predict the forgotten data allowing more forgotten data to be imputed in the training set, suggesting this data is more difficult to unlearn.} \JC{[delete.]}
In \textbf{Table\,\ref{tab: main_retrain_ratio}}, we examine the performance disparities between the \textit{\ourset} and the \textit{\randomset} in the task of {\MU} for image classification on CIFAR-10, when employing {\Retrain} at different  forgetting data ratios including $1\%$, $5\%$, $10\%$, and $20\%$.
 In terms of unlearning effectiveness, the chosen worst-case forget set consistently poses the greatest challenge for unlearning in all scenarios tested, as indicated by a significant drop in UA and MIA to nearly 0\% (see the `Worst-case' and `Diff' columns of Table\,\ref{tab: main_retrain_ratio}). 
 In addition, the variance in worst-case unlearning effectiveness performance (as measured by UA and MIA) remains significantly lower than that observed with random data forgetting at various forgetting data ratios. Furthermore, the utility of the unlearned model, as indicated by RA and TA, shows no loss when comparing unlearning on worst-case forget sets to random forget sets. Intriguingly, the TA of models unlearned with the worst-case forget set may even surpass those unlearned with random sets, hinting at a connection to coreset selection that will be further explored later. 
 In Appendix\,\ref{app: transferability},  we broaden our evaluation of the worst-case forget set across additional datasets and model architectures. The findings align with those presented in Table\,\ref{tab: main_retrain_ratio}.

\noindent 
\textbf{Reassessing  approximate unlearning under {\ourset}.}
Next, we examine the performance of approximate unlearning methods (\FT, \EUk, \CFk, \Scrub, \MUSparse, \RL, \BE, \BS, and \Salun) when applied to the {\ourset} scenario.
Ideally, an effective and robust approximate unlearning method should mirror the trend of {\Retrain}, \textit{i.e.}, maintaining a minimal performance discrepancy with exact unlearning.
However, evaluations using worst-case forget sets can reveal performance disparities in approximate unlearning (vs. {\Retrain}), which differ from those observed with random forget sets.

\begin{table}[t]
\caption{
Performance of approximate unlearning methods (including both relabeling-free and relabeling-based methods) under {\randomset}s and {\ourset}s on CIFAR-10 using ResNet-18 with forgetting ratio 10\%.
The result format follows Table\,\ref{tab: main_retrain_ratio}. Additionally, a performance gap against \textcolor{blue}{{\Retrain}} is provided in (\textcolor{blue}{$\bullet$}). 
The metric \textit{averaging (avg.) gap} is calculated by averaging the performance gaps measured in all metrics.
Note that the better performance of an MU method corresponds to the smaller performance gap with {\Retrain}.
}
\vspace*{-6.5mm}
\label{tab: main_approximate_unlearn}
\begin{center}
\resizebox{\textwidth}{!}{

\renewcommand\tabcolsep{5pt}

\begin{tabular}{c|ccccc|ccccc}
\toprule[1pt]
\midrule
\multirow{2}{*}{\textbf{Methods}} & \multicolumn{5}{c|}{\textbf{\RandomSet}}  & \multicolumn{5}{c}{\textbf{\OurSet}}  \\
                        & \multicolumn{1}{c|}{UA}   & \multicolumn{1}{c|}{MIA}     & \multicolumn{1}{c|}{RA} & \multicolumn{1}{c|}{TA} & \multicolumn{1}{c|}{Avg. Gap} 
                        & \multicolumn{1}{c|}{UA}   & \multicolumn{1}{c|}{MIA}     & \multicolumn{1}{c|}{RA}    & \multicolumn{1}{c|}{TA} & \multicolumn{1}{c}{Avg. Gap} \\
\midrule
{{\Retrain}} & $5.28_{\pm 0.33}$ & $12.86_{\pm 0.61}$ & $100.00_{\pm 0.00}$ & $94.38_{\pm 0.15}$ & \textcolor{blue}{0.00} & $0.00_{\pm 0.00}$ & $0.00_{\pm 0.00}$ & $100.00_{\pm 0.00}$ & $94.66_{\pm 0.09}$ & \textcolor{blue}{0.00}\\
\midrule
\rowcolor{Gray}
\multicolumn{11}{c}{\textbf{Relabeling-free}} \\
\midrule
{{\FT}} & $5.08_{\pm 0.39}$ (\textcolor{blue}{0.20}) & $10.96_{\pm 0.38}$ (\textcolor{blue}{1.90}) & $97.46_{\pm 0.52}$ (\textcolor{blue}{2.54}) & $91.02_{\pm 0.36}$ (\textcolor{blue}{3.36}) & \textcolor{blue}{2.00} & $0.00_{\pm 0.00}$ (\textcolor{blue}{0.00}) & $0.02_{\pm 0.03}$ (\textcolor{blue}{0.02}) & $97.63_{\pm 0.46}$ (\textcolor{blue}{2.37}) & $91.58_{\pm 0.40}$ (\textcolor{blue}{3.08}) & \textcolor{blue}{1.37}\\
\rowcolor{Gray}
{{\EUk}} & $2.34_{\pm 0.79}$ (\textcolor{blue}{2.94}) & $6.35_{\pm 0.89}$ (\textcolor{blue}{6.51}) & $97.52_{\pm 0.89}$ (\textcolor{blue}{2.48}) & $90.17_{\pm 0.88}$ (\textcolor{blue}{4.21}) & \textcolor{blue}{4.04} & $0.68_{\pm 0.56}$ (\textcolor{blue}{0.68}) & $5.02_{\pm 4.42}$ (\textcolor{blue}{5.02}) & $97.17_{\pm 0.86}$ (\textcolor{blue}{2.83}) & $90.08_{\pm 0.70}$ (\textcolor{blue}{4.58}) & \textcolor{blue}{3.28}\\
{{\CFk}} & $0.02_{\pm 0.02}$ (\textcolor{blue}{5.26}) & $0.76_{\pm 0.02}$ (\textcolor{blue}{12.10}) & $99.98_{\pm 0.00}$ (\textcolor{blue}{0.02}) & $94.45_{\pm 0.02}$ (\textcolor{blue}{0.07}) & \textcolor{blue}{4.36} & $0.00_{\pm 0.00}$ (\textcolor{blue}{0.00}) & $0.00_{\pm 0.00}$ (\textcolor{blue}{0.00}) & $99.98_{\pm 0.01}$ (\textcolor{blue}{0.02}) & $94.34_{\pm 0.05}$ (\textcolor{blue}{0.32}) & \textcolor{blue}{0.08}\\
\rowcolor{Gray}
{{\Scrub}} & $12.42_{\pm 19.82}$ (\textcolor{blue}{7.14}) & $22.43_{\pm 24.44}$ (\textcolor{blue}{9.57}) & $88.31_{\pm 19.78}$ (\textcolor{blue}{11.69}) & $83.15_{\pm 17.94}$ (\textcolor{blue}{11.23}) & \textcolor{blue}{9.91} & $0.01_{\pm 0.01}$ (\textcolor{blue}{0.01}) & $0.04_{\pm 0.03}$ (\textcolor{blue}{0.04}) & $98.65_{\pm 0.33}$ (\textcolor{blue}{1.35}) & $92.78_{\pm 0.30}$ (\textcolor{blue}{1.88}) & \textcolor{blue}{0.82}\\
{{\MUSparse}} & $4.34_{\pm 0.73}$ (\textcolor{blue}{0.94}) & $9.82_{\pm 1.04}$ (\textcolor{blue}{3.04}) & $97.70_{\pm 0.72}$ (\textcolor{blue}{2.30}) & $91.41_{\pm 0.68}$ (\textcolor{blue}{2.97}) & \textcolor{blue}{2.31} & $0.02_{\pm 0.03}$ (\textcolor{blue}{0.02}) & $0.11_{\pm 0.11}$ (\textcolor{blue}{0.11}) & $96.93_{\pm 0.73}$ (\textcolor{blue}{3.07}) & $90.96_{\pm 0.82}$ (\textcolor{blue}{3.70}) & \textcolor{blue}{1.72}\\
\midrule
\rowcolor{Gray}
\multicolumn{11}{c}{\textbf{Relabeling-based}} \\
\midrule
{{\RL}} & $3.59_{\pm 0.24}$ (\textcolor{blue}{1.69}) & $28.02_{\pm 2.47}$ (\textcolor{blue}{15.16}) & $99.97_{\pm 0.01}$ (\textcolor{blue}{0.03}) & $93.74_{\pm 0.12}$ (\textcolor{blue}{0.64}) & \textcolor{blue}{4.38} & $1.93_{\pm 1.5}$ (\textcolor{blue}{1.93}) & $96.70_{\pm 0.66}$ (\textcolor{blue}{96.70}) & $99.96_{\pm 0.01}$ (\textcolor{blue}{0.04}) & $93.83_{\pm 0.24}$ (\textcolor{blue}{0.83}) & \textcolor{blue}{24.88}\\
\rowcolor{Gray}
{{\BE}} & $1.19_{\pm 0.49}$ (\textcolor{blue}{4.09}) & $22.06_{\pm 0.61}$ (\textcolor{blue}{9.20}) & $98.77_{\pm 0.41}$ (\textcolor{blue}{1.23}) & $91.79_{\pm 0.32}$ (\textcolor{blue}{2.59}) & \textcolor{blue}{4.28} & $19.47_{\pm 2.12}$ (\textcolor{blue}{19.47}) & $81.45_{\pm 5.16}$ (\textcolor{blue}{81.45}) & $81.35_{\pm 2.76}$ (\textcolor{blue}{18.65}) & $75.41_{\pm 1.77}$ (\textcolor{blue}{19.25}) & \textcolor{blue}{34.70}\\
{{\BS}} & $5.72_{\pm 1.42}$ (\textcolor{blue}{0.44}) & $27.15_{\pm 1.41}$ (\textcolor{blue}{14.29}) & $94.29_{\pm 1.06}$ (\textcolor{blue}{5.71}) & $87.45_{\pm 1.06}$ (\textcolor{blue}{6.93}) & \textcolor{blue}{6.84} & $29.75_{\pm 6.39}$ (\textcolor{blue}{29.75}) & $74.88_{\pm 3.13}$ (\textcolor{blue}{74.88}) & $78.34_{\pm 0.98}$ (\textcolor{blue}{21.66}) & $72.07_{\pm 1.25}$ (\textcolor{blue}{22.59}) & \textcolor{blue}{37.22}\\
\rowcolor{Gray}
{{\Salun}} & $1.48_{\pm 0.14}$ (\textcolor{blue}{3.80}) & $16.19_{\pm 0.34}$ (\textcolor{blue}{3.33}) & $99.98_{\pm 0.01}$ (\textcolor{blue}{0.02}) & $93.95_{\pm 0.01}$ (\textcolor{blue}{0.43}) & \textcolor{blue}{1.89} & $0.96_{\pm 0.59}$ (\textcolor{blue}{0.96}) & $96.43_{\pm 0.33}$ (\textcolor{blue}{96.43}) & $99.98_{\pm 0.01}$ (\textcolor{blue}{0.02}) & $94.03_{\pm 0.08}$ (\textcolor{blue}{0.63}) & \textcolor{blue}{24.51}\\
\midrule
\bottomrule[1pt]
\end{tabular}
}
\end{center}
\vspace*{-11mm}
\end{table}

In \textbf{Table\,\ref{tab: main_approximate_unlearn}}, we present the performance of approximate unlearning methods under both random and worst-case forget sets, with the forgetting data ratio 10\%.
For comparison, we also include the performance of {\Retrain} and analyze the performance gap between approximate unlearning methods and {\Retrain} (see the column `\textit{Avg. Gap}'). Note that an ideal approximate unlearning method is expected to have a smaller performance gap with {\Retrain}.
Following Sec.\,\ref{sec: preliminary}, we categorize approximate unlearning methods into two groups: relabeling-free and relabeling-based, respectively. 
We highlight two main observations from Table\,\ref{tab: main_approximate_unlearn}.

\textit{First},  relabeling-free approximate unlearning methods ({\FT}, {\EUk}, {\CFk}, {\Scrub}, {\MUSparse}) generally follow the trend of {\Retrain} when evaluated on worst-case forget sets (\textit{i.e.}, inadequately erasure the influence of the worst-case forget set). This is evidenced by a significant decrease in UA and MIA, along with a narrowing performance gap with {\Retrain}. Moreover, consistent with  Table\,\ref{tab: main_retrain_ratio}, evaluations using the worst-case forget set reduce the high variance encountered in random data forgetting, as demonstrated by the {\Scrub} method.

\textit{Second}, different from relabeling-free methods, relabeling-based approximate unlearning ({\RL}, {\BE}, {\BS}, {\Salun}) 
exhibit a significant performance gap compared to {\Retrain}.
This gap is highlighted by an increase in {UA} for methods like {\BE} and {\BS} when applied to worst-case forget sets,  diverging from the performance of {\Retrain}.
A similar discrepancy is noted in {MIA}.
Crucially, all relabeling-based approaches seem to offer a \textit{false} sense of unlearning effectiveness under {\ourset}, as evidenced by a marked rise in MIA, suggesting the inefficacy of relabeling-based strategies in this context. 
%{\ourset} encapsulates an inherently more predictable subset of the training dataset, potentially inferable even when not directly observed during the training phase. This suggests a latent correlation with other training set data points. 
Recall that relabeling-based methods achieve unlearning by explicitly negating the forget set through relabeling. Therefore, this approach can cause significant changes in model behavior, especially when the forget set mainly consists of common-case data points.

\begin{wrapfigure}{r}{0.491\textwidth}
\centering
\vspace*{-8mm}
% \hspace*{-4mm}
\begin{tabular}{cc}
\includegraphics[width=0.230\textwidth]{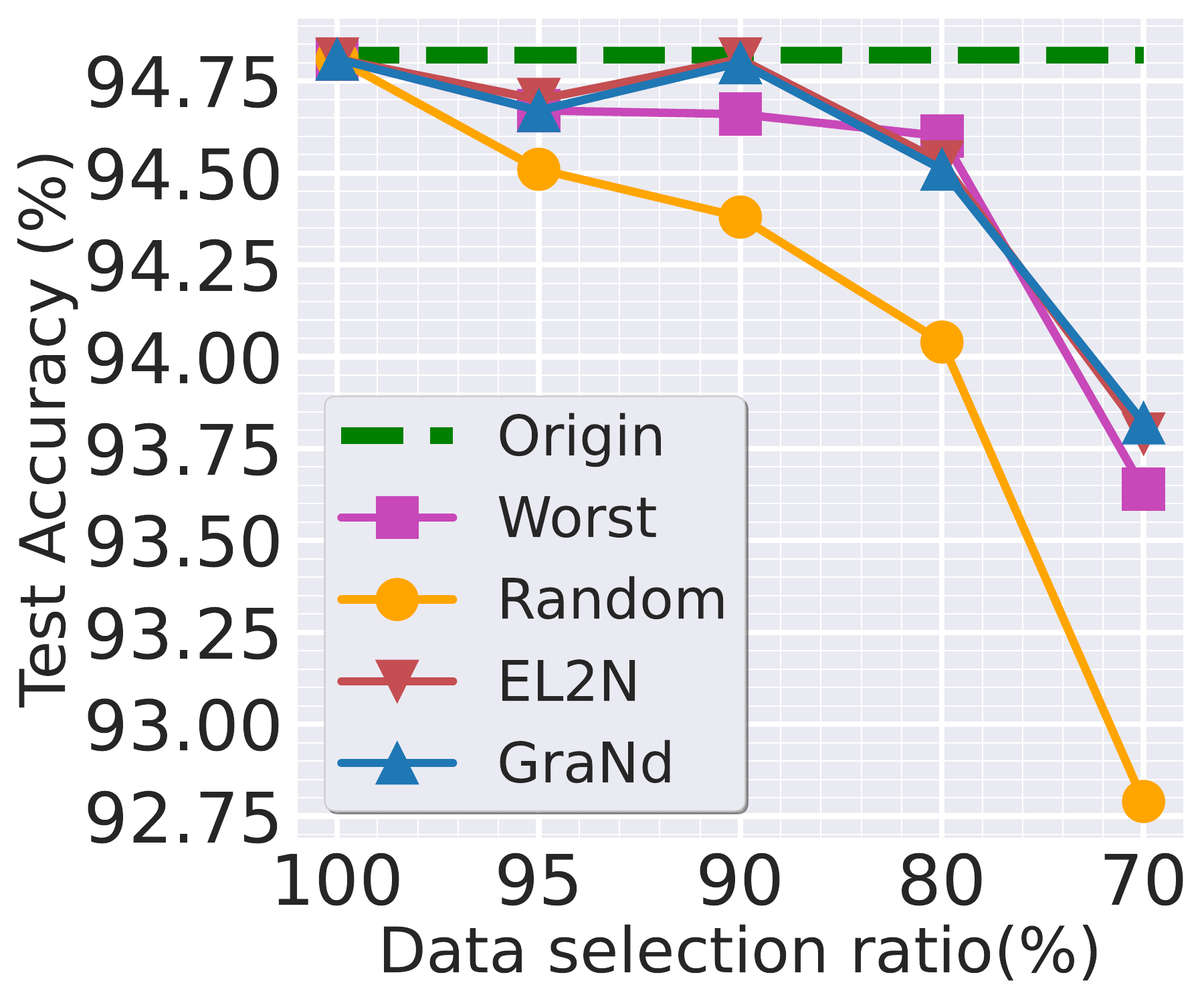} &
\includegraphics[width=0.230\textwidth]{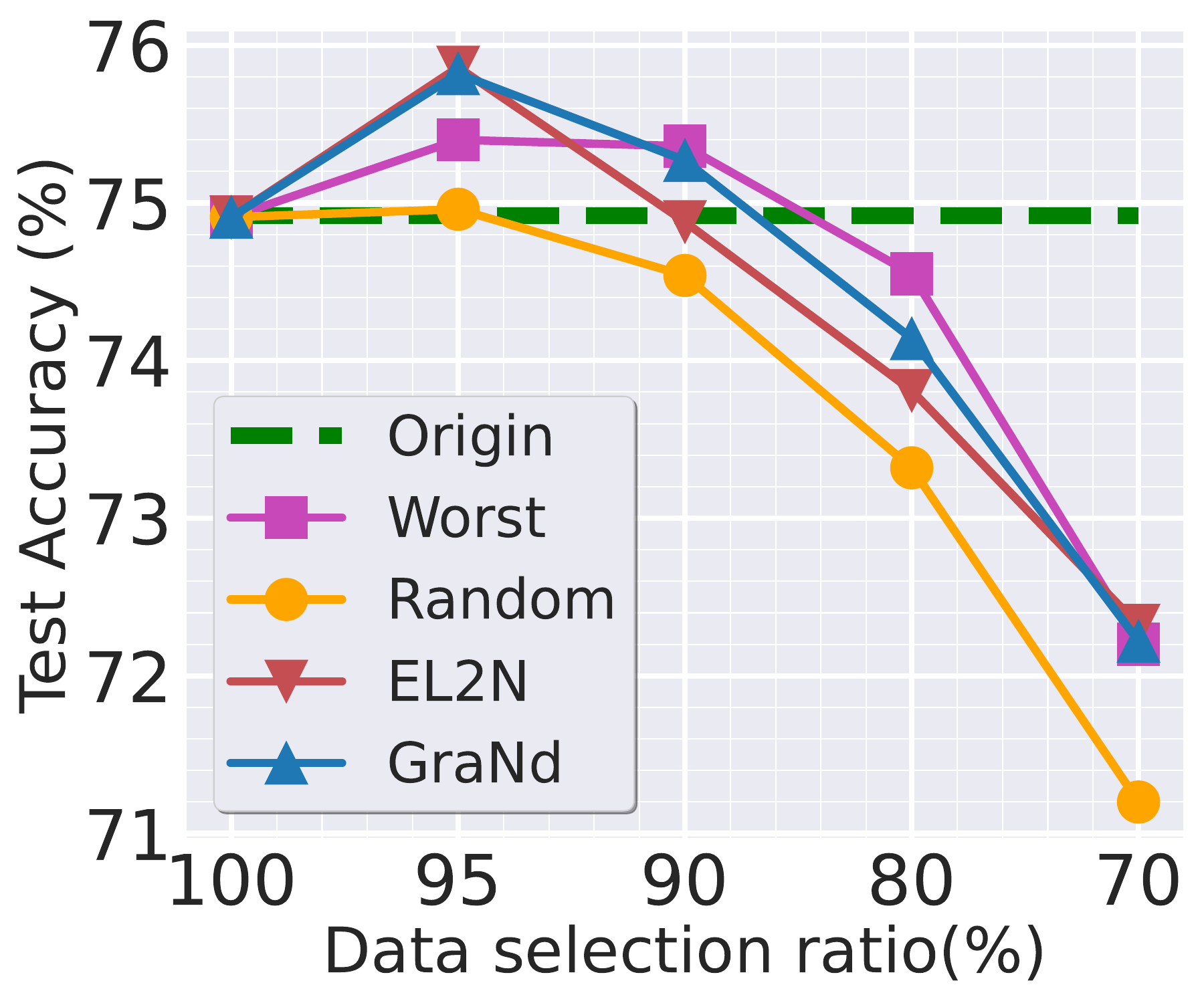} \\
\scriptsize{(a) CIFAR-10} & \scriptsize{(b) CIFAR-100}
\end{tabular}

\vspace{-2.5mm}
\caption{Performance of ResNet-18 trained on coresets of  (a) 
 CIFAR-10 and (b) CIFAR-100, determined by different approaches, including the complement of {\ourset} (Worst), random select (Random), EL2N and {GraNd}, at varying coreset ratios. The dashed line represents the model's performance trained on the full dataset (Origin).}
\vspace*{-7mm}
\label{fig: coreset}
\end{wrapfigure}
\noindent
\textbf{Analyzing the selected {\ourset} through a {coreset} lens.}
Our prior observation from Table\,\ref{tab: main_retrain_ratio}  indicates that TA of models unlearned with {\ourset}s could surpass those subjected to {\randomset}s.
This prompts us to investigate the characteristics of data typically selected within the worst-case forget sets.
In what follows, we explore this question through the lens of \textit{coreset},  given by a subset of the training data deemed sufficient for model training \cite{mirzasoleiman2020coresets,borsos2020coresets,zhang2024selectivity}. 
%This observation raises a pertinent question: \textit{Does the dataset, after the exclusion of {\ourset}, serve as an efficacious subset for model training?}
We hypothesize that the data points included in the worst-case forget set likely do \textit{not} constitute the coreset, as forgetting them poses challenges, possibly owing to their strong inherent correlation with data not selected for forgetting.
Motivated by this hypothesis, we explore whether the \textit{complement} of the worst-case forget set corresponds to the \textit{coreset}.
% To answer the above question, \textbf{Fig.\,\ref{fig: coreset}} contrasts the testing accuracy of an image classifier (ResNet-18) trained on the complement of the worst-case forget set (termed `Worst') against that trained on coresets identified through the moderate coreset selection method (termed `Moderate') \cite{xia2022moderate}.
% For comparison, we also present the performance using the model trained on the original full dataset (termed `Original') and the random set (termed `Random'). 
% As we can see, for both CIFAR-10 and CIFAR-100 datasets, training models on the complement of the worst-case forget set achieve superior TA across various data selection ratios, outperforming both random data selection and the coreset selection method, Moderate.
%
%\CF{
%To answer the above question, 
\textbf{Fig.\,\ref{fig: coreset}} contrasts the testing accuracy of an image classifier (ResNet-18) trained on the complement of the worst-case forget set (termed `Worst') against that trained on coresets identified through other coreset selection methods, including \textbf{EL2N} and \textbf{GraNd} \cite{paul2021deep}.
For comparison, we also present the performance using the model trained on the original full dataset (termed `Original') and the random set (termed ‘Random’). 
Training models on the complement of the worst-case forget set achieve TA comparable to the state-of-the-art coreset selection methods across various data selection ratios for both CIFAR-10 and CIFAR-100 datasets.
Furthermore, TA achieved by using the complement of the worst-case forget set on CIFAR-100 can even exceed the performance of the original model trained on the entire dataset at $90\%$ and $95\%$ selection ratios (\textit{i.e.}, $10\%$ and $5\%$ forgetting ratios). These observations suggest that the chosen worst-case forget set indeed does \textit{not} constitute a coreset, whereas its complement serves as a coreset.
From the coreset analysis perspective, identifying the worst-case forget set not only addresses the most challenging data to forget but also offers a method to \textit{attribute data influence} in model training based on their `unlearning difficulty' levels.

\begin{wrapfigure}{r}{0.25\textwidth}
\centering
% \vspace*{-8mm}
\includegraphics[width=0.25\textwidth]{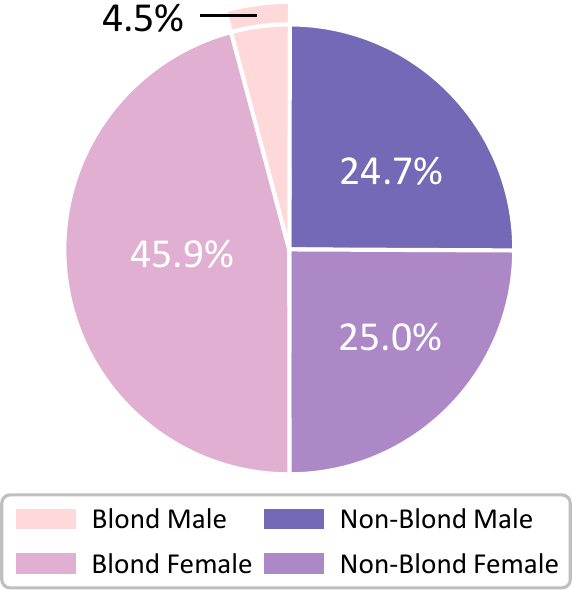}
\vspace*{-6mm}
\caption{
Composition of the {\ourset}  under CelebA.
}
\vspace*{-10mm}
\label{fig: ratios_celeba}
\end{wrapfigure}
\noindent 
\textbf{Case study: Selecting the worst-case forget set in a biased dataset.} 
The previous experiment results suggest that the identified worst-case forget set corresponds to the complement of the coreset (\textit{i.e.}, the non-coreset). We further 
explore this intriguing finding through a case study, identifying the worst-case forget set in a biased dataset created from CelebA \cite{liu2015deep}.
% \SL{
We consider this dataset for hair color prediction (Blond vs. Non-Blond), with
a spurious correlation with 
the `gender' attribute (Male vs. Female) \cite{sagawa2020investigation}.
% We wonder if the majority of the worst-case forget set comprises biased data points for hair color prediction (Blonde vs. Non-Blonde), which may exhibit a spurious correlation with the `gender' attribute (Male vs. Female) of the data \cite{sagawa2020investigation}.
\textbf{Fig.\,\ref{fig: ratios_celeba}} presents the composition 
of the selected worst-case forget set (with the data forgetting ratio 10\%), categorized into four groups based on the combination of the label and the `gender' attribute.
As we can see, within the chosen worst-case forget set, there exists a large portion of data points associated with (Blond + Female). 
It's worth noting that in CelebA, blonde hair is commonly correlated with females, making data points in the (Blond + Female) group relatively \textit{easy to learn}, acting as a non-coreset if forgetting part of the data points from this group.

\begin{figure}[htb]
\vspace*{-5mm}
\centering
\begin{minipage}{.49\textwidth}
\centering
\resizebox{\textwidth}{!}{
    \begin{tabular}{c|c|c}
        \toprule[1pt]
        \midrule
        \multirow{2}{*}{\scriptsize{\textbf{Model}}} & \multicolumn{2}{c}{\scriptsize{\textbf{Generation Condition}}} \\ 
        & \multicolumn{1}{c|}{\scriptsize{\Puw\texttt{:}
        % \texttt{ Dogs +  Rust}
        \begin{tabular}[c]{@{}c@{}}
        \vspace{-0.5mm}\tiny{\texttt{\textit{A painting of }\textcolor{Red}{Dogs}}} \\
        \tiny{\texttt{\textit{in} \textcolor{Blue}{Rust}\textit{ Style.}}}
        \end{tabular}
        }} 
        & \multicolumn{1}{c}{\scriptsize{\Pn\texttt{:}
        % \texttt{Trees + Ukiyoe}
        \begin{tabular}[c]{@{}c@{}}
        \vspace{-0.5mm}\tiny{\texttt{\textit{A painting of} \textcolor{Red}{Trees}}} \\
        \tiny{\texttt{\textit{in} \textcolor{Blue}{Ukiyoe}\textit{ Style.}}}
        \end{tabular}
        }}\\
        \midrule

        \begin{tabular}[c]{@{}c@{}}
        \vspace{-1.5mm}\tiny{\textbf{Original}} \\
        \vspace{-1.5mm}\tiny{\textbf{Diffusion}} \\
        \vspace{1mm}\tiny{\textbf{Model}}
        \end{tabular}
        &  
          \multicolumn{1}{m{0.49\textwidth}|}{ \includegraphics[width=0.235\textwidth]{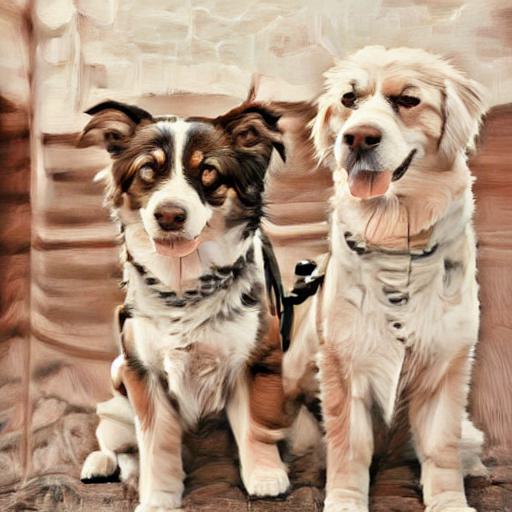} 
         \includegraphics[width=0.235\textwidth]{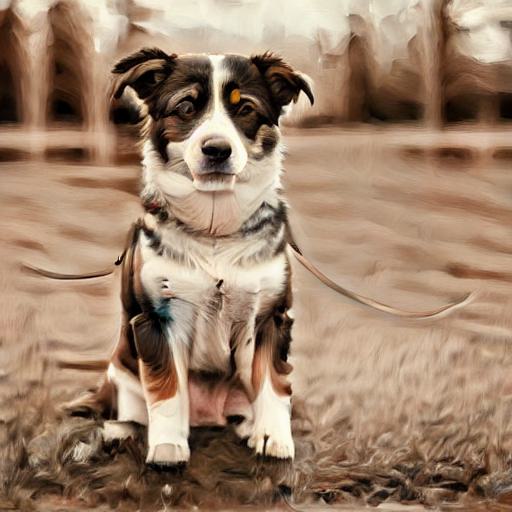}
         }&  
          \multicolumn{1}{m{0.49\textwidth}}{ \includegraphics[width=0.235\textwidth]{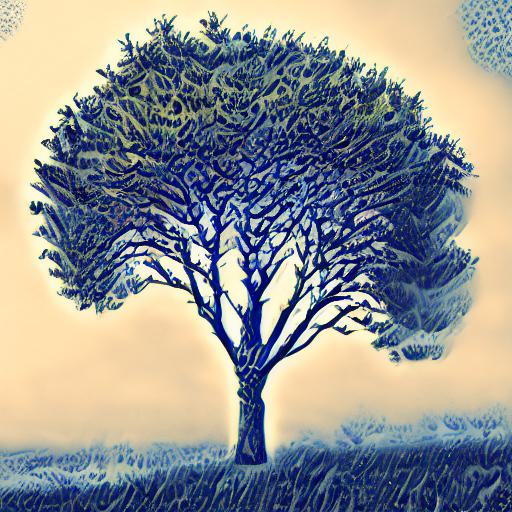} 
         \includegraphics[width=0.235\textwidth]{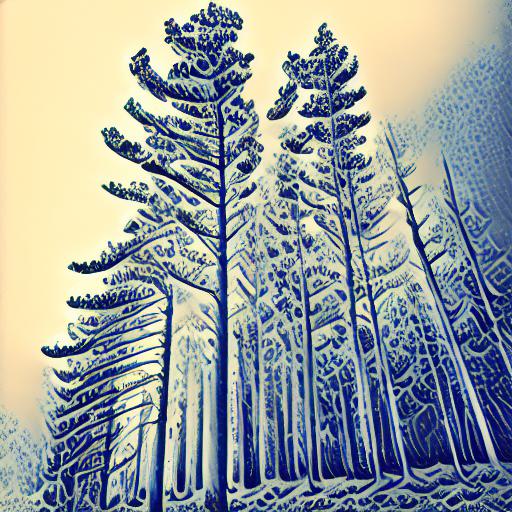}
         } \\ \midrule 
    
        \begin{tabular}[c]{@{}c@{}}
        \vspace{-1.5mm}\tiny{\textbf{Unlearned}} \\
        \vspace{-1.5mm}\tiny{\textbf{Diffusion}} \\
        \vspace{-1.5mm}\tiny{\textbf{Model}} \\
        \vspace{1mm}\tiny{\textbf{(Worst)}}
        \end{tabular}
        &  
          \multicolumn{1}{m{0.49\textwidth}|}{ \includegraphics[width=0.235\textwidth]{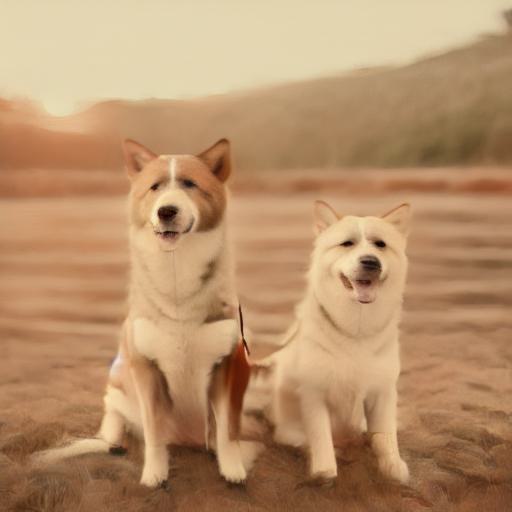} 
         \includegraphics[width=0.235\textwidth]{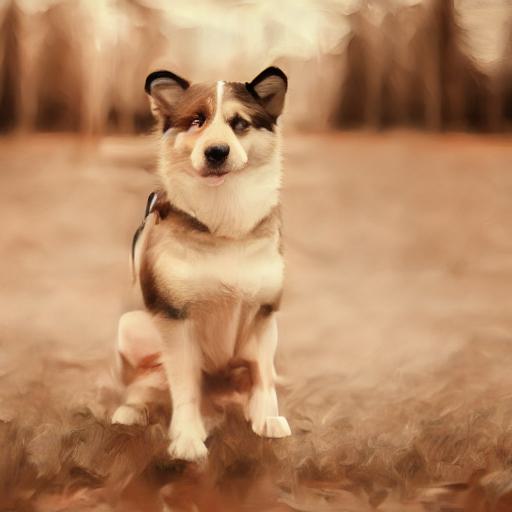}
         }&  
          \multicolumn{1}{m{0.49\textwidth}}{ \includegraphics[width=0.235\textwidth]{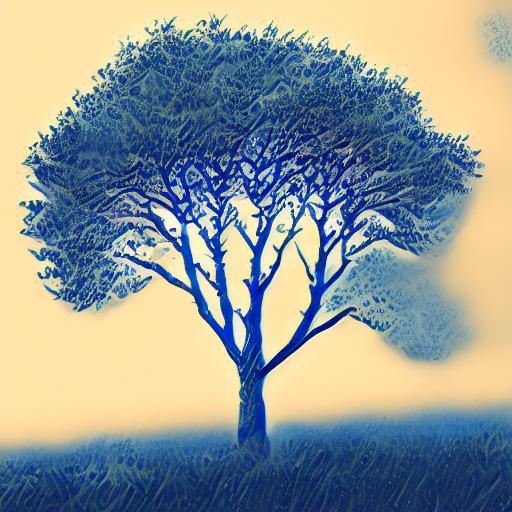} 
         \includegraphics[width=0.235\textwidth]{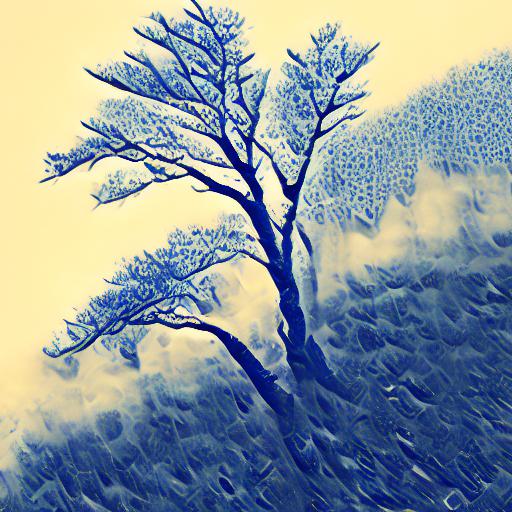} 
         } \\ \midrule 
        \bottomrule[1pt]
    \end{tabular}
}
\end{minipage}
\begin{minipage}{.49\textwidth}
\centering
\resizebox{\textwidth}{!}{
    \begin{tabular}{c|cc|cc}
        \toprule[1pt]
        \midrule
        \multirow{2}{*}{\scriptsize{\textbf{Model}}}
        & \multicolumn{4}{c}{\scriptsize{\textbf{Generation Condition}}} \\ 
        & \multicolumn{2}{c|}{\scriptsize{\Pur \texttt{:}
        % \texttt{Birds + Mosaic }
        \begin{tabular}[c]{@{}c@{}}
        \vspace{-0.5mm}\tiny{\texttt{\textit{A painting of} \textcolor{Red}{Birds}}} \\
        \tiny{\texttt{\textit{in} \textcolor{Blue}{Mosaic}\textit{ Style.}}}
        \end{tabular}
        }} 
        & \multicolumn{2}{c}{\scriptsize{\Pn \texttt{:}
        % \texttt{ Trees + Crayon}
        \begin{tabular}[c]{@{}c@{}}
        \vspace{-0.5mm}\tiny{\texttt{\textit{A painting of} \textcolor{Red}{Trees}}} \\
        \tiny{\texttt{\textit{in} \textcolor{Blue}{Crayon}\textit{ Style.}}}
        \end{tabular}
        }}\\
        \midrule
        
        \begin{tabular}[c]{@{}c@{}}
        \vspace{-1.5mm}\tiny{\textbf{Original}} \\
        \vspace{-1.5mm}\tiny{\textbf{Diffusion}} \\
        \vspace{1mm}\tiny{\textbf{Model}}
        \end{tabular}
         
         &  
          \multicolumn{2}{m{0.49\textwidth}|}{ \includegraphics[width=0.235\textwidth]{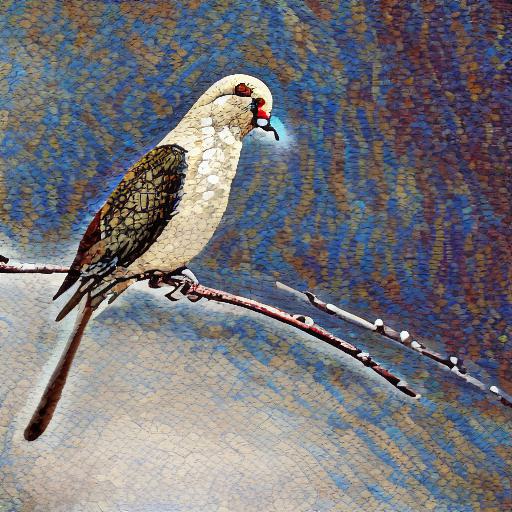} 
         \includegraphics[width=0.235\textwidth]{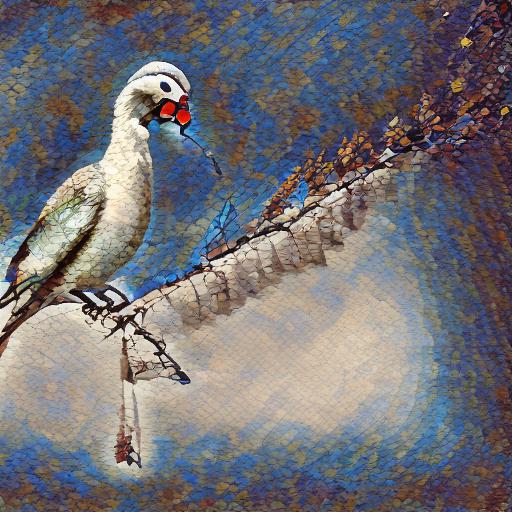}
         }&  
          \multicolumn{2}{m{0.49\textwidth}}{ \includegraphics[width=0.235\textwidth]{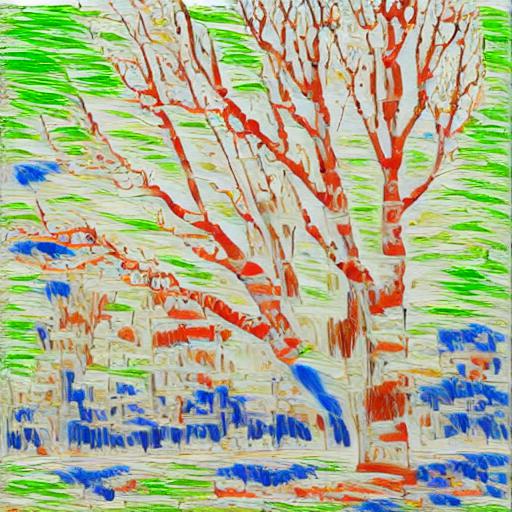} 
         \includegraphics[width=0.235\textwidth]{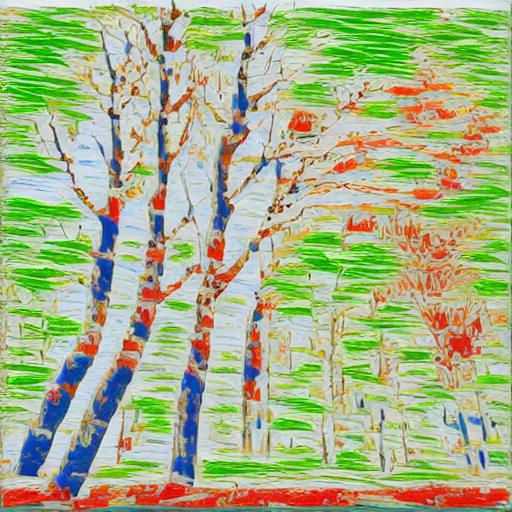}
         } \\ \midrule 
    
        \begin{tabular}[c]{@{}c@{}}
        \vspace{-1.5mm}\tiny{\textbf{Unlearned}} \\
        \vspace{-1.5mm}\tiny{\textbf{Diffusion}} \\
        \vspace{-1.5mm}\tiny{\textbf{Model}} \\
        \vspace{1mm}\tiny{\textbf{(Random)}}
        \end{tabular}
        
         &  
          \multicolumn{2}{m{0.49\textwidth}|}{ \includegraphics[width=0.235\textwidth]{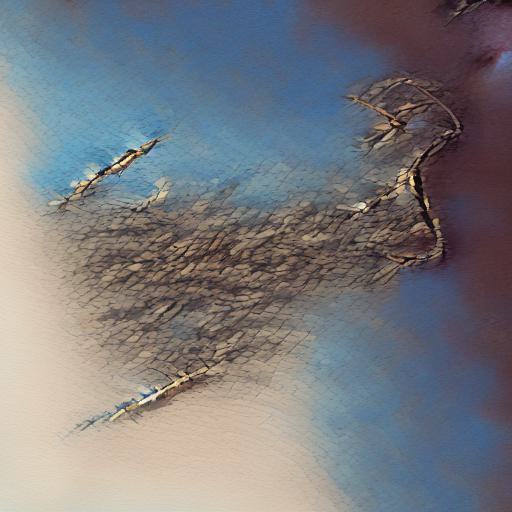} 
         \includegraphics[width=0.235\textwidth]{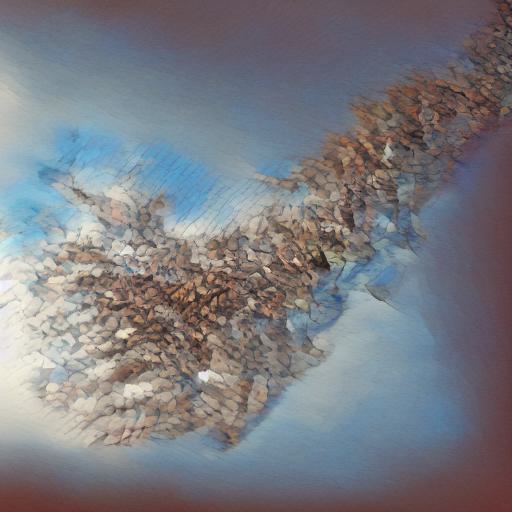}
         }&  
          \multicolumn{2}{m{0.49\textwidth}}{ \includegraphics[width=0.235\textwidth]{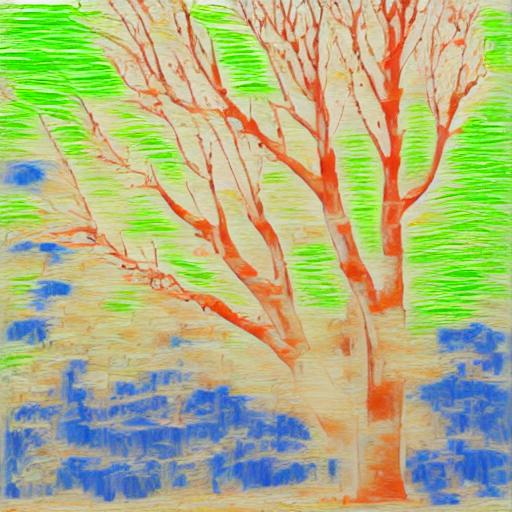} 
         \includegraphics[width=0.235\textwidth]{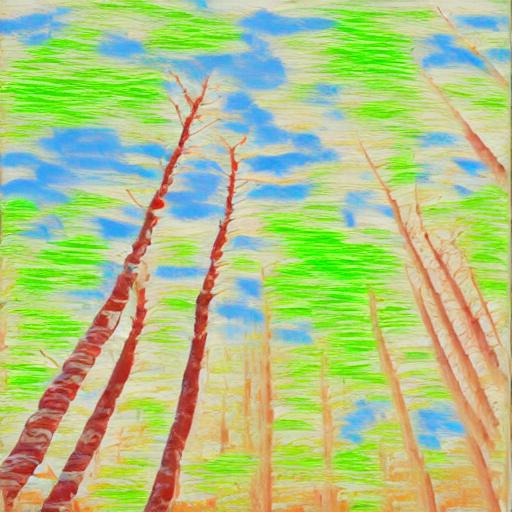} 
         } \\ \midrule 
        \bottomrule[1pt]
        % \caption{(b)}
        \end{tabular}
}
\end{minipage}
\vspace*{-2mm}
\caption{
Examples of image generation using the original SD model (w/o unlearning), the unlearned SD over the worst-case forgetting prompt set (Worst), and the unlearned SD over the random forget set (Random).  For each diffusion model, images are generated based on two conditions, an unlearned prompt ($\Puw$ or $\Pur$) and an unlearning-irrelevant normal prompt ($\Pn$). Here $\Puw$ and $\Pur$ indicate the prompt drawn from the worst-case forget set and the random forget set, respectively.
}
\label{fig: generation_examples}
\vspace*{-6mm}
\end{figure}
\noindent 
\textbf{An extended study: {\Ourset} on prompt-wise unlearning.}
Extending from data-wise forgetting, 
we further demonstrate the efficacy of our approach in prompt-wise forgetting for text-to-image generation. We utilize the stable diffusion (SD) model \cite{rombach2022high} on the UnlearnCanvas dataset, a benchmark dataset designed to evaluate the unlearning of painting styles and objects \cite{zhang2024unlearncanvas}. 
In {UnlearnCanvas}, a text prompt used as the condition of image generation is given by `\texttt{\textit{A painting of }\textcolor{Red}{[Object Name]}\textit{ in }\textcolor{Blue}{[Style Name]}\textit{ Style.}}'
We considered 10 objects and 10 styles (100 combinations in total) for prompts and selected 10\% of them to unlearn.
For designated prompts targeted for unlearning, we apply the Erased Stable Diffusion (ESD) \cite{gandikota2023erasing} technique; See Appendix\,\ref{app: prompt_experiment_setup}
for further implementation details. 
\textbf{Fig.\,\ref{fig: generation_examples}} presents examples of images generated using the pre-trained SD model, the unlearned model by forgetting a random prompt set, and the unlearned model by forgetting the identified worst-case forget set.
For each model, images are generated under two conditions, given from  (1) the \underline{u}nlearned prompt set {\Puwr} where $^{(\mathrm{w})}$ or $^{(\mathrm{r})}$ indicates the worst-case or the random forget set, and (2) the \underline{n}ormal prompt set {\Pn} irrelevant to forget sets.
As we can see, the unlearned diffusion model is \textit{unable to prevent} image generation based on prompts from the worst-case forget set ({\Puw}), resulting in similar image outputs to those of the original diffusion model.
%failed on the worst-case prompts set, resulting in the worst-case-unlearned model still generating images in the same style and object as the origin model when conditioned by the unlearned prompt in \Puw. 
In contrast, the diffusion model unlearned through random forgetting can avoid generating accurate images based on the unlearning-targeted prompts ({\Pur}) from the random forget set, displaying a significant deviation from the original model's outputs, indicating successful unlearning.
Furthermore, when conditioned on the normal, forgetting-irrelevant prompts ({\Pn}), both worst-case and random forgetting-oriented diffusion models perform well, generating the requested images. 
The above results indicate that erasing the influence of prompts from the worst-case forget set introduces new challenges of {\MU} for image generation. 
We refer readers to Appendix\,\ref{app: more_visualization} for more visualizations. 

\noindent \textbf{Additional results.} 
In Appendix\,\ref{app: uniqueness}, we examine the uniqueness of the worst-case forget set by mixing it with other randomly selected data points for unlearning. We also demonstrate
the effectiveness of 
{\ourset} in the scenario of  \textit{class-wise} forgetting on ImageNet \cite{deng2009imagenet}   in Appendix\,\ref{app: class_wise}.
% We also demonstrate the generality of the identified worst-case forget set across different models and unlearning methods. 
% Furthermore, we examine the uniqueness of the worst-case forget set by mixing it with other randomly selected data points for unlearning.
% , and show that the proposed method can be easily adapted to address the easiest-case unlearning as well.

\section{Conclusion and Discussion}
\label{sec: conclusion}

In this study, we delved into the challenge of pinpointing the worst-case forget set in {\MU}, introducing a fresh perspective that broadens the scope and enhances the effectiveness of  {\MU}  beyond conventional methods like random data forgetting. By employing BLO, we developed a structured approach to accurately identify these pivotal sets.
Through extensive experiments, we demonstrated the effectiveness of our proposed method in different data-model setups, showcasing its significance for improved reliability in MU evaluations. 

Although our worst-case performance assessment was inspired by the lack of robustness in random data forgetting, it also deepens the understanding of when MU becomes `easy' or `difficult' and the underlying reasons from a data selection-based interpretability perspective. Our results further encourage rethinking the role of data difficulty in unlearning.
For example, incorporating a curriculum based on these difficulty levels may significantly impact unlearning performance. We term the incorporation of curriculum learning \cite{bengio2009curriculum} into MU as \textit{curriculum unlearning}, which may show promise in improving unlearning effectiveness. Additionally, the process of identifying the worst-case forget set offers a way to attribute data influence by evaluating their unlearning difficulty. In this work, coreset selection emerges as a byproduct of this data attribution process, based on the assessment of unlearning difficulty levels.
Furthermore, the inability to retrain from scratch to unlearn the identified challenging forget set prompts a reevaluation of its appropriateness in defining `exact' unlearning.

% ---- Bibliography ----
%
% BibTeX users should specify bibliography style 'splncs04'.
% References will then be sorted and formatted in the correct style.
%

% \newpage
% \clearpage

\section*{Acknowledgement}
This research is supported by the ARO Award W911NF2310343. Additionally, the work of C. Fan, J. Liu, and S. Liu is partially supported by the NSF Grant IIS-2207052, and the work of Alfred Hero is partially supported by NSF-2246157. 
% Thanks to Naughty, Fries, Crescent, Catcat, Wula, and other cuties for appearing in Fig.\,\ref{fig: teaser}.

%\clearpage

\bibliography{refs/MU,refs/MU_LLM_ref_SLiu,refs/BLO}
\bibliographystyle{splncs04}

\newpage
\clearpage

\appendix

\setcounter{table}{0}
\setcounter{figure}{0}
\renewcommand{\thetable}{A\arabic{table}}
\renewcommand{\thefigure}{A\arabic{figure}}

\begin{center}
    \textbf{\Large Appendix}
\end{center}

\section{Additional Derivations}
\label{app: derivations}

\subsection{The closed-form projection operation}
\label{app: closed_form_derivations}

Recall from \eqref{eq: U-PGD} that 
$\mathrm{Proj}_{\mathbf w \in \gS} (\mathbf a)$ indicates the projection of a constant $\mathbf a$ onto the constraint set $\gS$. This projection operation is defined as solving the auxiliary minimization problem $\mathrm{Proj}_{\mathbf w \in \gS} (\mathbf a) = \argmin_{\mathbf w \in \gS} \| \mathbf w - \mathbf a \|_2^2$, where 
 $\gS = \{\bw|\bw \in [\mathbf 0, \mathbf 1 ], \mathbf{1}^\top\bw = m \}$. It is worth noting that we have relaxed the original binary constraint into its continuous counterpart,  with $\mathbf w \in [\mathbf 0, \mathbf 1] $.
 The relaxed constraint is given by the intersection of the box constraint $\bw \in [\mathbf 0, \mathbf 1 ]$ and the hyperplane $\mathbf{1}^\top\bw = m$. 

 According to \cite[Proposition\,1]{xu2019topology}, the solution of the above projection problem  yields
\begin{align}
    \mathrm{Proj}_{\mathbf w \in \gS} (\mathbf a) =P_{[\mathbf{0,1}]}[\mathbf{a}-\lambda \mathbf{1}],
\end{align}%
where the variable $\lambda$ is given by the root of the  equation $\mathbf{1}^\top P_{[\mathbf{0},\mathbf{1}]}[\mathbf{a}-\lambda \mathbf{1}] = m$, and $P_{[\mathbf{0,1}]}$ is an element-wise thresholding function
\begin{align}
P_{[\mathbf{0,1}]}[x_i] = \left \{ 
\begin{array}{ll}
   0,  &  x < 0 \\
   x,   &  x \in [0, 1] \\
   1, & x  > 1
\end{array}
\right.
\end{align}%
for the $i$th entry of a vector $\mathbf x$.
We also remark that finding the root of the non-linear equation with respect to $\lambda$ can be achieved using the bi-section method \cite{boyd2004convex}.

\subsection{Worst-case forget set identification in  class-wise and prompt-wise forgetting}
\label{app: data_prompt_closed_form_derivations}
For class-wise forgetting, the data selection variables $\bw$ in \eqref{eq: lower_level_MU_loss}-\eqref{eq: bilevel_MU} can be reinterpreted as class selection variables. Here, $w_i = 1$ indicates the selection of the $i$th class for targeted worst-case unlearning. Accordingly, the \textit{upper- and lower-level objectives} of the BLO problem \eqref{eq: bilevel_MU} can be modified to
\begin{align}
 &   f(\mathbf w, \thetau(\bw)) =      \sum_{ i } \left ( w_i \mathbb E_{\mathbf z \in \mathcal D_i} [\ell ( \thetau(\mathbf w); \mathbf z ) ] \right ) + \gamma \|\bw\|_2^2, \\
 & \ell_\MU(\btheta;  \bw) =   \sum_{
    i} \left (  w_i \mathbb E_{\mathbf z \in \mathcal D_i} [\ell_\mathrm{f} ( \btheta; \mathbf z ) ] + {(1 - w_i) \mathbb E_{\mathbf z \in \mathcal D_i} [\ell_\mathrm{r} ( \btheta; \mathbf z ) ] } \right ),
 \end{align}%
where $\mathcal D_i$ represents the dataset corresponding to class $i$,   $
\mathbb E_{\mathbf z \in \mathcal D_i} [\ell ( \btheta; \mathbf z ) ]
$ denotes the training loss over $\mathcal D_i$, and recalling that $\ell_\mathrm{r} = - \ell_\mathrm{f} = \ell $. 
With these specifications in place, the task of identifying the worst-case class-wise forget set can be similarly addressed by resolving the  BLO problem \eqref{eq: bilevel_MU}.

In the context of prompt-wise forgetting, we interpret the data selection variables $\bw$ as prompt selection variables. Here a prompt refers to a text condition used for text-to-image generation, and is known as a `concept' within {\MU}  for generative models  \cite{gandikota2023erasing}.
Thus, the unlearned generative model, when $w_i = 1$, corresponds to the scenario of removing the influence of the $i$th concept from the generative modeling process.
Extended from the concept erasing framework for diffusion models \cite{gandikota2023erasing}, identifying the worst-case prompt-wise forget set can be formulated under the same BLO structure \eqref{eq: bilevel_MU}.  The \textit{upper-level} objective function can be written as
 \begin{align}
 &   f(\mathbf w, \thetau(\bw)) =  \sum_{ i } \left ( w_i    \mathbb E_{t, \boldsymbol{\epsilon} } \left [ \| \boldsymbol{\epsilon} (\mathbf x_t | c_i, \thetau(\mathbf w)) -  \boldsymbol{\epsilon} (\mathbf x_t | c_i, \thetao) \|_2^2 \right ] \right ) + \gamma \|\bw\|_2^2, 
 \label{eq: upper_DM_unlearn}
 \end{align}%
where $\mathbf x_t$ represents the latent feature subject to standard Gaussian noise injection, $\boldsymbol \epsilon$, during the diffusion step $t$ through a forward diffusion process, and $\boldsymbol{\epsilon} (\mathbf x_t | c, \btheta)$ denotes the noise estimator for $\mathbf x_t$ within a diffusion model parameterized by $\btheta$ and conditioned on the text prompt $c$. The loss term $\| \boldsymbol{\epsilon} (\mathbf x_t | c_i, \thetau(\mathbf w)) -  \boldsymbol{\epsilon} (\mathbf x_t | c_i, \thetao) \|_2^2$
penalizes the mean squared error of image generation using the unlearned model $\thetau(\mathbf w)$ and the original diffusion model $\thetao$, respectively. Therefore, minimizing \eqref{eq: upper_DM_unlearn} challenges the unlearning efficacy of $\thetau(\mathbf w)$ regarding the concept $c_i$ to be erased (when $w_i = 1$), by steering its noise estimation accuracy towards that of the original model prior to unlearning.
Furthermore, we specify
the \textit{lower-level} objective function of \eqref{eq: bilevel_MU}  as the Erasing Stable Diffusion (ESD) loss, $\ell_\mathrm{ESD}(\btheta; \mathbf c_i)$, developed in \cite{gandikota2023erasing}. This loss function is designed to remove the influence of the concept $c_i$ from the diffusion model $\btheta$. Consequently, this lower-level objective function is given by $\ell_\mathrm{MU}(\btheta; \bw) = \sum_{i} \left( w_i \ell_\mathrm{ESD}(\btheta; \mathbf c_i) \right)$.

\section{Additional Implementation Details}
\label{app: experiment_setup}

\subsection{Worst-case forget set in data-wise unlearning}
\label{app: data_experiment_setup}

For the exact unlearning method {\Retrain}, the training process comprises $182$ epochs, utilizing the SGD optimizer with a cosine-scheduled learning rate initially set to $0.1$. For {\FT}\cite{warnecke2021machine}, {\RL}\cite{golatkar2020eternal}, {\EUk}\cite{goel2022towards}, {\CFk}\cite{goel2022towards}, and {\Scrub}\cite{kurmanji2023towards}, the unlearning process takes $10$ epochs, during which the optimal learning rate is searched within the range of $[10^{-4}, 10^{-1}]$, and $k = 1$ is set for {\EUk} and {\CFk}.
For {\MUSparse}\cite{jia2023model}, the unlearning-enabled model updating process also takes $10$ epochs, searching for the optimal sparse ratio in the range $[10^{-6}, 10^{-4}]$ and exploring the most appropriate learning rate within $[10^{-3}, 10^{-1}]$. Regarding the method {\BS}\cite{chen2023boundary}, the 
step size of {fast gradient sign method} (FGSM) is fixed at $0.1$. Both {\BS} and {\BE}\cite{chen2023boundary} undergo a $10$-epoch fine-tuning process, during which the optimal learning rate is sought within the interval $[10^{-8}, 10^{-4}]$. Finally, for {\Salun}\cite{fan2023salun}, we conducted a $10$-epoch fine-tuning phase, exploring learning rates within the range $[10^{-4}, 10^{-2}]$, and investigating sparsity ratios in the range $[0.1, 0.9]$.

\subsection{Worst-case forget set in prompt-wise unlearning}
\label{app: prompt_experiment_setup}
In the UnlearnCanvas benchmark dataset \cite{zhang2024unlearncanvas} for image generation, we select 10 objects and 10 styles, leading to 100 prompt combinations. The objects include Horses, Towers, Humans, Flowers, Birds, Trees, Waterfalls, Jellyfish, Sandwiches, and Dogs, while the styles feature Crayon, Ukiyoe, Mosaic, Sketch, Dadaism, Winter, Van Gogh, Rust, Glowing Sunset, and Red Blue Ink. We target 10\% of these combinations for the unlearning task.

 For prompt-wise worst-case forget set identification, we utilize the Erased Stable Diffusion (ESD) method combined with SignSGD, setting a learning rate of $10^{-5}$ for $1000$ iterations when specifying \eqref{eq: L-signSGD}. After identifying the worst-case forget prompts, we apply ESD again, this time with a learning rate of $3 \times 10^{-7}$ for $1000$ iterations to unlearn these prompts. During image generation, DDIM is specified using $100$ time steps and a conditional scale of $7.5$.

\begin{table}[t]
\caption{Performance of various unlearning methods under {\randomset}s and {\ourset}s on CIFAR-10 using ResNet-18 for different forgetting ratio (including 1\%, 5\%, 10\% and 20\%). The result format follows Table\,\ref{tab: main_approximate_unlearn}.}
\vspace*{-6mm}
\label{tab: more_ratio}
\begin{center}
\resizebox{\textwidth}{!}{

\renewcommand\tabcolsep{5pt}

\begin{tabular}{c|ccccc|ccccc}
\toprule[1pt]
\midrule
\multirow{2}{*}{\textbf{Methods}} & \multicolumn{5}{c|}{\textbf{\RandomSet}}  & \multicolumn{5}{c}{\textbf{\OurSet}}  \\
                        & \multicolumn{1}{c|}{UA}   & \multicolumn{1}{c|}{MIA}     & \multicolumn{1}{c|}{RA} & \multicolumn{1}{c|}{TA} & \multicolumn{1}{c|}{Avg. Gap} 
                        & \multicolumn{1}{c|}{UA}   & \multicolumn{1}{c|}{MIA}     & \multicolumn{1}{c|}{RA}    & \multicolumn{1}{c|}{TA} & \multicolumn{1}{c}{Avg. Gap} \\

\midrule
\rowcolor{Gray}
\multicolumn{11}{c}{\textbf{1\%-Data Forgetting}} \\
\midrule
{{\Retrain}} & $5.85_{\pm 0.69}$  & $12.89_{\pm 1.27}$  & $99.96_{\pm 0.00}$  & $93.17_{\pm 0.15}$  & \textcolor{blue}{0.00} & $0.00_{\pm 0.00}$  & $0.00_{\pm 0.00}$  & $99.95_{\pm 0.02}$  & $93.45_{\pm 0.17}$  & \textcolor{blue}{0.00}\\
\midrule
{{\FT}} & $8.93_{\pm 2.74}$ (\textcolor{blue}{3.08}) & $14.40_{\pm 1.98}$ (\textcolor{blue}{1.51}) & $94.52_{\pm 1.79}$ (\textcolor{blue}{5.44}) & $88.53_{\pm 1.69}$ (\textcolor{blue}{4.64}) & \textcolor{blue}{3.67} & $0.13_{\pm 0.27}$ (\textcolor{blue}{0.13}) & $0.13_{\pm 0.18}$ (\textcolor{blue}{0.13}) & $90.69_{\pm 6.71}$ (\textcolor{blue}{9.26}) & $85.51_{\pm 5.76}$ (\textcolor{blue}{7.94}) & \textcolor{blue}{4.36}\\
{{\EUk}} & $1.60_{\pm 1.10}$ (\textcolor{blue}{4.25}) & $5.60_{\pm 1.49}$ (\textcolor{blue}{7.29}) & $97.79_{\pm 0.74}$ (\textcolor{blue}{2.17}) & $90.58_{\pm 0.76}$ (\textcolor{blue}{2.59}) & \textcolor{blue}{4.07} & $0.09_{\pm 0.11}$ (\textcolor{blue}{0.09}) & $1.69_{\pm 1.04}$ (\textcolor{blue}{1.69}) & $97.76_{\pm 0.92}$ (\textcolor{blue}{2.19}) & $90.49_{\pm 0.88}$ (\textcolor{blue}{2.96}) & \textcolor{blue}{1.73}\\
{{\CFk}} & $0.00_{\pm 0.00}$ (\textcolor{blue}{5.85}) & $0.44_{\pm 0.36}$ (\textcolor{blue}{12.45}) & $99.98_{\pm 0.00}$ (\textcolor{blue}{0.02}) & $94.33_{\pm 0.08}$ (\textcolor{blue}{1.16}) & \textcolor{blue}{4.87} & $0.00_{\pm 0.00}$ (\textcolor{blue}{0.00}) & $0.00_{\pm 0.00}$ (\textcolor{blue}{0.00}) & $99.98_{\pm 0.00}$ (\textcolor{blue}{0.03}) & $94.40_{\pm 0.07}$ (\textcolor{blue}{0.95}) & \textcolor{blue}{0.24}\\
{{\MUSparse}} & $5.89_{\pm 1.81}$ (\textcolor{blue}{0.04}) & $12.33_{\pm 1.84}$ (\textcolor{blue}{0.56}) & $96.52_{\pm 0.85}$ (\textcolor{blue}{3.44}) & $90.57_{\pm 0.66}$ (\textcolor{blue}{2.60}) & \textcolor{blue}{1.66} & $0.00_{\pm 0.00}$ (\textcolor{blue}{0.00}) & $0.09_{\pm 0.18}$ (\textcolor{blue}{0.09}) & $92.08_{\pm 3.44}$ (\textcolor{blue}{7.87}) & $86.91_{\pm 2.88}$ (\textcolor{blue}{6.54}) & \textcolor{blue}{3.62}\\
\midrule
% \rowcolor{Gray}
% \multicolumn{11}{c}{\textbf{Relabeling-based}} \\
% \midrule
{{\RL}} & $7.26_{\pm 1.47}$ (\textcolor{blue}{1.41}) & $43.85_{\pm 2.67}$ (\textcolor{blue}{30.96}) & $99.99_{\pm 0.00}$ (\textcolor{blue}{0.03}) & $94.11_{\pm 0.04}$ (\textcolor{blue}{0.94}) & \textcolor{blue}{8.33} & $0.00_{\pm 0.00}$ (\textcolor{blue}{0.00}) & $84.37_{\pm 5.61}$ (\textcolor{blue}{84.37}) & $100.00_{\pm 0.00}$ (\textcolor{blue}{0.05}) & $94.46_{\pm 0.09}$ (\textcolor{blue}{1.01}) & \textcolor{blue}{21.36}\\
{{\BE}} & $0.00_{\pm 0.00}$ (\textcolor{blue}{5.85}) & $0.98_{\pm 0.30}$ (\textcolor{blue}{11.91}) & $99.97_{\pm 0.01}$ (\textcolor{blue}{0.01}) & $94.26_{\pm 0.09}$ (\textcolor{blue}{1.09}) & \textcolor{blue}{4.72} & $3.47_{\pm 1.54}$ (\textcolor{blue}{3.47}) & $16.36_{\pm 6.93}$ (\textcolor{blue}{16.36}) & $92.37_{\pm 0.48}$ (\textcolor{blue}{7.58}) & $85.55_{\pm 0.57}$ (\textcolor{blue}{7.90}) & \textcolor{blue}{8.83}\\
{{\BS}} & $0.00_{\pm 0.00}$ (\textcolor{blue}{5.85}) & $0.98_{\pm 0.30}$ (\textcolor{blue}{11.91}) & $99.97_{\pm 0.01}$ (\textcolor{blue}{0.01}) & $94.27_{\pm 0.11}$ (\textcolor{blue}{1.10}) & \textcolor{blue}{4.72} & $3.29_{\pm 1.44}$ (\textcolor{blue}{3.29}) & $13.69_{\pm 5.09}$ (\textcolor{blue}{13.69}) & $92.34_{\pm 0.52}$ (\textcolor{blue}{7.61}) & $85.48_{\pm 0.60}$ (\textcolor{blue}{7.97}) & \textcolor{blue}{8.14}\\
{{\Salun}} & $1.26_{\pm 0.55}$ (\textcolor{blue}{4.59}) & $17.33_{\pm 1.01}$ (\textcolor{blue}{4.44}) & $99.99_{\pm 0.01}$ (\textcolor{blue}{0.03}) & $94.28_{\pm 0.07}$ (\textcolor{blue}{1.11}) & \textcolor{blue}{2.54} & $0.00_{\pm 0.00}$ (\textcolor{blue}{0.00}) & $72.59_{\pm 2.22}$ (\textcolor{blue}{72.59}) & $100.00_{\pm 0.00}$ (\textcolor{blue}{0.05}) & $94.45_{\pm 0.14}$ (\textcolor{blue}{1.00}) & \textcolor{blue}{18.41}\\

\midrule
\rowcolor{Gray}
\multicolumn{11}{c}{\textbf{5\%-Data Forgetting}} \\
\midrule
{{\Retrain}} & $5.92_{\pm 0.44}$ & $13.00_{\pm 0.55}$ & $100.00_{\pm 0.00}$ & $94.51_{\pm 0.07}$& \textcolor{blue}{0.00} & $0.00_{\pm 0.00}$ & $0.02_{\pm 0.02}$  & $100.00_{\pm 0.00}$ & $94.67_{\pm 0.08}$ & \textcolor{blue}{0.00}\\
\midrule
{{\FT}} & $5.17_{\pm 0.73}$ (\textcolor{blue}{0.75}) & $11.32_{\pm 0.94}$ (\textcolor{blue}{1.68}) & $97.08_{\pm 0.44}$ (\textcolor{blue}{2.92}) & $90.71_{\pm 0.38}$ (\textcolor{blue}{3.80}) & \textcolor{blue}{2.29} & $0.01_{\pm 0.02}$ (\textcolor{blue}{0.01}) & $0.02_{\pm 0.04}$ (\textcolor{blue}{0.00}) & $97.39_{\pm 0.48}$ (\textcolor{blue}{2.61}) & $91.10_{\pm 0.55}$ (\textcolor{blue}{3.57}) & \textcolor{blue}{1.55}\\
{{\EUk}} & $2.13_{\pm 0.75}$ (\textcolor{blue}{3.79}) & $6.07_{\pm 1.10}$ (\textcolor{blue}{6.93}) & $97.81_{\pm 0.86}$ (\textcolor{blue}{2.19}) & $90.53_{\pm 0.82}$ (\textcolor{blue}{3.98}) & \textcolor{blue}{4.22} & $0.16_{\pm 0.12}$ (\textcolor{blue}{0.16}) & $2.11_{\pm 1.17}$ (\textcolor{blue}{2.09}) & $97.61_{\pm 0.78}$ (\textcolor{blue}{2.39}) & $90.48_{\pm 0.74}$ (\textcolor{blue}{4.19}) & \textcolor{blue}{2.21}\\
{{\CFk}} & $0.04_{\pm 0.06}$ (\textcolor{blue}{5.88}) & $0.74_{\pm 0.04}$ (\textcolor{blue}{12.26}) & $99.99_{\pm 0.00}$ (\textcolor{blue}{0.01}) & $94.47_{\pm 0.02}$ (\textcolor{blue}{0.04}) & \textcolor{blue}{4.55} & $0.00_{\pm 0.00}$ (\textcolor{blue}{0.00}) & $0.00_{\pm 0.00}$ (\textcolor{blue}{0.02}) & $99.98_{\pm 0.00}$ (\textcolor{blue}{0.02}) & $94.40_{\pm 0.04}$ (\textcolor{blue}{0.27}) & \textcolor{blue}{0.08}\\
{{\MUSparse}} & $4.63_{\pm 1.01}$ (\textcolor{blue}{1.29}) & $10.09_{\pm 1.00}$ (\textcolor{blue}{2.91}) & $97.13_{\pm 0.75}$ (\textcolor{blue}{2.87}) & $90.92_{\pm 0.65}$ (\textcolor{blue}{3.59}) & \textcolor{blue}{2.66} & $0.00_{\pm 0.00}$ (\textcolor{blue}{0.00}) & $0.01_{\pm 0.02}$ (\textcolor{blue}{0.01}) & $97.12_{\pm 0.58}$ (\textcolor{blue}{2.88}) & $91.27_{\pm 0.59}$ (\textcolor{blue}{3.40}) & \textcolor{blue}{1.57}\\
% \midrule
% \rowcolor{Gray}
% \multicolumn{11}{c}{\textbf{Relabeling-based}} \\
\midrule
{{\RL}} & $5.47_{\pm 0.45}$ (\textcolor{blue}{0.45}) & $35.38_{\pm 0.50}$ (\textcolor{blue}{22.38}) & $99.97_{\pm 0.01}$ (\textcolor{blue}{0.03}) & $93.70_{\pm 0.10}$ (\textcolor{blue}{0.81}) & \textcolor{blue}{5.92} & $0.15_{\pm 0.18}$ (\textcolor{blue}{0.15}) & $95.57_{\pm 0.80}$ (\textcolor{blue}{95.55}) & $99.98_{\pm 0.00}$ (\textcolor{blue}{0.02}) & $94.08_{\pm 0.02}$ (\textcolor{blue}{0.59}) & \textcolor{blue}{24.08}\\
{{\BE}} & $0.36_{\pm 0.10}$ (\textcolor{blue}{5.56}) & $18.77_{\pm 1.22}$ (\textcolor{blue}{5.77}) & $99.73_{\pm 0.04}$ (\textcolor{blue}{0.27}) & $93.07_{\pm 0.17}$ (\textcolor{blue}{1.44}) & \textcolor{blue}{3.26} & $38.19_{\pm 5.76}$ (\textcolor{blue}{38.19}) & $85.74_{\pm 1.92}$ (\textcolor{blue}{85.72}) & $79.07_{\pm 2.41}$ (\textcolor{blue}{20.93}) & $72.56_{\pm 2.43}$ (\textcolor{blue}{22.11}) & \textcolor{blue}{41.74}\\
{{\BS}} & $1.65_{\pm 0.67}$ (\textcolor{blue}{4.27}) & $17.59_{\pm 1.65}$ (\textcolor{blue}{4.59}) & $98.56_{\pm 0.65}$ (\textcolor{blue}{1.44}) & $91.93_{\pm 0.55}$ (\textcolor{blue}{2.58}) & \textcolor{blue}{3.22} & $39.62_{\pm 9.10}$ (\textcolor{blue}{39.62}) & $84.07_{\pm 5.29}$ (\textcolor{blue}{84.05}) & $72.30_{\pm 7.03}$ (\textcolor{blue}{27.70}) & $66.74_{\pm 6.86}$ (\textcolor{blue}{27.93}) & \textcolor{blue}{44.82}\\
{{\Salun}} & $0.67_{\pm 0.04}$ (\textcolor{blue}{5.25}) & $12.87_{\pm 1.27}$ (\textcolor{blue}{0.13}) & $100.00_{\pm 0.00}$ (\textcolor{blue}{0.00}) & $94.13_{\pm 0.03}$ (\textcolor{blue}{0.38}) & \textcolor{blue}{1.44} & $0.09_{\pm 0.07}$ (\textcolor{blue}{0.09}) & $93.23_{\pm 0.25}$ (\textcolor{blue}{93.21}) & $100.00_{\pm 0.00}$ (\textcolor{blue}{0.00}) & $94.20_{\pm 0.13}$ (\textcolor{blue}{0.47}) & \textcolor{blue}{23.44}\\

\midrule
\rowcolor{Gray}
\multicolumn{11}{c}{\textbf{10\%-Data Forgetting}} \\
\midrule
{{\Retrain}} & $5.28_{\pm 0.33}$ & $12.86_{\pm 0.61}$ & $100.00_{\pm 0.00}$ & $94.38_{\pm 0.15}$ & \textcolor{blue}{0.00} & $0.00_{\pm 0.00}$ & $0.00_{\pm 0.00}$ & $100.00_{\pm 0.00}$ & $94.66_{\pm 0.09}$ & \textcolor{blue}{0.00}\\
\midrule
{{\FT}} & $5.08_{\pm 0.39}$ (\textcolor{blue}{0.20}) & $10.96_{\pm 0.38}$ (\textcolor{blue}{1.90}) & $97.46_{\pm 0.52}$ (\textcolor{blue}{2.54}) & $91.02_{\pm 0.36}$ (\textcolor{blue}{3.36}) & \textcolor{blue}{2.00} & $0.00_{\pm 0.00}$ (\textcolor{blue}{0.00}) & $0.02_{\pm 0.03}$ (\textcolor{blue}{0.02}) & $97.63_{\pm 0.46}$ (\textcolor{blue}{2.37}) & $91.58_{\pm 0.40}$ (\textcolor{blue}{3.08}) & \textcolor{blue}{1.37}\\
{{\EUk}} & $2.34_{\pm 0.79}$ (\textcolor{blue}{2.94}) & $6.35_{\pm 0.89}$ (\textcolor{blue}{6.51}) & $97.52_{\pm 0.89}$ (\textcolor{blue}{2.48}) & $90.17_{\pm 0.88}$ (\textcolor{blue}{4.21}) & \textcolor{blue}{4.04} & $0.68_{\pm 0.56}$ (\textcolor{blue}{0.68}) & $5.02_{\pm 4.42}$ (\textcolor{blue}{5.02}) & $97.17_{\pm 0.86}$ (\textcolor{blue}{2.83}) & $90.08_{\pm 0.70}$ (\textcolor{blue}{4.58}) & \textcolor{blue}{3.28}\\
{{\CFk}} & $0.02_{\pm 0.02}$ (\textcolor{blue}{5.26}) & $0.76_{\pm 0.02}$ (\textcolor{blue}{12.10}) & $99.98_{\pm 0.00}$ (\textcolor{blue}{0.02}) & $94.45_{\pm 0.02}$ (\textcolor{blue}{0.07}) & \textcolor{blue}{4.36} & $0.00_{\pm 0.00}$ (\textcolor{blue}{0.00}) & $0.00_{\pm 0.00}$ (\textcolor{blue}{0.00}) & $99.98_{\pm 0.01}$ (\textcolor{blue}{0.02}) & $94.34_{\pm 0.05}$ (\textcolor{blue}{0.32}) & \textcolor{blue}{0.08}\\
{{\MUSparse}} & $4.34_{\pm 0.73}$ (\textcolor{blue}{0.94}) & $9.82_{\pm 1.04}$ (\textcolor{blue}{3.04}) & $97.70_{\pm 0.72}$ (\textcolor{blue}{2.30}) & $91.41_{\pm 0.68}$ (\textcolor{blue}{2.97}) & \textcolor{blue}{2.31} & $0.02_{\pm 0.03}$ (\textcolor{blue}{0.02}) & $0.11_{\pm 0.11}$ (\textcolor{blue}{0.11}) & $96.93_{\pm 0.73}$ (\textcolor{blue}{3.07}) & $90.96_{\pm 0.82}$ (\textcolor{blue}{3.70}) & \textcolor{blue}{1.72}\\
% \midrule
% \rowcolor{Gray}
% \multicolumn{11}{c}{\textbf{Relabeling-based}} \\
\midrule
{{\RL}} & $3.59_{\pm 0.24}$ (\textcolor{blue}{1.69}) & $28.02_{\pm 2.47}$ (\textcolor{blue}{15.16}) & $99.97_{\pm 0.01}$ (\textcolor{blue}{0.03}) & $93.74_{\pm 0.12}$ (\textcolor{blue}{0.64}) & \textcolor{blue}{4.38} & $1.93_{\pm 1.5}$ (\textcolor{blue}{1.93}) & $96.70_{\pm 0.66}$ (\textcolor{blue}{96.70}) & $99.96_{\pm 0.01}$ (\textcolor{blue}{0.04}) & $93.83_{\pm 0.24}$ (\textcolor{blue}{0.83}) & \textcolor{blue}{24.88}\\
% \rowcolor{Gray}
{{\BE}} & $1.19_{\pm 0.49}$ (\textcolor{blue}{4.09}) & $22.06_{\pm 0.61}$ (\textcolor{blue}{9.20}) & $98.77_{\pm 0.41}$ (\textcolor{blue}{1.23}) & $91.79_{\pm 0.32}$ (\textcolor{blue}{2.59}) & \textcolor{blue}{4.28} & $19.47_{\pm 2.12}$ (\textcolor{blue}{19.47}) & $81.45_{\pm 5.16}$ (\textcolor{blue}{81.45}) & $81.35_{\pm 2.76}$ (\textcolor{blue}{18.65}) & $75.41_{\pm 1.77}$ (\textcolor{blue}{19.25}) & \textcolor{blue}{34.70}\\
{{\BS}} & $5.72_{\pm 1.42}$ (\textcolor{blue}{0.44}) & $27.15_{\pm 1.41}$ (\textcolor{blue}{14.29}) & $94.29_{\pm 1.06}$ (\textcolor{blue}{5.71}) & $87.45_{\pm 1.06}$ (\textcolor{blue}{6.93}) & \textcolor{blue}{6.84} & $29.75_{\pm 6.39}$ (\textcolor{blue}{29.75}) & $74.88_{\pm 3.13}$ (\textcolor{blue}{74.88}) & $78.34_{\pm 0.98}$ (\textcolor{blue}{21.66}) & $72.07_{\pm 1.25}$ (\textcolor{blue}{22.59}) & \textcolor{blue}{37.22}\\
% \rowcolor{Gray}
{{\Salun}} & $1.48_{\pm 0.14}$ (\textcolor{blue}{3.80}) & $16.19_{\pm 0.34}$ (\textcolor{blue}{3.33}) & $99.98_{\pm 0.01}$ (\textcolor{blue}{0.02}) & $93.95_{\pm 0.01}$ (\textcolor{blue}{0.43}) & \textcolor{blue}{1.89} & $0.96_{\pm 0.59}$ (\textcolor{blue}{0.96}) & $96.43_{\pm 0.33}$ (\textcolor{blue}{96.43}) & $99.98_{\pm 0.01}$ (\textcolor{blue}{0.02}) & $94.03_{\pm 0.08}$ (\textcolor{blue}{0.63}) & \textcolor{blue}{24.51}\\
\midrule
\rowcolor{Gray}
\multicolumn{11}{c}{\textbf{20\%-Data Forgetting}} \\
\midrule
{{\Retrain}} & $5.76_{\pm 0.20}$ & $14.34_{\pm 0.40}$ & $100.00_{\pm 0.00}$  & $94.04_{\pm 0.08}$ & \textcolor{blue}{0.00} & $0.00_{\pm 0.00}$ & $0.03_{\pm 0.01}$ & $100.00_{\pm 0.00}$ & $94.60_{\pm 0.08}$ & \textcolor{blue}{0.00}\\
\midrule
% \rowcolor{Gray}
% \multicolumn{11}{c}{\textbf{Relabeling-free}} \\
% \midrule
{{\FT}} & $5.46_{\pm 0.42}$ (\textcolor{blue}{0.30}) & $11.40_{\pm 0.78}$ (\textcolor{blue}{2.94}) & $97.10_{\pm 0.42}$ (\textcolor{blue}{2.90}) & $90.32_{\pm 0.41}$ (\textcolor{blue}{3.72}) & \textcolor{blue}{2.46} & $0.14_{\pm 0.13}$ (\textcolor{blue}{0.14}) & $0.27_{\pm 0.15}$ (\textcolor{blue}{0.24}) & $96.56_{\pm 1.12}$ (\textcolor{blue}{3.44}) & $90.62_{\pm 1.07}$ (\textcolor{blue}{3.98}) & \textcolor{blue}{1.95}\\
{{\EUk}} & $3.08_{\pm 1.19}$ (\textcolor{blue}{2.68}) & $7.43_{\pm 1.29}$ (\textcolor{blue}{6.91}) & $96.70_{\pm 1.31}$ (\textcolor{blue}{3.30}) & $89.37_{\pm 1.23}$ (\textcolor{blue}{4.67}) & \textcolor{blue}{4.39} & $1.76_{\pm 1.25}$ (\textcolor{blue}{1.76}) & $7.39_{\pm 4.89}$ (\textcolor{blue}{7.36}) & $95.75_{\pm 1.26}$ (\textcolor{blue}{4.25}) & $88.96_{\pm 1.05}$ (\textcolor{blue}{5.64}) & \textcolor{blue}{4.75}\\
{{\CFk}} & $0.03_{\pm 0.01}$ (\textcolor{blue}{5.73}) & $0.68_{\pm 0.10}$ (\textcolor{blue}{13.66}) & $99.99_{\pm 0.00}$ (\textcolor{blue}{0.01}) & $94.45_{\pm 0.06}$ (\textcolor{blue}{0.41}) & \textcolor{blue}{4.95} & $0.00_{\pm 0.00}$ (\textcolor{blue}{0.00}) & $0.00_{\pm 0.00}$ (\textcolor{blue}{0.03}) & $99.97_{\pm 0.01}$ (\textcolor{blue}{0.03}) & $94.29_{\pm 0.03}$ (\textcolor{blue}{0.31}) & \textcolor{blue}{0.09}\\
{{\MUSparse}} & $3.83_{\pm 0.59}$ (\textcolor{blue}{1.93}) & $8.76_{\pm 0.92}$ (\textcolor{blue}{5.58}) & $97.99_{\pm 0.55}$ (\textcolor{blue}{2.01}) & $91.30_{\pm 0.64}$ (\textcolor{blue}{2.74}) & \textcolor{blue}{3.06} & $0.07_{\pm 0.07}$ (\textcolor{blue}{0.07}) & $0.14_{\pm 0.14}$ (\textcolor{blue}{0.11}) & $97.25_{\pm 0.93}$ (\textcolor{blue}{2.75}) & $91.22_{\pm 0.76}$ (\textcolor{blue}{3.38}) & \textcolor{blue}{1.58}\\
\midrule
% \rowcolor{Gray}
% \multicolumn{11}{c}{\textbf{Relabeling-based}} \\
% \midrule
{{\RL}} & $2.44_{\pm 0.04}$ (\textcolor{blue}{3.32}) & $22.21_{\pm 0.43}$ (\textcolor{blue}{7.87}) & $99.97_{\pm 0.00}$ (\textcolor{blue}{0.03}) & $93.44_{\pm 0.04}$ (\textcolor{blue}{0.60}) & \textcolor{blue}{2.95} & $3.84_{\pm 2.01}$ (\textcolor{blue}{3.84}) & $97.54_{\pm 0.12}$ (\textcolor{blue}{97.51}) & $99.92_{\pm 0.01}$ (\textcolor{blue}{0.08}) & $93.07_{\pm 0.40}$ (\textcolor{blue}{1.53}) & \textcolor{blue}{25.74}\\
{{\BE}} & $11.54_{\pm 1.44}$ (\textcolor{blue}{5.78}) & $29.84_{\pm 1.08}$ (\textcolor{blue}{15.50}) & $88.20_{\pm 1.48}$ (\textcolor{blue}{11.80}) & $80.67_{\pm 1.35}$ (\textcolor{blue}{13.37}) & \textcolor{blue}{11.61} & $25.05_{\pm 2.89}$ (\textcolor{blue}{25.05}) & $82.25_{\pm 4.88}$ (\textcolor{blue}{82.22}) & $76.88_{\pm 2.77}$ (\textcolor{blue}{23.12}) & $71.07_{\pm 2.46}$ (\textcolor{blue}{23.53}) & \textcolor{blue}{38.48}\\
{{\BS}} & $20.69_{\pm 1.59}$ (\textcolor{blue}{14.93}) & $32.96_{\pm 2.16}$ (\textcolor{blue}{18.62}) & $78.87_{\pm 1.91}$ (\textcolor{blue}{21.13}) & $72.23_{\pm 1.70}$ (\textcolor{blue}{21.81}) & \textcolor{blue}{19.12} & $40.61_{\pm 2.26}$ (\textcolor{blue}{40.61}) & $75.05_{\pm 6.78}$ (\textcolor{blue}{75.02}) & $72.17_{\pm 1.16}$ (\textcolor{blue}{27.83}) & $64.53_{\pm 1.12}$ (\textcolor{blue}{30.07}) & \textcolor{blue}{43.38}\\
{{\Salun}} & $1.31_{\pm 0.08}$ (\textcolor{blue}{4.45}) & $17.15_{\pm 0.85}$ (\textcolor{blue}{2.81}) & $99.98_{\pm 0.01}$ (\textcolor{blue}{0.02}) & $93.67_{\pm 0.17}$ (\textcolor{blue}{0.37}) & \textcolor{blue}{1.91} & $2.82_{\pm 1.42}$ (\textcolor{blue}{2.82}) & $97.09_{\pm 0.27}$ (\textcolor{blue}{97.06}) & $99.95_{\pm 0.01}$ (\textcolor{blue}{0.05}) & $93.36_{\pm 0.38}$ (\textcolor{blue}{1.24}) & \textcolor{blue}{25.29}\\

\midrule
\bottomrule[1pt]
\end{tabular}
}
\vspace{-9mm}
\end{center}
\end{table}

\begin{table}[t]
\caption{{Performance of various unlearning methods under {\randomset}s and {\ourset}s on CIFAR-100 and Tiny ImageNet using ResNet-18 for forgetting ratio 10\%. The result format follows Table\,\ref{tab: main_approximate_unlearn}.}}
\vspace*{-6mm}
\label{tab: more_dataset}

\begin{center}
\resizebox{\textwidth}{!}{

\renewcommand\tabcolsep{5pt}

\begin{tabular}{c|ccccc|ccccc}
\toprule[1pt]
\midrule
\multirow{2}{*}{\textbf{Methods}} & \multicolumn{5}{c|}{\textbf{\RandomSet}}  & \multicolumn{5}{c}{\textbf{\OurSet}}  \\
                        & \multicolumn{1}{c|}{UA}   & \multicolumn{1}{c|}{MIA}     & \multicolumn{1}{c|}{RA} & \multicolumn{1}{c|}{TA} & \multicolumn{1}{c|}{Avg. Gap} 
                        & \multicolumn{1}{c|}{UA}   & \multicolumn{1}{c|}{MIA}     & \multicolumn{1}{c|}{RA}    & \multicolumn{1}{c|}{TA} & \multicolumn{1}{c}{Avg. Gap} \\
\midrule
\rowcolor{Gray}
\multicolumn{11}{c}{\textbf{CIFAR-100}} \\
\midrule
{{\Retrain}} & $25.06_{\pm 0.25}$ & $49.98_{\pm 0.91}$ & $99.98_{\pm 0.00}$ & $74.54_{\pm 0.07}$ & \textcolor{blue}{0.00} & $0.13_{\pm 0.04}$ & $1.11_{\pm 0.19}$ & $99.97_{\pm 0.00}$ & $75.36_{\pm 0.31}$ & \textcolor{blue}{0.00}\\
\midrule
{{\FT}} & $23.10_{\pm 0.97}$ (\textcolor{blue}{1.96}) & $30.47_{\pm 0.87}$ (\textcolor{blue}{19.51}) & $90.44_{\pm 1.19}$ (\textcolor{blue}{9.54}) & $64.42_{\pm 0.71}$ (\textcolor{blue}{10.12}) & \textcolor{blue}{10.28} & $0.66_{\pm 0.25}$ (\textcolor{blue}{0.53}) & $1.08_{\pm 0.31}$ (\textcolor{blue}{0.03}) & $90.74_{\pm 0.54}$ (\textcolor{blue}{9.23}) & $65.77_{\pm 0.51}$ (\textcolor{blue}{9.59}) & \textcolor{blue}{4.85}\\
{{\EUk}} & $12.55_{\pm 0.63}$ (\textcolor{blue}{12.51}) & $15.04_{\pm 0.84}$ (\textcolor{blue}{34.94}) & $87.61_{\pm 0.35}$ (\textcolor{blue}{12.37}) & $58.77_{\pm 0.58}$ (\textcolor{blue}{15.77}) & \textcolor{blue}{18.90} & $4.23_{\pm 1.45}$ (\textcolor{blue}{4.10}) & $4.94_{\pm 1.64}$ (\textcolor{blue}{3.83}) & $86.87_{\pm 0.14}$ (\textcolor{blue}{13.10}) & $58.65_{\pm 0.19}$ (\textcolor{blue}{16.71}) & \textcolor{blue}{9.44}\\
{{\CFk}} & $0.02_{\pm 0.02}$ (\textcolor{blue}{25.04}) & $2.36_{\pm 0.27}$ (\textcolor{blue}{47.62}) & $99.98_{\pm 0.00}$ (\textcolor{blue}{0.00}) & $75.34_{\pm 0.10}$ (\textcolor{blue}{0.80}) & \textcolor{blue}{18.36} & $0.00_{\pm 0.00}$ (\textcolor{blue}{0.13}) & $0.12_{\pm 0.01}$ (\textcolor{blue}{0.99}) & $99.98_{\pm 0.00}$ (\textcolor{blue}{0.01}) & $75.22_{\pm 0.06}$ (\textcolor{blue}{0.14}) & \textcolor{blue}{0.32}\\
{{\MUSparse}} & $27.30_{\pm 1.49}$ (\textcolor{blue}{2.24}) & $33.11_{\pm 1.06}$ (\textcolor{blue}{16.87}) & $87.17_{\pm 2.03}$ (\textcolor{blue}{12.81}) & $63.12_{\pm 1.49}$ (\textcolor{blue}{11.42}) & \textcolor{blue}{10.84} & $1.53_{\pm 0.75}$ (\textcolor{blue}{1.40}) & $1.64_{\pm 0.82}$ (\textcolor{blue}{0.53}) & $87.19_{\pm 0.64}$ (\textcolor{blue}{12.78}) & $64.45_{\pm 0.45}$ (\textcolor{blue}{10.91}) & \textcolor{blue}{6.40}\\
\midrule
{{\RL}} & $47.18_{\pm 0.00}$ (\textcolor{blue}{22.12}) & $91.71_{\pm 0.00}$ (\textcolor{blue}{41.73}) & $99.88_{\pm 0.00}$ (\textcolor{blue}{0.10}) & $67.31_{\pm 0.00}$ (\textcolor{blue}{7.23}) & \textcolor{blue}{17.79} & $62.64_{\pm 0.00}$ (\textcolor{blue}{62.51}) & $97.18_{\pm 0.00}$ (\textcolor{blue}{96.07}) & $99.67_{\pm 0.00}$ (\textcolor{blue}{0.30}) & $66.22_{\pm 0.00}$ (\textcolor{blue}{9.14}) & \textcolor{blue}{42.00}\\
{{\BE}} & $26.39_{\pm 2.38}$ (\textcolor{blue}{1.33}) & $24.43_{\pm 1.51}$ (\textcolor{blue}{25.55}) & $76.04_{\pm 2.25}$ (\textcolor{blue}{23.94}) & $42.31_{\pm 1.34}$ (\textcolor{blue}{32.23}) & \textcolor{blue}{20.76} & $32.10_{\pm 1.10}$ (\textcolor{blue}{31.97}) & $31.18_{\pm 0.72}$ (\textcolor{blue}{30.07}) & $78.42_{\pm 2.00}$ (\textcolor{blue}{21.55}) & $47.03_{\pm 1.46}$ (\textcolor{blue}{28.33}) & \textcolor{blue}{27.98}\\
{{\BS}} & $8.44_{\pm 0.65}$ (\textcolor{blue}{16.62}) & $19.24_{\pm 1.45}$ (\textcolor{blue}{30.74}) & $93.17_{\pm 0.59}$ (\textcolor{blue}{6.81}) & $62.75_{\pm 0.27}$ (\textcolor{blue}{11.79}) & \textcolor{blue}{16.49} & $20.64_{\pm 0.57}$ (\textcolor{blue}{20.51}) & $27.10_{\pm 1.48}$ (\textcolor{blue}{25.99}) & $80.81_{\pm 0.31}$ (\textcolor{blue}{19.16}) & $52.69_{\pm 0.24}$ (\textcolor{blue}{22.67}) & \textcolor{blue}{22.08}\\
{{\Salun}} & $24.22_{\pm 0.00}$ (\textcolor{blue}{0.84}) & $77.76_{\pm 0.00}$ (\textcolor{blue}{27.78}) & $99.84_{\pm 0.00}$ (\textcolor{blue}{0.14}) & $67.64_{\pm 0.00}$ (\textcolor{blue}{6.90}) & \textcolor{blue}{8.92} & $44.76_{\pm 0.00}$ (\textcolor{blue}{44.63}) & $89.40_{\pm 0.00}$ (\textcolor{blue}{88.29}) & $99.52_{\pm 0.00}$ (\textcolor{blue}{0.45}) & $67.20_{\pm 0.00}$ (\textcolor{blue}{8.16}) & \textcolor{blue}{35.38}\\
% \midrule
\midrule
\rowcolor{Gray}
\multicolumn{11}{c}{\textbf{Tiny ImageNet}} \\
\midrule
{{\Retrain}} & $36.40_{\pm 0.25}$ & $63.77_{\pm 0.02}$ & $99.98_{\pm 0.00}$ & $63.67_{\pm 0.34}$ & \textcolor{blue}{0.00} & $0.78_{\pm 0.06}$  & $4.80_{\pm 0.25}$ & $99.98_{\pm 0.00}$ & $64.87_{\pm 0.19}$  & \textcolor{blue}{0.00}\\
\midrule
{{\FT}} & $14.41_{\pm 0.24}$ (\textcolor{blue}{21.99}) & $25.48_{\pm 0.51}$ (\textcolor{blue}{38.29}) & $98.72_{\pm 0.03}$ (\textcolor{blue}{1.26}) & $62.01_{\pm 0.20}$ (\textcolor{blue}{1.66}) & \textcolor{blue}{15.80} & $0.05_{\pm 0.00}$ (\textcolor{blue}{0.73}) & $0.14_{\pm 0.00}$ (\textcolor{blue}{4.66}) & $98.36_{\pm 0.00}$ (\textcolor{blue}{1.62}) & $61.87_{\pm 0.00}$ (\textcolor{blue}{3.00}) & \textcolor{blue}{2.50}\\
{{\EUk}} & $16.77_{\pm 0.08}$ (\textcolor{blue}{19.63}) & $23.66_{\pm 2.70}$ (\textcolor{blue}{40.11}) & $84.48_{\pm 0.48}$ (\textcolor{blue}{15.50}) & $57.70_{\pm 0.42}$ (\textcolor{blue}{5.97}) & \textcolor{blue}{20.30} & $0.20_{\pm 0.02}$ (\textcolor{blue}{0.58}) & $0.23_{\pm 0.03}$ (\textcolor{blue}{4.57}) & $83.59_{\pm 0.27}$ (\textcolor{blue}{16.39}) & $58.51_{\pm 0.34}$ (\textcolor{blue}{6.36}) & \textcolor{blue}{6.98}\\
{{\CFk}} & $13.48_{\pm 0.30}$ (\textcolor{blue}{22.92}) & $22.49_{\pm 1.80}$ (\textcolor{blue}{41.28}) & $87.98_{\pm 0.59}$ (\textcolor{blue}{12.00}) & $60.29_{\pm 0.31}$ (\textcolor{blue}{3.38}) & \textcolor{blue}{19.90} & $0.10_{\pm 0.03}$ (\textcolor{blue}{0.68}) & $0.11_{\pm 0.04}$ (\textcolor{blue}{4.69}) & $86.85_{\pm 0.21}$ (\textcolor{blue}{13.13}) & $60.37_{\pm 0.11}$ (\textcolor{blue}{4.50}) & \textcolor{blue}{5.75}\\
{{\MUSparse}} & $15.19_{\pm 0.24}$ (\textcolor{blue}{21.21}) & $26.39_{\pm 0.54}$ (\textcolor{blue}{37.38}) & $98.61_{\pm 0.04}$ (\textcolor{blue}{1.37}) & $61.78_{\pm 0.21}$ (\textcolor{blue}{1.89}) & \textcolor{blue}{15.46} & $0.11_{\pm 0.03}$ (\textcolor{blue}{0.67}) & $0.17_{\pm 0.05}$ (\textcolor{blue}{4.63}) & $98.15_{\pm 0.04}$ (\textcolor{blue}{1.83}) & $61.35_{\pm 0.12}$ (\textcolor{blue}{3.52}) & \textcolor{blue}{2.66}\\
\midrule
{{\RL}} & $29.13_{\pm 0.49}$ (\textcolor{blue}{7.27}) & $42.62_{\pm 1.89}$ (\textcolor{blue}{21.15}) & $96.25_{\pm 0.09}$ (\textcolor{blue}{3.73}) & $58.99_{\pm 0.21}$ (\textcolor{blue}{4.68}) & \textcolor{blue}{9.21} & $37.04_{\pm 0.51}$ (\textcolor{blue}{36.26}) & $47.84_{\pm 1.21}$ (\textcolor{blue}{43.04}) & $95.39_{\pm 0.08}$ (\textcolor{blue}{4.59}) & $56.64_{\pm 0.29}$ (\textcolor{blue}{8.23}) & \textcolor{blue}{23.03}\\
{{\BE}} & $47.41_{\pm 0.44}$ (\textcolor{blue}{11.01}) & $29.65_{\pm 0.23}$ (\textcolor{blue}{34.12}) & $53.14_{\pm 1.00}$ (\textcolor{blue}{46.84}) & $36.07_{\pm 0.41}$ (\textcolor{blue}{27.60}) & \textcolor{blue}{29.89} & $29.94_{\pm 0.77}$ (\textcolor{blue}{29.16}) & $39.67_{\pm 7.84}$ (\textcolor{blue}{34.87}) & $34.66_{\pm 0.26}$ (\textcolor{blue}{65.32}) & $26.63_{\pm 0.27}$ (\textcolor{blue}{38.24}) & \textcolor{blue}{41.90}\\
{{\BS}} & $30.32_{\pm 0.91}$ (\textcolor{blue}{6.08}) & $25.45_{\pm 1.02}$ (\textcolor{blue}{38.32}) & $70.48_{\pm 0.80}$ (\textcolor{blue}{29.50}) & $47.00_{\pm 0.53}$ (\textcolor{blue}{16.67}) & \textcolor{blue}{22.64} & $24.32_{\pm 0.63}$ (\textcolor{blue}{23.54}) & $14.40_{\pm 0.55}$ (\textcolor{blue}{9.60}) & $54.39_{\pm 0.55}$ (\textcolor{blue}{45.59}) & $37.54_{\pm 0.20}$ (\textcolor{blue}{27.33}) & \textcolor{blue}{26.52}\\
{{\Salun}} & $26.18_{\pm 0.50}$ (\textcolor{blue}{10.22}) & $38.02_{\pm 1.42}$ (\textcolor{blue}{25.75}) & $95.90_{\pm 0.07}$ (\textcolor{blue}{4.08}) & $59.20_{\pm 0.16}$ (\textcolor{blue}{4.47}) & \textcolor{blue}{11.13} & $25.84_{\pm 0.10}$ (\textcolor{blue}{25.06}) & $37.03_{\pm 0.25}$ (\textcolor{blue}{32.23}) & $95.94_{\pm 0.06}$ (\textcolor{blue}{4.04}) & $59.12_{\pm 0.15}$ (\textcolor{blue}{5.75}) & \textcolor{blue}{16.77}\\
\midrule

\bottomrule[1pt]
\end{tabular}
}
\end{center}
\vspace*{-5mm}
\end{table}

\begin{table}[t]
\caption{{Performance of various unlearning methods under {\randomset}s and {\ourset}s on CIFAR-10 using ResNet-50 and VGG-16 for forgetting ratio 10\%. The result format follows Table\,\ref{tab: main_approximate_unlearn}.}}
\vspace{-6mm}
\label{tab: more_model}
\begin{center}
\resizebox{\textwidth}{!}{
\renewcommand\tabcolsep{5pt}
\begin{tabular}{c|ccccc|ccccc}
\toprule[1pt]
\midrule
\multirow{2}{*}{\textbf{Methods}} & \multicolumn{5}{c|}{\textbf{\RandomSet}}  & \multicolumn{5}{c}{\textbf{\OurSet}}  \\
                        & \multicolumn{1}{c|}{UA}   & \multicolumn{1}{c|}{MIA}     & \multicolumn{1}{c|}{RA} & \multicolumn{1}{c|}{TA} & \multicolumn{1}{c|}{Avg. Gap} 
                        & \multicolumn{1}{c|}{UA}   & \multicolumn{1}{c|}{MIA}     & \multicolumn{1}{c|}{RA}    & \multicolumn{1}{c|}{TA} & \multicolumn{1}{c}{Avg. Gap} \\
\midrule
\rowcolor{Gray}
\multicolumn{11}{c}{\textbf{ResNet-50}} \\
\midrule
{{\Retrain}} & $5.56_{\pm 0.35}$ & $11.68_{\pm 0.66}$ & $100.00_{\pm 0.00}$ & $94.17_{\pm 0.01}$ & \textcolor{blue}{0.00} & $0.00_{\pm 0.00}$ & $0.00_{\pm 0.00}$ & $100.00_{\pm 0.00}$& $94.04_{\pm 0.30}$ & \textcolor{blue}{0.00}\\
% \midrule
% \rowcolor{Gray}
% \multicolumn{11}{c}{\textbf{Relabeling-free}} \\
\midrule
{{\FT}} & $4.48_{\pm 0.20}$ (\textcolor{blue}{1.08}) & $10.05_{\pm 0.71}$ (\textcolor{blue}{1.63}) & $98.13_{\pm 0.38}$ (\textcolor{blue}{1.87}) & $91.47_{\pm 0.23}$ (\textcolor{blue}{2.70}) & \textcolor{blue}{1.82} & $0.01_{\pm 0.01}$ (\textcolor{blue}{0.01}) & $0.04_{\pm 0.03}$ (\textcolor{blue}{0.04}) & $97.55_{\pm 0.33}$ (\textcolor{blue}{2.45}) & $91.38_{\pm 0.49}$ (\textcolor{blue}{2.66}) & \textcolor{blue}{1.29}\\
{{\EUk}} & $4.54_{\pm 0.16}$ (\textcolor{blue}{1.02}) & $7.94_{\pm 0.18}$ (\textcolor{blue}{3.74}) & $95.53_{\pm 0.21}$ (\textcolor{blue}{4.47}) & $87.34_{\pm 0.08}$ (\textcolor{blue}{6.83}) & \textcolor{blue}{4.02} & $1.59_{\pm 0.67}$ (\textcolor{blue}{1.59}) & $3.68_{\pm 1.46}$ (\textcolor{blue}{3.68}) & $95.56_{\pm 0.47}$ (\textcolor{blue}{4.44}) & $87.61_{\pm 0.21}$ (\textcolor{blue}{6.43}) & \textcolor{blue}{4.04}\\
{{\CFk}} & $0.01_{\pm 0.01}$ (\textcolor{blue}{5.55}) & $0.53_{\pm 0.06}$ (\textcolor{blue}{11.15}) & $100.00_{\pm 0.00}$ (\textcolor{blue}{0.00}) & $94.16_{\pm 0.00}$ (\textcolor{blue}{0.01}) & \textcolor{blue}{4.18} & $0.00_{\pm 0.00}$ (\textcolor{blue}{0.00}) & $0.01_{\pm 0.01}$ (\textcolor{blue}{0.01}) & $99.99_{\pm 0.00}$ (\textcolor{blue}{0.01}) & $94.06_{\pm 0.04}$ (\textcolor{blue}{0.02}) & \textcolor{blue}{0.01}\\
{{\MUSparse}} & $2.38_{\pm 0.12}$ (\textcolor{blue}{3.18}) & $7.49_{\pm 0.39}$ (\textcolor{blue}{4.19}) & $98.91_{\pm 0.01}$ (\textcolor{blue}{1.09}) & $92.53_{\pm 0.07}$ (\textcolor{blue}{1.64}) & \textcolor{blue}{2.53} & $0.00_{\pm 0.00}$ (\textcolor{blue}{0.00}) & $0.03_{\pm 0.03}$ (\textcolor{blue}{0.03}) & $98.34_{\pm 0.15}$ (\textcolor{blue}{1.66}) & $92.08_{\pm 0.02}$ (\textcolor{blue}{1.96}) & \textcolor{blue}{0.91}\\
\midrule
{{\RL}} & $7.24_{\pm 0.46}$ (\textcolor{blue}{1.68}) & $43.07_{\pm 4.11}$ (\textcolor{blue}{31.39}) & $99.79_{\pm 0.02}$ (\textcolor{blue}{0.21}) & $92.52_{\pm 0.08}$ (\textcolor{blue}{1.65}) & \textcolor{blue}{8.73} & $1.01_{\pm 0.47}$ (\textcolor{blue}{1.01}) & $95.76_{\pm 0.86}$ (\textcolor{blue}{95.76}) & $99.82_{\pm 0.07}$ (\textcolor{blue}{0.18}) & $93.25_{\pm 0.25}$ (\textcolor{blue}{0.79}) & \textcolor{blue}{24.44}\\
{{\BE}} & $3.54_{\pm 0.54}$ (\textcolor{blue}{2.02}) & $15.70_{\pm 0.71}$ (\textcolor{blue}{4.02}) & $96.36_{\pm 0.12}$ (\textcolor{blue}{3.64}) & $89.75_{\pm 0.10}$ (\textcolor{blue}{4.42}) & \textcolor{blue}{3.52} & $19.24_{\pm 1.14}$ (\textcolor{blue}{19.24}) & $47.57_{\pm 0.61}$ (\textcolor{blue}{47.57}) & $87.59_{\pm 0.52}$ (\textcolor{blue}{12.41}) & $79.63_{\pm 0.39}$ (\textcolor{blue}{14.41}) & \textcolor{blue}{23.41}\\
{{\BS}} & $3.62_{\pm 0.52}$ (\textcolor{blue}{1.94}) & $10.75_{\pm 0.71}$ (\textcolor{blue}{0.93}) & $96.27_{\pm 0.27}$ (\textcolor{blue}{3.73}) & $89.72_{\pm 0.26}$ (\textcolor{blue}{4.45}) & \textcolor{blue}{2.76} & $14.27_{\pm 1.13}$ (\textcolor{blue}{14.27}) & $37.49_{\pm 1.12}$ (\textcolor{blue}{37.49}) & $88.56_{\pm 0.15}$ (\textcolor{blue}{11.44}) & $81.18_{\pm 0.36}$ (\textcolor{blue}{12.86}) & \textcolor{blue}{19.02}\\
{{\Salun}} & $1.67_{\pm 0.15}$ (\textcolor{blue}{3.89}) & $19.44_{\pm 1.95}$ (\textcolor{blue}{7.76}) & $99.94_{\pm 0.01}$ (\textcolor{blue}{0.06}) & $93.44_{\pm 0.07}$ (\textcolor{blue}{0.73}) & \textcolor{blue}{3.11} & $0.34_{\pm 0.13}$ (\textcolor{blue}{0.34}) & $92.83_{\pm 0.59}$ (\textcolor{blue}{92.83}) & $98.99_{\pm 0.26}$ (\textcolor{blue}{1.01}) & $92.08_{\pm 0.20}$ (\textcolor{blue}{1.96}) & \textcolor{blue}{24.04}\\
% \midrule
\midrule
\rowcolor{Gray}
\multicolumn{11}{c}{\textbf{VGG-16}} \\
\midrule
{{\Retrain}} & $6.76_{\pm 0.43}$ & $11.77_{\pm 0.27}$ & $99.99_{\pm 0.00}$ & $93.28_{\pm 0.15}$ & \textcolor{blue}{0.00} & $0.01_{\pm 0.01}$ & $0.07_{\pm 0.04}$ & $99.99_{\pm 0.00}$ & $93.43_{\pm 0.13}$ & \textcolor{blue}{0.00}\\
% \midrule
% \rowcolor{Gray}
% \multicolumn{11}{c}{\textbf{Relabeling-free}} \\
\midrule
{{\FT}} & $3.91_{\pm 0.68}$ (\textcolor{blue}{2.85}) & $8.75_{\pm 0.94}$ (\textcolor{blue}{3.02}) & $98.31_{\pm 0.35}$ (\textcolor{blue}{1.68}) & $90.62_{\pm 0.38}$ (\textcolor{blue}{2.66}) & \textcolor{blue}{2.55} & $0.07_{\pm 0.07}$ (\textcolor{blue}{0.06}) & $0.28_{\pm 0.29}$ (\textcolor{blue}{0.21}) & $97.36_{\pm 0.45}$ (\textcolor{blue}{2.63}) & $90.04_{\pm 0.30}$ (\textcolor{blue}{3.39}) & \textcolor{blue}{1.57}\\
{{\EUk}} & $15.79_{\pm 7.41}$ (\textcolor{blue}{9.03}) & $19.61_{\pm 5.29}$ (\textcolor{blue}{7.84}) & $83.63_{\pm 8.02}$ (\textcolor{blue}{16.36}) & $76.36_{\pm 6.99}$ (\textcolor{blue}{16.92}) & \textcolor{blue}{12.54} & $2.25_{\pm 2.18}$ (\textcolor{blue}{2.24}) & $3.08_{\pm 1.86}$ (\textcolor{blue}{3.01}) & $83.51_{\pm 9.61}$ (\textcolor{blue}{16.48}) & $77.57_{\pm 7.75}$ (\textcolor{blue}{15.86}) & \textcolor{blue}{9.40}\\
{{\CFk}} & $0.02_{\pm 0.00}$ (\textcolor{blue}{6.74}) & $0.33_{\pm 0.06}$ (\textcolor{blue}{11.44}) & $99.99_{\pm 0.00}$ (\textcolor{blue}{0.00}) & $93.59_{\pm 0.01}$ (\textcolor{blue}{0.31}) & \textcolor{blue}{4.62} & $0.00_{\pm 0.00}$ (\textcolor{blue}{0.01}) & $0.00_{\pm 0.00}$ (\textcolor{blue}{0.07}) & $99.98_{\pm 0.01}$ (\textcolor{blue}{0.01}) & $93.54_{\pm 0.07}$ (\textcolor{blue}{0.11}) & \textcolor{blue}{0.05}\\
{{\MUSparse}} & $4.48_{\pm 0.43}$ (\textcolor{blue}{2.28}) & $9.76_{\pm 0.32}$ (\textcolor{blue}{2.01}) & $97.68_{\pm 0.16}$ (\textcolor{blue}{2.31}) & $90.61_{\pm 0.13}$ (\textcolor{blue}{2.67}) & \textcolor{blue}{2.32} & $0.04_{\pm 0.04}$ (\textcolor{blue}{0.03}) & $0.17_{\pm 0.19}$ (\textcolor{blue}{0.10}) & $97.56_{\pm 0.12}$ (\textcolor{blue}{2.43}) & $90.36_{\pm 0.15}$ (\textcolor{blue}{3.07}) & \textcolor{blue}{1.41}\\
\midrule
{{\RL}} & $2.71_{\pm 0.29}$ (\textcolor{blue}{4.05}) & $14.01_{\pm 2.29}$ (\textcolor{blue}{2.24}) & $99.97_{\pm 0.00}$ (\textcolor{blue}{0.02}) & $92.92_{\pm 0.04}$ (\textcolor{blue}{0.36}) & \textcolor{blue}{1.67} & $3.50_{\pm 4.09}$ (\textcolor{blue}{3.49}) & $95.96_{\pm 0.85}$ (\textcolor{blue}{95.89}) & $99.89_{\pm 0.02}$ (\textcolor{blue}{0.10}) & $92.65_{\pm 0.49}$ (\textcolor{blue}{0.78}) & \textcolor{blue}{25.06}\\
{{\BE}} & $11.73_{\pm 4.40}$ (\textcolor{blue}{4.97}) & $26.33_{\pm 10.56}$ (\textcolor{blue}{14.56}) & $88.22_{\pm 4.15}$ (\textcolor{blue}{11.77}) & $80.53_{\pm 3.41}$ (\textcolor{blue}{12.75}) & \textcolor{blue}{11.01} & $49.28_{\pm 22.21}$ (\textcolor{blue}{49.27}) & $67.18_{\pm 8.23}$ (\textcolor{blue}{67.11}) & $77.53_{\pm 1.63}$ (\textcolor{blue}{22.46}) & $68.02_{\pm 3.24}$ (\textcolor{blue}{25.41}) & \textcolor{blue}{41.06}\\
{{\BS}} & $7.46_{\pm 3.27}$ (\textcolor{blue}{0.70}) & $8.61_{\pm 1.14}$ (\textcolor{blue}{3.16}) & $92.86_{\pm 3.01}$ (\textcolor{blue}{7.13}) & $84.23_{\pm 2.88}$ (\textcolor{blue}{9.05}) & \textcolor{blue}{5.01} & $52.39_{\pm 31.85}$ (\textcolor{blue}{52.38}) & $53.78_{\pm 35.44}$ (\textcolor{blue}{53.71}) & $75.38_{\pm 2.55}$ (\textcolor{blue}{24.61}) & $65.64_{\pm 4.96}$ (\textcolor{blue}{27.79}) & \textcolor{blue}{39.62}\\
{{\Salun}} & $6.38_{\pm 0.49}$ (\textcolor{blue}{0.38}) & $18.66_{\pm 1.66}$ (\textcolor{blue}{6.89}) & $99.76_{\pm 0.14}$ (\textcolor{blue}{0.23}) & $91.88_{\pm 0.41}$ (\textcolor{blue}{1.40}) & \textcolor{blue}{2.22} & $3.21_{\pm 4.48}$ (\textcolor{blue}{3.20}) & $94.61_{\pm 0.26}$ (\textcolor{blue}{94.54}) & $99.29_{\pm 0.17}$ (\textcolor{blue}{0.70}) & $91.46_{\pm 0.56}$ (\textcolor{blue}{1.97}) & \textcolor{blue}{25.10}\\
\midrule

\bottomrule[1pt]
\end{tabular}
}
\end{center}
\vspace*{-8mm}
\end{table}

\section{Additional Experiment Results}
\label{app: experiment_results}

\subsection{{Additional results of Table\,\ref{tab: main_approximate_unlearn}}}

As an expansion of Table\,\ref{tab: main_approximate_unlearn}, 
 \textbf{Table\,\ref{tab: more_ratio}} 
 details the performance of various approximate unlearning methods for both random and worst-case forget sets, with data forgetting ratios of 1\%, 5\%, 10\%, and 20\% on CIFAR-10. For worst-case forget sets, relabeling-free unlearning methods often follow a performance trend similar to Retrain. On the other hand, relabeling-based unlearning methods display a significant performance discrepancy from Retrain, highlighting the impact of the unlearning strategy on method efficacy.

\subsection{Additional results on CIFAR-100 and Tiny ImageNet}
In \textbf{Table\,\ref{tab: more_dataset}}, we present the performance of various unlearning methods under random and worst-case forget sets at a 10\% forgetting data ratio on the additional datasets, CIFAR-100 \cite{krizhevsky2009learning} and Tiny ImageNet \cite{le2015tiny}. When subjected to evaluation on worst-case forget sets, relabeling-free approximate unlearning methods consistently display a performance trend akin to that of \Retrain. However, in sharp contrast to their relabeling-free counterparts, relabeling-based approximate unlearning methods manifest a discernible performance gap when compared to \Retrain.

\subsection{Additional results on different model architectures}

In \textbf{Table\,\ref{tab: more_model}}, we comprehensively assess the performance of diverse unlearning techniques under both random and worst-case forget sets scenarios, employing a 10\% forgetting data ratio on CIFAR-10. This evaluation encompasses a broadened range of model architectures, including ResNet-50 \cite{he2016deep} and VGG-16 \cite{simonyan2014very}. When evaluating the methods on worst-case forget set, the relabeling-free approximate unlearning methods consistently exhibit a performance trend that closely resembles that of {\Retrain}. Conversely, relabeling-based approximate unlearning methods demonstrate a notable performance discrepancy when compared to {\Retrain}.

\begin{figure}[tb]
    \centering
    \begin{minipage}{0.49\textwidth}
        \centering
        \vspace*{5.5mm}
        \includegraphics[width=\textwidth]{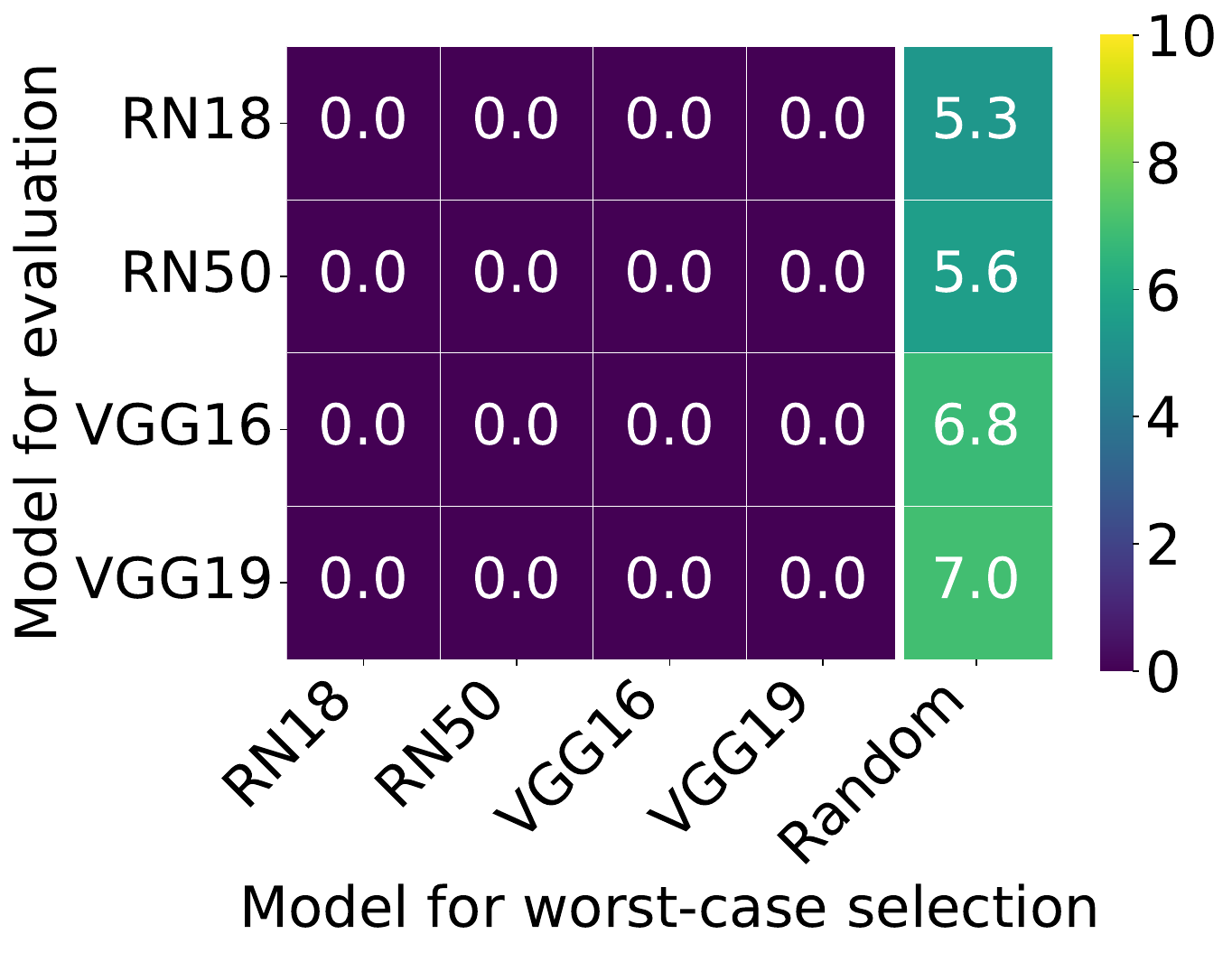}
        \vspace*{-3.5mm}
        \captionof{figure}{
        UA of {\Retrain} on CIFAR-10 using various models with 10\% forgetting ratio. The rightmost column represents UA on \randomset, while other columns depict UA on \ourset.
        }
        % \vspace{-9mm}
        \label{fig: transfer_model}
    \end{minipage} \hfill
    \begin{minipage}{0.49\textwidth}
        \centering
        \includegraphics[width=\textwidth]{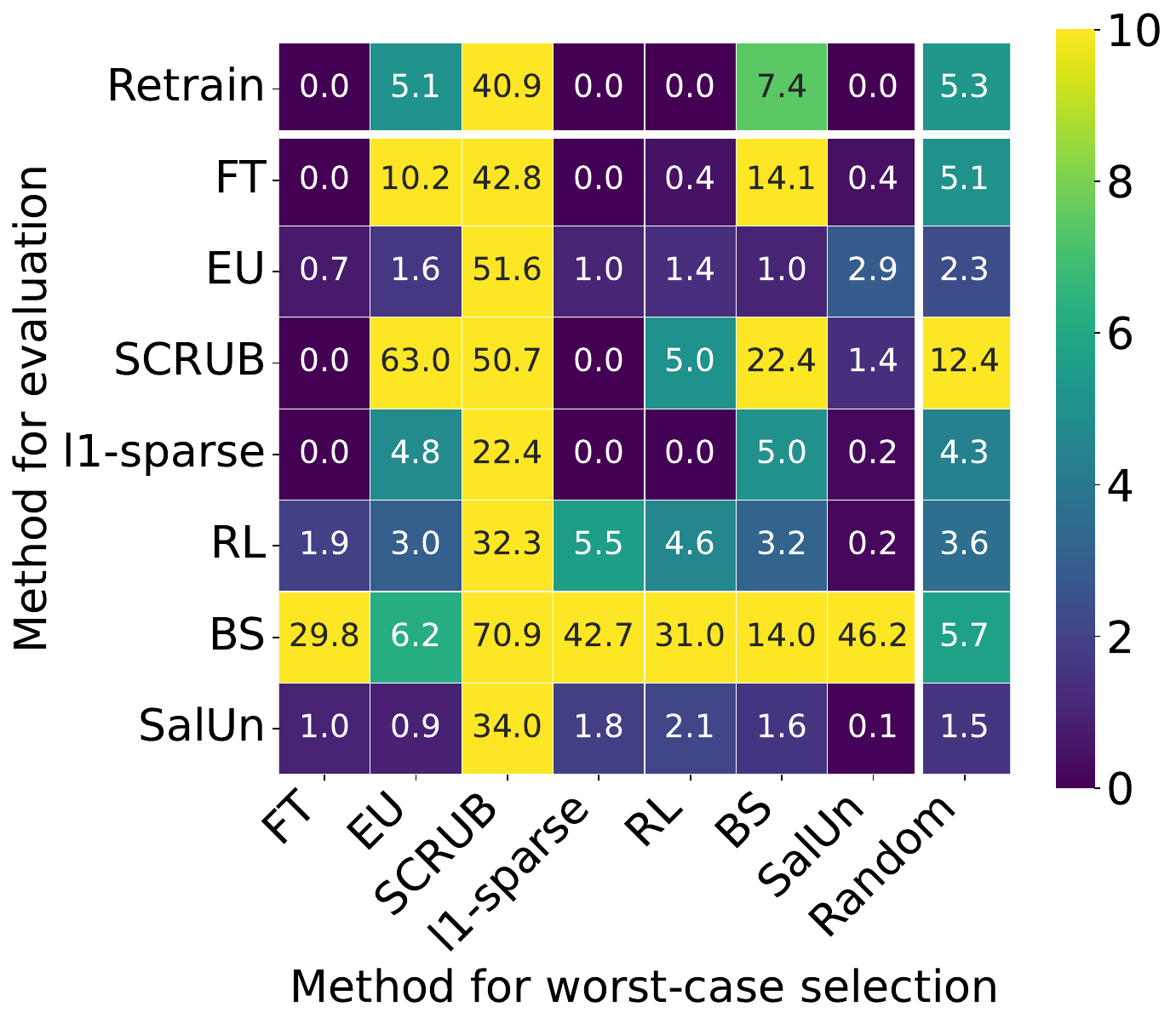}
        % \vspace*{-2mm}
        \captionof{figure}{
        \footnotesize{
        UA of unlearning methods on CIFAR-10 using ResNet-18 with a 10\% forgetting ratio. The rightmost column shows UA on \randomset; others show UA on \ourset.
        }
        }
        \label{fig: transfer_method}
    \end{minipage}
    \vspace{-5mm}
\end{figure}

\subsection{Transferability of {\ourset}s between different models and methods}
\label{app: transferability}

\begin{wraptable}{r}{0.5\textwidth}
    \centering
    \vspace*{-11.5mm}
    \caption{
    Performance of various unlearning methods on CIFAR-10 using ResNet-18 with a 10\% forgetting ratio under {\ourset}s obtained using RL. The result format follows Table\,\ref{tab: main_approximate_unlearn}. (\textcolor{black}{$\bullet$}) after {\Retrain} in {\OurSet} indicates the difference from {\RandomSet}.
    }
    \vspace*{2mm}
    \label{tab: rl_lower_level}
    \resizebox{0.5\textwidth}{!}{
    \renewcommand\tabcolsep{5pt}
    \begin{tabular}{c|ccccc}
    \toprule[1pt]
    \midrule
    \textbf{Methods} & \multicolumn{1}{c|}{\textbf{UA}} & \multicolumn{1}{c|}{\textbf{MIA}} & \multicolumn{1}{c|}{\textbf{RA}} & \multicolumn{1}{c|}{\textbf{TA}} & \multicolumn{1}{c}{\textbf{Avg. Gap}} \\
    \midrule
    \rowcolor{Gray}
    \multicolumn{6}{c}{\textbf{\RandomSet}} \\
    \midrule
    \Retrain & $5.28$ & $12.86$ & $100.00$ & $94.38$ & N/A \\
    \midrule
    \rowcolor{Gray}
    \multicolumn{6}{c}{\textbf{\OurSet}} \\
    \midrule
    \Retrain & $0.02$ (\textcolor{Green}{$ 5.26 \blacktriangledown$}) & $0.16$ (\textcolor{Green}{$ 12.70 \blacktriangledown$}) & $100.00$ (\textcolor{Blue}{$ 0.00 - $}) & $94.67$ (\textcolor{Red}{$ 0.29 \blacktriangle$}) & \textcolor{blue}{$0$} \\
    \midrule
    \rowcolor{Gray}
    \FT & $0.44$ (\textcolor{blue}{$0.42$}) & $0.44$ (\textcolor{blue}{$0.29$}) & $92.94$ (\textcolor{blue}{$7.06$}) & $87.93$ (\textcolor{blue}{$6.74$}) & \textcolor{blue}{$3.63$} \\
    \EUk & $1.36$ (\textcolor{blue}{$1.33$}) & $11.93$ (\textcolor{blue}{$11.78$}) & $97.99$ (\textcolor{blue}{$2.01$}) & $90.75$ (\textcolor{blue}{$3.92$}) & \textcolor{blue}{$4.76$} \\
    \rowcolor{Gray}
    \Scrub & $5.04$ (\textcolor{blue}{$5.02$}) & $20.71$ (\textcolor{blue}{$20.56$}) & $89.92$ (\textcolor{blue}{$10.08$}) & $85.70$ (\textcolor{blue}{$8.97$}) & \textcolor{blue}{$11.16$} \\
    \MUSparse & $0.04$ (\textcolor{blue}{$0.02$}) & $0.13$ (\textcolor{blue}{$0.02$}) & $97.59$ (\textcolor{blue}{$2.41$}) & $91.62$ (\textcolor{blue}{$3.05$}) & \textcolor{blue}{$1.38$} \\
    \midrule
    \rowcolor{Gray}
    \RL & $4.62$ (\textcolor{blue}{$4.60$}) & $98.36$ (\textcolor{blue}{$98.20$}) & $99.96$ (\textcolor{blue}{$0.04$}) & $93.59$ (\textcolor{blue}{$1.08$}) & \textcolor{blue}{$25.98$} \\
    \BE & $38.38$ (\textcolor{blue}{$38.36$}) & $96.64$ (\textcolor{blue}{$96.49$}) & $77.00$ (\textcolor{blue}{$23.00$}) & $71.42$ (\textcolor{blue}{$23.25$}) & \textcolor{blue}{$45.27$} \\
    \rowcolor{Gray}
    \BS & $31.02$ (\textcolor{blue}{$31.00$}) & $96.82$ (\textcolor{blue}{$96.67$}) & $81.88$ (\textcolor{blue}{$18.12$}) & $75.62$ (\textcolor{blue}{$19.05$}) & \textcolor{blue}{$41.21$} \\
    \Salun & $2.09$ (\textcolor{blue}{$2.07$}) & $97.40$ (\textcolor{blue}{$97.24$}) & $99.98$ (\textcolor{blue}{$0.02$}) & $94.00$ (\textcolor{blue}{$0.67$}) & \textcolor{blue}{$25.00$} \\
    \midrule
    \bottomrule[1pt]
    \end{tabular}
    }
    \vspace*{-7mm}
\end{wraptable}

In this section, we validate the transferability of {\ourset}s across a wider range of model architectures and methods.
Concerning the transferability between models, we leverage a diverse range of models for the selection process, including ResNet-18, ResNet-50 \cite{he2016deep}, VGG-16, and VGG-19 \cite{simonyan2014very}. Conversely, for the evaluation process, we employ various models and adopt {\Retrain} as the corresponding unlearning method. The UA (unlearning accuracy) results are exhibited in \textbf{Fig.\,\ref{fig: transfer_model}}. Notably, when the {\ourset} is selected using one model and subsequently undergoes unlearning with another model, the UA shows significantly lower than that of \randomset. This observation clearly demonstrates the transferability of {\ourset} across diverse models.

Regarding the transferability between methods, we employ various approximate unlearning objectives in the selection process to specify the lower-level optimization problem \eqref{eq: L-signSGD}, while utilizing different unlearning methods during evaluation. The UA results are illustrated in \textbf{Fig.\,\ref{fig: transfer_method}}. As evident from the figure, the columns corresponding to the four methods, \FT, \MUSparse, \RL, and \Salun, exhibit deeper shades than the column for random, indicating lower UA values. Consequently, \FT, \MUSparse, \RL, and {\Salun} are more suitable for addressing the lower-level problem. In \textbf{Table\,\ref{tab: rl_lower_level}}, we further test the unlearning methods under the {\ourset} obtained
using RL to perform the lower-level unlearning process in BLO. The results are consistent with Fig.\,\ref{fig: transfer_method}.

\subsection{From worst-case unlearning to easiest-case unlearning}
\label{app: uniqueness}
\begin{table*}[t]
\caption{{Performance of approximate unlearning methods under {\randomset} and easiest-case forget set on CIFAR-10 using ResNet-18 with forgetting ratio 10\%.
The result format follows Table\,\ref{tab: main_approximate_unlearn}.}}

\vspace*{-8mm}
\label{tab: easy_to_unlearn}
\begin{center}
\resizebox{\textwidth}{!}{

\renewcommand\tabcolsep{5pt}
\begin{tabular}{c|ccccc|ccccc}
\toprule[1pt]
\midrule
\multirow{2}{*}{\textbf{Methods}} & \multicolumn{5}{c|}{\textbf{\RandomSet}} & \multicolumn{5}{c}{\textbf{Easiest-Case Forget Set}}  \\
                        & \multicolumn{1}{c|}{UA}   & \multicolumn{1}{c|}{MIA}     & \multicolumn{1}{c|}{RA} & \multicolumn{1}{c|}{TA} & \multicolumn{1}{c|}{Avg. Gap} 
                        & \multicolumn{1}{c|}{UA}   & \multicolumn{1}{c|}{MIA}     & \multicolumn{1}{c|}{RA}    & \multicolumn{1}{c|}{TA} & \multicolumn{1}{c}{Avg. Gap} \\
\midrule
{{\Retrain}} & $5.28_{\pm 0.33}$ & $12.86_{\pm 0.61}$ & $100.00_{\pm 0.00}$ & $94.38_{\pm 0.15}$ & \textcolor{blue}{0.00} & $43.18_{\pm 1.06}$  & $67.72_{\pm 0.87}$  & $100.00_{\pm 0.00}$  & $93.15_{\pm 0.14}$  & \textcolor{blue}{0.00}\\
\midrule
\rowcolor{Gray}
\multicolumn{11}{c}{\textbf{Relabeling-free}} \\
\midrule
{{\FT}} & $5.08_{\pm 0.39}$ (\textcolor{blue}{0.20}) & $10.96_{\pm 0.38}$ (\textcolor{blue}{1.90}) & $97.46_{\pm 0.52}$ (\textcolor{blue}{2.54}) & $91.02_{\pm 0.36}$ (\textcolor{blue}{3.36}) & \textcolor{blue}{2.00} & $28.86_{\pm 0.99}$ (\textcolor{blue}{14.32}) & $49.39_{\pm 1.83}$ (\textcolor{blue}{18.33}) & $98.49_{\pm 0.40}$ (\textcolor{blue}{1.51}) & $90.90_{\pm 0.43}$ (\textcolor{blue}{2.25}) & \textcolor{blue}{9.10}\\
{{\EUk}} & $2.34_{\pm 0.79}$ (\textcolor{blue}{2.94}) & $6.35_{\pm 0.89}$ (\textcolor{blue}{6.51}) & $97.52_{\pm 0.89}$ (\textcolor{blue}{2.48}) & $90.17_{\pm 0.88}$ (\textcolor{blue}{4.21}) & \textcolor{blue}{4.04} & $9.48_{\pm 4.75}$ (\textcolor{blue}{33.70}) & $19.27_{\pm 6.13}$ (\textcolor{blue}{48.45}) & $97.87_{\pm 1.01}$ (\textcolor{blue}{2.13}) & $89.81_{\pm 1.17}$ (\textcolor{blue}{3.34}) & \textcolor{blue}{21.90}\\
{{\CFk}} & $0.02_{\pm 0.02}$ (\textcolor{blue}{5.26}) & $0.76_{\pm 0.02}$ (\textcolor{blue}{12.10}) & $99.98_{\pm 0.00}$ (\textcolor{blue}{0.02}) & $94.45_{\pm 0.02}$ (\textcolor{blue}{0.07}) & \textcolor{blue}{4.36} & $0.10_{\pm 0.03}$ (\textcolor{blue}{43.08}) & $3.09_{\pm 0.13}$ (\textcolor{blue}{64.63}) & $99.99_{\pm 0.00}$ (\textcolor{blue}{0.01}) & $94.40_{\pm 0.03}$ (\textcolor{blue}{1.25}) & \textcolor{blue}{27.24}\\
% {{\Scrub}} & $59.30_{\pm 3.51}$ (\textcolor{blue}{16.12}) & $75.77_{\pm 5.03}$ (\textcolor{blue}{8.05}) & $82.10_{\pm 3.91}$ (\textcolor{blue}{17.90}) & $75.88_{\pm 3.44}$ (\textcolor{blue}{17.27}) & \textcolor{blue}{14.84}\\
{{\MUSparse}} & $4.34_{\pm 0.73}$ (\textcolor{blue}{0.94}) & $9.82_{\pm 1.04}$ (\textcolor{blue}{3.04}) & $97.70_{\pm 0.72}$ (\textcolor{blue}{2.30}) & $91.41_{\pm 0.68}$ (\textcolor{blue}{2.97}) & \textcolor{blue}{2.31} & $26.26_{\pm 1.24}$ (\textcolor{blue}{16.92}) & $46.56_{\pm 1.75}$ (\textcolor{blue}{21.16}) & $98.55_{\pm 0.14}$ (\textcolor{blue}{1.45}) & $90.75_{\pm 0.43}$ (\textcolor{blue}{2.40}) & \textcolor{blue}{10.48}\\
\midrule
\rowcolor{Gray}
\multicolumn{11}{c}{\textbf{Relabeling-based}} \\
\midrule
{{\RL}} & $3.59_{\pm 0.24}$ (\textcolor{blue}{1.69}) & $28.02_{\pm 2.47}$ (\textcolor{blue}{15.16}) & $99.97_{\pm 0.01}$ (\textcolor{blue}{0.03}) & $93.74_{\pm 0.12}$ (\textcolor{blue}{0.64}) & \textcolor{blue}{4.38} & $22.84_{\pm 2.76}$ (\textcolor{blue}{20.34}) & $92.51_{\pm 0.98}$ (\textcolor{blue}{24.79}) & $99.95_{\pm 0.01}$ (\textcolor{blue}{0.05}) & $92.73_{\pm 0.25}$ (\textcolor{blue}{0.42}) & \textcolor{blue}{11.40}\\
{{\BE}} & $1.19_{\pm 0.49}$ (\textcolor{blue}{4.09}) & $22.06_{\pm 0.61}$ (\textcolor{blue}{9.20}) & $98.77_{\pm 0.41}$ (\textcolor{blue}{1.23}) & $91.79_{\pm 0.32}$ (\textcolor{blue}{2.59}) & \textcolor{blue}{4.28} & $13.21_{\pm 1.48}$ (\textcolor{blue}{29.97}) & $33.93_{\pm 3.58}$ (\textcolor{blue}{33.79}) & $97.19_{\pm 1.59}$ (\textcolor{blue}{2.81}) & $88.65_{\pm 1.31}$ (\textcolor{blue}{4.50}) & \textcolor{blue}{17.77}\\
{{\BS}} & $5.72_{\pm 1.42}$ (\textcolor{blue}{0.44}) & $27.15_{\pm 1.41}$ (\textcolor{blue}{14.29}) & $94.29_{\pm 1.06}$ (\textcolor{blue}{5.71}) & $87.45_{\pm 1.06}$ (\textcolor{blue}{6.93}) & \textcolor{blue}{6.84} & $17.55_{\pm 1.00}$ (\textcolor{blue}{25.63}) & $34.93_{\pm 3.03}$ (\textcolor{blue}{32.79}) & $96.33_{\pm 0.55}$ (\textcolor{blue}{3.67}) & $87.46_{\pm 0.48}$ (\textcolor{blue}{5.69}) & \textcolor{blue}{16.94}\\
{{\Salun}} & $1.48_{\pm 0.14}$ (\textcolor{blue}{3.80}) & $16.19_{\pm 0.34}$ (\textcolor{blue}{3.33}) & $99.98_{\pm 0.01}$ (\textcolor{blue}{0.02}) & $93.95_{\pm 0.01}$ (\textcolor{blue}{0.43}) & \textcolor{blue}{1.89} & $16.76_{\pm 1.71}$ (\textcolor{blue}{26.42}) & $88.71_{\pm 0.94}$ (\textcolor{blue}{20.99}) & $99.95_{\pm 0.01}$ (\textcolor{blue}{0.05}) & $92.85_{\pm 0.19}$ (\textcolor{blue}{0.30}) & \textcolor{blue}{11.94}\\
\midrule
\bottomrule[1pt]
\end{tabular}
}
\end{center}
\vspace*{-11mm}
\end{table*}

By considering the opposite objective function of the upper-level optimization in \eqref{eq: bilevel_MU}, we can obtain the problem of selecting the easiest-case forget sets through BLO:
\begin{align}
    \begin{array}{ll}
    \displaystyle \min_{\bw \in \gS} & 
   \displaystyle   \sum_{ \mathbf z_i \in  \train} - [w_i \ell ( \thetau(\mathbf w); \mathbf z_i ) ] + \gamma \|\bw\|_2^2 ;
    \st ~
   \displaystyle  \thetau(\mathbf w) = \argmin_{\btheta} ~ \ell_\MU(\btheta;   \bw),
    \end{array}
    \label{eq: easiest_bilevel_MU}
\end{align}%
\textbf{Table\,\ref{tab: easy_to_unlearn}} presents the performance of identified easiest-case forget sets.  We draw two observations. 
\textit{First}, for \Retrain, UA and MIA on easiest-case forget sets are significantly higher than those on \randomset. 
\textit{Second}, for approximate unlearning methods, Avg. Gap on easiest-case forget set is much higher than that on the \randomset. This is due to the significantly lower UA of approximate unlearning methods on easiest-case forget set compared to that of \Retrain. This suggests that the current approximate unlearning methods are not yet effective enough, even for data in easiest-case forget set, and cannot accurately forget them.

\subsection{Uniqueness and mixture of {\ourset}}

\begin{wrapfigure}{r}{0.491\textwidth}
\centering
\vspace{-8mm}
% \hspace*{-4mm}
\begin{tabular}{cc}
\vspace{-1mm}
\includegraphics[width=0.230\textwidth]{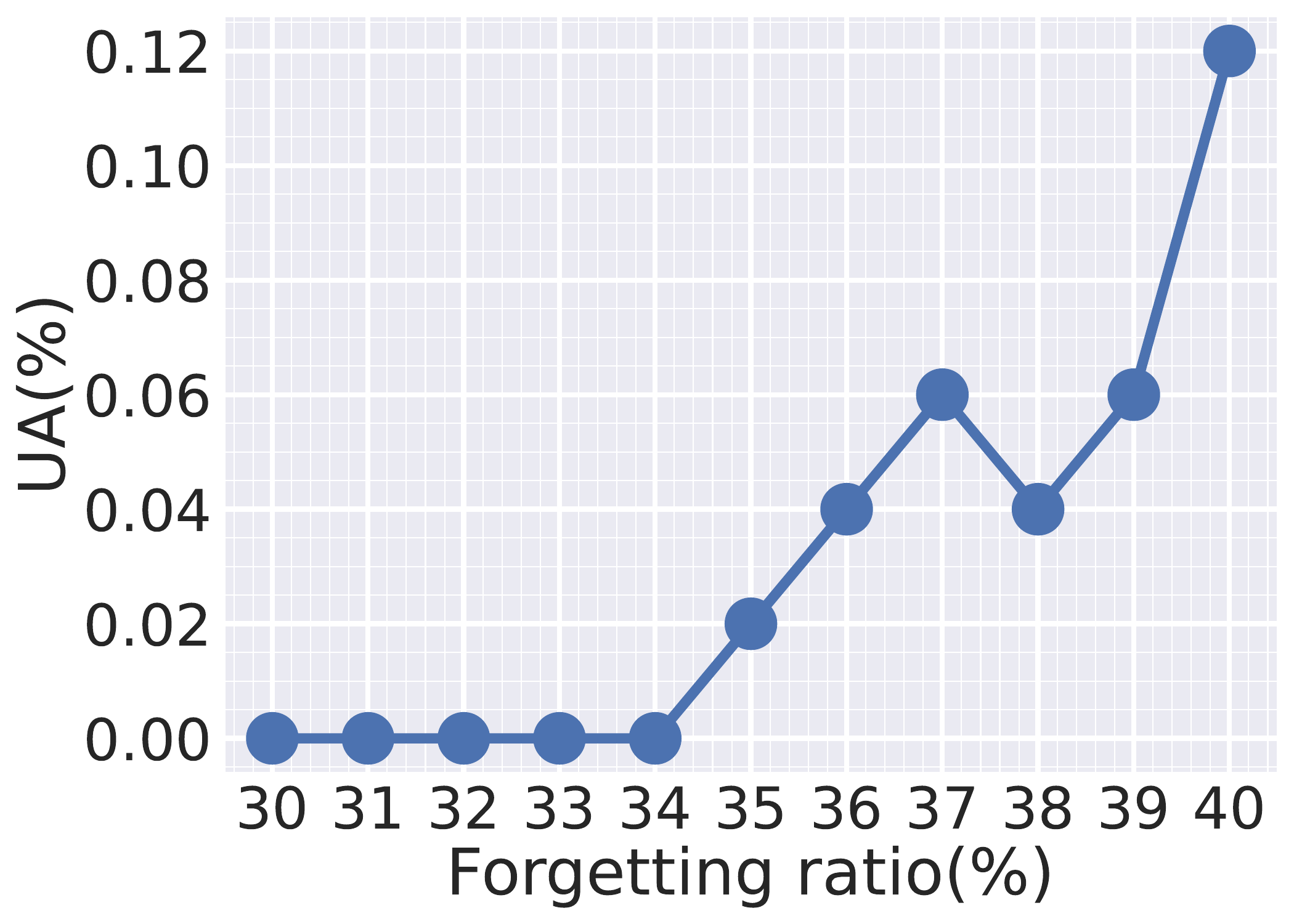} &
\includegraphics[width=0.230\textwidth]{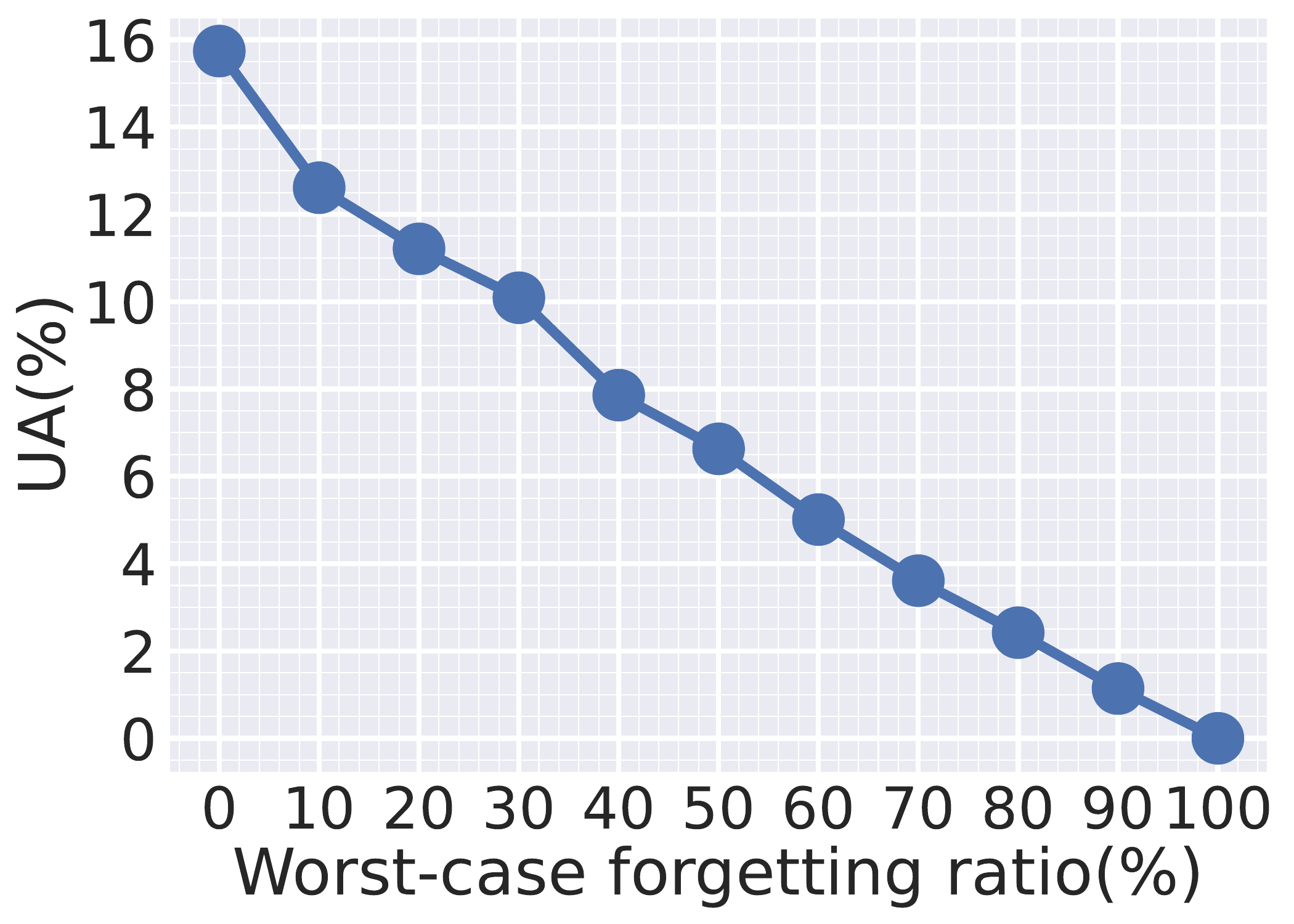} \\
\scriptsize{(a) Uniqueness} & \scriptsize{(b) Mixture}
\end{tabular}
\vspace{-2.5mm}
\caption{
\footnotesize{
UA of {\Retrain} on CIFAR-10 using ResNet-18. (a) UA under worst-case forgetting scenarios at different forgetting data ratios. (b) UA under a mixture of random and worst-case forgetting scenarios at different mixture ratios.
}
}
\vspace{-7mm}
\label{fig: uni_mix}
\end{wrapfigure}

To verify the uniqueness of {\ourset}, we identified {\ourset} for different forgetting data ratios and performed unlearning using {\Retrain}. We found that a \textbf{maximal} set with zero UA can exist. As shown in \textbf{Fig.\,\ref{fig: uni_mix}-(a)}, with appropriately defined set sizes (up to $34\%$ of the entire dataset), our method consistently identifies a worst-case forget set with $0$ UA. 

Furthermore, any subset of this set will also exhibit the worst-case property. \textbf{Fig.\,\ref{fig: uni_mix}-(b)} illustrates that including any part of the worst-case set complicates the unlearning process. When the forget set represents a $34\%$ ratio comprising a mix of {\ourset} and {\randomset} and unlearning is performed using {\Retrain}, the unlearning becomes increasingly difficult as the proportion of worst-case {\randomset} increases, which is indicated by the decrease in UA. This highlights the importance and significance of worst-case forgetting.

\subsection{Identifying {\ourset} in class-wise forgetting.}
\label{app: class_wise}
\begin{wraptable}{r}{0.50\textwidth}
\vspace{-12mm}
\caption{
Performance of various MU methods on ImageNet, ResNet-18. The content format follows Table\,\ref{tab: rl_lower_level}.}

\vspace*{-6mm}
\label{tab: main_imagenet}
\begin{center}
\resizebox{0.5\textwidth}{!}{
\renewcommand\tabcolsep{5pt}
\begin{tabular}{c|ccccc}
\toprule[1pt]
\midrule
\textbf{Methods} & \multicolumn{1}{c|}{\textbf{UA}}   & \multicolumn{1}{c|}{\textbf{MIA}}     & \multicolumn{1}{c|}{\textbf{RA}}    & \multicolumn{1}{c|}{\textbf{TA}} & \multicolumn{1}{c}{\textbf{Avg. Gap}} \\
\midrule
\rowcolor{Gray}
\multicolumn{6}{c}{\textbf{\RandomSet}} \\
\midrule
{{\Retrain}} & $72.92$ & $98.78$ & $65.90$ & $66.03$ & N/A \\ 
\midrule
\rowcolor{Gray}
\multicolumn{6}{c}{\textbf{\OurSet}} \\
\midrule
{{\Retrain}} & $45.92$ (\textcolor{Green}{$ 27.00 \blacktriangledown$}) & $98.56$ (\textcolor{Green}{$ 0.22 \blacktriangledown$})  & $66.64$ (\textcolor{Red}{$ 0.74 \blacktriangle$}) & $66.68$ (\textcolor{Red}{$ 0.65 \blacktriangle$}) & \textcolor{blue}{0.00}\\
\midrule
\rowcolor{Gray}
{{\FT}} & $36.40$ (\textcolor{blue}{9.52}) & $96.71$ (\textcolor{blue}{1.85}) & $65.40$ (\textcolor{blue}{1.24}) & $65.73$ (\textcolor{blue}{0.95}) & \textcolor{blue}{3.39} \\

{{\MUSparse}} & $38.50$ (\textcolor{blue}{7.42}) & $95.57$ (\textcolor{blue}{2.99}) & $64.16$ (\textcolor{blue}{2.48}) & $65.00$ (\textcolor{blue}{1.68}) & \textcolor{blue}{3.64} \\

% \midrule
% \rowcolor{Gray}
% \multicolumn{6}{c}{\textbf{Relabeling-based}} \\
\rowcolor{Gray}
{{\RL}} & $99.89$ (\textcolor{blue}{53.97}) & $99.01$ (\textcolor{blue}{0.45}) & $39.70$ (\textcolor{blue}{26.94}) & $44.04$ (\textcolor{blue}{22.64}) & \textcolor{blue}{26.00} \\
\midrule
\bottomrule[1pt]
\end{tabular}
}
\end{center}
\vspace*{-11mm}
\end{wraptable}

Extended from data-wise forgetting, 
\textbf{Table\,\ref{tab: main_imagenet}} showcases the effectiveness of our proposal in class-wise forgetting for image classification on the ImageNet dataset \cite{deng2009imagenet}.
Recall that the data selection variables are now interpreted as class selection variables. In this experiment, our objective is to eliminate the influence of 10\% of the ImageNet classes on classification performance.

To avoid completely eliminating the prediction head for the forgetting class in the model (ResNet-18), we define a class removal as the elimination of 90\% of its data points.
Consistent with our previous observations in class-wise forgetting, we can observe from Table\,\ref{tab: main_imagenet} that our identified {\ourset} constitutes a more challenging subset for the erasure of data influence as compared to {\randomset}, evidenced by a significant decline in UA of {\Retrain} {from $72.92\%$ to $45.92\%$}. A smaller reduction in MIA performance compared to data-wise forgetting suggests that class-wise forgetting presents a relatively simpler challenge.
In addition, by examining the performance of representative approximate unlearning methods ({\FT}, {\MUSparse}, and {\RL}), we observe that relabeling-free unlearning methods exhibit performances akin to {\Retrain} under the worst-case forget set,   whereas relabeling-based methods demonstrate substantial discrepancies in UA, consistent with our observations in Table\,\ref{tab: main_approximate_unlearn}.

\begin{figure}[htb]  
\centering
% \vspace{-5mm}
\includegraphics[width=\textwidth]{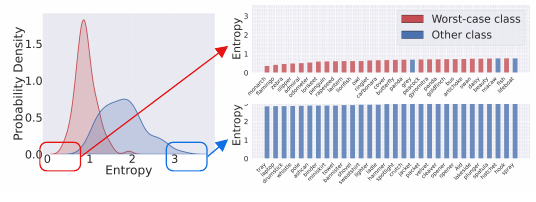}
\vspace{-10mm}
\caption{{
Average entropy of worst-case forget classes vs. that of other classes on ImageNet using ResNet-18. The number of worst-case forget classes is 100.}}
\label{fig: entropy_imagenet_full}
\end{figure}

\begin{figure}[htb]  
\centering 
% \vspace{-5mm}
\includegraphics[width=\textwidth]{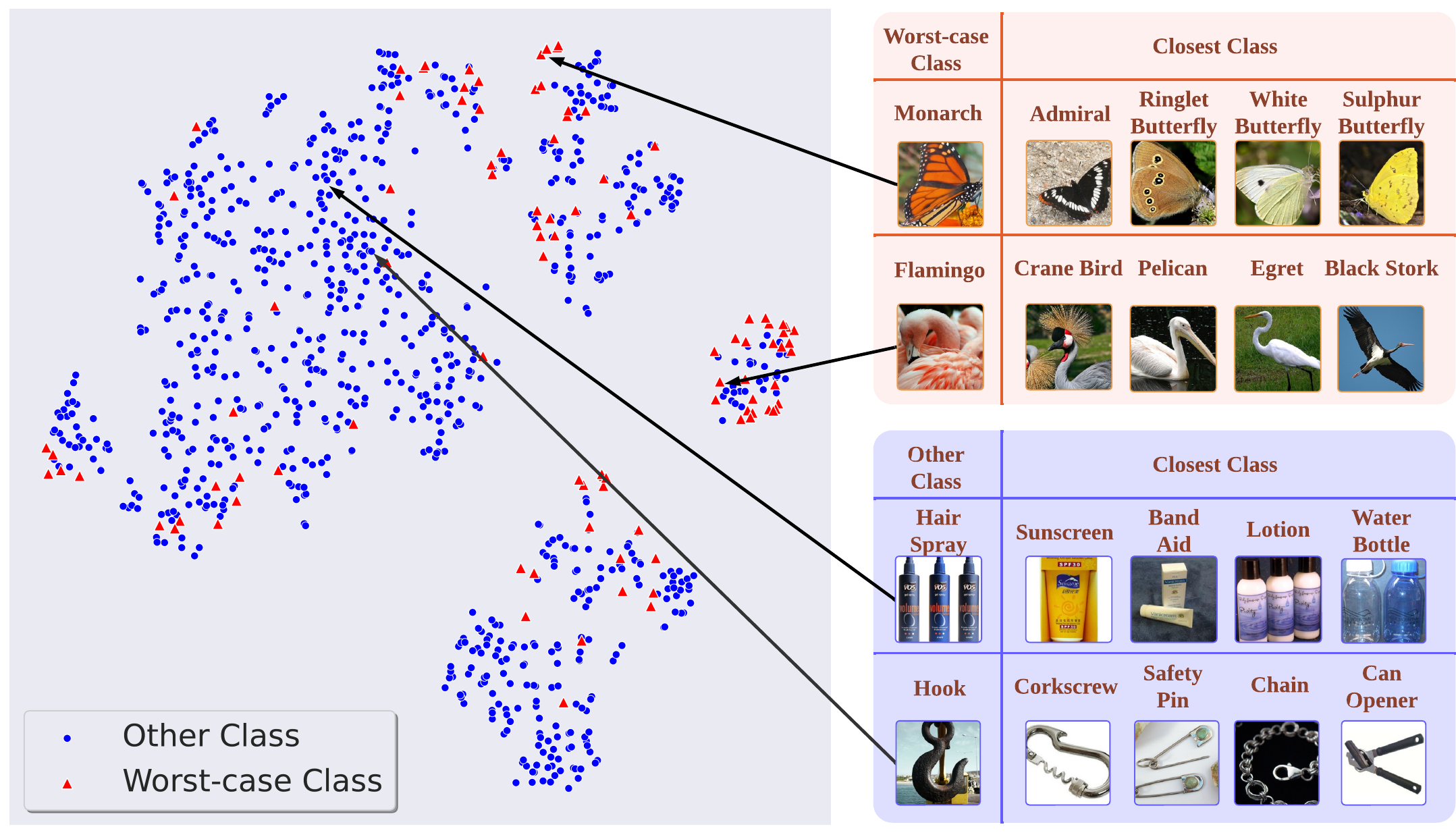}
\vspace{-5mm}

\caption{
{T-SNE for all classes in the learned feature space, with an additional display on the right side showcasing two worst-case classes and two others, along with their four closest classes.}
}

% \caption{
% \CF{T-SNE for all classes \SL{based on their learned features?}, with an additional display on the right side showcasing two worst-case classes and two others, along with their four closest classes. \JC{[Change the images]}}
% }
\label{fig: tsne_imagenet}
\end{figure}

Moreover, \textbf{Fig.\,\ref{fig: entropy_imagenet_full}} portrays the class-wise entropy for ImageNet classes with-in {\ourset} in comparison to other classes. This visualization elucidates a predilection for selecting low-entropy classes as the worst-case scenarios for unlearning, suggesting that these classes are ostensibly simpler to learn. Furthermore, the worst-case forget class is primarily composed of animals and insects. In \textbf{Fig.\,\ref{fig: tsne_imagenet}}, we use t-SNE to show the relationship between worst-case classes and other classes. As we can see, the worst-case class primarily resides on the periphery of the distribution.
% \SL{[I did not fully understand the following sentences.]} Further elaboration is provided through visual representations of the classes encompassed by {\ourset} and those external to it, in addition to the closest classes within ImageNet's feature space. Highlightably,  the classes within the {\ourset} pronounced differences characterizing to the nearest classes, in contrast, other classes appear virtually similar to the nearest classes. 

\subsection{{Additional results of Fig.\,\ref{fig: generation_examples}}
}
\label{app: more_visualization}
% \JC{[Show all the selected prompts][Failed prompts in worst-case can also be included to illustrate we ``successful'' unlearned][show random selected for comparison (?)]}
In \textbf{Tables\,\ref{fig: generation_examples_appendix_pre}}-\textbf{\ref{fig: generation_examples_appendix_more}}, we present more examples using the original stable diffusion model (w/o unlearning), the unlearned diffusion model over the worst-case forgetting prompt set (Worst). For each diffusion model, images are generated based on an unlearned prompt from the worst-case forget set. It is evident that the unlearned diffusion model is still capable of generating corresponding images for prompts from the worst-case forget set.

\section{Broader Impacts and Limitations}
\label{app: experiment_results: broader_limitation}
{\OurSet} represents a novel perspective in evaluating data privacy and security. This set strikes a balance between data influence erasure and model utility, offering a robust assessment of the effectiveness of existing unlearning methods from an adversarial standpoint. It also provides a deeper understanding of datasets from the perspective of machine unlearning.

However, it is crucial to acknowledge the limitations of {\ourset}. While {\ourset} has demonstrated its effectiveness in various scenarios, including data-wise, class-wise, and prompt-wise, the effectiveness of unlearning methods for language models on {\ourset} remains an area worthy of further exploration. 

\begin{table}[t]
    \centering
    % \vspace{-2mm}
    \caption{{Examples of image generation using the original stable diffusion model (w/o unlearning), the unlearned diffusion model over the worst-case forgetting prompt set (Worst). For each diffusion model, images are generated based on an unlearned prompt from the worst-case forget set.}}
    \label{fig: generation_examples_appendix_pre}
    \begin{center}
    \resizebox{1.0\textwidth}{!}{
    \renewcommand\tabcolsep{5pt}
    \begin{tabular}{c|c}
        \toprule[1pt]
        \midrule
 
        {\scriptsize{\textbf{Model}}} & \multicolumn{1}{c}{\scriptsize{\textbf{Generation Condition}}} \\ 
        \midrule
        \rowcolor{Gray}
        \multicolumn{2}{c}{~~~~~~~~~~~~\scriptsize{\Puw\texttt{:}
\scriptsize{\texttt{\textit{A painting of }\textcolor{Red}{Dogs}} \texttt{\textit{in} \textcolor{Blue}{Van Gogh}\textit{ Style.}}}
        }}\\
        
        \midrule
        \begin{tabular}[c]{@{}c@{}}
        \vspace{-1.5mm}\tiny{\textbf{Original}} \\
        \vspace{-1.5mm}\tiny{\textbf{Diffusion}} \\
        \vspace{1mm}\tiny{\textbf{Model}}
        \end{tabular} & \multicolumn{1}{m{0.9\textwidth}}{
        \includegraphics[width=0.084\textwidth]{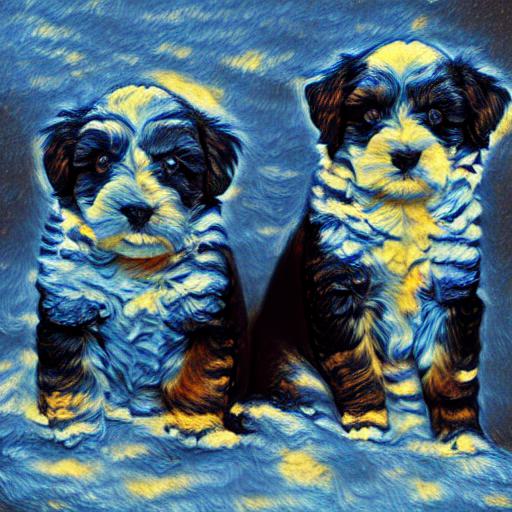} 
        \includegraphics[width=0.084\textwidth]{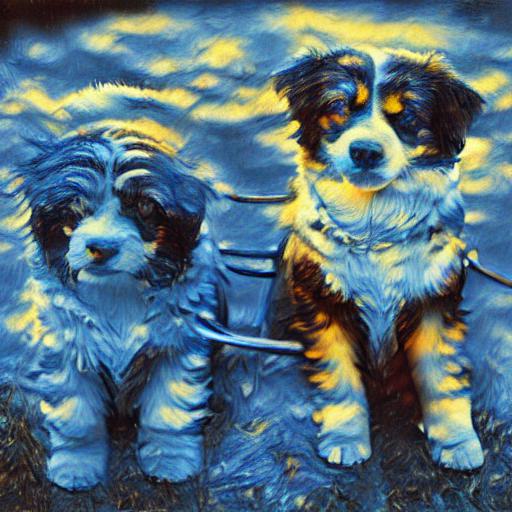} 
        \includegraphics[width=0.084\textwidth]{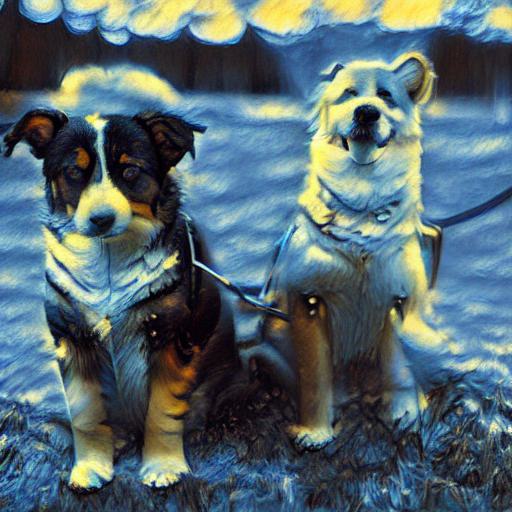} 
        \includegraphics[width=0.084\textwidth]{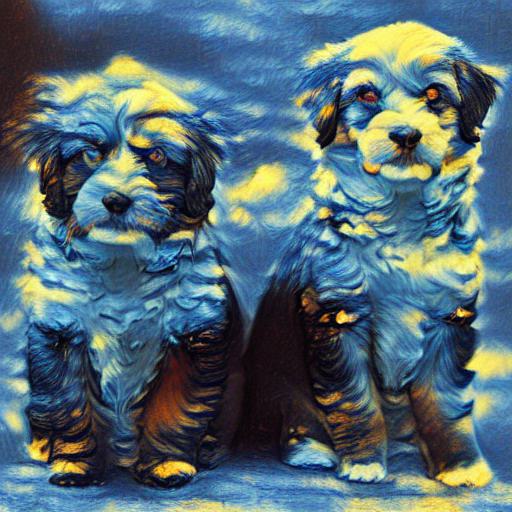} 
        \includegraphics[width=0.084\textwidth]{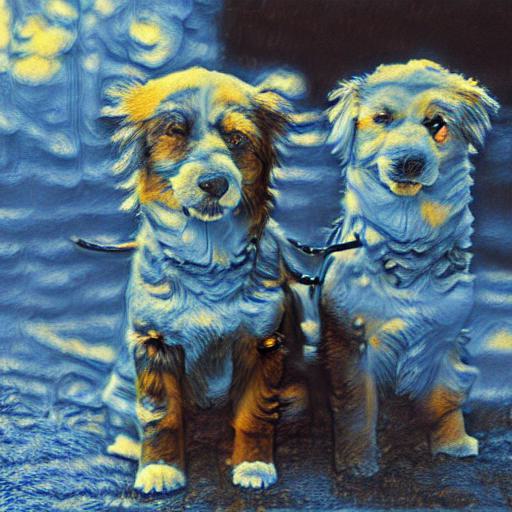} 
        \includegraphics[width=0.084\textwidth]{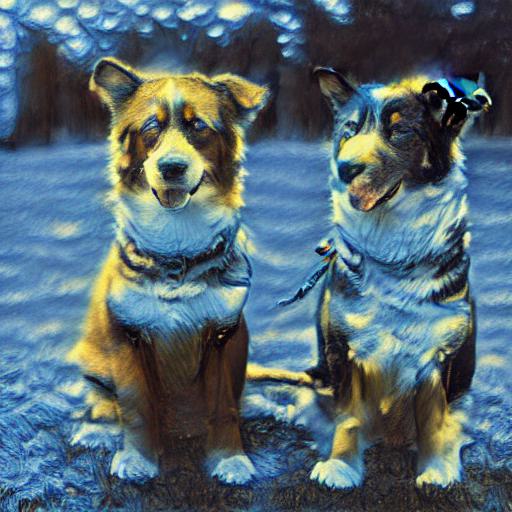} 
        \includegraphics[width=0.084\textwidth]{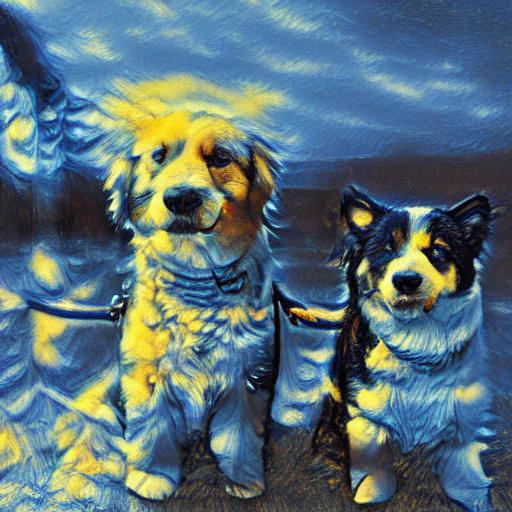} 
        \includegraphics[width=0.084\textwidth]{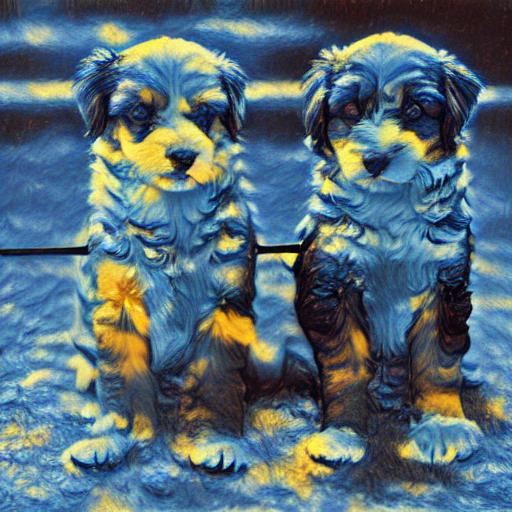} 
        \includegraphics[width=0.084\textwidth]{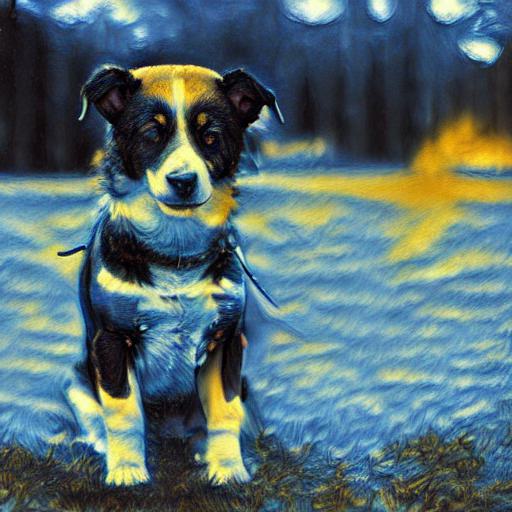} 
        \includegraphics[width=0.084\textwidth]{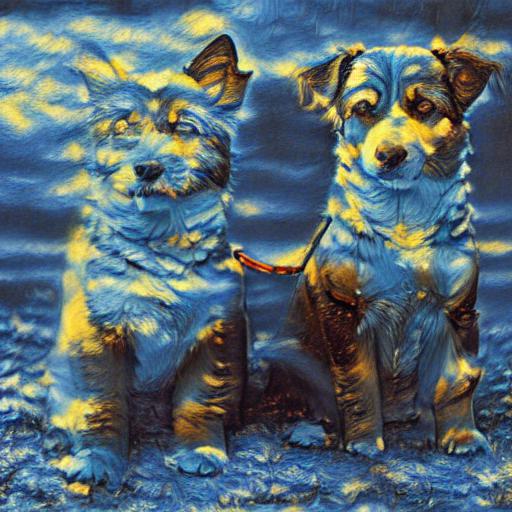}} \\ 
        \midrule 
         \begin{tabular}[c]{@{}c@{}}
        \vspace{-1.5mm}\tiny{\textbf{Unlearned}} \\
        \vspace{-1.5mm}\tiny{\textbf{Diffusion}} \\
        \vspace{-1.5mm}\tiny{\textbf{Model}} \\
        \vspace{1mm}\tiny{\textbf{(Worst)}}
        \end{tabular} & \multicolumn{1}{m{0.9\textwidth}}{
        \includegraphics[width=0.084\textwidth]{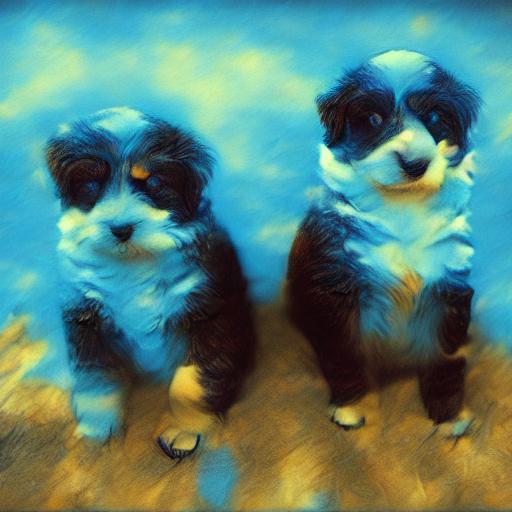} 
        \includegraphics[width=0.084\textwidth]{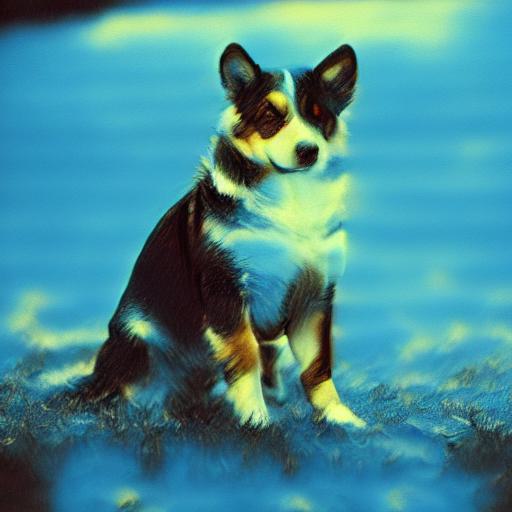} 
        \includegraphics[width=0.084\textwidth]{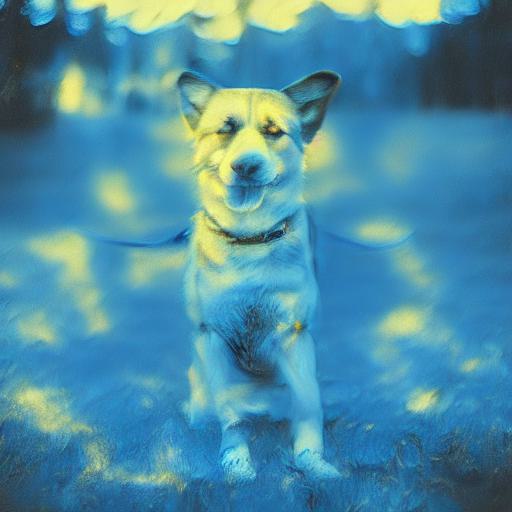} 
        \includegraphics[width=0.084\textwidth]{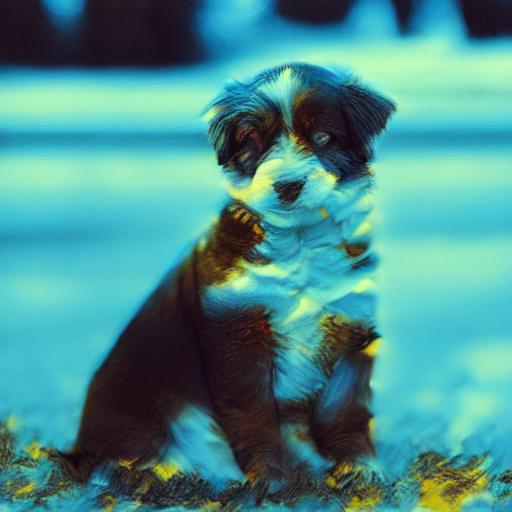} 
        \includegraphics[width=0.084\textwidth]{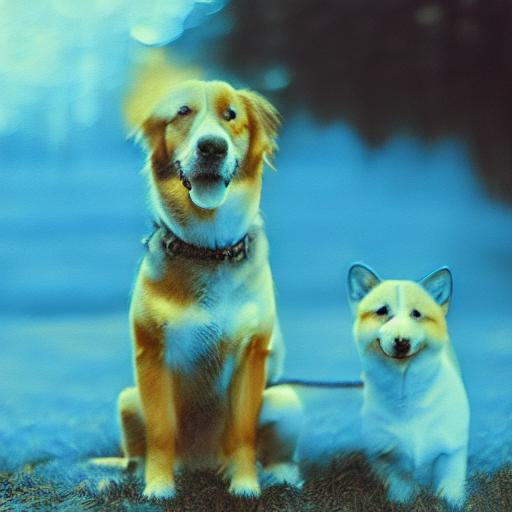} 
        \includegraphics[width=0.084\textwidth]{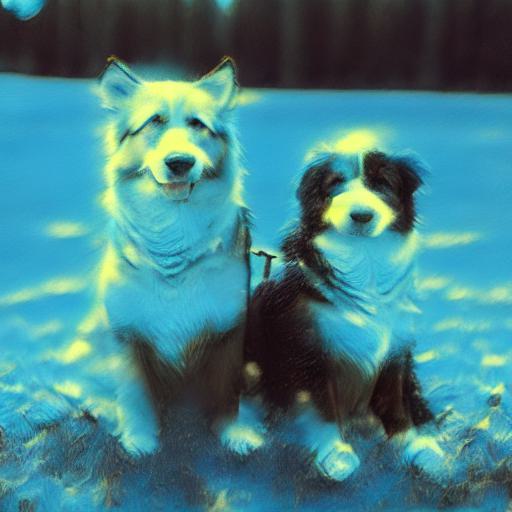} 
        \includegraphics[width=0.084\textwidth]{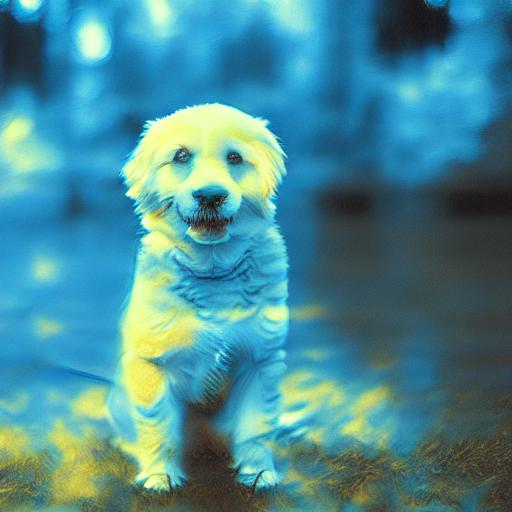} 
        \includegraphics[width=0.084\textwidth]{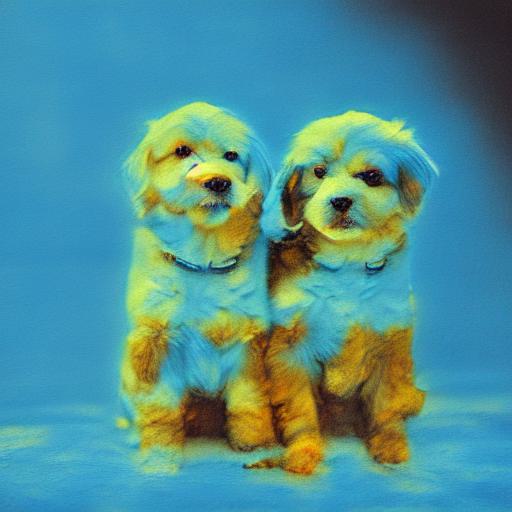} 
        \includegraphics[width=0.084\textwidth]{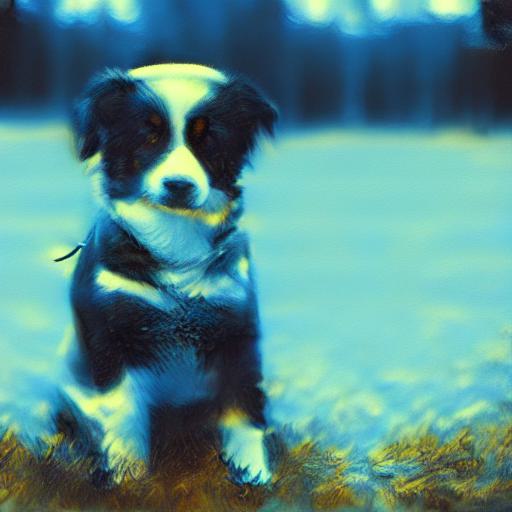} 
        \includegraphics[width=0.084\textwidth]{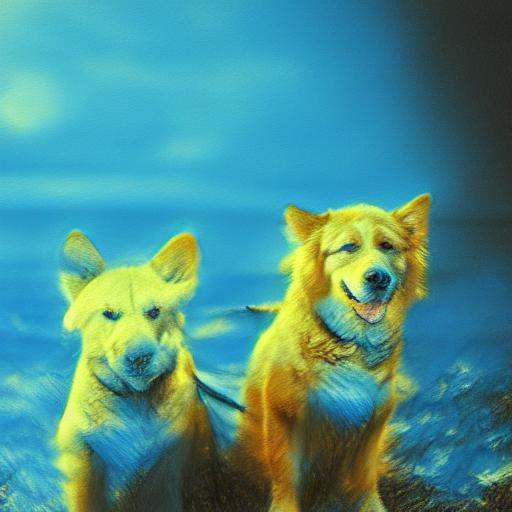}} \\ 
        \midrule
        \rowcolor{Gray}
        \multicolumn{2}{c}{~~~~~~~~~~~~\scriptsize{\Puw\texttt{:}
\scriptsize{\texttt{\textit{A painting of }\textcolor{Red}{Dogs}} \texttt{\textit{in} \textcolor{Blue}{Rust}\textit{ Style.}}}
}}\\
\midrule
        \begin{tabular}[c]{@{}c@{}}
        \vspace{-1.5mm}\tiny{\textbf{Original}} \\
        \vspace{-1.5mm}\tiny{\textbf{Diffusion}} \\
        \vspace{1mm}\tiny{\textbf{Model}}
        \end{tabular} & \multicolumn{1}{m{0.9\textwidth}}{
 \includegraphics[width=0.084\textwidth]{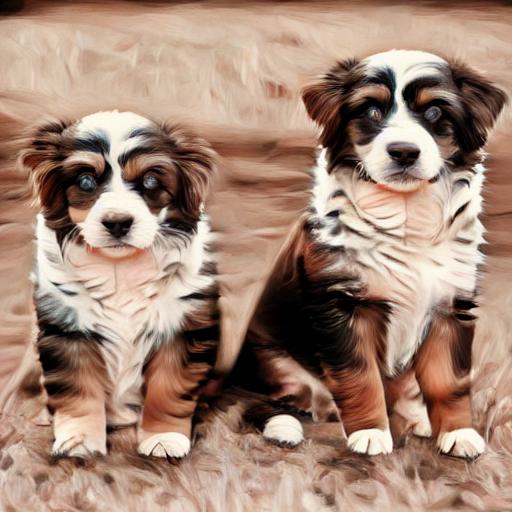} 
 \includegraphics[width=0.084\textwidth]{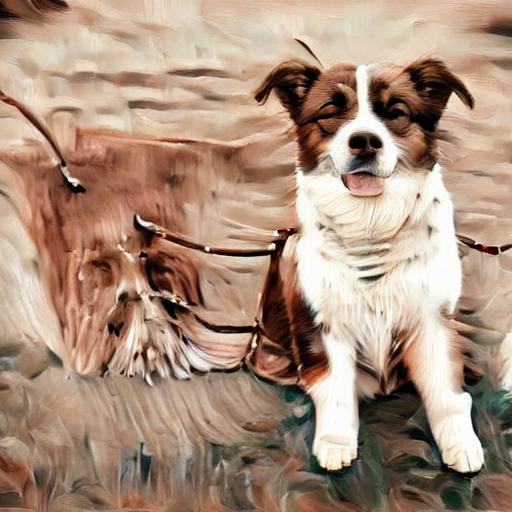} 
 \includegraphics[width=0.084\textwidth]{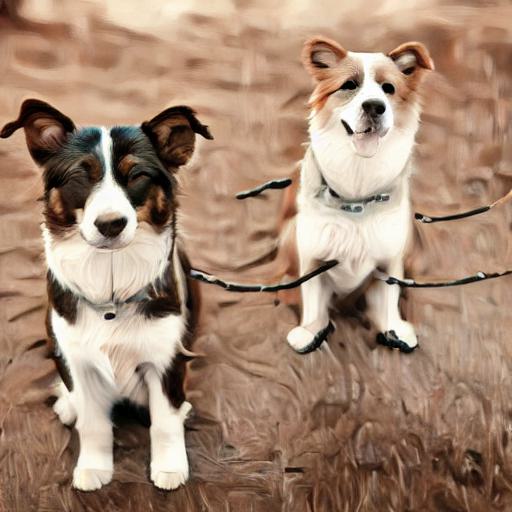} 
 \includegraphics[width=0.084\textwidth]{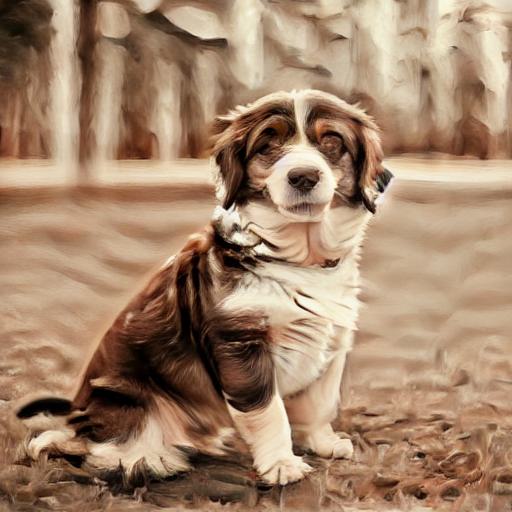} 
 \includegraphics[width=0.084\textwidth]{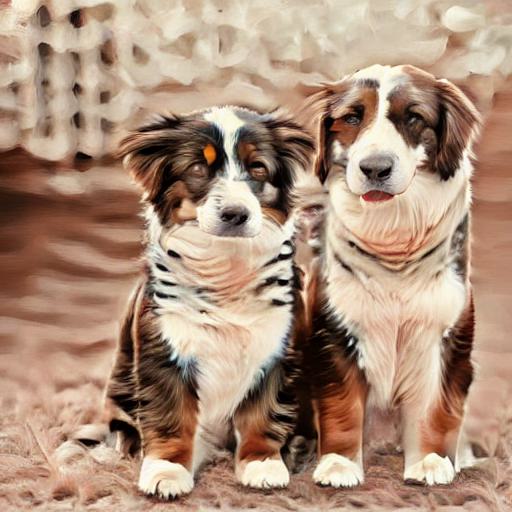} 
 \includegraphics[width=0.084\textwidth]{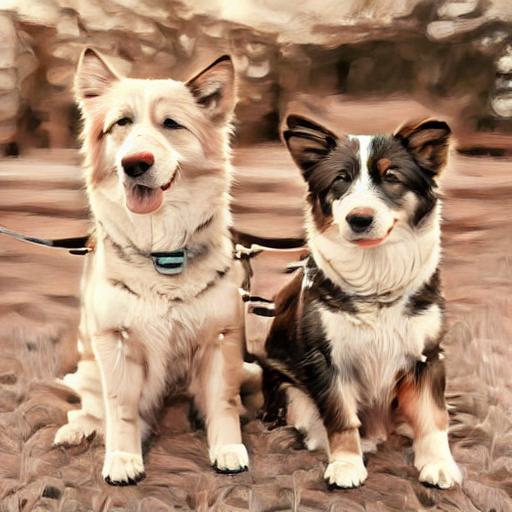} 
 \includegraphics[width=0.084\textwidth]{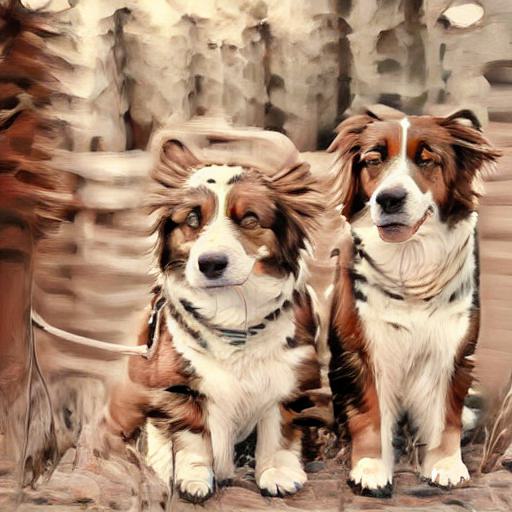} 
 \includegraphics[width=0.084\textwidth]{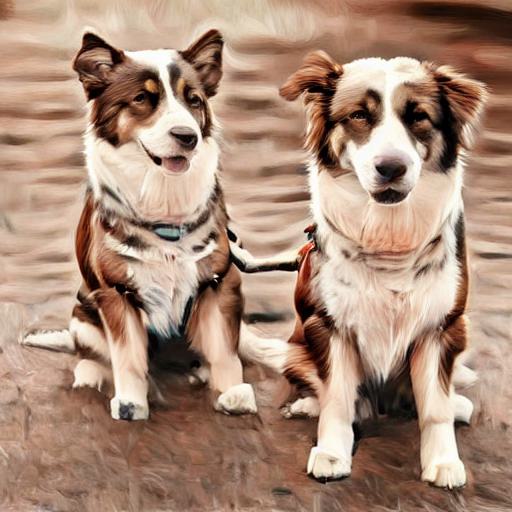} 
 \includegraphics[width=0.084\textwidth]{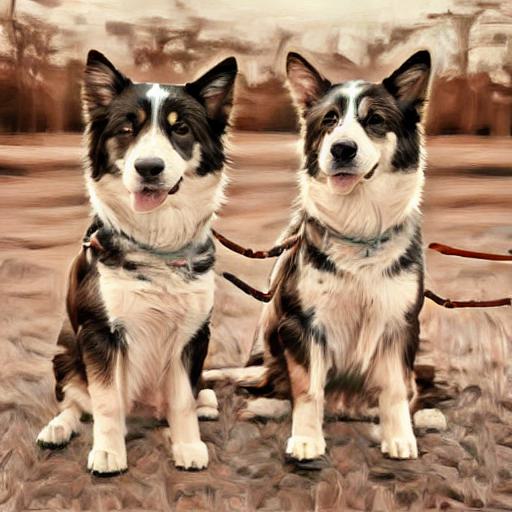} 
 \includegraphics[width=0.084\textwidth]{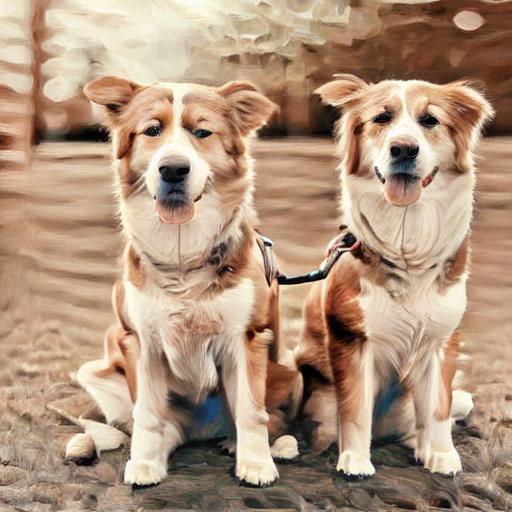}} \\ 
\midrule 
         \begin{tabular}[c]{@{}c@{}}
        \vspace{-1.5mm}\tiny{\textbf{Unlearned}} \\
        \vspace{-1.5mm}\tiny{\textbf{Diffusion}} \\
        \vspace{-1.5mm}\tiny{\textbf{Model}} \\
        \vspace{1mm}\tiny{\textbf{(Worst)}}
        \end{tabular} & \multicolumn{1}{m{0.9\textwidth}}{
 \includegraphics[width=0.084\textwidth]{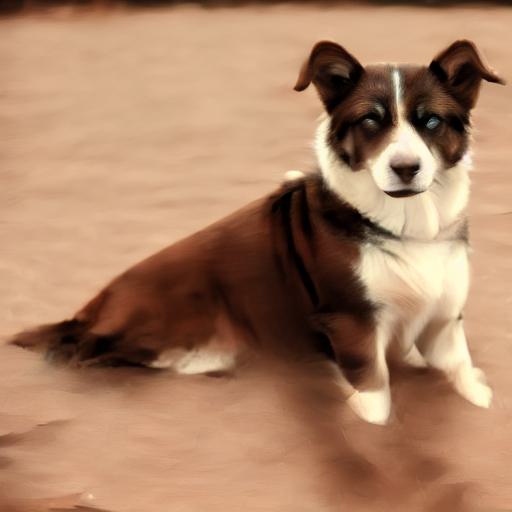} 
 \includegraphics[width=0.084\textwidth]{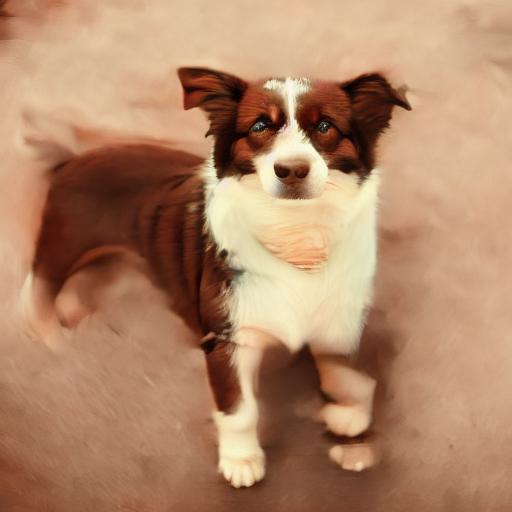} 
 \includegraphics[width=0.084\textwidth]{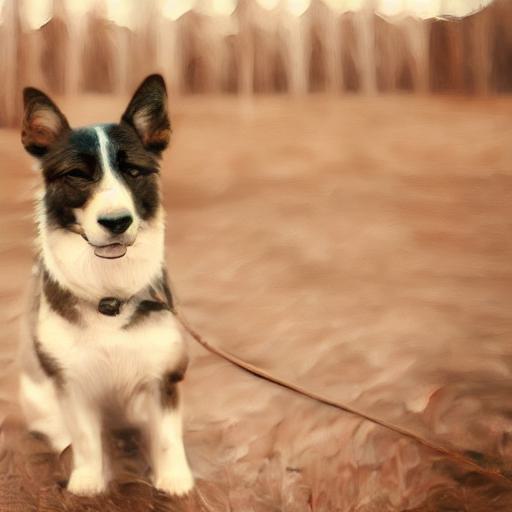} 
 \includegraphics[width=0.084\textwidth]{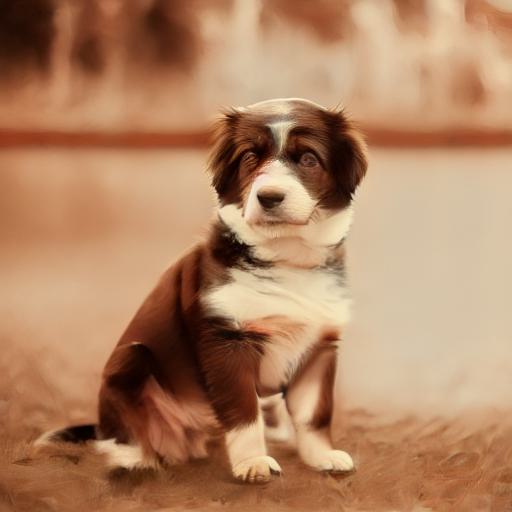} 
 \includegraphics[width=0.084\textwidth]{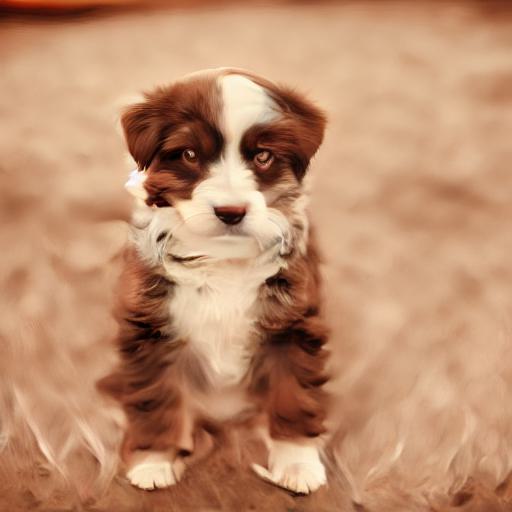} 
 \includegraphics[width=0.084\textwidth]{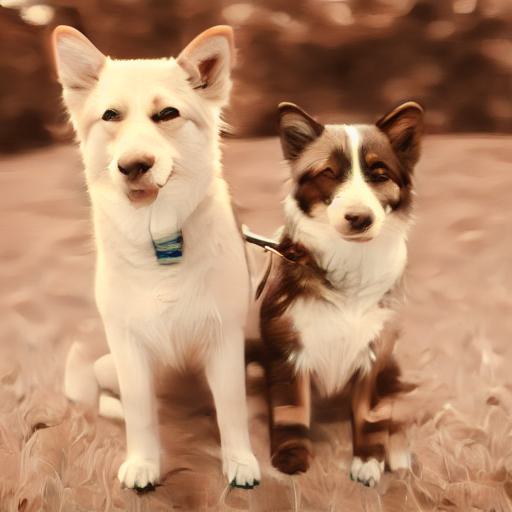} 
 \includegraphics[width=0.084\textwidth]{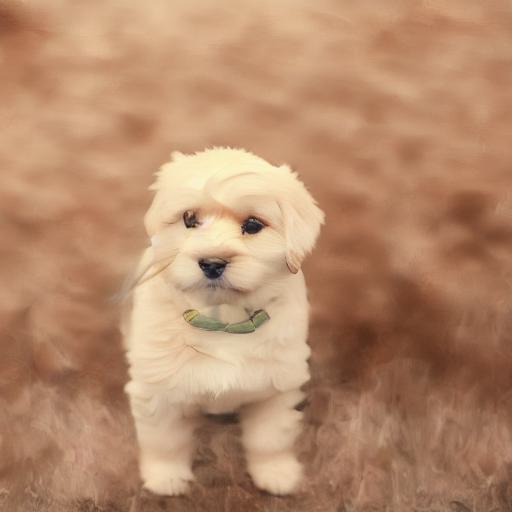} 
 \includegraphics[width=0.084\textwidth]{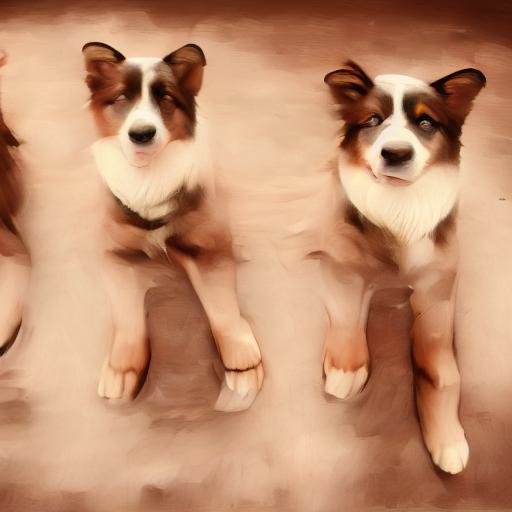} 
 \includegraphics[width=0.084\textwidth]{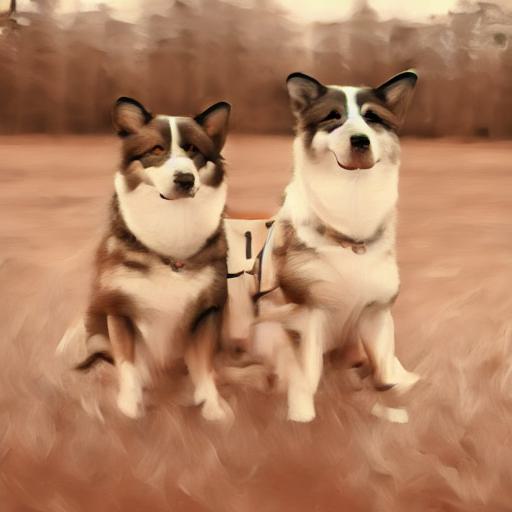} 
 \includegraphics[width=0.084\textwidth]{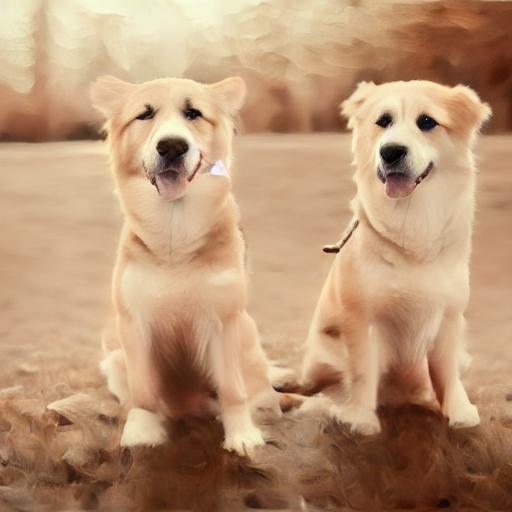}} \\ 
\midrule

        \bottomrule[1pt]
    \end{tabular}
    }
    \end{center}
\end{table}

\begin{table}[t]
    \centering
    % \vspace{-2mm}
    \caption{{Examples of image generation using the original stable diffusion model (w/o unlearning), the unlearned diffusion model over the worst-case forgetting prompt set (Worst). For each diffusion model, images are generated based on an unlearned prompt from the worst-case forget set.}}
    % \vspace{-5mm}
    \label{fig: generation_examples_appendix}
    \begin{center}
    \resizebox{1.0\textwidth}{!}{
    \renewcommand\tabcolsep{5pt}
    \begin{tabular}{c|c}
        \toprule[1pt]
        \midrule
 
        {\scriptsize{\textbf{Model}}} & \multicolumn{1}{c}{\scriptsize{\textbf{Generation Condition}}} \\ 
        \midrule
        \rowcolor{Gray}
        \multicolumn{2}{c}{~~~~~~~~~~~~\scriptsize{\Puw\texttt{:}
\scriptsize{\texttt{\textit{A painting of }\textcolor{Red}{Waterfalls}} \texttt{\textit{in} \textcolor{Blue}{Rust}\textit{ Style.}}}
}}\\
\midrule
        \begin{tabular}[c]{@{}c@{}}
        \vspace{-1.5mm}\tiny{\textbf{Original}} \\
        \vspace{-1.5mm}\tiny{\textbf{Diffusion}} \\
        \vspace{1mm}\tiny{\textbf{Model}}
        \end{tabular} & \multicolumn{1}{m{0.9\textwidth}}{
 \includegraphics[width=0.084\textwidth]{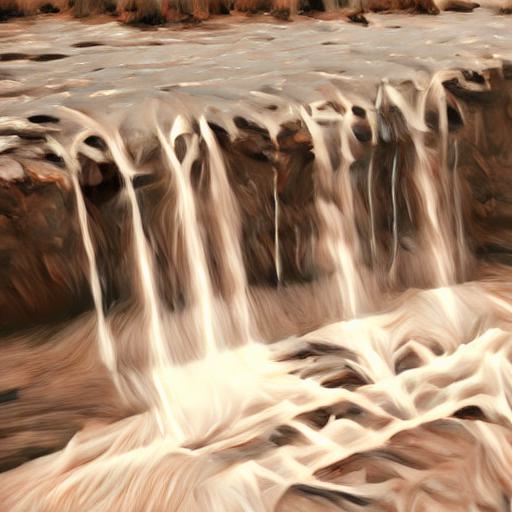} 
 \includegraphics[width=0.084\textwidth]{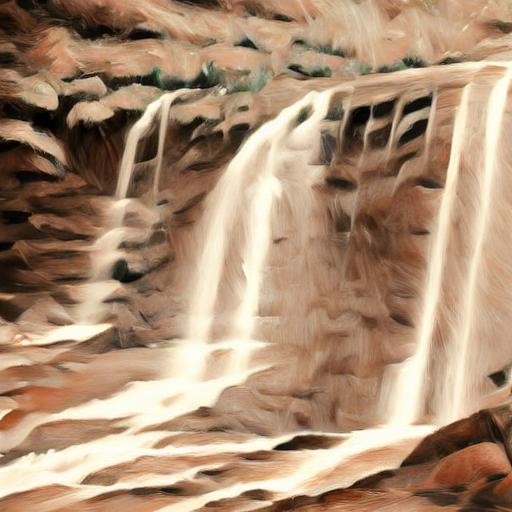} 
 \includegraphics[width=0.084\textwidth]{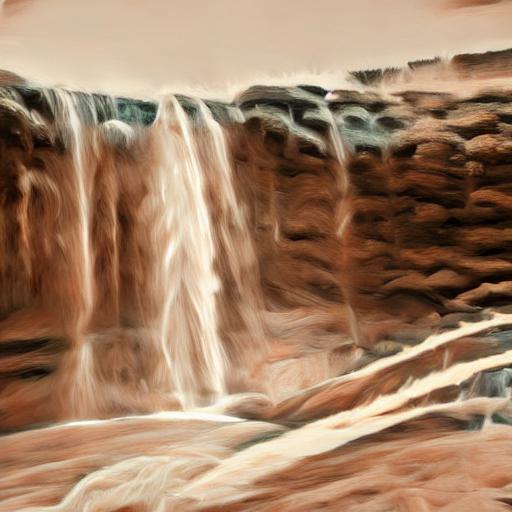} 
 \includegraphics[width=0.084\textwidth]{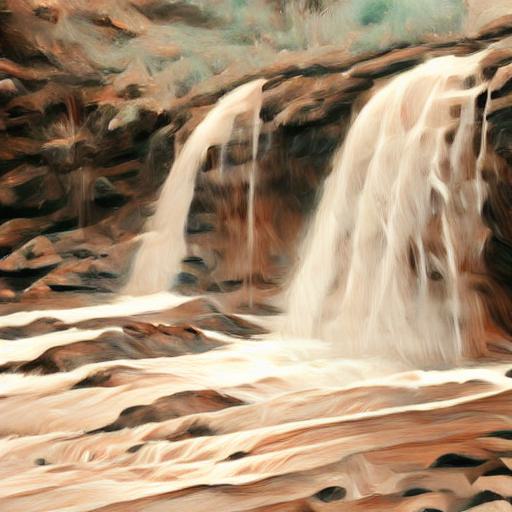} 
 \includegraphics[width=0.084\textwidth]{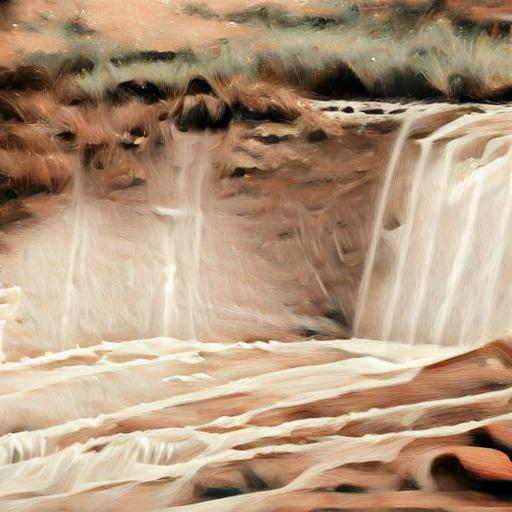} 
 \includegraphics[width=0.084\textwidth]{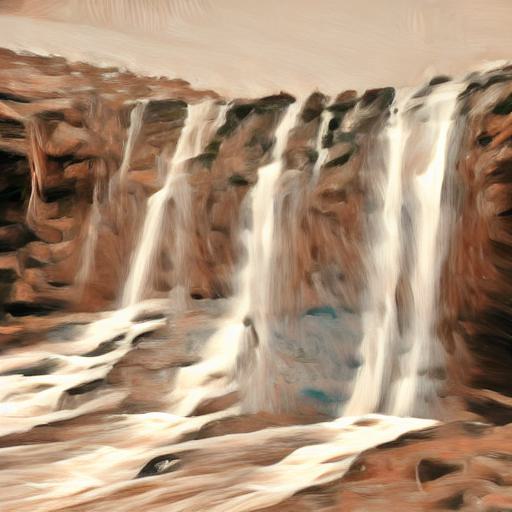} 
 \includegraphics[width=0.084\textwidth]{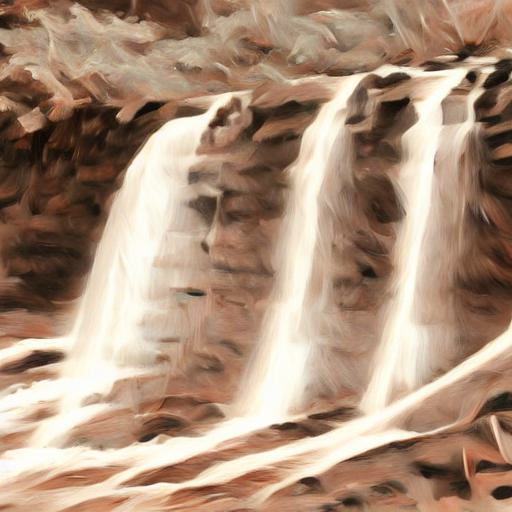} 
 \includegraphics[width=0.084\textwidth]{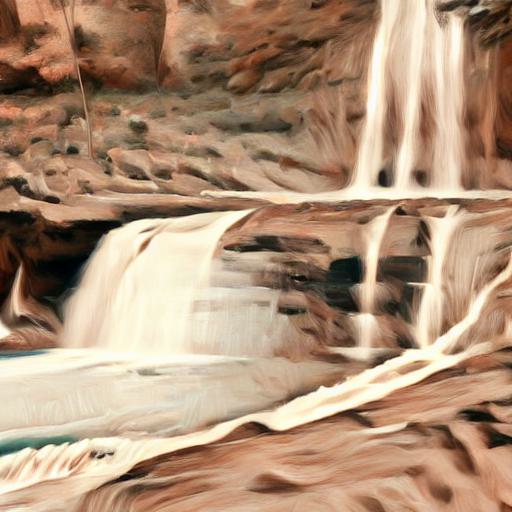} 
 \includegraphics[width=0.084\textwidth]{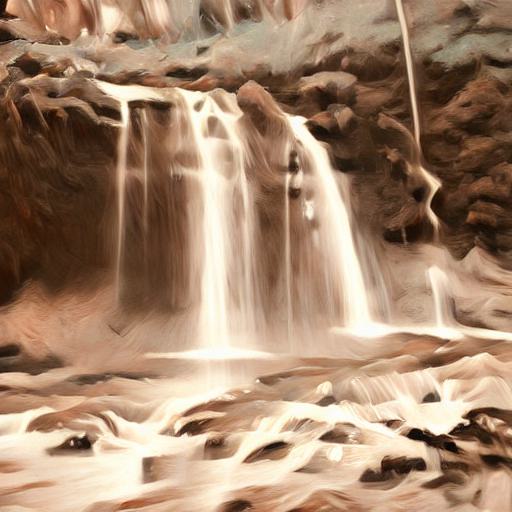} 
 \includegraphics[width=0.084\textwidth]{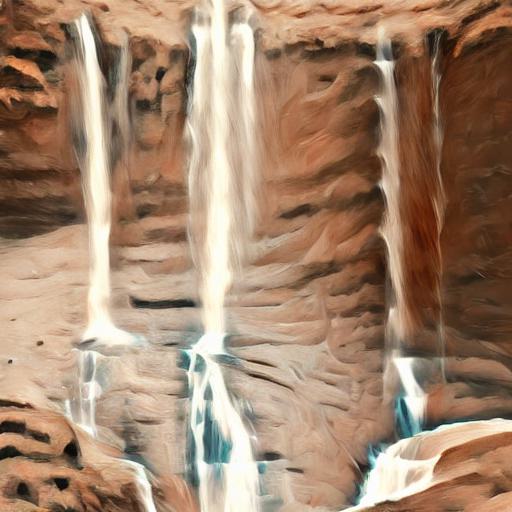}} \\ 
\midrule 
         \begin{tabular}[c]{@{}c@{}}
        \vspace{-1.5mm}\tiny{\textbf{Unlearned}} \\
        \vspace{-1.5mm}\tiny{\textbf{Diffusion}} \\
        \vspace{-1.5mm}\tiny{\textbf{Model}} \\
        \vspace{1mm}\tiny{\textbf{(Worst)}}
        \end{tabular} & \multicolumn{1}{m{0.9\textwidth}}{
 \includegraphics[width=0.084\textwidth]{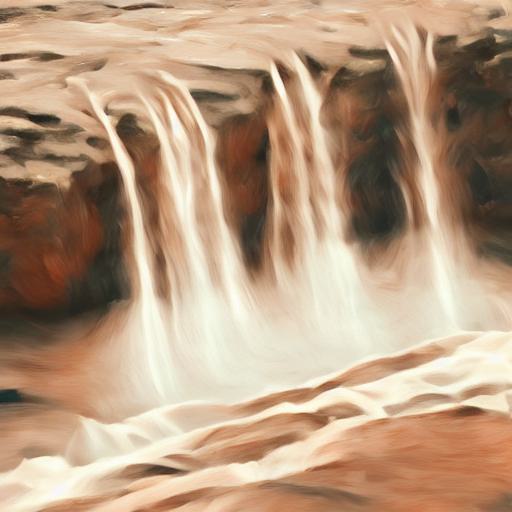} 
 \includegraphics[width=0.084\textwidth]{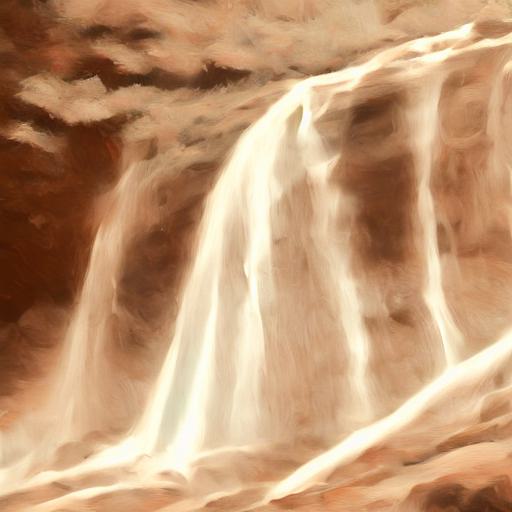} 
 \includegraphics[width=0.084\textwidth]{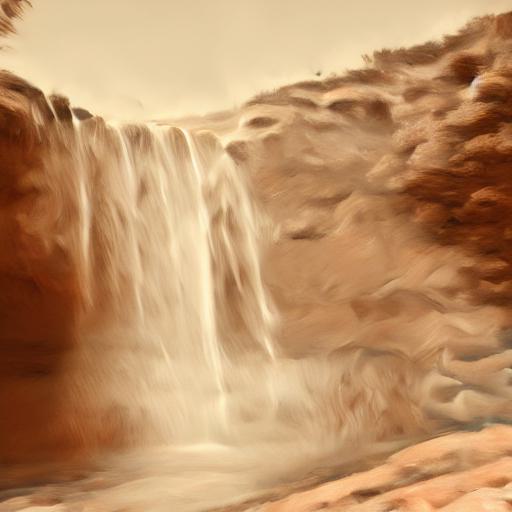} 
 \includegraphics[width=0.084\textwidth]{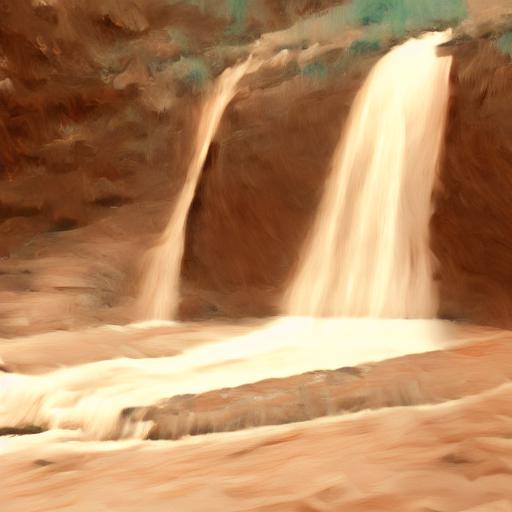} 
 \includegraphics[width=0.084\textwidth]{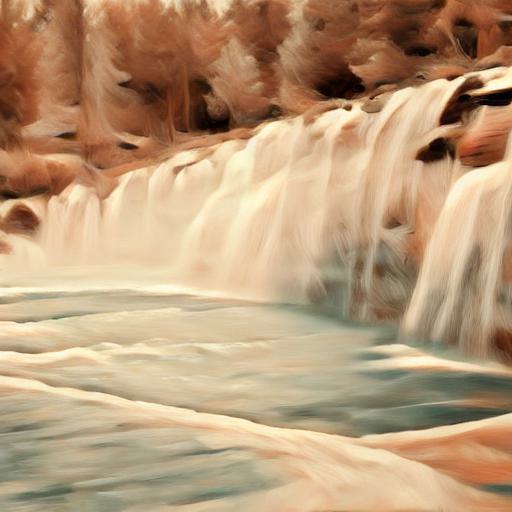} 
 \includegraphics[width=0.084\textwidth]{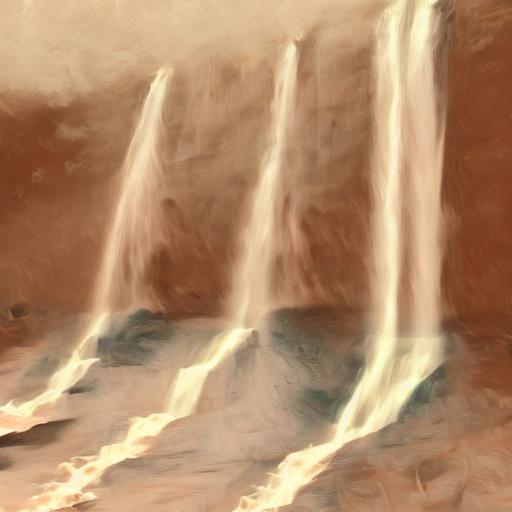} 
 \includegraphics[width=0.084\textwidth]{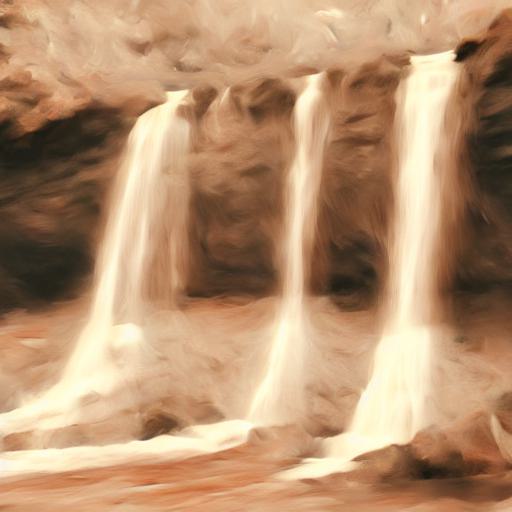} 
 \includegraphics[width=0.084\textwidth]{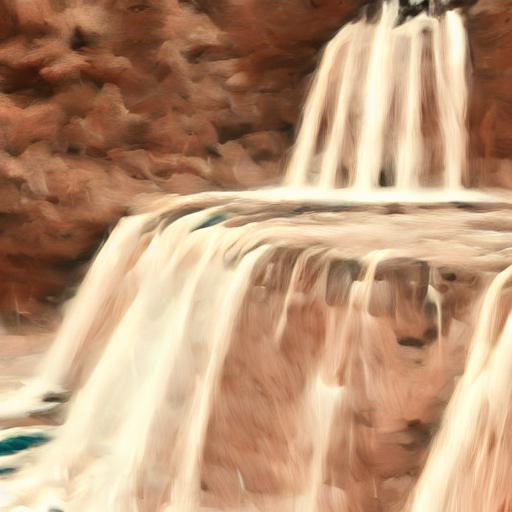} 
 \includegraphics[width=0.084\textwidth]{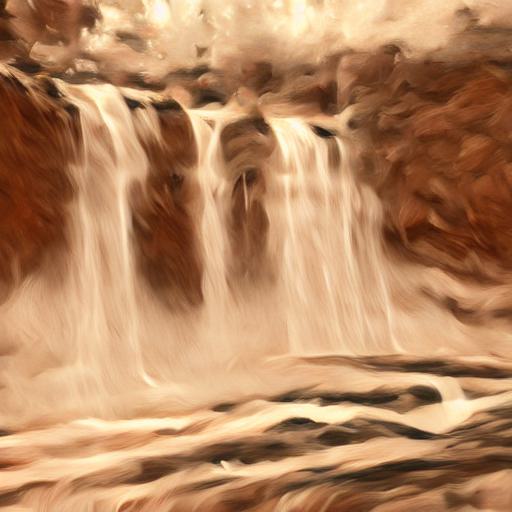} 
 \includegraphics[width=0.084\textwidth]{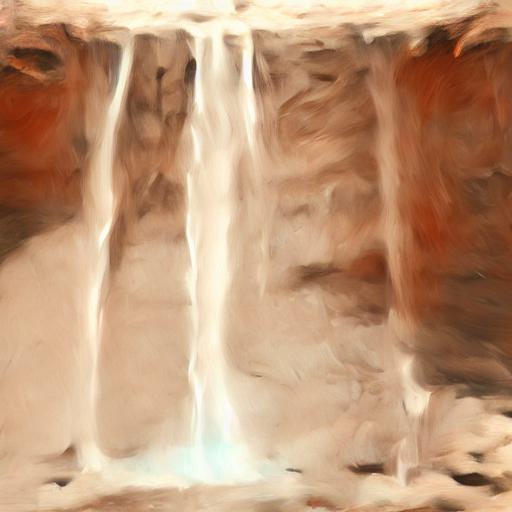}} \\ 
\midrule 
        \rowcolor{Gray}
        \multicolumn{2}{c}{~~~~~~~~~~~~\scriptsize{\Puw\texttt{:}
\scriptsize{\texttt{\textit{A painting of }\textcolor{Red}{Horses}} \texttt{\textit{in} \textcolor{Blue}{Winter}\textit{ Style.}}}
}}\\
\midrule
        \begin{tabular}[c]{@{}c@{}}
        \vspace{-1.5mm}\tiny{\textbf{Original}} \\
        \vspace{-1.5mm}\tiny{\textbf{Diffusion}} \\
        \vspace{1mm}\tiny{\textbf{Model}}
        \end{tabular} & \multicolumn{1}{m{0.9\textwidth}}{
 \includegraphics[width=0.084\textwidth]{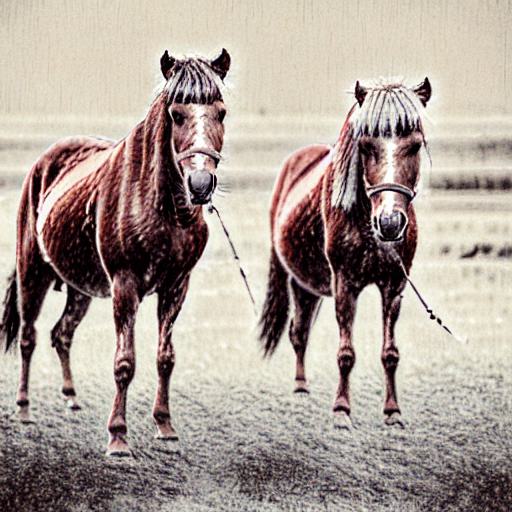} 
 \includegraphics[width=0.084\textwidth]{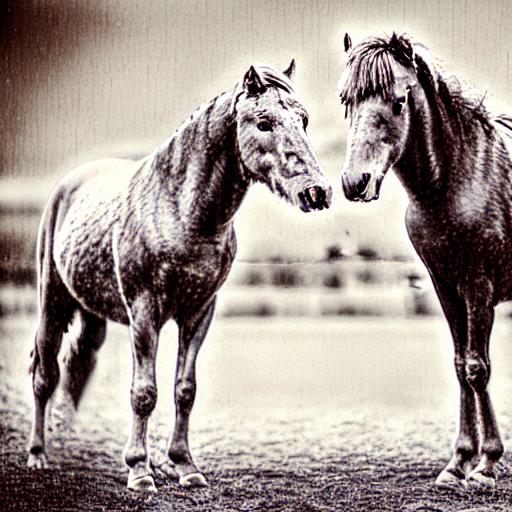} 
 \includegraphics[width=0.084\textwidth]{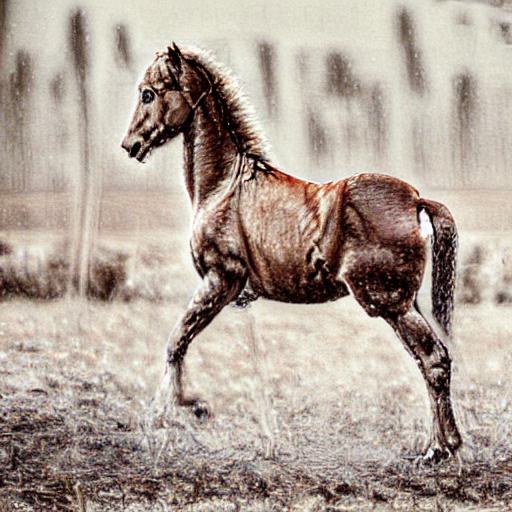} 
 \includegraphics[width=0.084\textwidth]{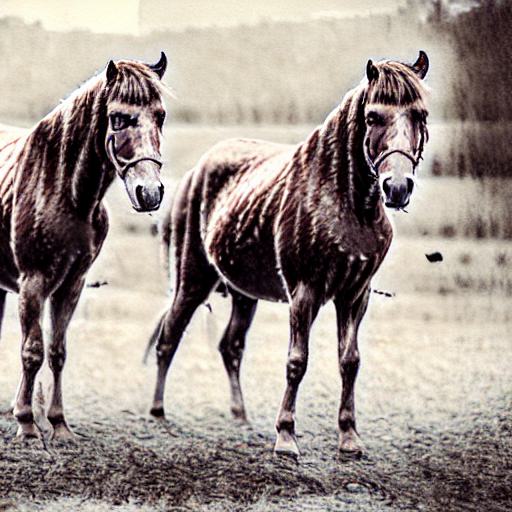} 
 \includegraphics[width=0.084\textwidth]{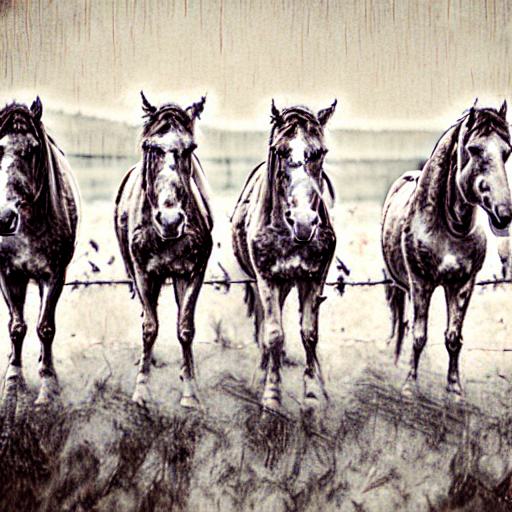} 
 \includegraphics[width=0.084\textwidth]{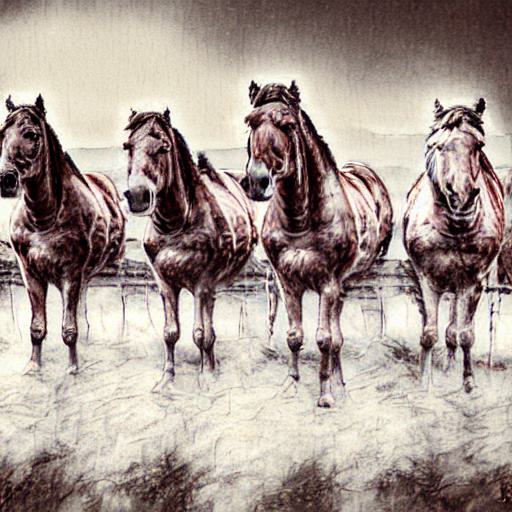} 
 \includegraphics[width=0.084\textwidth]{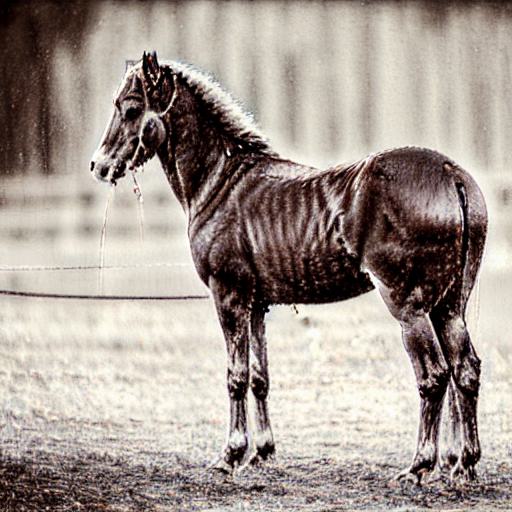} 
 \includegraphics[width=0.084\textwidth]{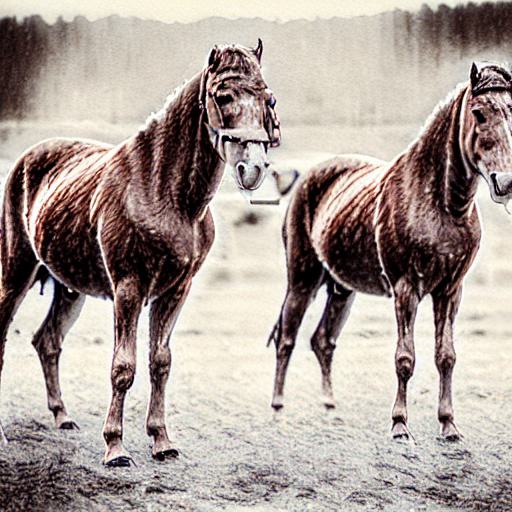} 
 \includegraphics[width=0.084\textwidth]{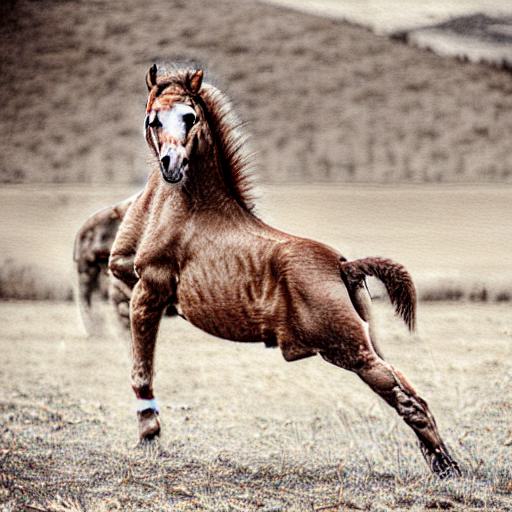} 
 \includegraphics[width=0.084\textwidth]{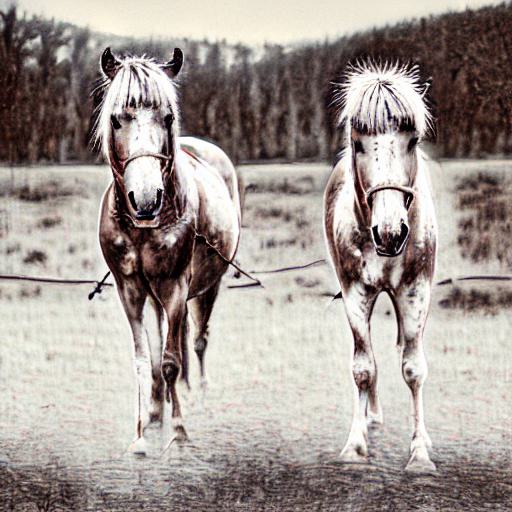}} \\ 
\midrule 
         \begin{tabular}[c]{@{}c@{}}
        \vspace{-1.5mm}\tiny{\textbf{Unlearned}} \\
        \vspace{-1.5mm}\tiny{\textbf{Diffusion}} \\
        \vspace{-1.5mm}\tiny{\textbf{Model}} \\
        \vspace{1mm}\tiny{\textbf{(Worst)}}
        \end{tabular} & \multicolumn{1}{m{0.9\textwidth}}{ 
 \includegraphics[width=0.084\textwidth]{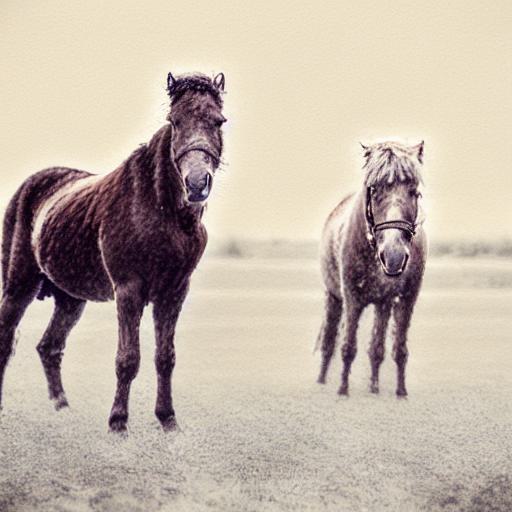} 
 \includegraphics[width=0.084\textwidth]{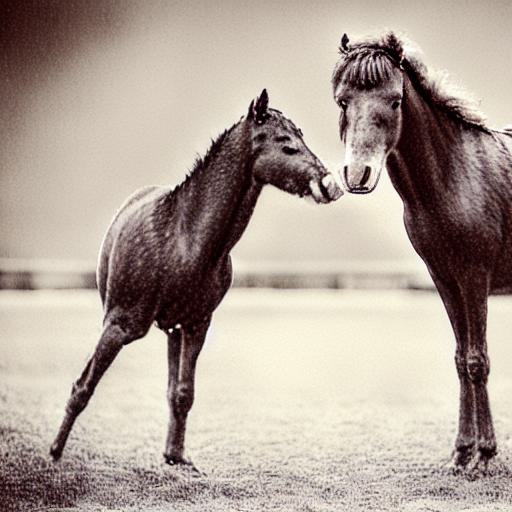} 
 \includegraphics[width=0.084\textwidth]{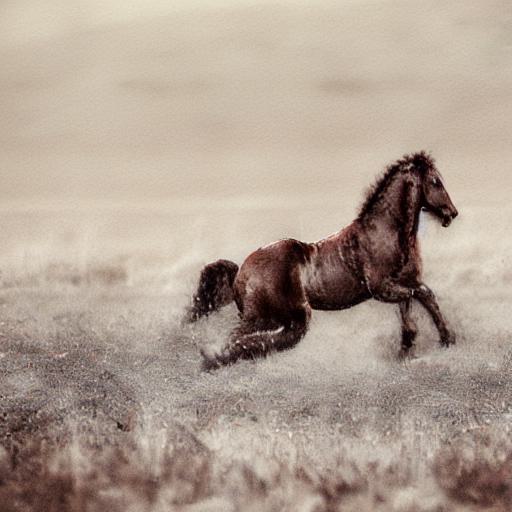} 
 \includegraphics[width=0.084\textwidth]{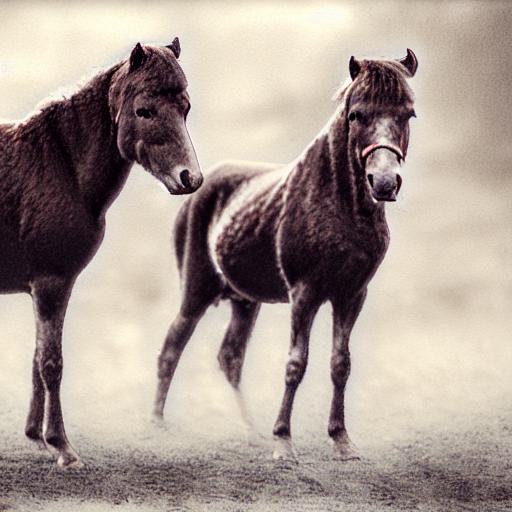} 
 \includegraphics[width=0.084\textwidth]{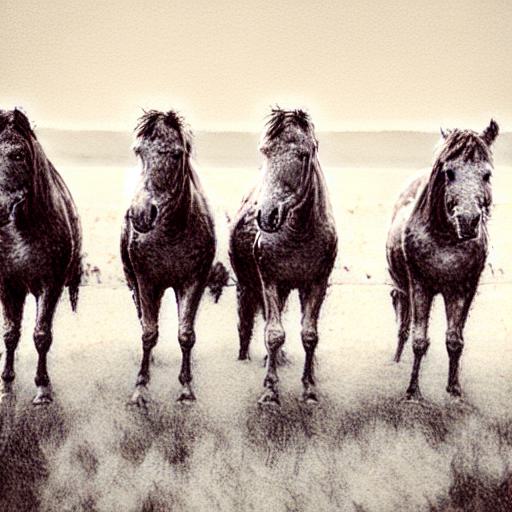} 
 \includegraphics[width=0.084\textwidth]{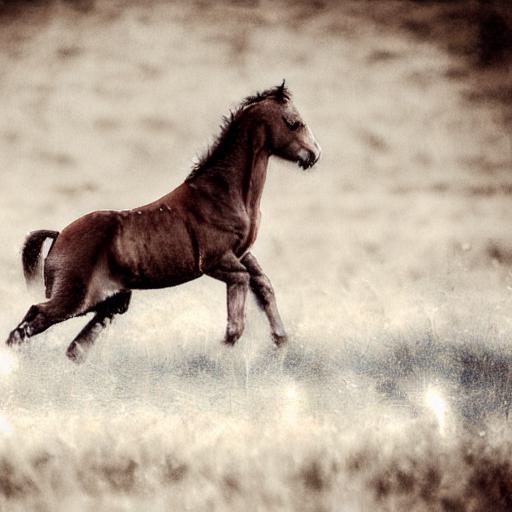} 
 \includegraphics[width=0.084\textwidth]{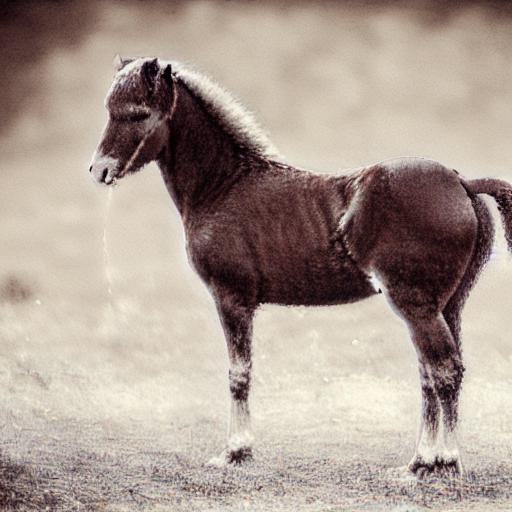} 
 \includegraphics[width=0.084\textwidth]{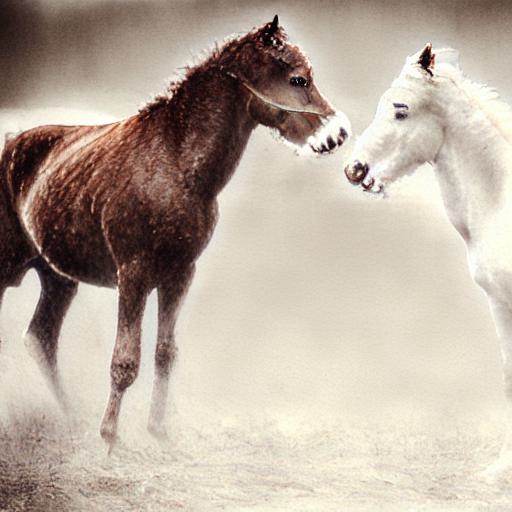} 
 \includegraphics[width=0.084\textwidth]{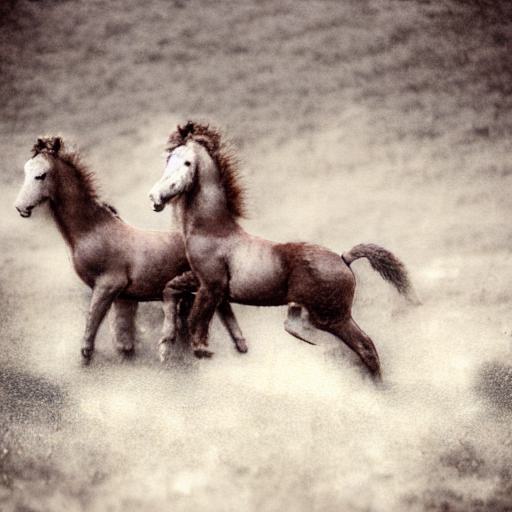} 
 \includegraphics[width=0.084\textwidth]{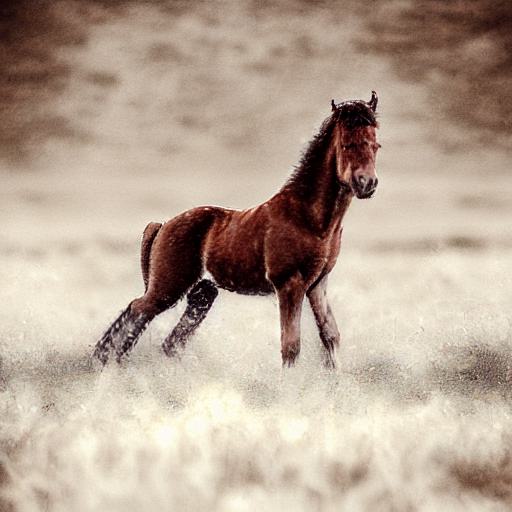}} \\ 
\midrule 
        \rowcolor{Gray}
        \multicolumn{2}{c}{~~~~~~~~~~~~\scriptsize{\Puw\texttt{:}
\scriptsize{\texttt{\textit{A painting of }\textcolor{Red}{Horses}} \texttt{\textit{in} \textcolor{Blue}{Van Gogh}\textit{ Style.}}}
}}\\
\midrule
        \begin{tabular}[c]{@{}c@{}}
        \vspace{-1.5mm}\tiny{\textbf{Original}} \\
        \vspace{-1.5mm}\tiny{\textbf{Diffusion}} \\
        \vspace{1mm}\tiny{\textbf{Model}}
        \end{tabular} & \multicolumn{1}{m{0.9\textwidth}}{
 \includegraphics[width=0.084\textwidth]{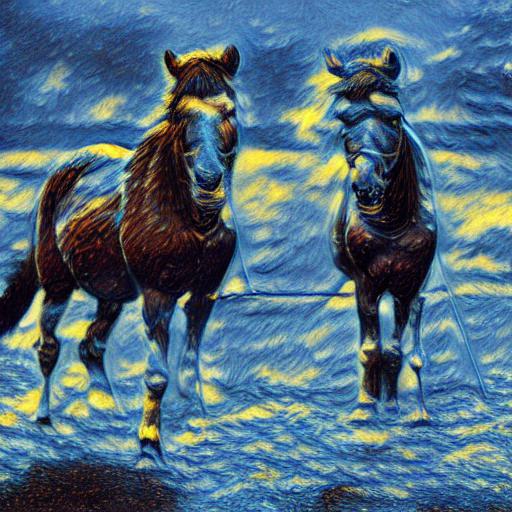} 
 \includegraphics[width=0.084\textwidth]{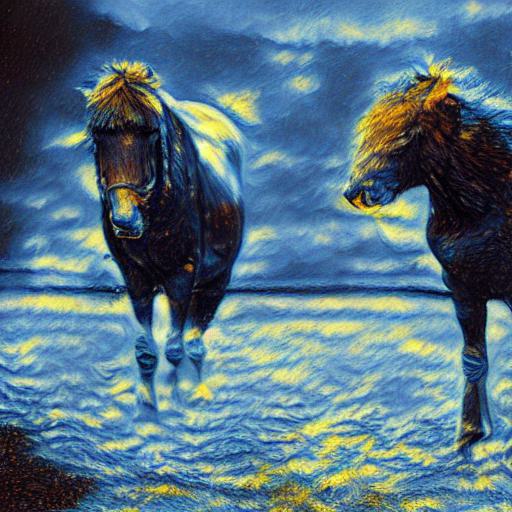} 
 \includegraphics[width=0.084\textwidth]{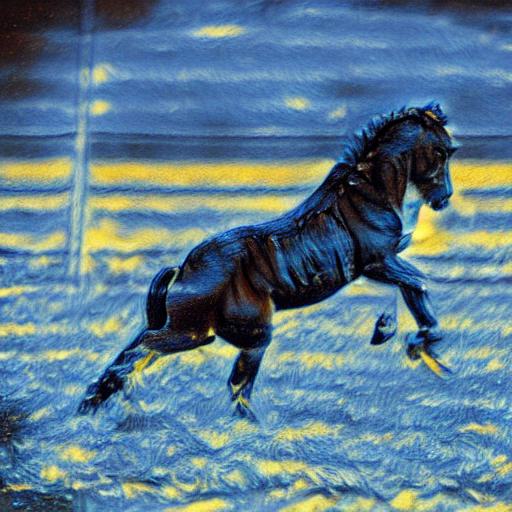} 
 \includegraphics[width=0.084\textwidth]{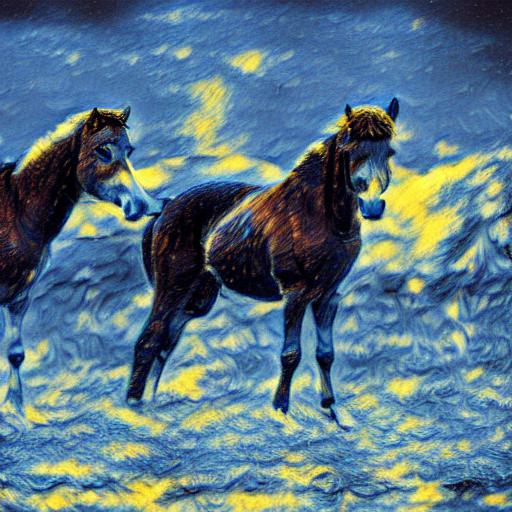} 
 \includegraphics[width=0.084\textwidth]{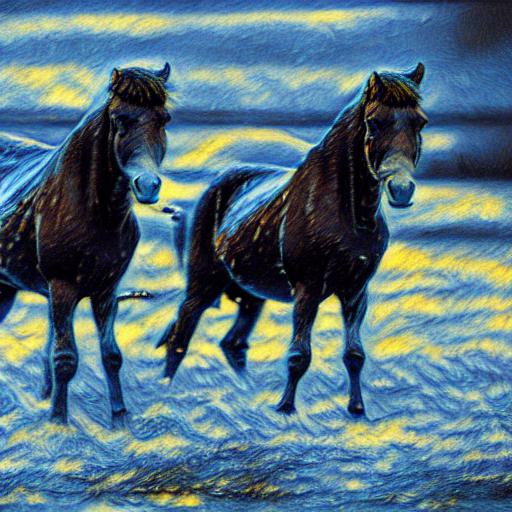} 
 \includegraphics[width=0.084\textwidth]{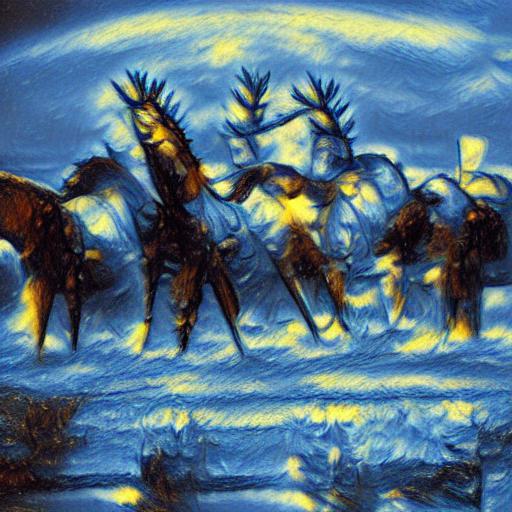} 
 \includegraphics[width=0.084\textwidth]{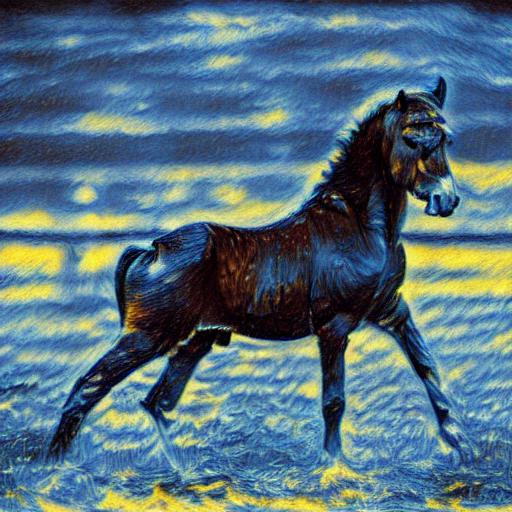} 
 \includegraphics[width=0.084\textwidth]{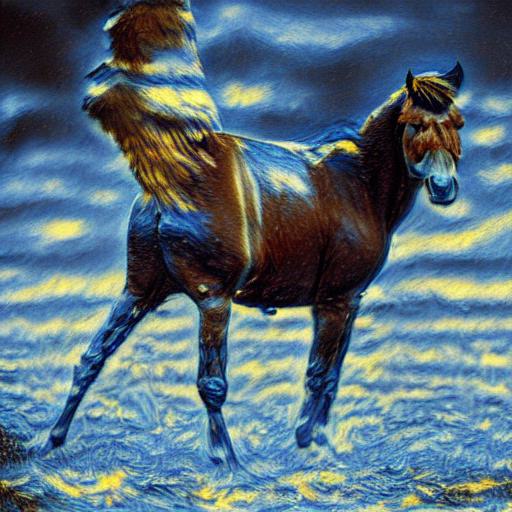} 
 \includegraphics[width=0.084\textwidth]{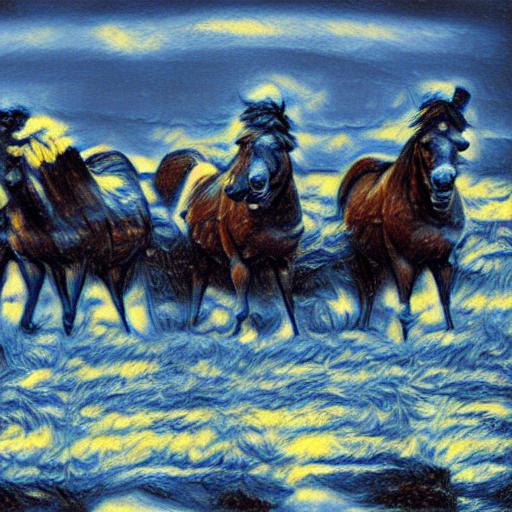} 
 \includegraphics[width=0.084\textwidth]{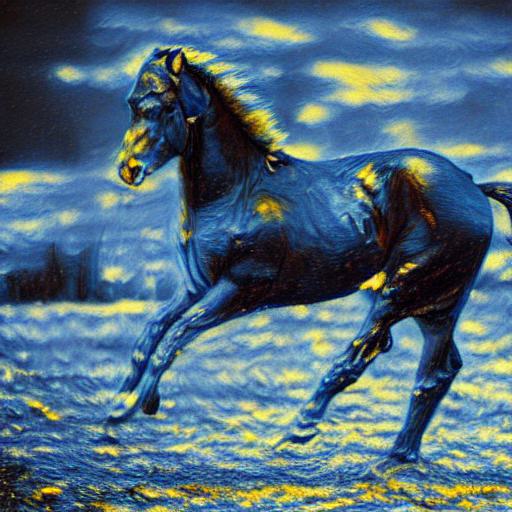}} \\ 
\midrule 
           \begin{tabular}[c]{@{}c@{}}
        \vspace{-1.5mm}\tiny{\textbf{Unlearned}} \\
        \vspace{-1.5mm}\tiny{\textbf{Diffusion}} \\
        \vspace{-1.5mm}\tiny{\textbf{Model}} \\
        \vspace{1mm}\tiny{\textbf{(Worst)}}
        \end{tabular} & \multicolumn{1}{m{0.9\textwidth}}{ 
 \includegraphics[width=0.084\textwidth]{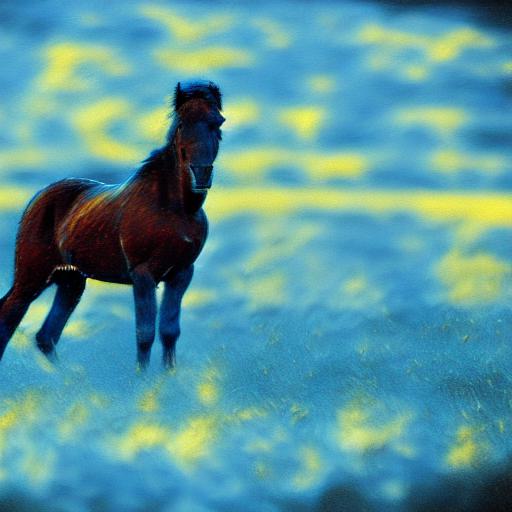} 
 \includegraphics[width=0.084\textwidth]{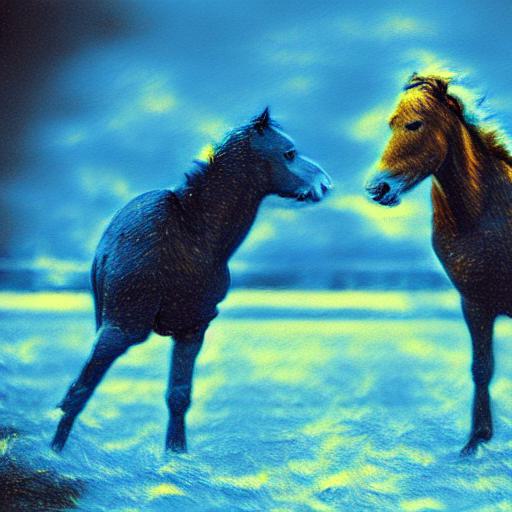} 
 \includegraphics[width=0.084\textwidth]{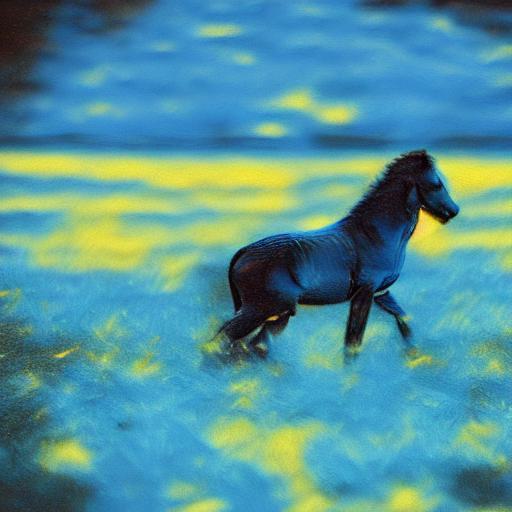} 
 \includegraphics[width=0.084\textwidth]{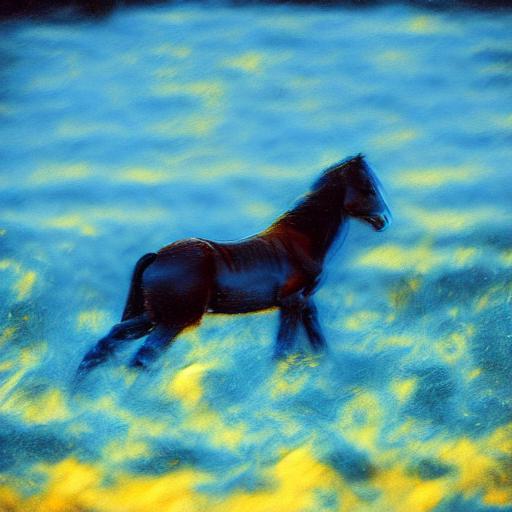} 
 \includegraphics[width=0.084\textwidth]{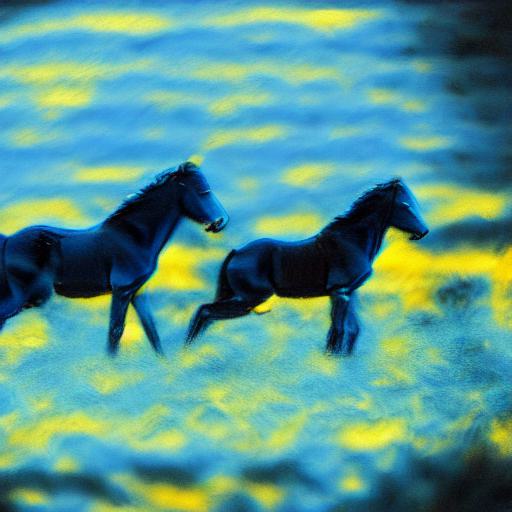} 
 \includegraphics[width=0.084\textwidth]{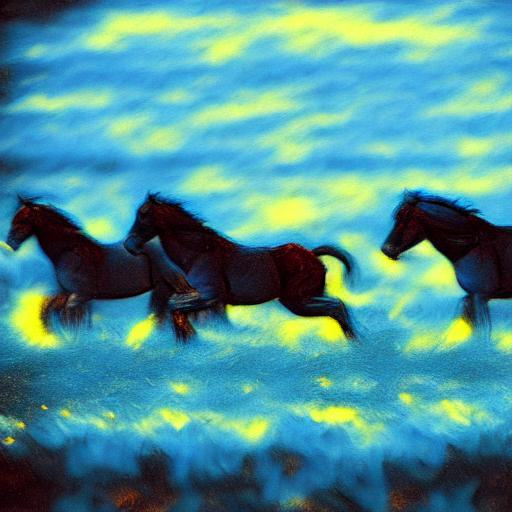} 
 \includegraphics[width=0.084\textwidth]{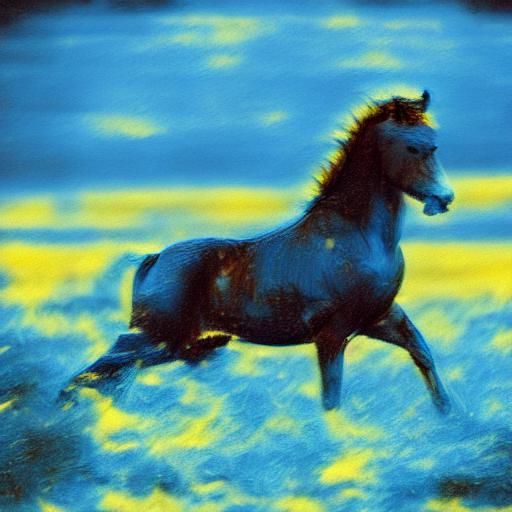} 
 \includegraphics[width=0.084\textwidth]{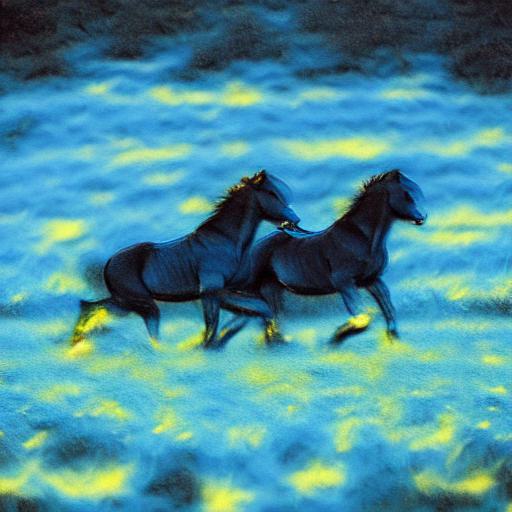} 
 \includegraphics[width=0.084\textwidth]{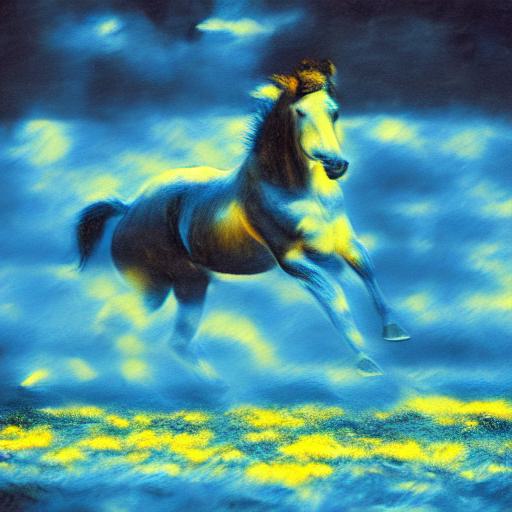} 
 \includegraphics[width=0.084\textwidth]{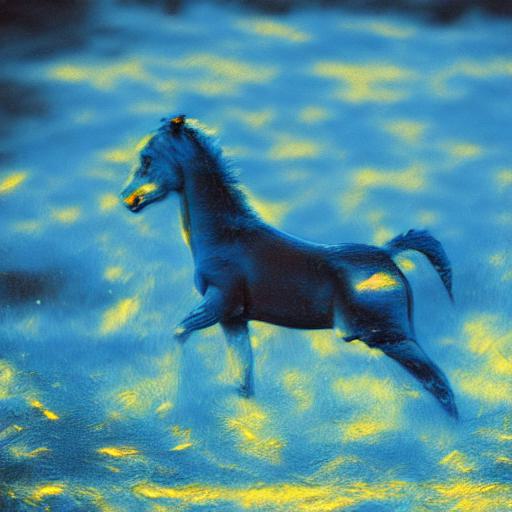}}

 \\ 
\midrule 
        \rowcolor{Gray}
        \multicolumn{2}{c}{~~~~~~~~~~~~\scriptsize{\Puw\texttt{:}
\scriptsize{\texttt{\textit{A painting of }\textcolor{Red}{Horses}} \texttt{\textit{in} \textcolor{Blue}{Rust}\textit{ Style.}}}
}}\\
\midrule
        \begin{tabular}[c]{@{}c@{}}
        \vspace{-1.5mm}\tiny{\textbf{Original}} \\
        \vspace{-1.5mm}\tiny{\textbf{Diffusion}} \\
        \vspace{1mm}\tiny{\textbf{Model}}
        \end{tabular} & \multicolumn{1}{m{0.9\textwidth}}{
 \includegraphics[width=0.084\textwidth]{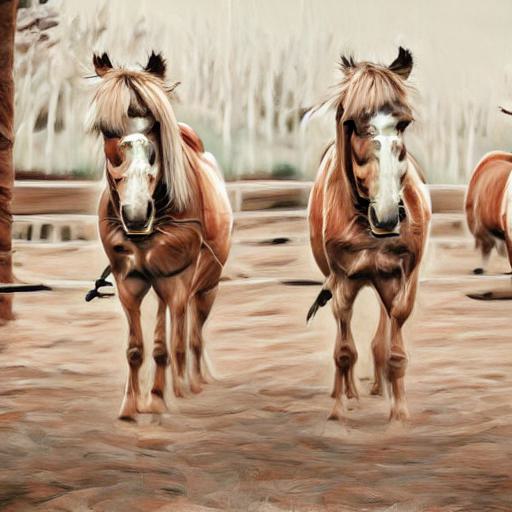} 
 \includegraphics[width=0.084\textwidth]{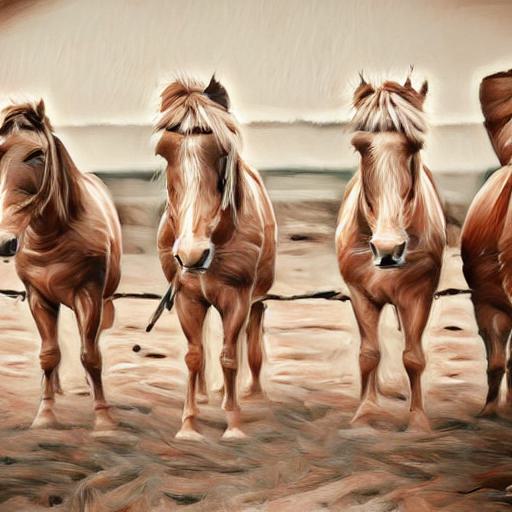} 
 \includegraphics[width=0.084\textwidth]{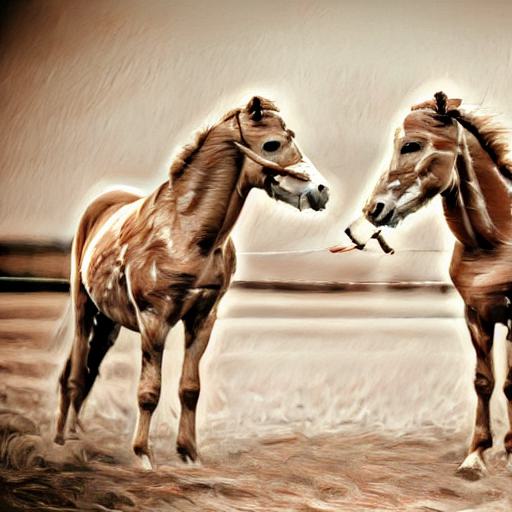} 
 \includegraphics[width=0.084\textwidth]{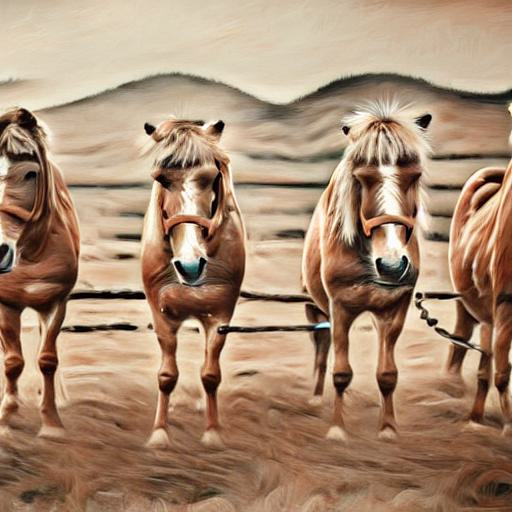} 
 \includegraphics[width=0.084\textwidth]{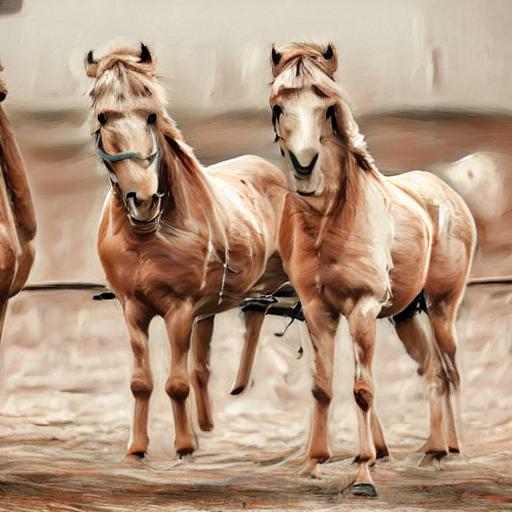} 
 \includegraphics[width=0.084\textwidth]{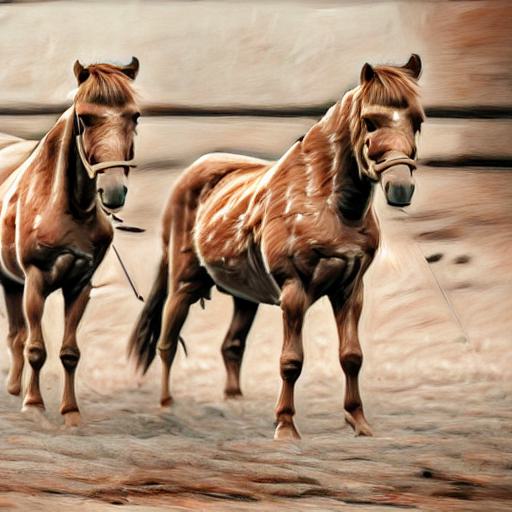} 
 \includegraphics[width=0.084\textwidth]{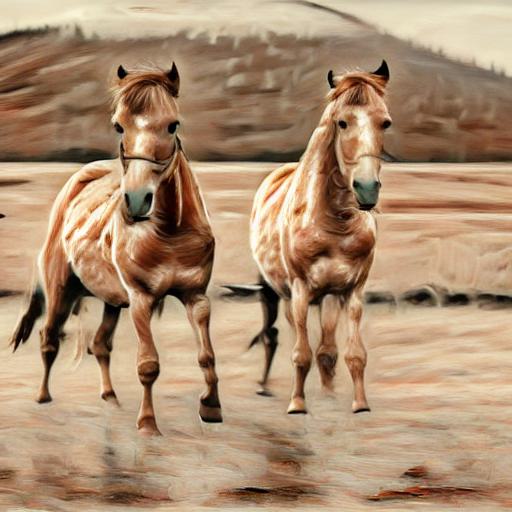} 
 \includegraphics[width=0.084\textwidth]{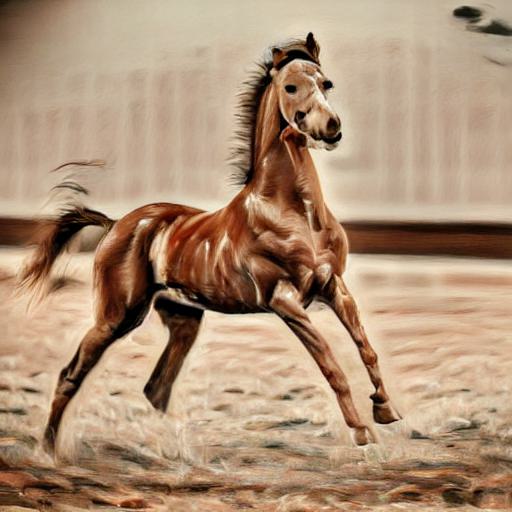} 
 \includegraphics[width=0.084\textwidth]{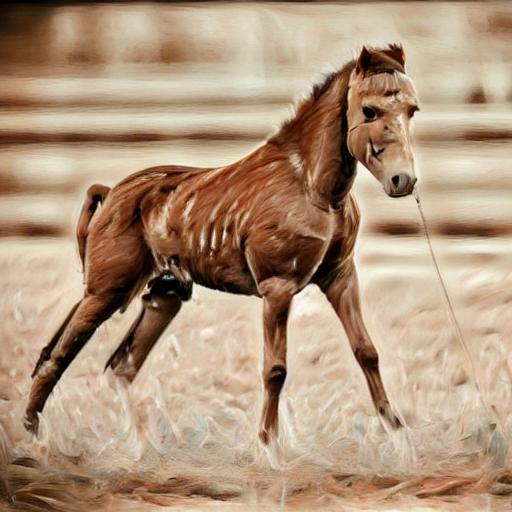} 
 \includegraphics[width=0.084\textwidth]{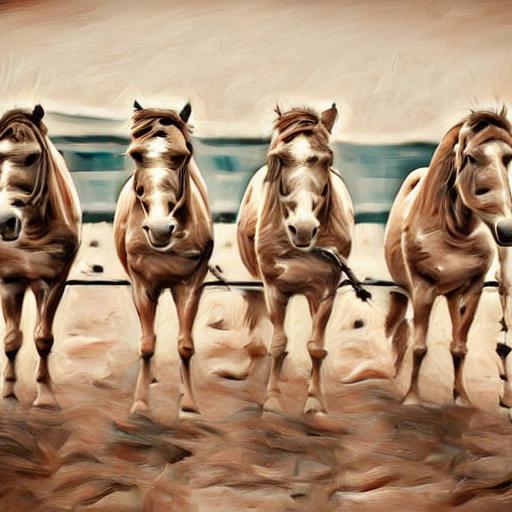}}
 \\ 
\midrule 
         \begin{tabular}[c]{@{}c@{}}
        \vspace{-1.5mm}\tiny{\textbf{Unlearned}} \\
        \vspace{-1.5mm}\tiny{\textbf{Diffusion}} \\
        \vspace{-1.5mm}\tiny{\textbf{Model}} \\
        \vspace{1mm}\tiny{\textbf{(Worst)}}
        \end{tabular} & \multicolumn{1}{m{0.9\textwidth}}{
 \includegraphics[width=0.084\textwidth]{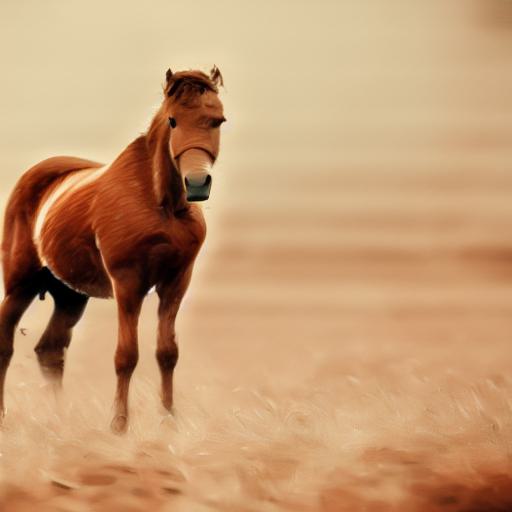} 
 \includegraphics[width=0.084\textwidth]{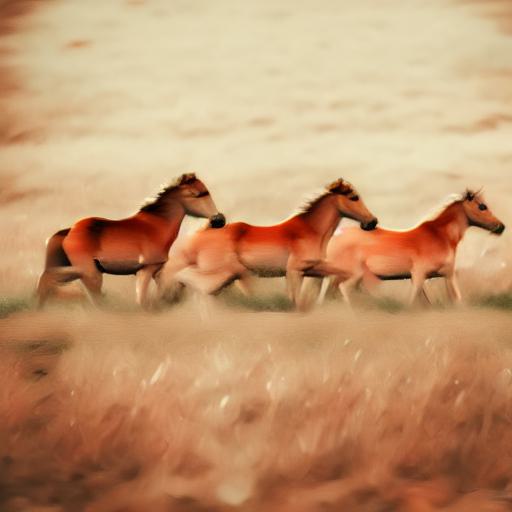} 
 \includegraphics[width=0.084\textwidth]{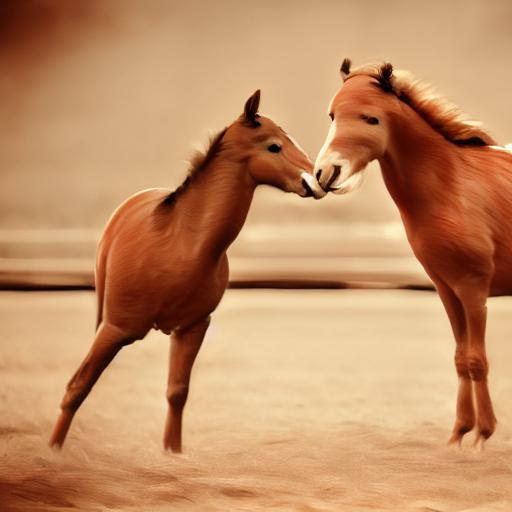} 
 \includegraphics[width=0.084\textwidth]{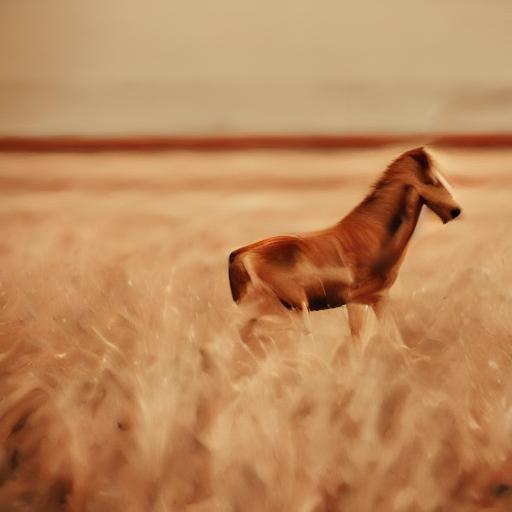} 
 \includegraphics[width=0.084\textwidth]{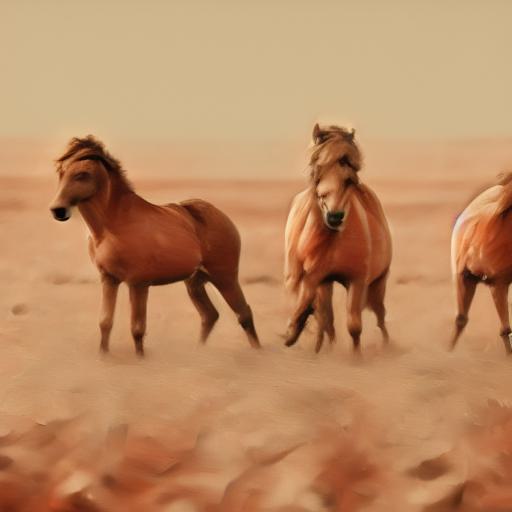} 
 \includegraphics[width=0.084\textwidth]{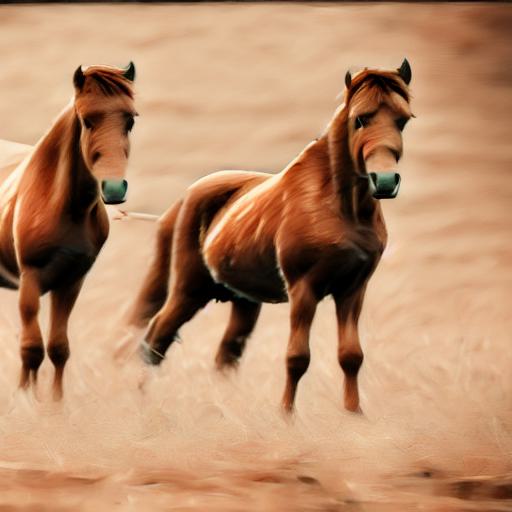} 
 \includegraphics[width=0.084\textwidth]{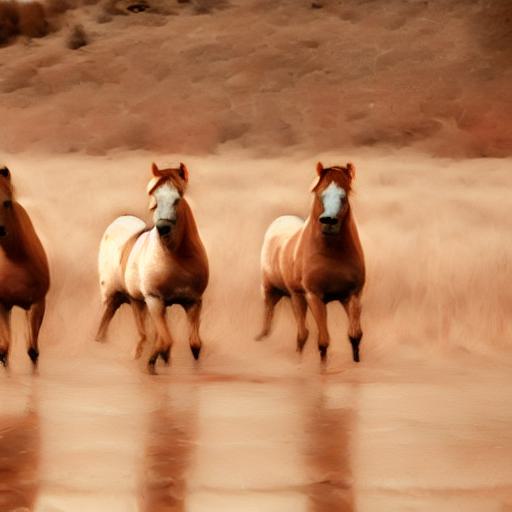} 
 \includegraphics[width=0.084\textwidth]{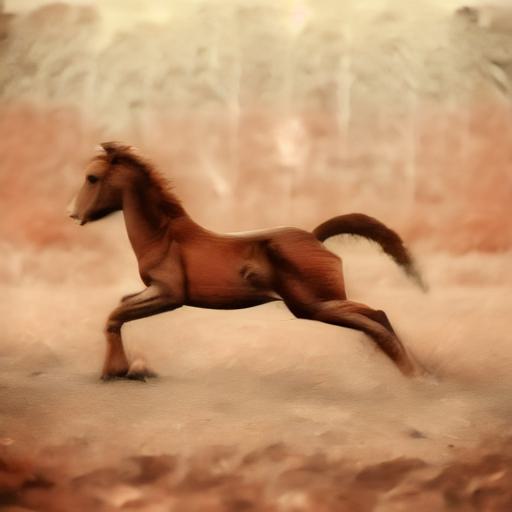} 
 \includegraphics[width=0.084\textwidth]{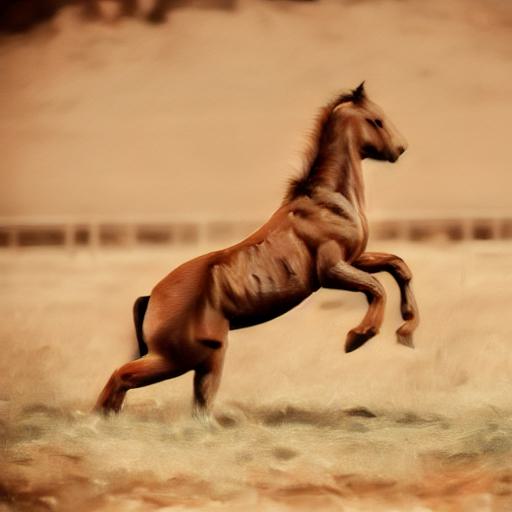} 
 \includegraphics[width=0.084\textwidth]{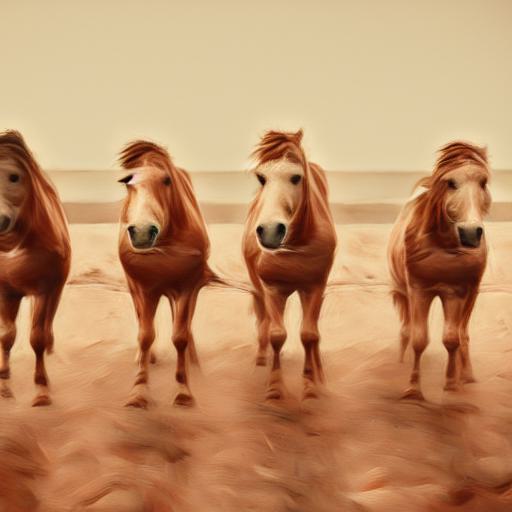}}
 \\ 
        \midrule
        \rowcolor{Gray}
        \multicolumn{2}{c}{~~~~~~~~~~~~\scriptsize{\Puw\texttt{:}
\scriptsize{\texttt{\textit{A painting of }\textcolor{Red}{Human}} \texttt{\textit{in} \textcolor{Blue}{Winter}\textit{ Style.}}}
}}\\
\midrule
        \begin{tabular}[c]{@{}c@{}}
        \vspace{-1.5mm}\tiny{\textbf{Original}} \\
        \vspace{-1.5mm}\tiny{\textbf{Diffusion}} \\
        \vspace{1mm}\tiny{\textbf{Model}}
        \end{tabular} & \multicolumn{1}{m{0.9\textwidth}}{
 \includegraphics[width=0.084\textwidth]{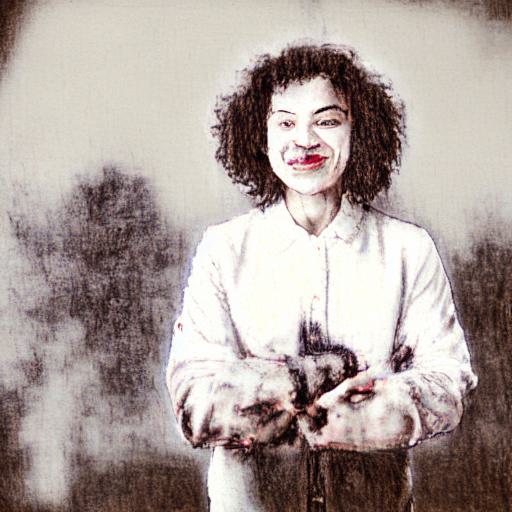} 
 \includegraphics[width=0.084\textwidth]{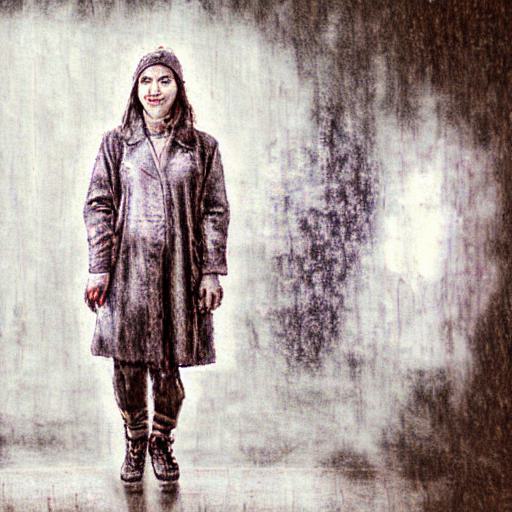} 
 \includegraphics[width=0.084\textwidth]{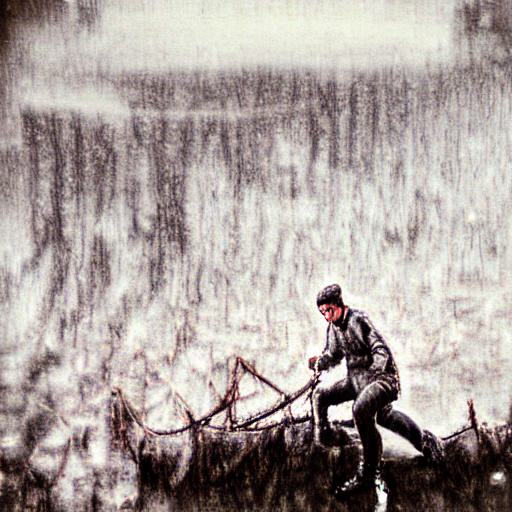} 
 \includegraphics[width=0.084\textwidth]{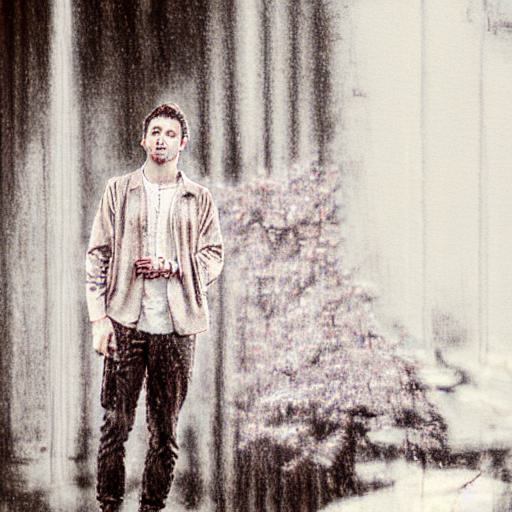} 
 \includegraphics[width=0.084\textwidth]{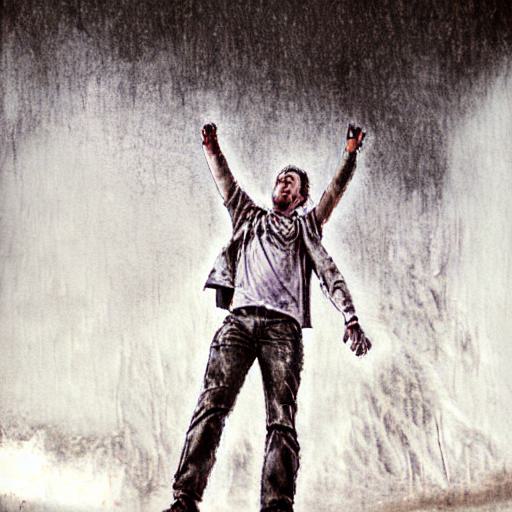} 
 \includegraphics[width=0.084\textwidth]{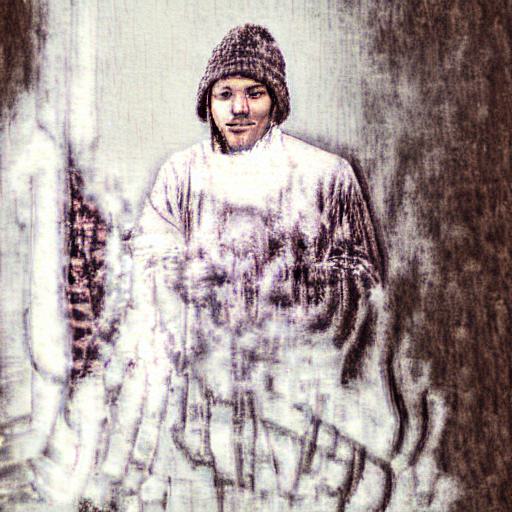} 
 \includegraphics[width=0.084\textwidth]{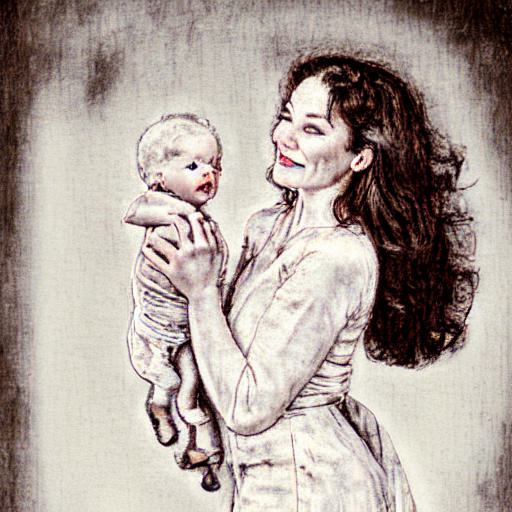} 
 \includegraphics[width=0.084\textwidth]{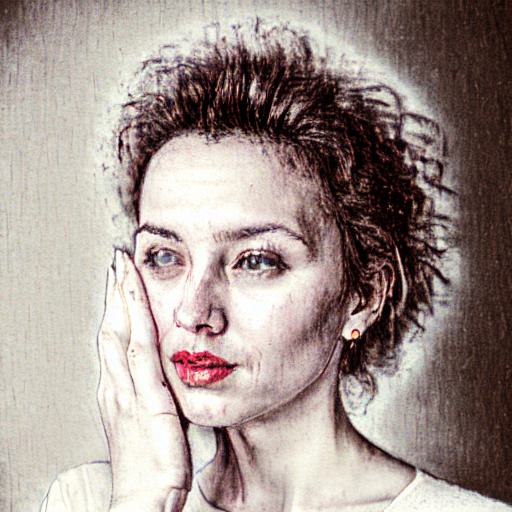} 
 \includegraphics[width=0.084\textwidth]{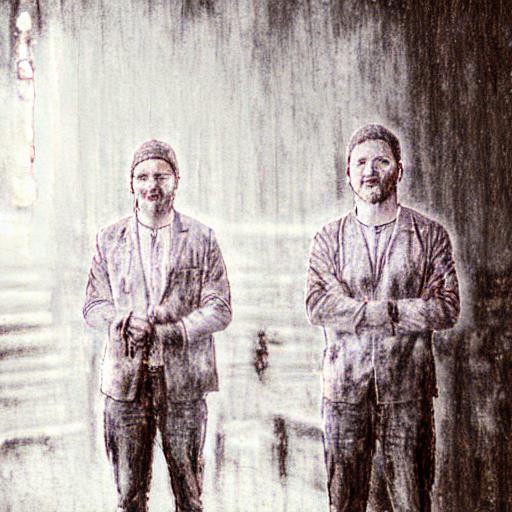} 
 \includegraphics[width=0.084\textwidth]{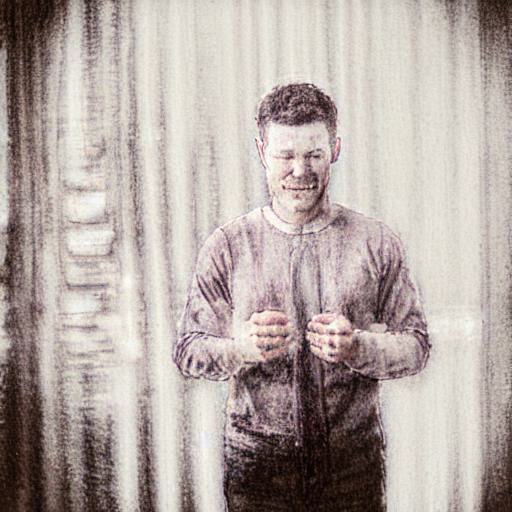}} \\ 
\midrule 
         \begin{tabular}[c]{@{}c@{}}
        \vspace{-1.5mm}\tiny{\textbf{Unlearned}} \\
        \vspace{-1.5mm}\tiny{\textbf{Diffusion}} \\
        \vspace{-1.5mm}\tiny{\textbf{Model}} \\
        \vspace{1mm}\tiny{\textbf{(Worst)}}
        \end{tabular} & \multicolumn{1}{m{0.9\textwidth}}{
 \includegraphics[width=0.084\textwidth]{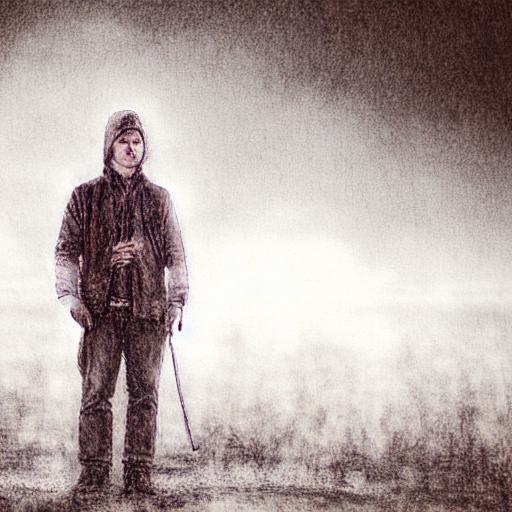} 
 \includegraphics[width=0.084\textwidth]{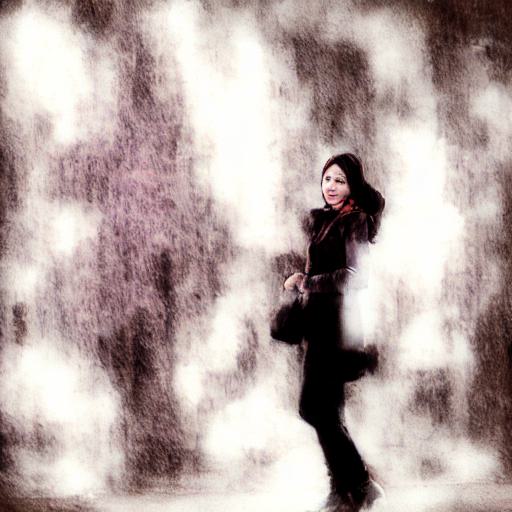} 
 \includegraphics[width=0.084\textwidth]{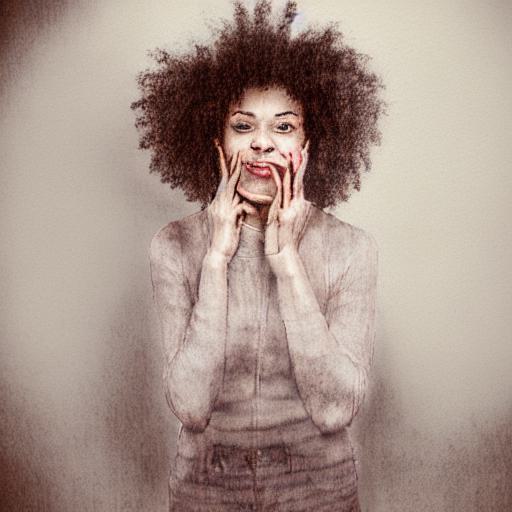} 
 \includegraphics[width=0.084\textwidth]{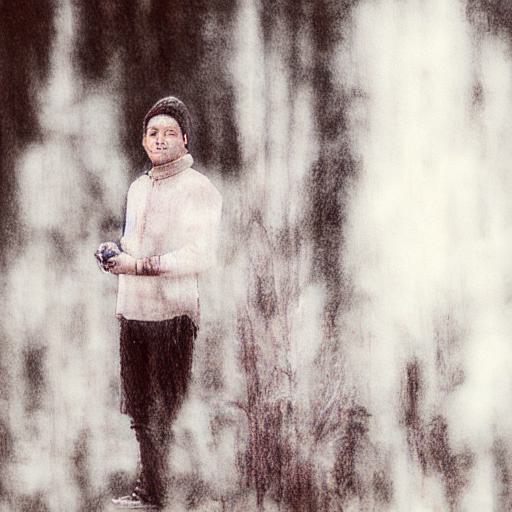} 
 \includegraphics[width=0.084\textwidth]{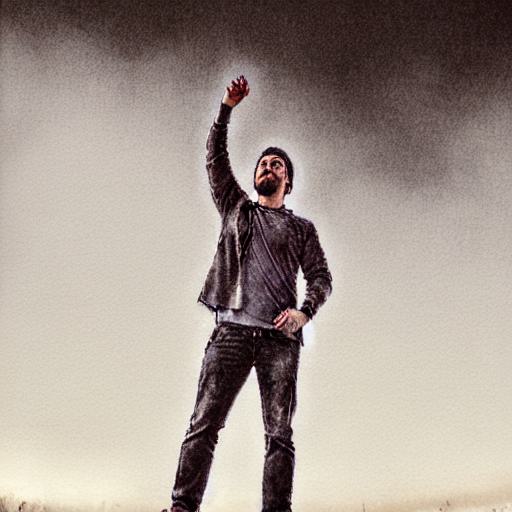} 
 \includegraphics[width=0.084\textwidth]{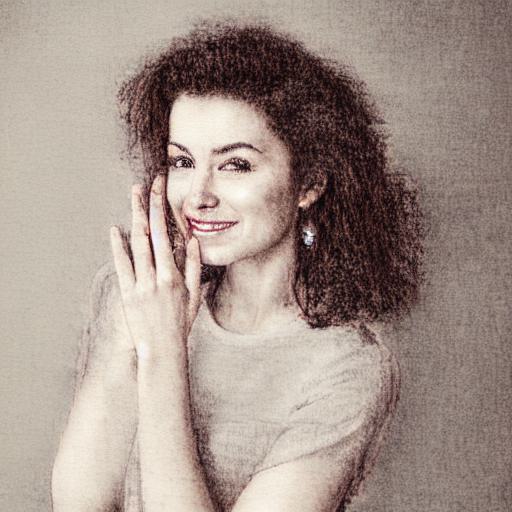} 
 \includegraphics[width=0.084\textwidth]{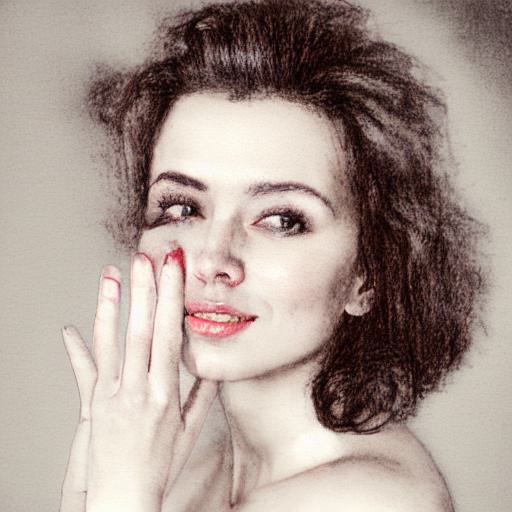} 
 \includegraphics[width=0.084\textwidth]{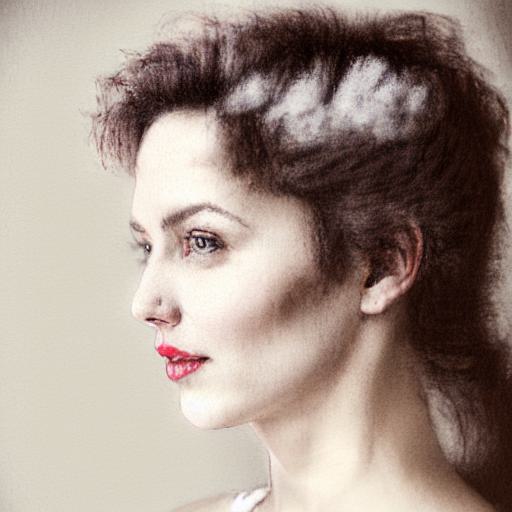} 
 \includegraphics[width=0.084\textwidth]{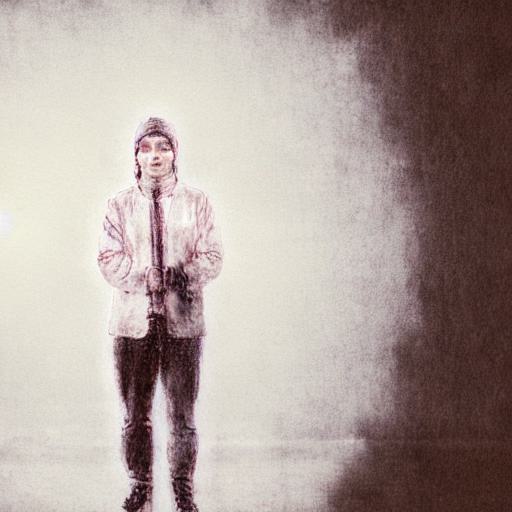} 
 \includegraphics[width=0.084\textwidth]{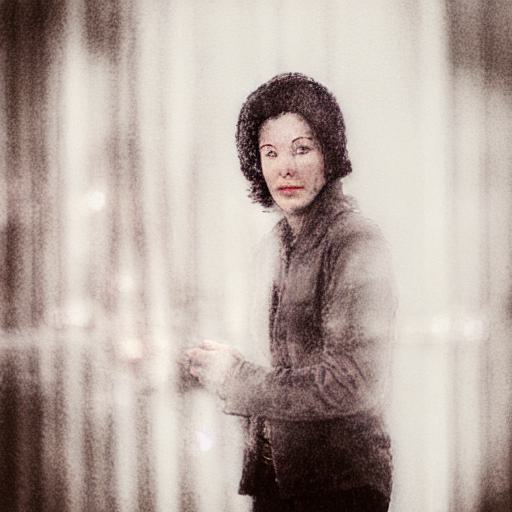}} \\ 
\midrule

        \bottomrule[1pt]
    \end{tabular}
    }
    \end{center}
\end{table}

\begin{table}[t]
    \centering
    % \vspace{-2mm}
    \caption{{Examples of image generation using the original stable diffusion model (w/o unlearning), the unlearned diffusion model over the worst-case forgetting prompt set (Worst). For each diffusion model, images are generated based on an unlearned prompt from the worst-case forget set. 
    %Extended results from  Table\,\ref{fig: generation_examples_appendix_more} with different generation conditions.
    }}
    %\vspace{-5mm}
    \label{fig: generation_examples_appendix_more}
    \begin{center}
    \resizebox{1.0\textwidth}{!}{
    \renewcommand\tabcolsep{5pt}
    \begin{tabular}{c|c}
        \toprule[1pt]
        \midrule
 
        {\scriptsize{\textbf{Model}}} & \multicolumn{1}{c}{\scriptsize{\textbf{Generation Condition}}} \\
        \midrule
        \rowcolor{Gray}
        \multicolumn{2}{c}{~~~~~~~~~~~~\scriptsize{\Puw\texttt{:}
\scriptsize{\texttt{\textit{A painting of }\textcolor{Red}{Human}} \texttt{\textit{in} \textcolor{Blue}{Van Gogh}\textit{ Style.}}}
}}\\
\midrule
        \begin{tabular}[c]{@{}c@{}}
        \vspace{-1.5mm}\tiny{\textbf{Original}} \\
        \vspace{-1.5mm}\tiny{\textbf{Diffusion}} \\
        \vspace{1mm}\tiny{\textbf{Model}}
        \end{tabular} & \multicolumn{1}{m{0.9\textwidth}}{
 \includegraphics[width=0.084\textwidth]{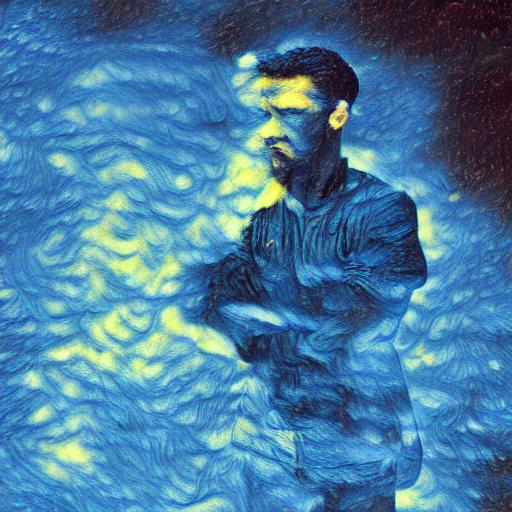} 
 \includegraphics[width=0.084\textwidth]{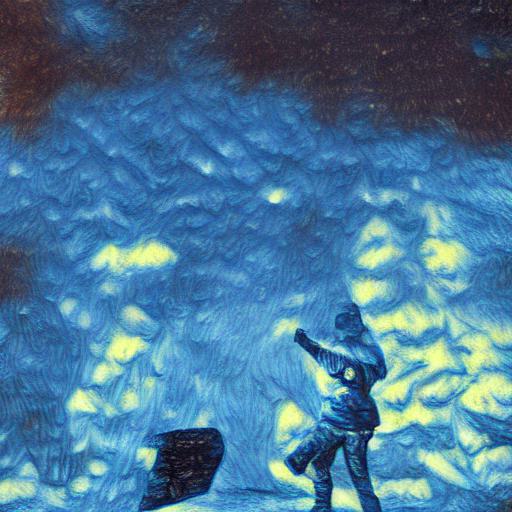} 
 \includegraphics[width=0.084\textwidth]{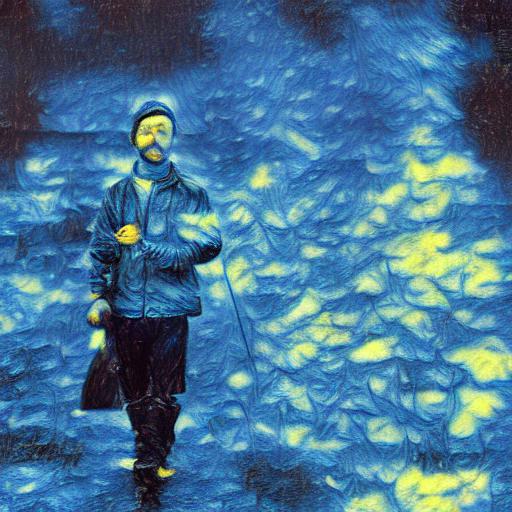} 
 \includegraphics[width=0.084\textwidth]{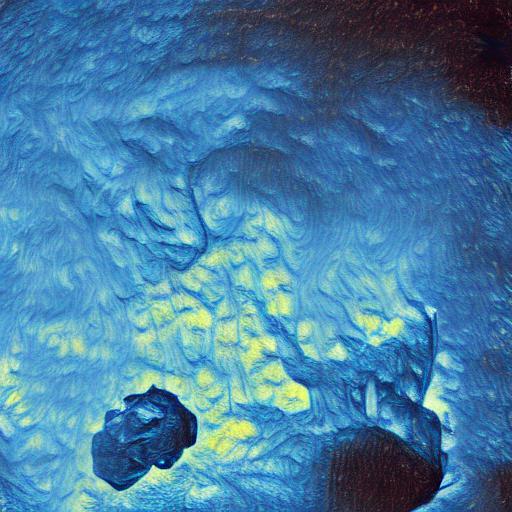} 
 \includegraphics[width=0.084\textwidth]{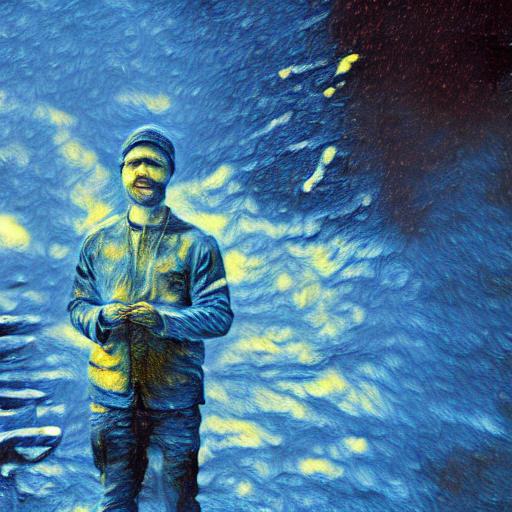} 
 \includegraphics[width=0.084\textwidth]{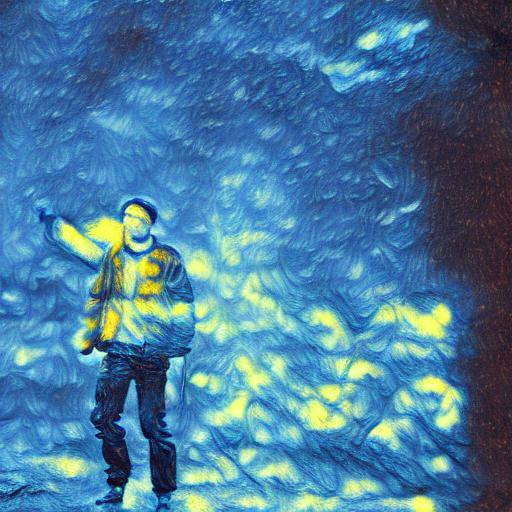} 
 \includegraphics[width=0.084\textwidth]{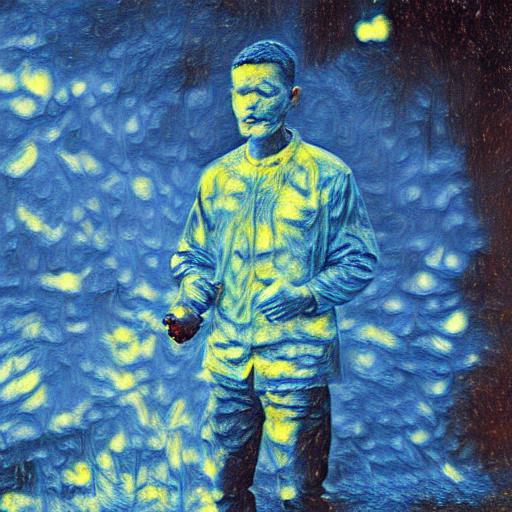} 
 \includegraphics[width=0.084\textwidth]{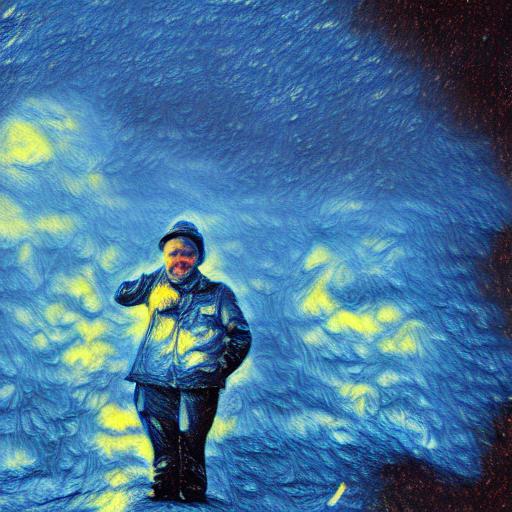} 
 \includegraphics[width=0.084\textwidth]{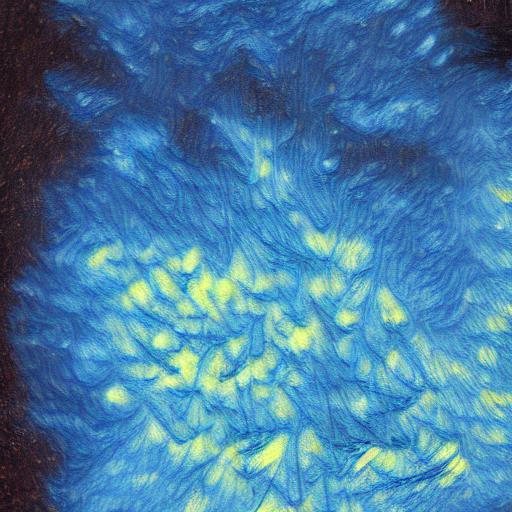} 
 \includegraphics[width=0.084\textwidth]{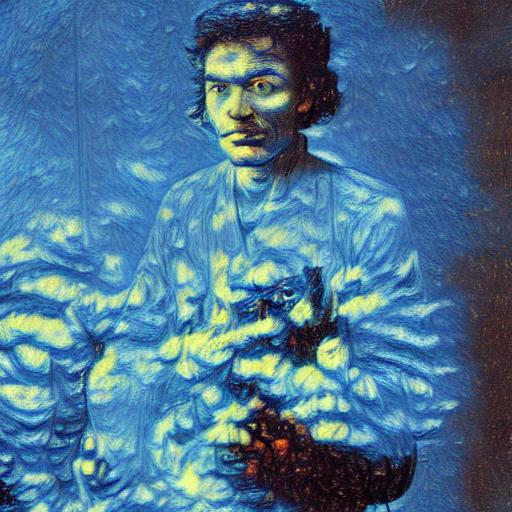}} \\ 
\midrule 
         \begin{tabular}[c]{@{}c@{}}
        \vspace{-1.5mm}\tiny{\textbf{Unlearned}} \\
        \vspace{-1.5mm}\tiny{\textbf{Diffusion}} \\
        \vspace{-1.5mm}\tiny{\textbf{Model}} \\
        \vspace{1mm}\tiny{\textbf{(Worst)}}
        \end{tabular} & \multicolumn{1}{m{0.9\textwidth}}{
 \includegraphics[width=0.084\textwidth]{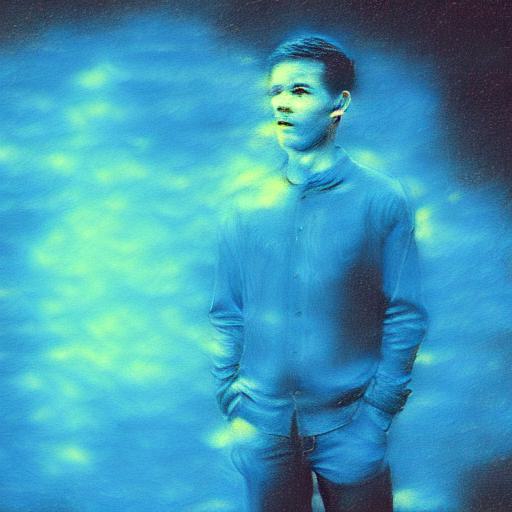} 
 \includegraphics[width=0.084\textwidth]{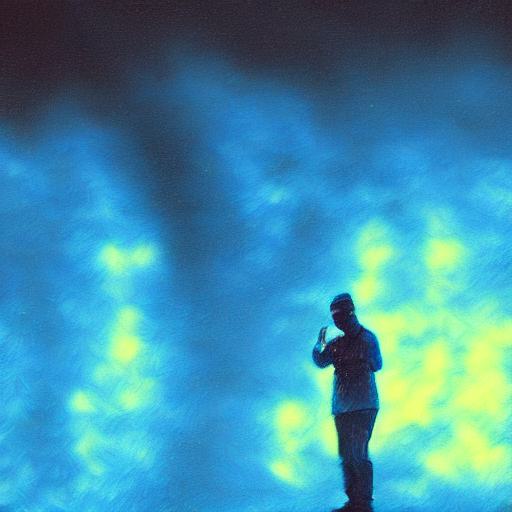} 
 \includegraphics[width=0.084\textwidth]{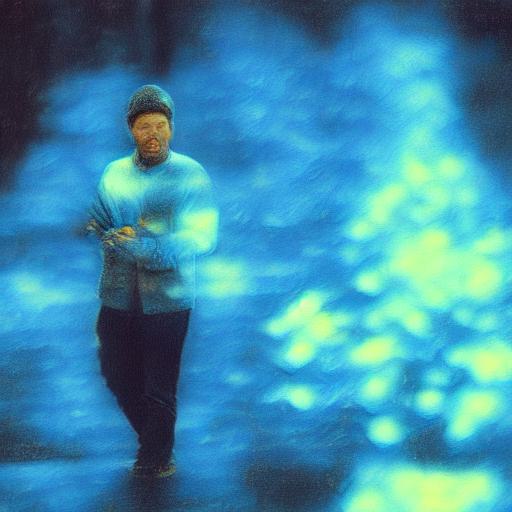} 
 \includegraphics[width=0.084\textwidth]{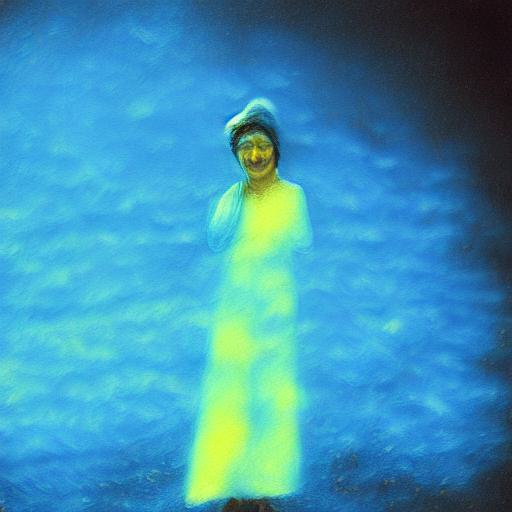} 
 \includegraphics[width=0.084\textwidth]{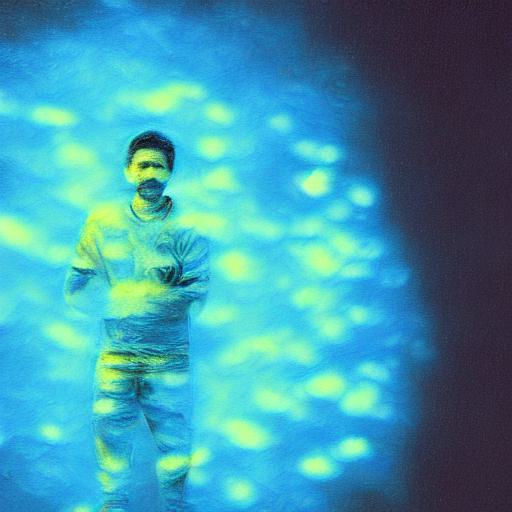} 
 \includegraphics[width=0.084\textwidth]{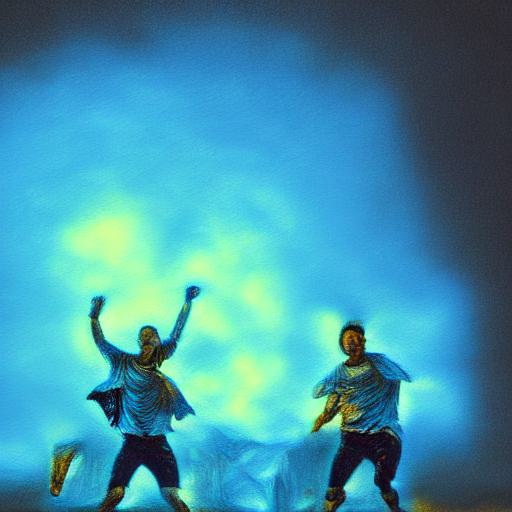} 
 \includegraphics[width=0.084\textwidth]{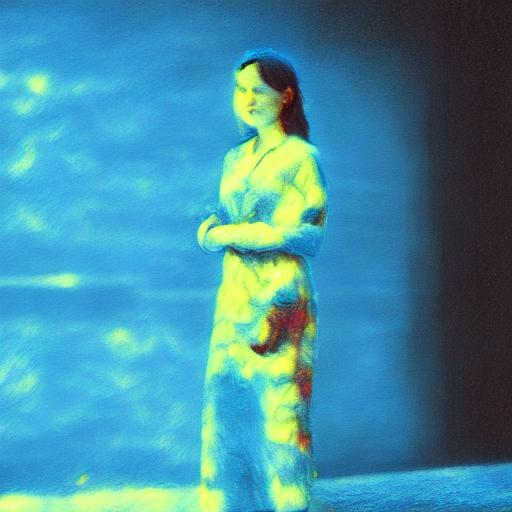} 
 \includegraphics[width=0.084\textwidth]{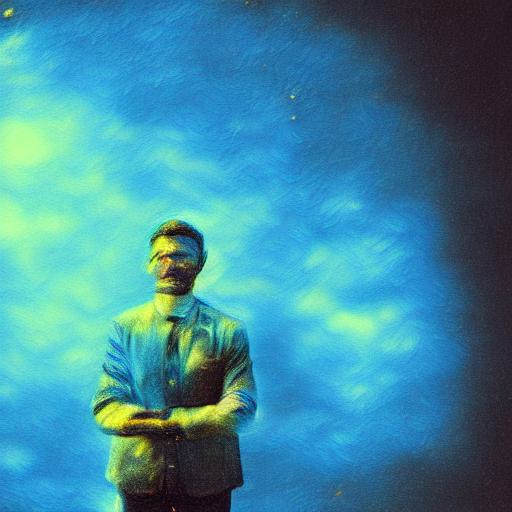} 
 \includegraphics[width=0.084\textwidth]{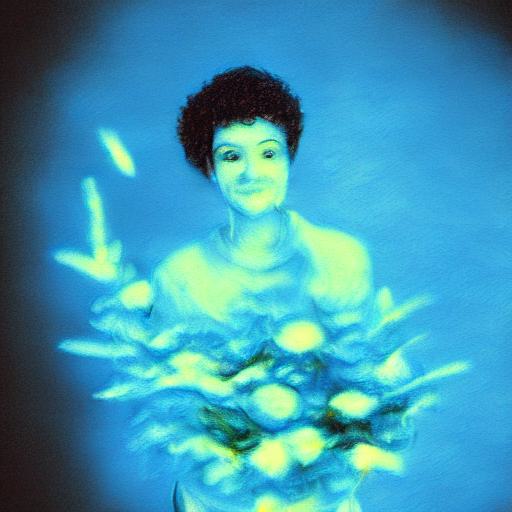} 
 \includegraphics[width=0.084\textwidth]{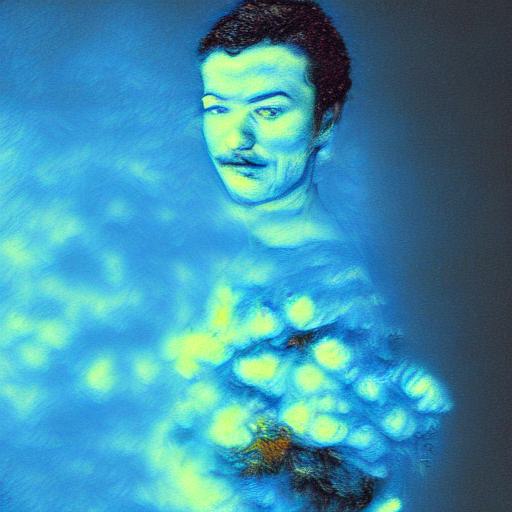}} \\ 
\midrule 
        \rowcolor{Gray}
        \multicolumn{2}{c}{~~~~~~~~~~~~\scriptsize{\Puw\texttt{:}
\scriptsize{\texttt{\textit{A painting of }\textcolor{Red}{Human}} \texttt{\textit{in} \textcolor{Blue}{Rust}\textit{ Style.}}}
}}\\
\midrule
        \begin{tabular}[c]{@{}c@{}}
        \vspace{-1.5mm}\tiny{\textbf{Original}} \\
        \vspace{-1.5mm}\tiny{\textbf{Diffusion}} \\
        \vspace{1mm}\tiny{\textbf{Model}}
        \end{tabular} & \multicolumn{1}{m{0.9\textwidth}}{
 \includegraphics[width=0.084\textwidth]{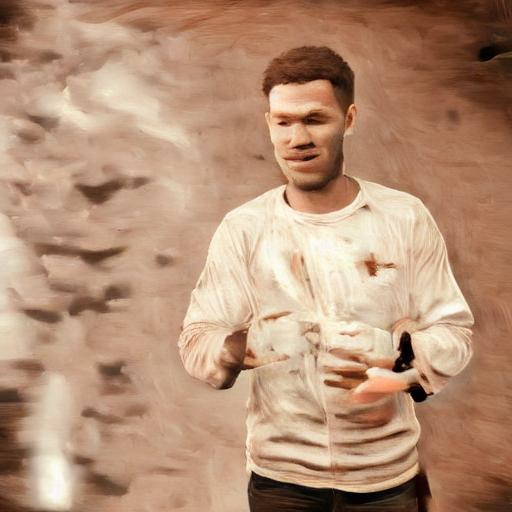} 
 \includegraphics[width=0.084\textwidth]{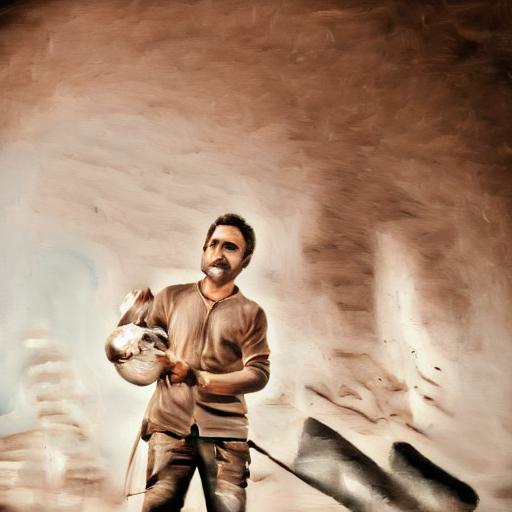} 
 \includegraphics[width=0.084\textwidth]{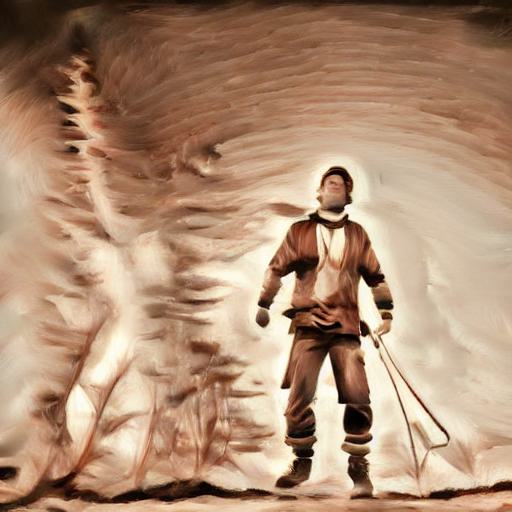} 
 \includegraphics[width=0.084\textwidth]{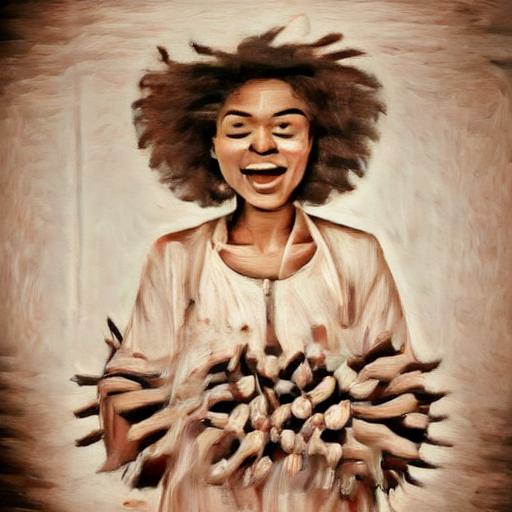} 
 \includegraphics[width=0.084\textwidth]{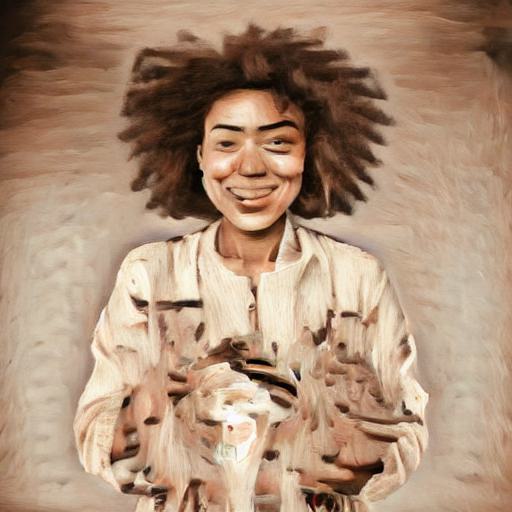} 
 \includegraphics[width=0.084\textwidth]{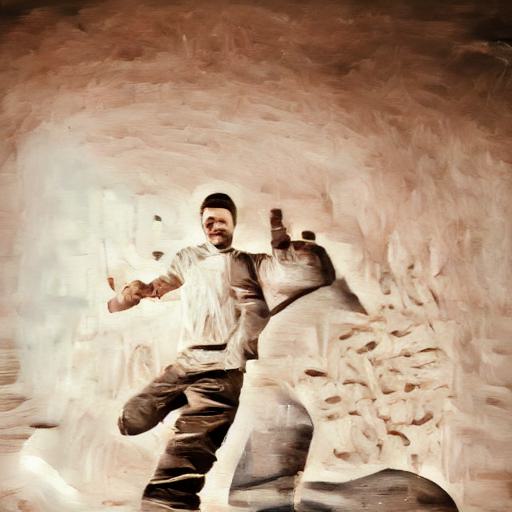} 
 \includegraphics[width=0.084\textwidth]{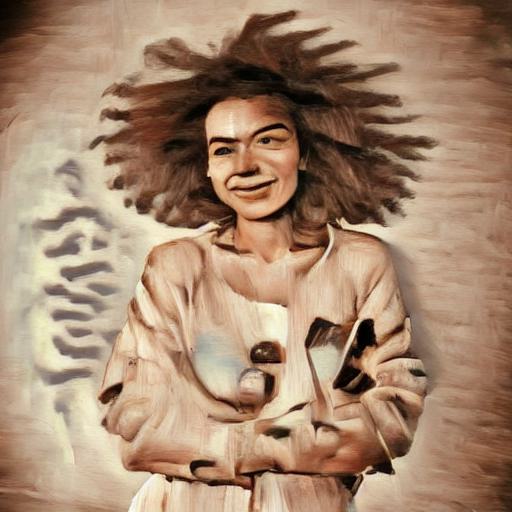} 
 \includegraphics[width=0.084\textwidth]{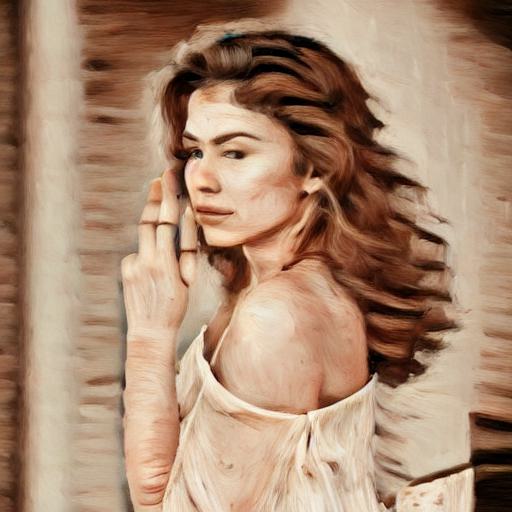} 
 \includegraphics[width=0.084\textwidth]{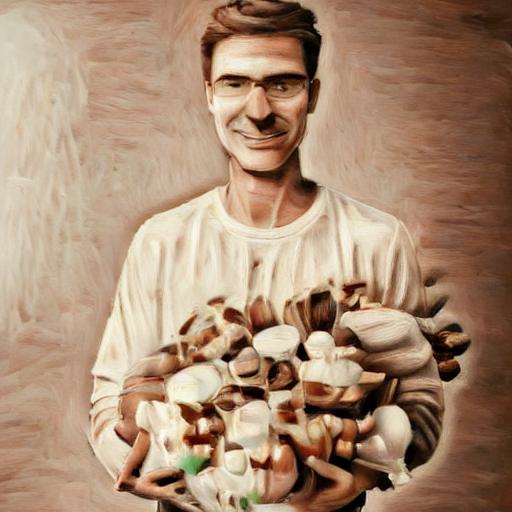} 
 \includegraphics[width=0.084\textwidth]{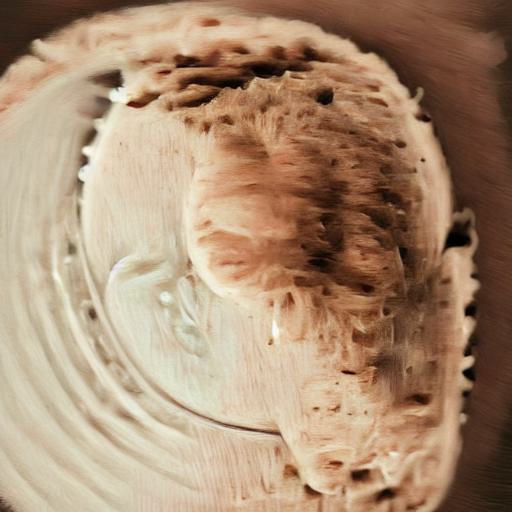}} \\ 
\midrule 
         \begin{tabular}[c]{@{}c@{}}
        \vspace{-1.5mm}\tiny{\textbf{Unlearned}} \\
        \vspace{-1.5mm}\tiny{\textbf{Diffusion}} \\
        \vspace{-1.5mm}\tiny{\textbf{Model}} \\
        \vspace{1mm}\tiny{\textbf{(Worst)}}
        \end{tabular} & \multicolumn{1}{m{0.9\textwidth}}{ 
 \includegraphics[width=0.084\textwidth]{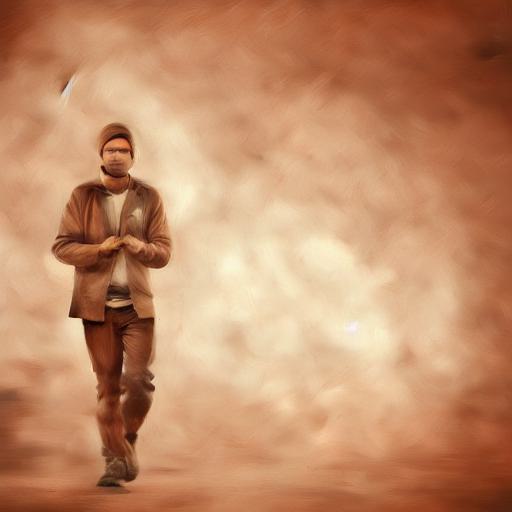} 
 \includegraphics[width=0.084\textwidth]{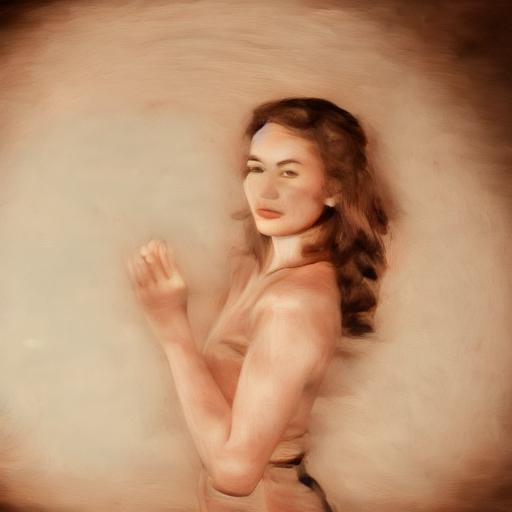} 
 \includegraphics[width=0.084\textwidth]{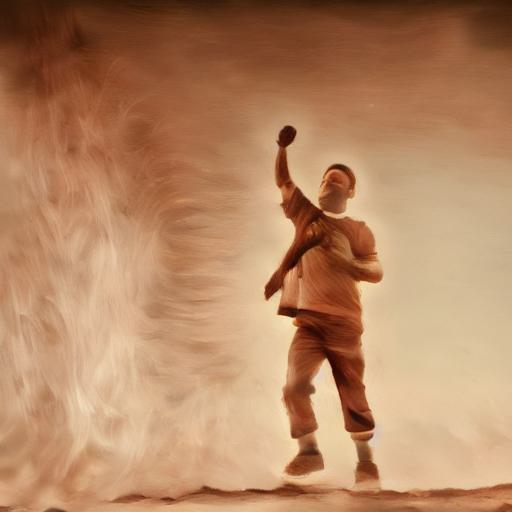} 
 \includegraphics[width=0.084\textwidth]{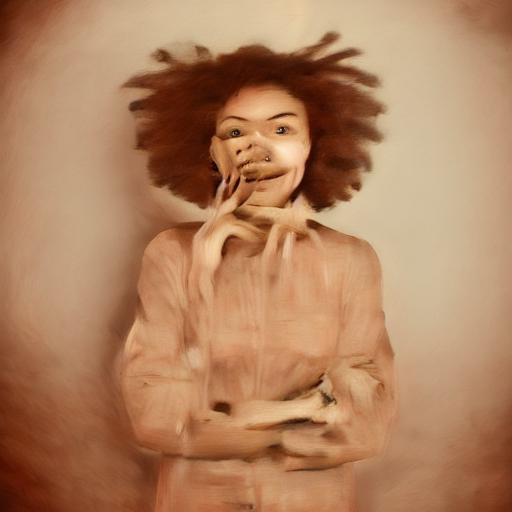} 
 \includegraphics[width=0.084\textwidth]{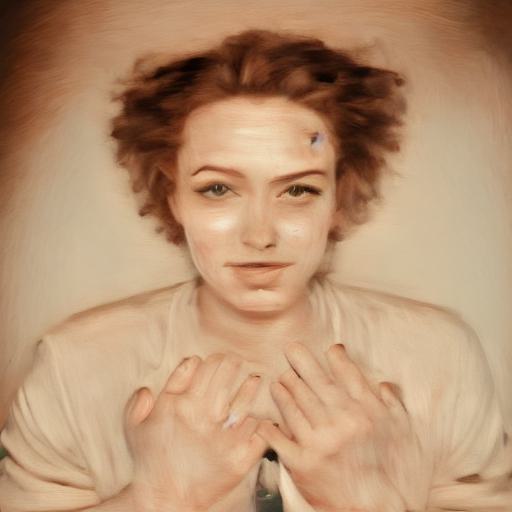} 
 \includegraphics[width=0.084\textwidth]{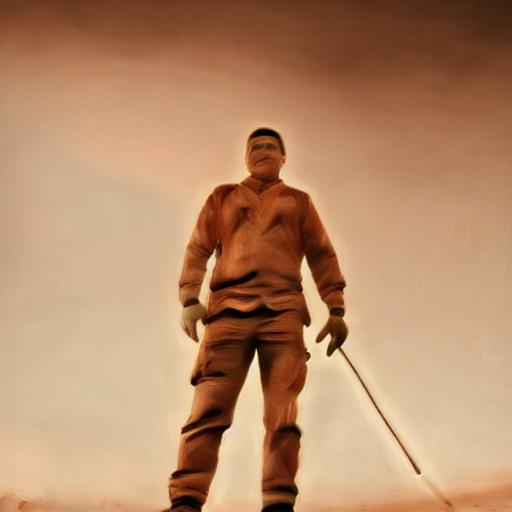} 
 \includegraphics[width=0.084\textwidth]{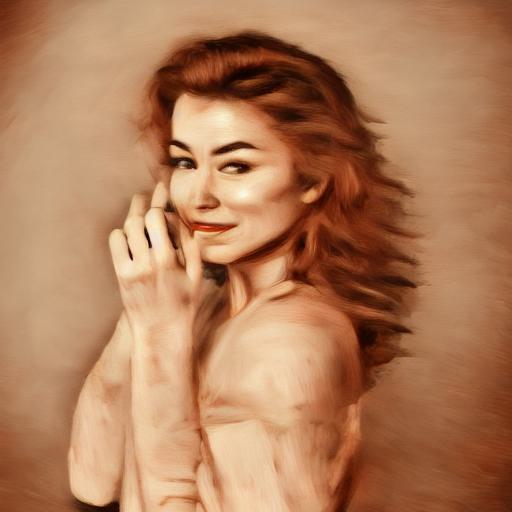} 
 \includegraphics[width=0.084\textwidth]{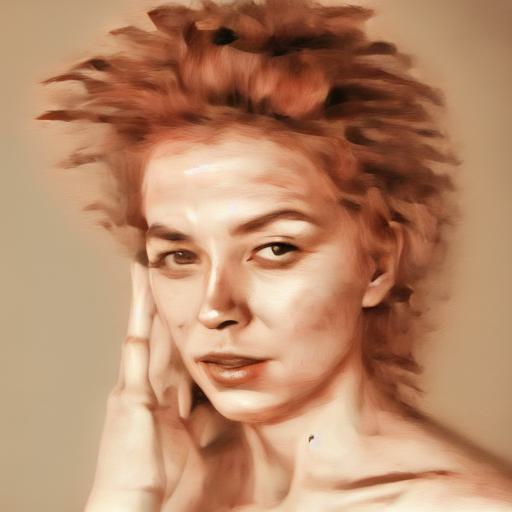} 
 \includegraphics[width=0.084\textwidth]{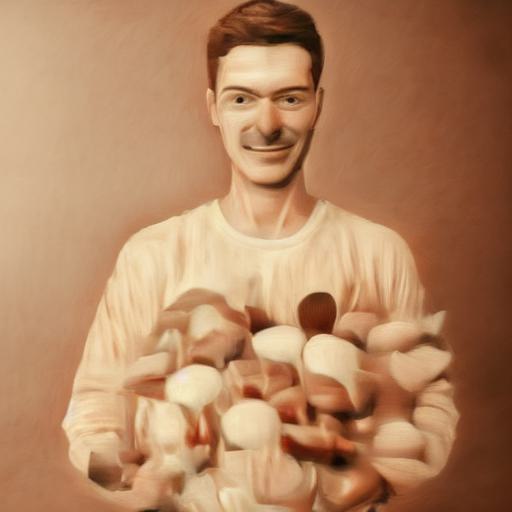} 
 \includegraphics[width=0.084\textwidth]{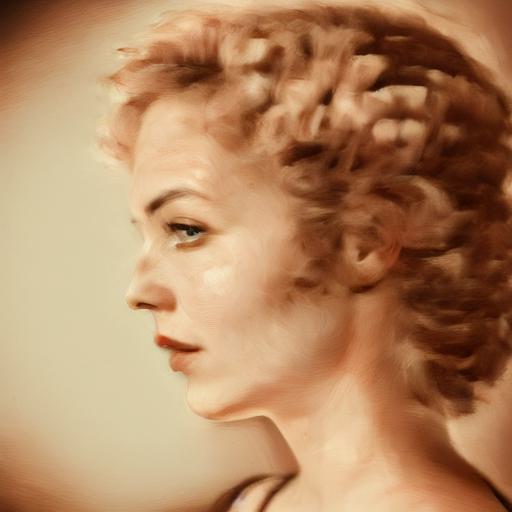}} \\ 
\midrule 
        \rowcolor{Gray}
        \multicolumn{2}{c}{~~~~~~~~~~~~\scriptsize{\Puw\texttt{:}
\scriptsize{\texttt{\textit{A painting of }\textcolor{Red}{Dogs}} \texttt{\textit{in} \textcolor{Blue}{Winter}\textit{ Style.}}}
}}\\
\midrule
        \begin{tabular}[c]{@{}c@{}}
        \vspace{-1.5mm}\tiny{\textbf{Original}} \\
        \vspace{-1.5mm}\tiny{\textbf{Diffusion}} \\
        \vspace{1mm}\tiny{\textbf{Model}}
        \end{tabular} & \multicolumn{1}{m{0.9\textwidth}}{
 \includegraphics[width=0.084\textwidth]{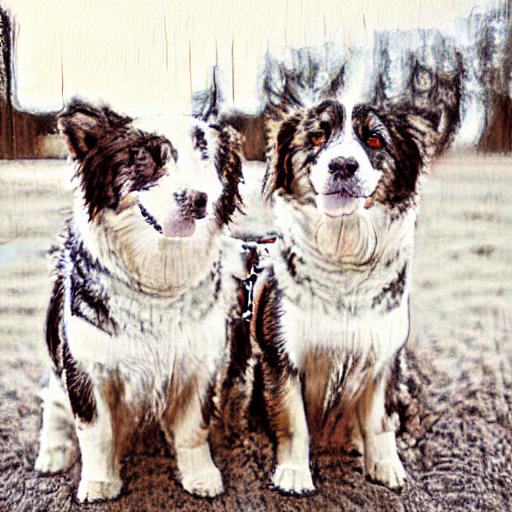} 
 \includegraphics[width=0.084\textwidth]{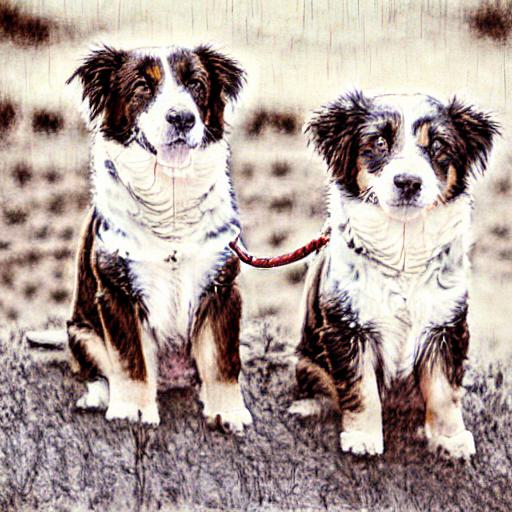} 
 \includegraphics[width=0.084\textwidth]{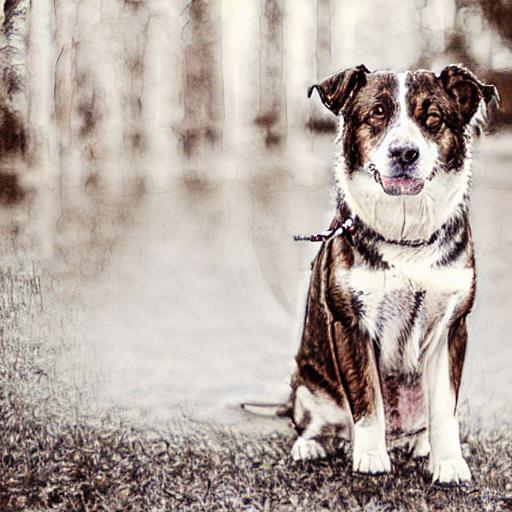} 
 \includegraphics[width=0.084\textwidth]{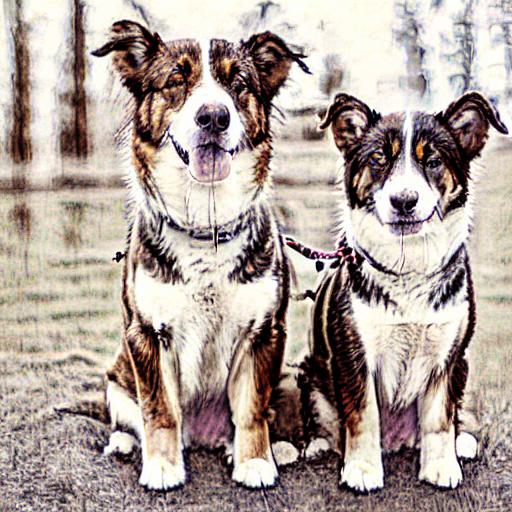} 
 \includegraphics[width=0.084\textwidth]{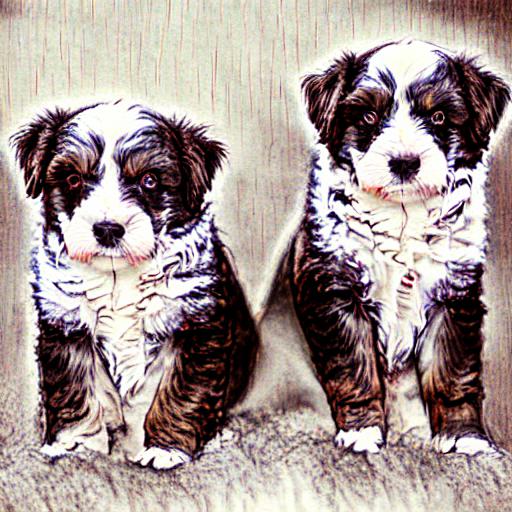} 
 \includegraphics[width=0.084\textwidth]{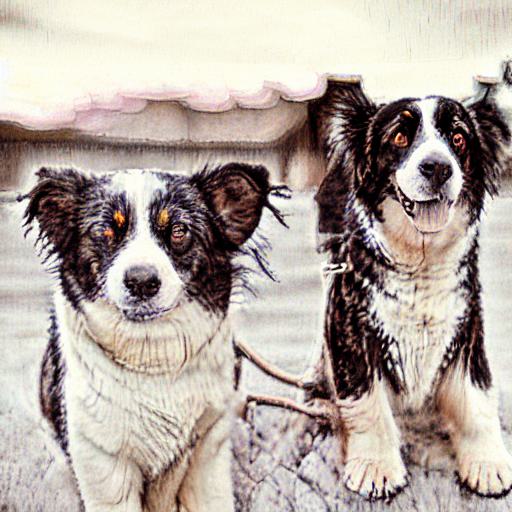} 
 \includegraphics[width=0.084\textwidth]{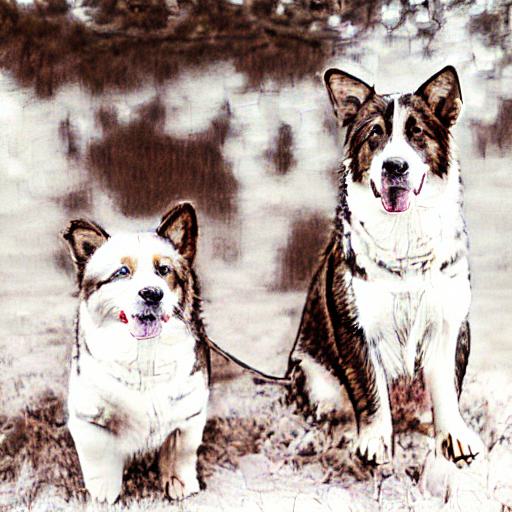} 
 \includegraphics[width=0.084\textwidth]{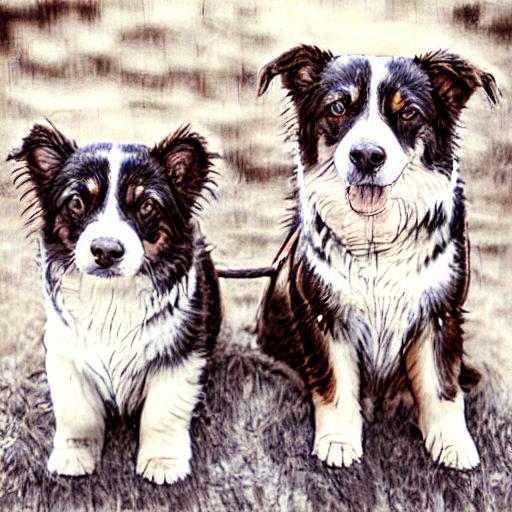} 
 \includegraphics[width=0.084\textwidth]{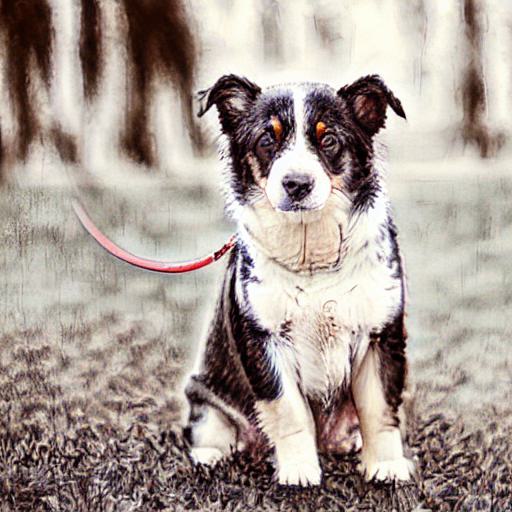} 
 \includegraphics[width=0.084\textwidth]{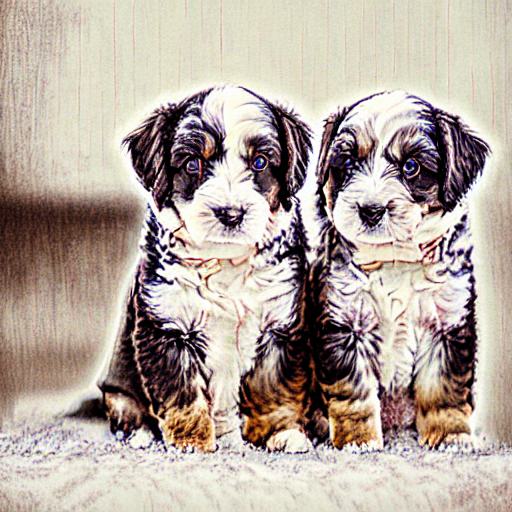}} \\ 
\midrule 
           \begin{tabular}[c]{@{}c@{}}
        \vspace{-1.5mm}\tiny{\textbf{Unlearned}} \\
        \vspace{-1.5mm}\tiny{\textbf{Diffusion}} \\
        \vspace{-1.5mm}\tiny{\textbf{Model}} \\
        \vspace{1mm}\tiny{\textbf{(Worst)}}
        \end{tabular} & \multicolumn{1}{m{0.9\textwidth}}{ 
 \includegraphics[width=0.084\textwidth]{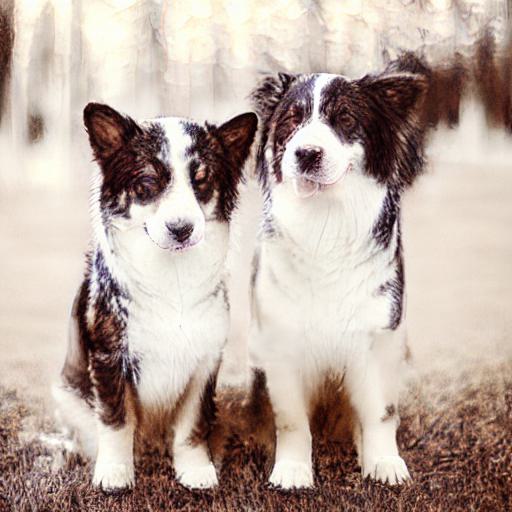} 
 \includegraphics[width=0.084\textwidth]{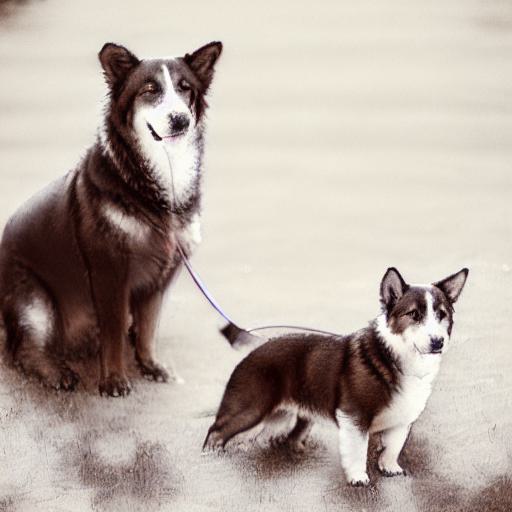} 
 \includegraphics[width=0.084\textwidth]{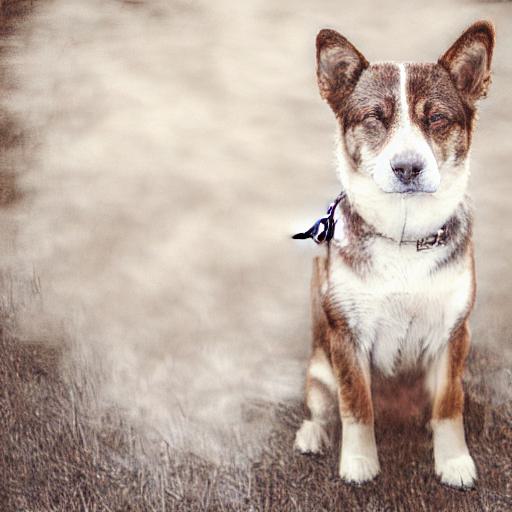} 
 \includegraphics[width=0.084\textwidth]{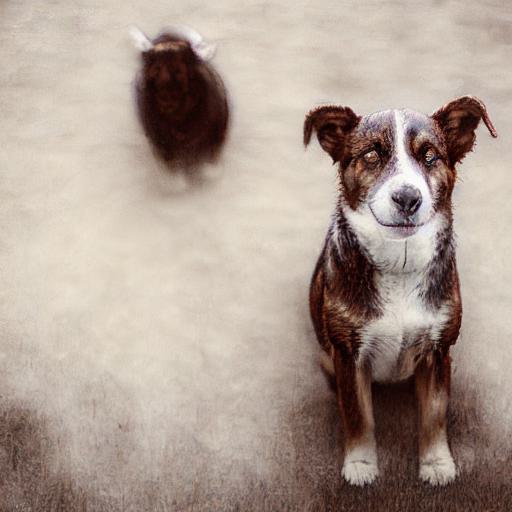} 
 \includegraphics[width=0.084\textwidth]{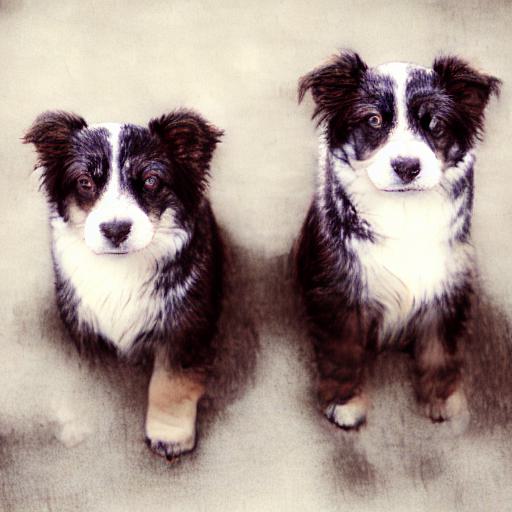} 
 \includegraphics[width=0.084\textwidth]{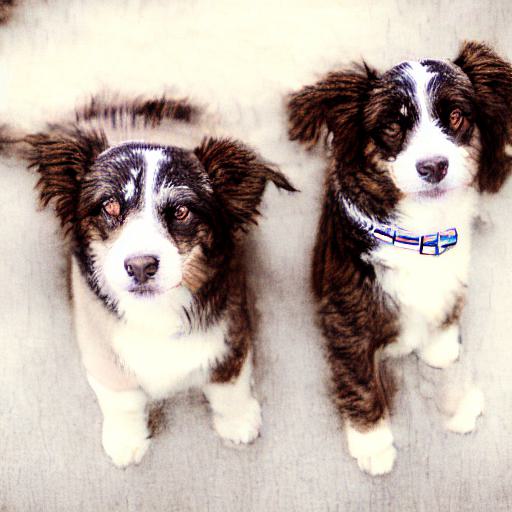} 
 \includegraphics[width=0.084\textwidth]{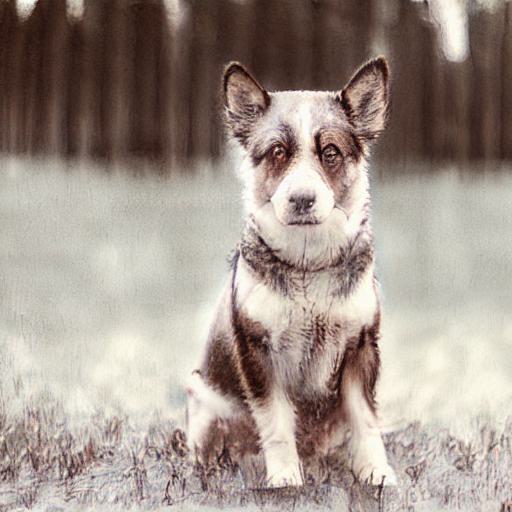} 
 \includegraphics[width=0.084\textwidth]{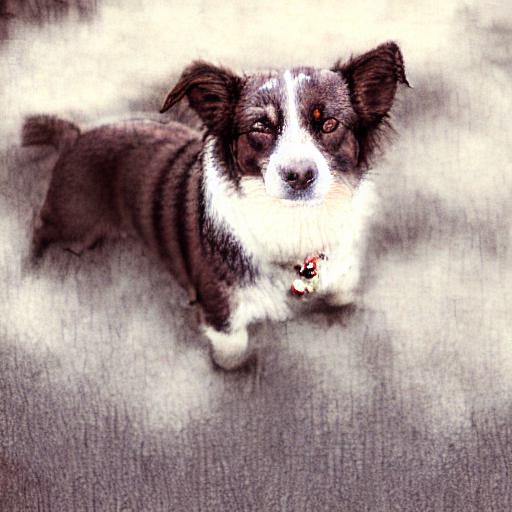} 
 \includegraphics[width=0.084\textwidth]{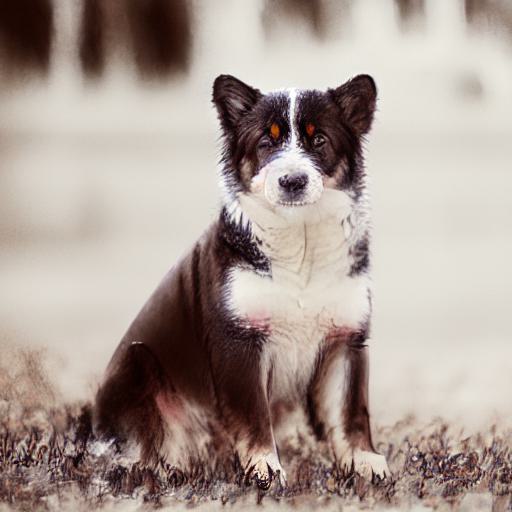} 
 \includegraphics[width=0.084\textwidth]{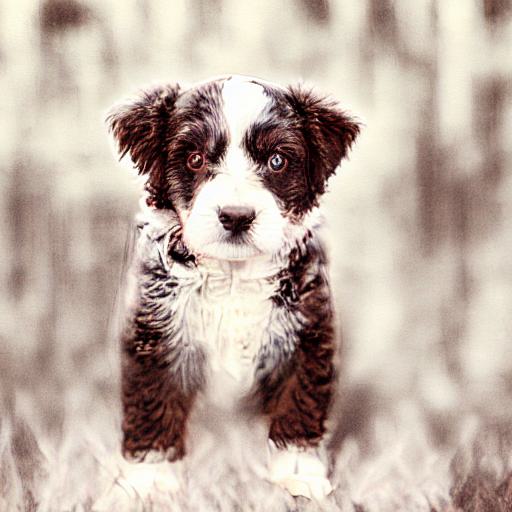}}
 \\ 
\midrule 

        \bottomrule[1pt]
    \end{tabular}
    }
    \end{center}
\end{table}

\end{document}